\documentclass{article}
\usepackage{natbib}

\usepackage{microtype}
\usepackage{graphicx}
\usepackage{booktabs} 
\usepackage{subcaption}
\usepackage{placeins}
\usepackage{hyperref}

\usepackage[accepted]{icml2025}

\usepackage{amsmath}
\usepackage{amssymb}
\usepackage{mathtools}
\usepackage{amsthm}

\usepackage[capitalize,noabbrev]{cleveref}

\theoremstyle{plain}

\theoremstyle{definition}

\theoremstyle{remark}

\usepackage[textsize=tiny]{todonotes}

\icmltitlerunning{First Hallucination Tokens Are Different from Conditional Ones}

\begin{document}
\twocolumn[
\icmltitle{First Hallucination Tokens Are Different from Conditional Ones}




\begin{icmlauthorlist}
\icmlauthor{Jakob Snel}{yyy}
\icmlauthor{Seong Joon Oh}{yyy,aaa}
\end{icmlauthorlist}

\icmlaffiliation{yyy}{University of Tübingen, Germany}
\icmlaffiliation{aaa}{Tübingen AI Center}

\icmlcorrespondingauthor{Jakob Snel}{jakob.snel@student.uni-tuebingen.de}

\icmlkeywords{Machine Learning, Uncertainty Estimation, Hallucination, Retrieval Augmented Generation, Token}

\vskip 0.3in
]

\begingroup
\let\clearpage\relax

\printAffiliationsAndNotice{}

\begin{abstract}

Large Language Models (LLMs) hallucinate, and detecting these cases is key to ensuring trust. While many approaches address hallucination detection at the response or span level, recent work explores token-level detection, enabling more fine-grained intervention. However, the distribution of hallucination signal across sequences of hallucinated tokens remains unexplored. We leverage token-level annotations from the RAGTruth corpus and find that the first hallucinated token is far more detectable than later ones. This structural property holds across models, suggesting that first hallucination tokens play a key role in token-level hallucination detection. Our code is available at \href{https://github.com/jakobsnl/RAGTruth_Xtended}{github.com/jakobsnl/RAGTruth\_Xtended}.
\end{abstract}

\section{Introduction}
Foundation models are transforming scientific research and society \citep{brown2020language, hallu_det_tree_prop}. However, their increasing capabilities raise critical questions about their responsible application, especially in terms of reliability and the potential to generate untruthful content \citep{hallu_det_tree_prop, halogen, token_benchmark, rawte2023surveyhallucinationlargefoundation}. 
The hallucination phenomenon, where LLMs produce non-factual or contradictory content, poses a key challenge for building trustworthy AI systems \citep{hallu_det_tree_prop, kaddour2023challengesapplicationslargelanguage}. Such errors can mislead users and undermine trust in critical applications \citep{robust_hallu_det, rawte2023surveyhallucinationlargefoundation}.
Although there are initiatives to alleviate hallucinations, LLMs are still fundamentally trained to approximate patterns in their training data. This makes hallucinations an inherent risk \citep{kirchhoff}. As a consequence, the need to detect hallucinated outputs is evident.\\
Detection methods vary regarding the granularity at which hallucinations are identified. While prior work has advanced response-level and span-level detection \citep{semantic_entropy, token_level_uncertainty, token_prob_hallu_det}, the majority is not designed to operate on token-level. Yet, token-level detection is increasingly important for enabling real-time filtering, targeted correction, and improved interpretability \citep{token_prob_hallu_det}. This shift is reflected in recent contributions \citep{token_benchmark, word_level_hallu}. However, a detailed understanding of how hallucination signals vary across tokens in a hallucinated span is lacking. The recently published large-scale corpus RAGTruth provides novel token-level hallucination annotations  that enable this  investigation \citep{ragtruth}.
\begin{figure}[h!]
\begin{center}
\centerline{\includegraphics[width=\columnwidth]{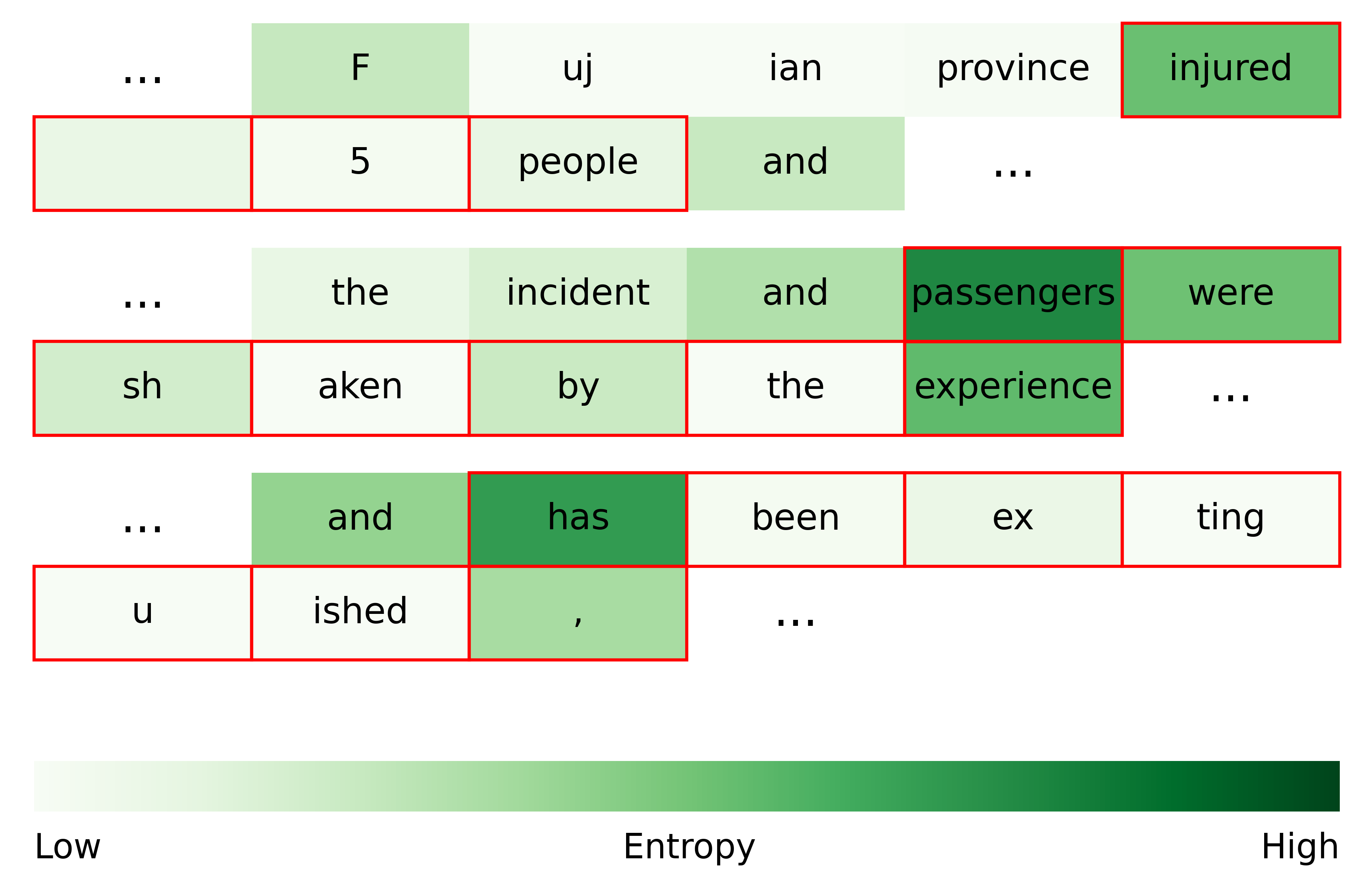}}
\caption{\textbf{First Hallucination Tokens Are Different:} We visualise three tokenised model responses from RAGTruth, overlaid with normalised logit entropy magnitudes. Tokens that are annotated as hallucination are highlighted with red outlines. The first hallucinated token exhibits higher entropy characteristics compared to conditional hallucinated tokens. This pattern holds consistently across different models, hallucination positions, and contexts. [model: llama-2-13b-chat, id: 214, 64, 730]}
\label{Sampled logit Llama 7B}
\end{center}
\end{figure}
\vspace{-14pt}
\FloatBarrier
Despite recent progress, current methods often overlook how the token-level hallucination signal evolves within a span. We argue that a token's corresponding hallucination signal depends on its position within the hallucinated sequence of tokens, the hallucinated span.
To investigate this, we hypothesise that the first token carries a stronger hallucination signal and achieves higher detection accuracy than subsequent, conditionally generated ones. 
We validate this hypothesis through a position-aware analysis using RAGTruth’s token-level annotations. Aware of various established hallucination signals such as intrinsic uncertainty \citep{hallu_det_tree_prop, agrawal2024languagemodelsknowtheyre}, internal representations \citep{chen2024insidellmsinternalstates, LLM_hallu_facts, early_det, token_level_uncertainty, Logits}, and external models as judges \citep{early_det, RAG_hallu_det, robust_hallu_det, kirchhoff}, we demonstrate that our hypothesis already holds for light-weight logit-based signals.
Therefore, we augment RAGTruth with reproduced output logits.
Our findings consistently support our hypothesis across models and contexts. This reveals a structural property of hallucination that improves understanding of token-level signals and supports more interpretable, fine-grained, and potentially real-time detection methods.

\section{Methodology}
\label{methodology}

This section outlines our approach for investigating whether hallucination detection signals vary systematically with token position in a hallucinated span.
\paragraph{Terminology.}
We define a hallucinated span as a contiguous sequence of tokens in a model-generated response that is annotated as hallucinated. A token's in-span index refers to its position within such a span, while the span index refers to the order of the hallucinated span within the response. Lastly, we refer to all subsequent tokens within the same hallucinated span as conditional tokens, reflecting their generation conditioned on the preceding hallucinated content.
\paragraph{Hypothesis.}
We hypothesise that the first token carries a stronger hallucination signal.  To test this, we analyse the detectability and separability of different token-level logit-derived signals.

Our methodology comprises three components: (1) enriching the RAGTruth dataset with model-generated logits for each response token; (2) categorising tokens by their position within hallucinated spans and across hallucination contexts; and (3) computing detectability and separability metrics for a range of logit-derived signals. The following subsections provide a detailed description of each component.

\subsection{RAGTruth Dataset}

Our dataset is a modified version of the RAGTruth corpus \citep{ragtruth}. RAGTruth provides large-scale token-level annotations of responses from a diverse range of state-of-the-art large language models (LLMs). 
For this work, we extract the token-level annotations and complement them with response token logits for all responses across the dataset.\\
As this reproduction step requires model access, we restrict our work to the publicly available Mistral-7b-Instruct \citep{jiang2023mistral7b}, Llama-2-7B-chat, Llama-2-13B-chat, and Llama-2-70B-chat \citep{zhao2025surveylargelanguagemodels}.

\subsection{Hallucination Token Positions}
\label{cat}

To evaluate how hallucination signals vary with token position, we categorise tokens to enable both detectability and separability analysis. This includes a basic split between non-hallucinated tokens ($\mathcal{T}_{\text{non}}$) and hallucinated tokens ($\mathcal{T}$), as well as more granular groupings that capture positional attributes within hallucinated spans.

\subsubsection{In-Span Index}
\label{inspan_index_main}

We group hallucinated tokens $\mathcal{T}$ by their positional index within a hallucination span. Let \( N \) be the maximum hallucination span length in the dataset. Then for each index \( k \) in a hallucinated span:

\vspace{-9pt}
{\small
\begin{equation}
\mathcal{T}_{k} = \left\{ t_i \mid t_i \text{ is the } k\text{th token in a span}, \; k = 0, 1, \dots, N \right\}
\end{equation}
}

where each \( \mathcal{T}_{k} \) corresponds to tokens at a specific positional index within hallucinated spans. To uncover inter-group differences, we analyse each set independently.

\subsubsection{Span Index}
\label{span_index_main}

In addition to analysing token position within hallucination spans, we examine whether detectability and separability patterns apply across different hallucination spans within the same response. Specifically, we validate that hallucination signals are consistent across different span positions in the response, for example, whether it is the first hallucinated span or a later one.\\
To test this, we differentiate hallucinated tokens by their span index within the response. Hallucination tokens are grouped according to their response-wide index of the hallucination span they are part of. Let a response contain \( M \) hallucination spans, denoted as \( S_1, S_2, \dots, S_M \), where each span \( S_j \) consists of a sequence of hallucinated tokens. We define the set of tokens belonging to the \( j \)-th hallucination span and in-span index \( k \) as:

{\small
\begin{equation}
\mathcal{T}_{k}^{(j)} = \{ t_{ki} \mid t_{ki} \in T_{kj} \}, \quad j = 1, \dots, M
\end{equation}
}

However, the distribution of sample sizes across hallucination span indices is not balanced, with later spans containing fewer tokens. To mitigate this imbalance, we introduce a binned grouping strategy.
Let $j \in \mathbb{N}$ denote the span index, starting from 0. We aggregate tokens from spans with $j \geq 2$ into $\mathcal{T}^{\text{third+}}$, and define $\mathcal{T}^{\text{all}}$ as the union of all hallucinated spans. This categorisation enables us to compare positional signal strength across early and later hallucination occurrences while maintaining sufficient sample sizes\footnote{As later hallucination spans are less frequent, we bin them.}, as shown in Table \ref{span_dist}.

\begin{table}[h]
\vskip 0 in
\begin{center}
\begin{small} 
\begin{tabular}{p{0.74cm}p{1.29cm}p{1.29cm}p{1.29cm}p{1.37cm}} 
\hline
 & llama-2-7b-chat & llama-2-13b-chat & llama-2-70b-chat & mistral-7b-instruct \\
\hline
$\mathcal{T}^{\text{all}}$ & 1832 & 1677 & 1395 & 1953 \\
$\mathcal{T}^{\text{first}}$ & 1012 & 697 & 744 & 1026 \\
$\mathcal{T}^{\text{second}}$ & 460 & 414 & 346 & 533 \\
$\mathcal{T}^{\text{third+}}$ & 360 & 566 & 305 & 394 \\
\cmidrule{1-5}
$\mathcal{T}_{\text{no}}$ & 1133 & 1288 & 1570 & 1012 \\
\hline
\end{tabular}
\end{small}
\end{center}
\caption{Model-wise distribution of hallucination span counts across response-wide span indices. $\mathcal{T}^{\text{all}}$ is the count of all dataset-wide hallucination spans, while $\mathcal{T}_{\text{no}}$ is the count of responses free of hallucination.}
\label{span_dist}
\vspace{-3pt}
\end{table}
\subsection{Detectability}
\label{det}
Following prior work, we frame token-level hallucination detection as a binary classification task: predicting whether a given token is hallucinated or not.
We hypothesise that the detectability of hallucinated tokens varies systematically with their position in a span. To test this, we compare each positional subgroup in our categorisation against non-hallucinated tokens.

We quantify detectability using the area under the receiver operating characteristic curve (AUROC) computed over a set of scalar signals derived from non-hallucinated ($y = 0$) and hallucinated ($y = 1$) outputs. These include commonly used uncertainty measures such as entropy, perplexity, sampled probability, and logit \citep{early_det, entropy_new, memberinference_joon}, as well as auxiliary signals like logit vector mean, variance, and L2 distance. Signals are computed per token and grouped according to the positional categories defined in Section \ref{cat}.\\
To capture both global and local trends, we compute AUROC scores across the entire dataset as well as per response (more details are provided in Appendix \ref{app:det}). This response-level perspective reflects a real-time usage setting, enabling us to assess the consistency of inter-positional patterns across independent model responses.

\subsection{Separability}
\label{sep}

We verify that the observed positional patterns are specific to hallucinated tokens rather than generic artifacts of token position. To do this, we examine whether similar signal patterns also appear in non-hallucinated spans. As part of this verification, the separability analysis includes non-hallucinated and hallucinated tokens for comparison.

We extract two subsets from the non-hallucinated token set \(\mathcal{T}^{\text{non}}\): \(\mathcal{T}^{\text{no}}\), containing tokens from hallucination-free responses, and \(\mathcal{T}^{\text{pre}}\), containing pre-hallucination tokens\footnote{For simplicity, we assume that non-hallucinated tokens preceding the first hallucination span exhibit similar patterns to those following it.}. For both subsets, we exclude the initial $<$start$>$ and the first generated token as their logits, regardless of whether the token is hallucinated or not, differ from those of conditional tokens. This follows from the Min-K probability and entropy distributions of \( \mathcal{T}^{\text{pre}} \) and \( \mathcal{T}^{\text{no}} \) in Appendix \ref{plt:mink:perc}. We bin the remainder by their position, following the in-span index logic from Section \ref{cat}. This yields six token groups for qualitative comparison with hallucinated tokens.

To measure distributional separability across these subsets, we adopt metrics from Membership Inference Attacks (MIA), which similarly rely on confidence-based features to distinguish in vs out-of-distribution behaviour. We use Min-K probability as our primary separability metric due to its proven efficacy in MIA contexts \citep{memberinference_joon, mink_p}, and also compute the Min-K entropy to support our findings. This yields a family of scores that reflect how token response logits vary across categories and positions. See Appendix \ref{app:metrics:glob:sep} for implementation details.

\section{Results}
\label{results}
Our experimental setup tests whether the detectability and separability of hallucinated tokens vary with their position within hallucinated spans. We specifically investigated two questions: (1) How does the detectability and separability of hallucinated tokens vary depending on their position within a hallucinated span? (2) Which logit-derived signals most reliably detect hallucinated tokens and separate them from truthful tokens?\\
Although our analysis covers all in-span token positions, we concentrate on the first nine token indices, as the median hallucination span length ranges between six and eight tokens, depending on the model.

\subsection{First Hallucination Tokens Are Better Detectable than Conditional Ones — Globally...}
Hallucination tokens at the in-span index 0 appear to be more distinguishable than conditional tokens (see Figure \ref{AUROC:plot}). In our simplified detection setup, the detection accuracy of conditional hallucination tokens is slightly higher than 0.5, regardless of the signal. In contrast, the first hallucination token appears to be strongly distinguishable, as indicated by entropy and perplexity, yielding AUROC scores close to 0.8 across all models (see Appendix \ref{plt:auroc}).

\begin{figure*}[!t]
\centering
\includegraphics[width=\textwidth]{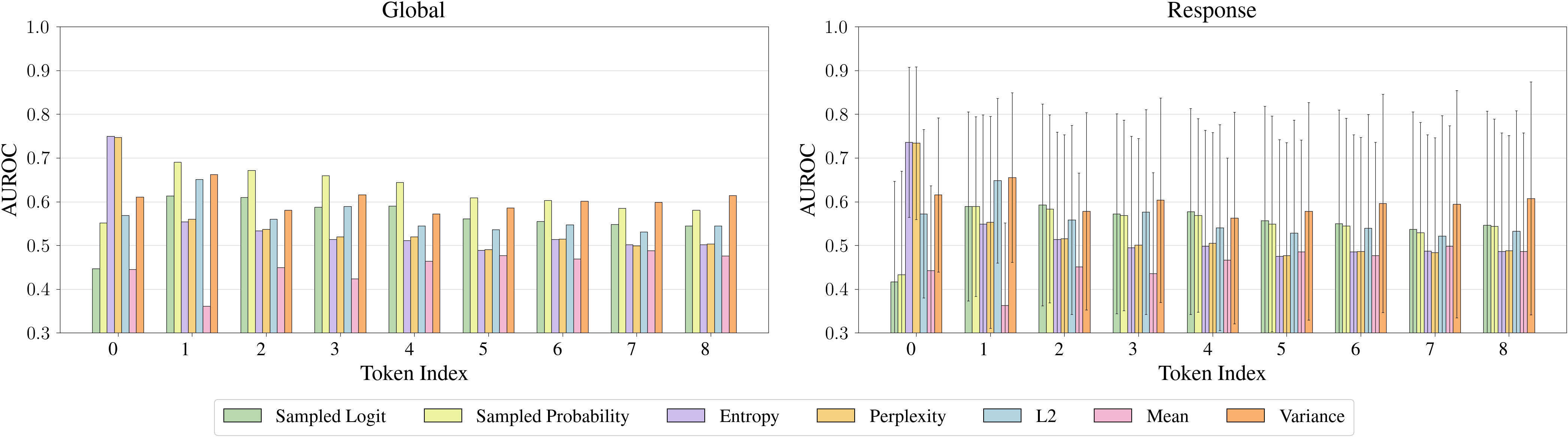}
\caption{\textbf{First Hallucination Tokens Are Better Detectable:} We show AUROC scores per signal and in-span hallucination token index across all hallucination spans. We report both global and averaged response-level scores. For the latter, we add error bars to account for the score distribution across different responses. Per analysis level and model, we invert AUROC scores that are, averaged over all indices, below 0.5 on $\mathcal{T}^{\text{all}}$. [llama-2-13b-chat; all]}
\label{AUROC:plot}
\vspace{0pt}
\end{figure*}

\begin{figure}[t]
\centering
\centerline{\includegraphics[width=\columnwidth]{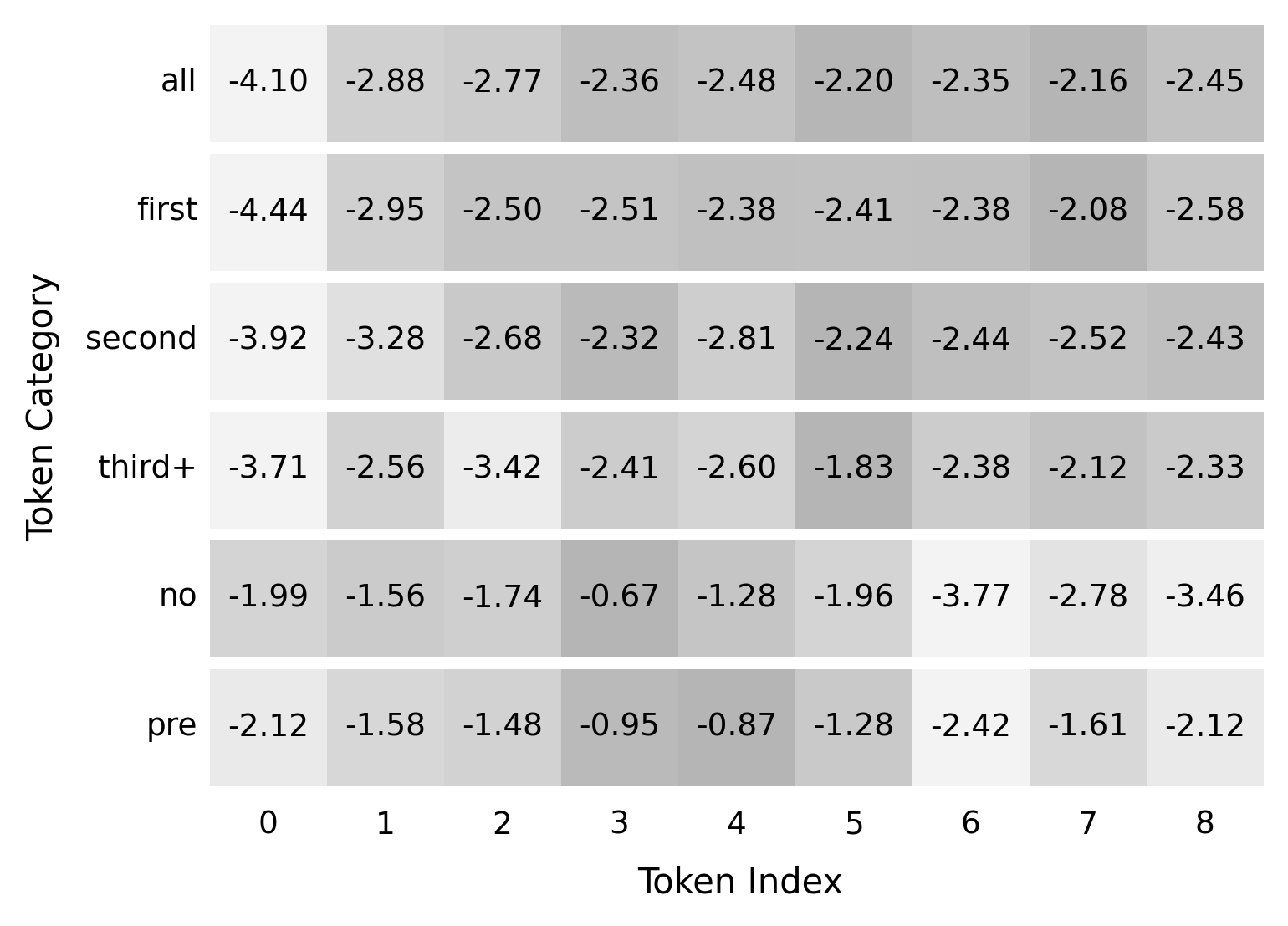}}
\caption{\textbf{First Hallucination Tokens Exhibit Greater Separability:} Min-10 probability distribution across different token categories and indices. Grey magnitudes are normalised across the entire category, while the numerical scores are not. Separability patterns are consistent across all percentiles in the range of 10 to 100 concerning token rankings (see appendix \ref{plt:mink:perc}). As the contrast is the greatest for the 10th percentile threshold, we choose it for visualisation. [llama-2-13b-chat; all]}
\label{Mink_plot}
\vspace{-20pt}
\end{figure}

\subsection{...and Locally}
At the response level, AUROC trends mirror global findings. However, the error bars in Figure \ref{AUROC:plot} reveal high variability in the first-token detectability. It becomes evident that, at least for raw token logit entropy and perplexity signals, the first token detectability is not stable but varies heavily. 

\subsection{First Hallucination Tokens Exhibit Greater Separability Than Conditional Ones}
Min-K Probabilities further support the enhanced detection scores for first hallucination tokens, which consistently exhibit lower scores than conditional ones (see Figure \ref{Mink_plot}). This pattern is consistent across models and percentiles (see Appendix \ref{plt:mink:sep})

\subsection{Entropy Most Effectively Identifies First Hallucinated Tokens}
Among all logit-derived hallucination signals tested, entropy yields the most pronounced separation between first and conditional hallucination tokens. This is reinforced by the larger score gap for Min-K entropy than for Min-K probability (see Appendix \ref{plt:mink:sep}). As a consequence, we conclude that logit entropy is the signal from our set that best reflects whether the first token is hallucinated.

\subsection{Individual Logit-Derived Signals Are Not Robust Across Token Indices} 
While our analysis reveals clear positional trends in hallucination detectability, it also highlights a key limitation: none of the evaluated logit-derived signals consistently detects hallucinated tokens across all in-span indices. Some, such as logit entropy, perform well for early tokens but degrade in later positions, while others show inconsistent or noisy behavior. This suggests that no single logit-derived feature is sufficient for robust, position-invariant hallucination detection.

\section{Conclusion}
\label{conclusion}
Our qualitative detectability and separability analysis reveals that first hallucination tokens are systematically more distinguishable than subsequent, conditionally generated ones. This pattern is evident across multiple logit-derived uncertainty signals, with logit entropy providing the clearest signal for first-token hallucination detection. However, no hallucination signal achieves robust performance across all in-span positions, highlighting the limitations of current logit-based methods.
These findings motivate several directions for future work. First, we hypothesise that richer model internals, such as hidden states, may amplify the observed positional effects and improve detection reliability.
Second, a complementary investigation into the last token in each span may reveal symmetric patterns and help characterise hallucination span boundaries with more precision. 
As our results also indicate that no single logit-derived metric consistently captures the hallucination signal across all token positions, they suggest the need for more robust or composite detection signals.
We leave these extensions for future work.

\subsection{Limitations}
\label{app:limit}
First, our approach assumes the accuracy of the hallucination span annotations provided in RAGTruth. Given the subjective nature of hallucination annotation, this could skew results either positively or negatively.

We provide a qualitative analysis based on a simplified detection setup, rather than a deployable classifier. Therefore, we leave for future work whether the observed patterns directly improve fine-grained hallucination token detection in practice.

Additionally, our analysis focuses specifically on intra-hallucination token patterns, particularly detectability and separability within hallucination spans. However, prior and subsequent tokens outside the hallucinated spans might also carry predictive signals for hallucination, as suggested in recent studies \citep{li2024dawndarkempiricalstudy, ragtruth, fine_hallu_det}.

Lastly, we neglect the hallucination taxonomy introduced in RAGTruth. While \citet{ragtruth} distinguish between Evident Conflict, Subtle Conflict, Evident Introduction of Baseless Information, and Subtle Introduction of Baseless Information across different task categories (QA, Data-to-Text, Summarisation), we treat all hallucinations uniformly. This decision promotes generality across hallucination types but leaves open whether the observed patterns are consistent across specific hallucination categories and tasks \citep{li2024dawndarkempiricalstudy, halogen}.

\section*{Impact Statement}
This work contributes to the responsible development of foundation models by advancing the diagnostic understanding of hallucinations at the token level. By uncovering structural patterns of hallucination signal in hallucinated spans, it provides insight into where and how hallucinations emerge, supporting the development of interpretable token-level detection methods.
As this is a knowledge-oriented analysis rather than a deployed system, it poses minimal direct societal risk. Instead, it lays the groundwork for building more trustworthy and accountable language models by improving the understanding of their failure modes.
\endgroup
\newpage
\bibliography{rta}
\bibliographystyle{icml2025}
\newpage
\appendix
\onecolumn

\section{Appendix}

\begingroup
\let\clearpage\relax

\subsection{Extended Methodology}
\label{app:methods}

\subsubsection{Detectability}
\label{app:det}
For each in-span index \( k \in \{0, 1, \dots, N\} \), feature \( f \), and span group \( g \in \{\text{first}, \text{second}, \text{third+}, \text{all}\} \), we define the global detectability as:

{\small
\vspace{-15pt}
\begin{equation}
\text{AUROC}_f(k, g) = \text{AUROC}\left( \mathcal{F}^{\text{non}}, \mathcal{F}_{k}^{(g)} \right)
\end{equation}
\vspace{-15pt}
}

where $\mathcal{F}^{\text{non}}$ and $\mathcal{F}_{k}^{(g)}$ denote the feature sets for non-hallucinated and hallucinated tokens across each models dataset.

For each response \( r \), we define the local detectability as:

{\small
\begin{equation}
\text{AUROC}_f^{(r)}(k, g) = \text{AUROC}\left( \mathcal{F}_{\text{non}}^{(r)}, \mathcal{F}_{k}^{(g, r)} \right)
\end{equation}
}

where $\mathcal{F}_{\text{non}}^{(r)}$ and $\mathcal{F}_{k}^{(g, r)}$ denote the feature sets for non-hallucinated and hallucinated tokens within the same model response $r$, respectively. 

\subsubsection{Separability}
\label{app:metrics:glob:sep}
We apply our qualitative separability analysis across six distinct token groups:

{\small
\vspace{-10pt}
\begin{equation}
\mathcal{T}_k^{\text{all}}, \mathcal{T}_k^{\text{first}}, \mathcal{T}_k^{\text{second}}, \mathcal{T}_k^{\text{third+}}, \mathcal{T}_k^{\text{pre}}, \mathcal{T}_k^{\text{no}}
\end{equation}
\vspace{-15pt}
}

\noindent where $k$ denotes the in-span token index for hallucination, and in-response index for non-hallucination. 

Min-K is defined as the \( K \)-th smallest value among the metric scores of a given token group \cite{mink_p}. 
Let \( f(t) \) be a scalar metric computed per token \( t \), and let \( r \in \{10, 20, \dots, 100\} \) be the percentile threshold. For each group \( \mathcal{T}_g \in \mathcal{G} \), and for a fixed in-span position \( k \), we compute:

{\small
\begin{equation}
\text{MIN-K}_r\left(\mathcal{T}_{k}^{g}\right) = \text{K}_{r\%}\left( \{ f(t_i) \mid t_i \in \mathcal{T}_{k}^{g} \} \right)
\end{equation}
}

\noindent where \( \text{K}_{r\%} \) denotes the \( r \)-th percentile of the sorted metric values.

\newpage

\endgroup

\newpage
\begingroup
\let\clearpage\relax
\subsection{Detectability}
\label{plt:auroc}
\begin{figure}[!h]
    \centering
    \begin{subfigure}{0.85\textwidth}
        \centering
        \includegraphics[width=\linewidth]{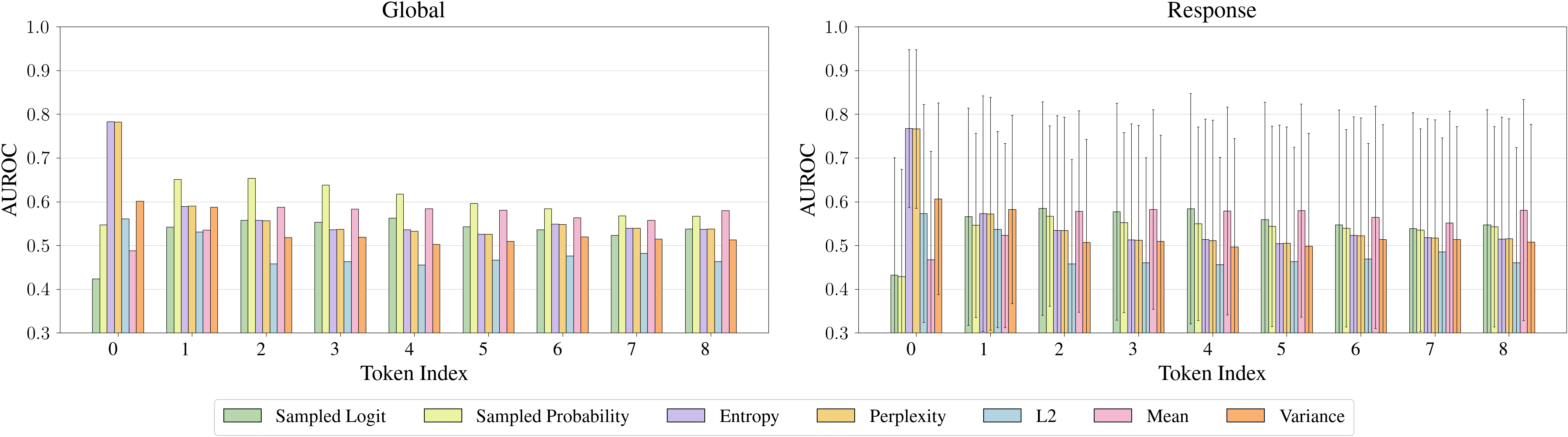}
        \subcaption{LLaMA-2-7B-chat}
        \label{det:all:llama-7b}
    \end{subfigure}
    \medskip
    \begin{subfigure}{0.85\textwidth}
        \centering
        \includegraphics[width=\linewidth]{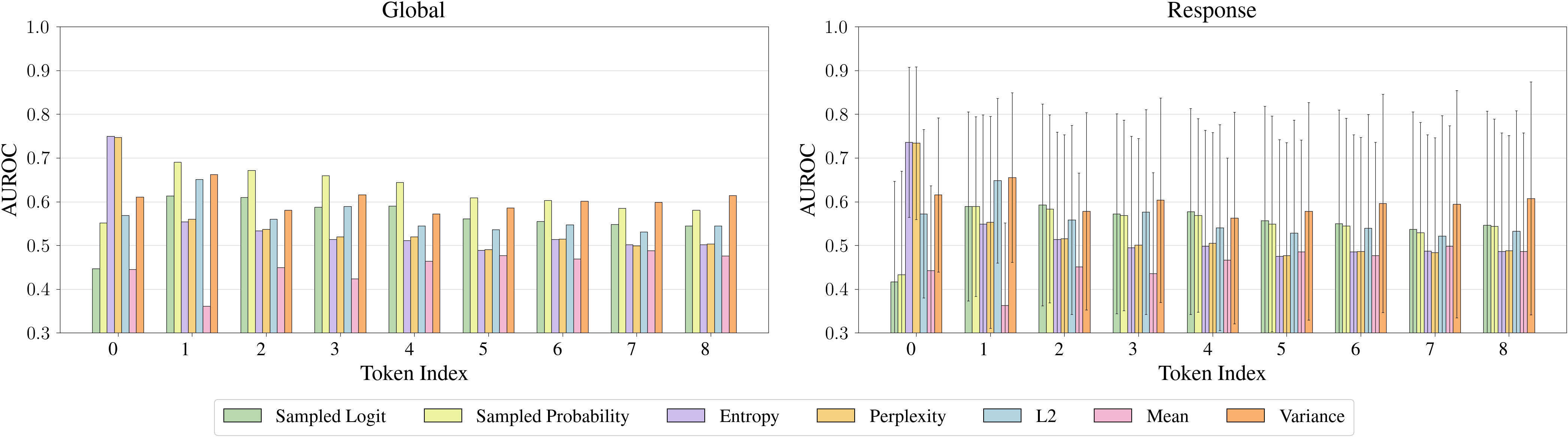}
        \subcaption{LLaMA-2-13B-chat}
        \label{det:all:llama-13b}
    \end{subfigure}
    \medskip
    \begin{subfigure}{0.85\textwidth}
        \centering
        \includegraphics[width=\linewidth]{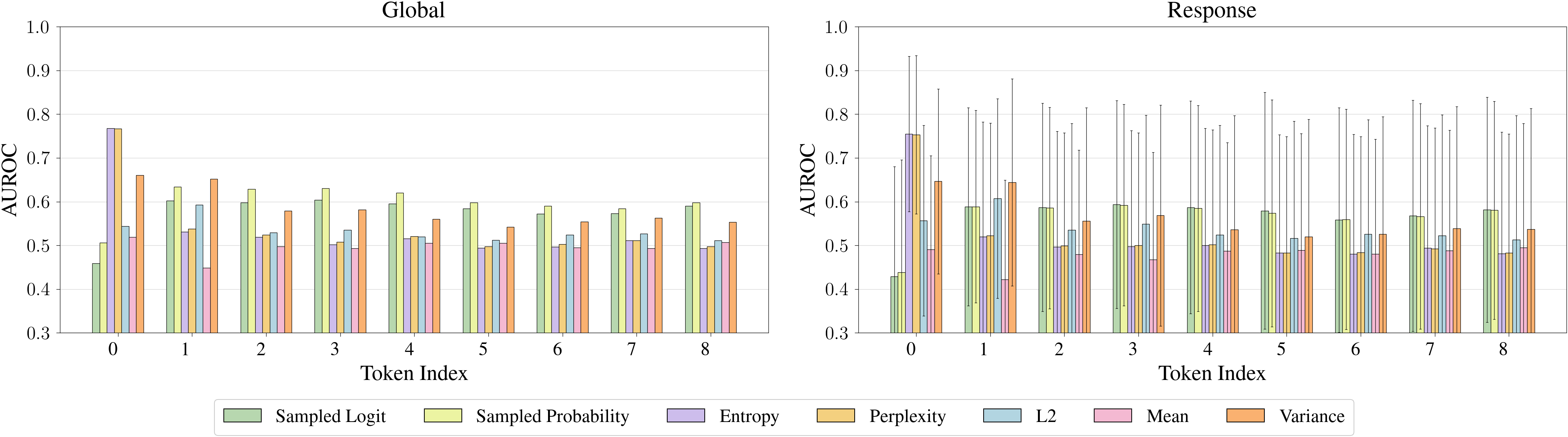}
        \subcaption{LLaMA-2-70B-chat}
        \label{det:all:llama-70b}
    \end{subfigure}
    \medskip
    \begin{subfigure}{0.85\textwidth}
        \centering
        \includegraphics[width=\linewidth]{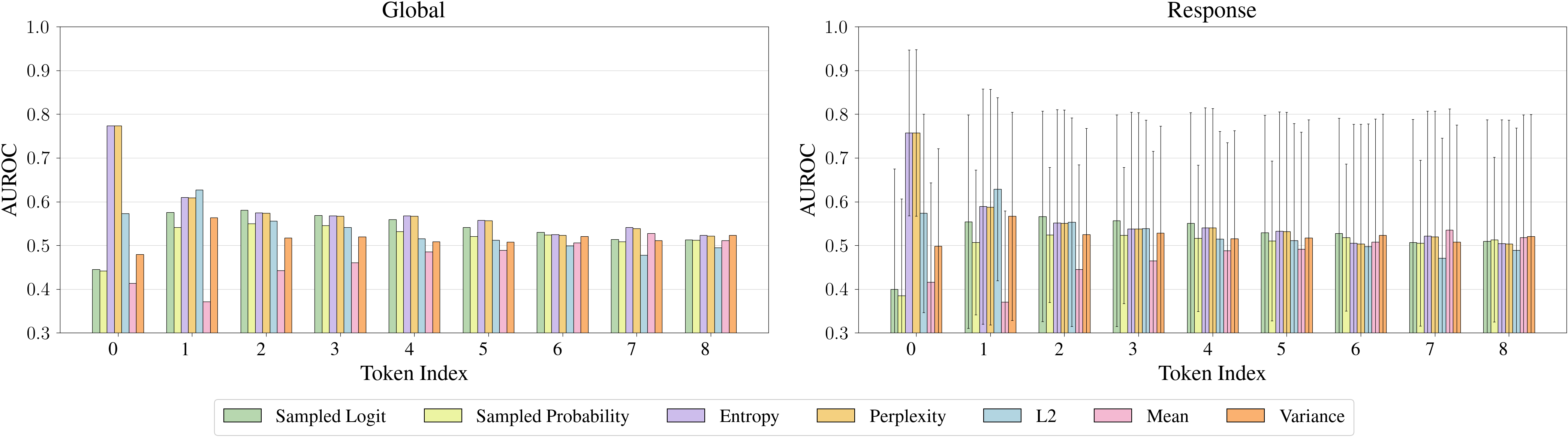}
        \subcaption{Mistral-7B-instruct}
        \label{det:all:mistral-7b}
    \end{subfigure}
    \caption{\textbf{[all]} AUROC per signal and in-span hallucination token indices from all hallucination spans at both global and response level.}
    \label{det:all}
\end{figure}
\begin{figure}[p]
    \vspace*{\fill}
    \centering
    \begin{subfigure}{0.8\textwidth}
        \centering
        \includegraphics[width=\linewidth]{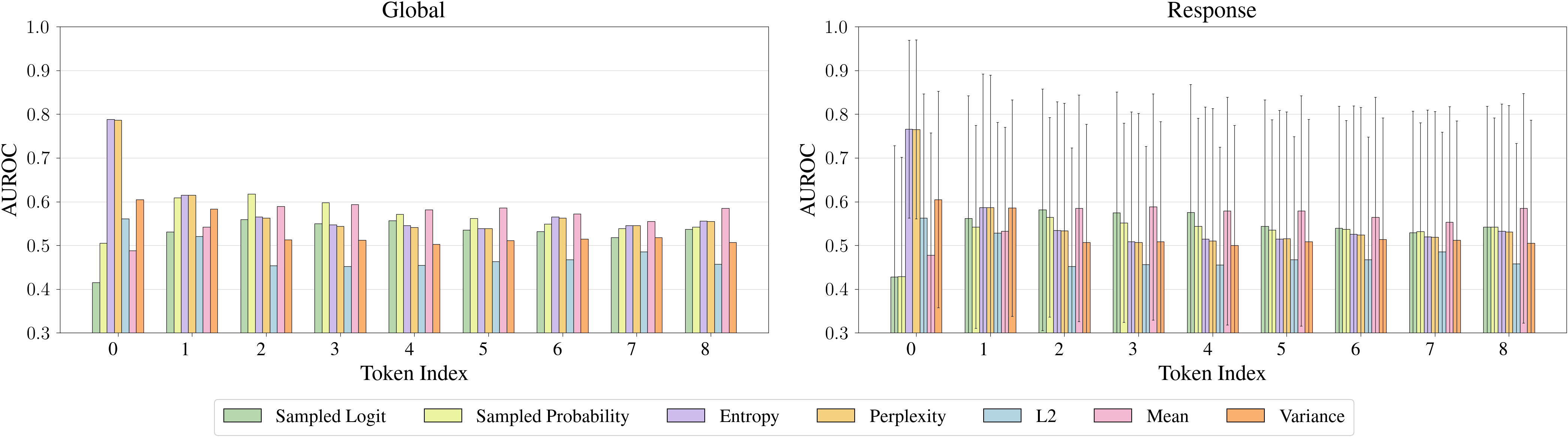}
        \subcaption{LLaMA-2-7B-chat}
        \label{det:first:llama-7b}
    \end{subfigure}
    \medskip
    \begin{subfigure}{0.8\textwidth}
        \centering
        \includegraphics[width=\linewidth]{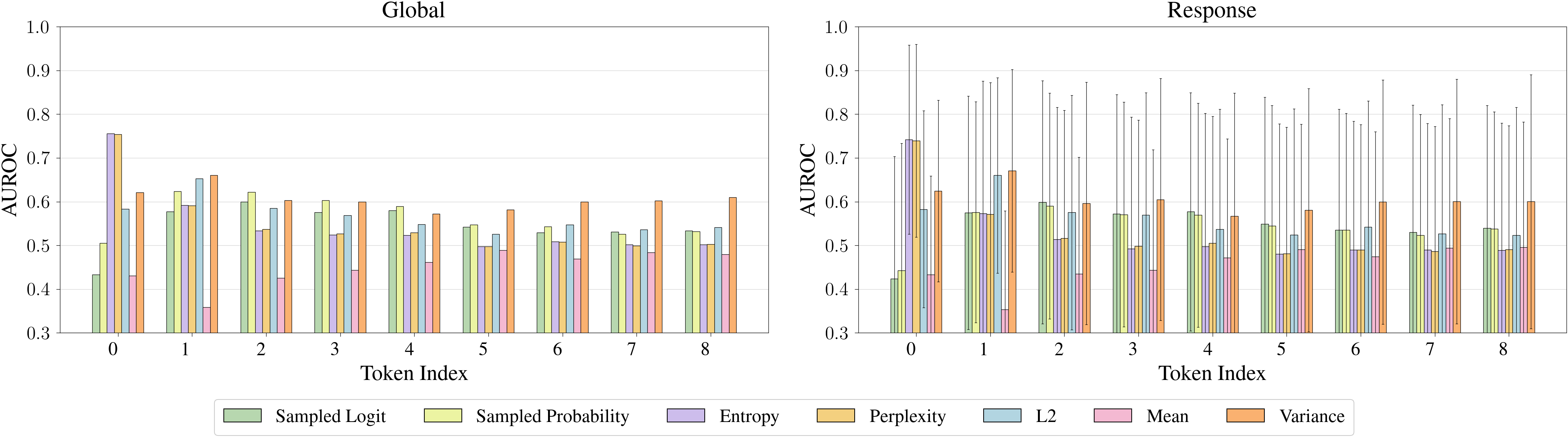}
        \subcaption{LLaMA-2-13B-chat}
        \label{det:first:llama-13b}
    \end{subfigure}
    \medskip
    \begin{subfigure}{0.8\textwidth}
        \centering
        \includegraphics[width=\linewidth]{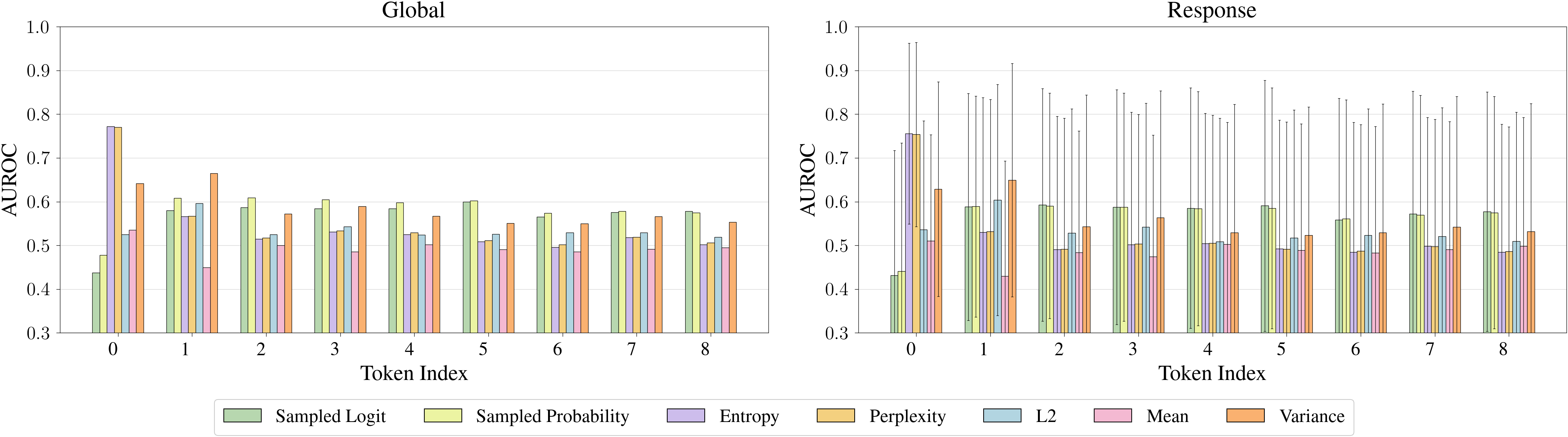}
        \subcaption{LLaMA-2-70B-chat}
        \label{det:first:llama-70b}
    \end{subfigure}
    \medskip
    \begin{subfigure}{0.8\textwidth}
        \centering
        \includegraphics[width=\linewidth]{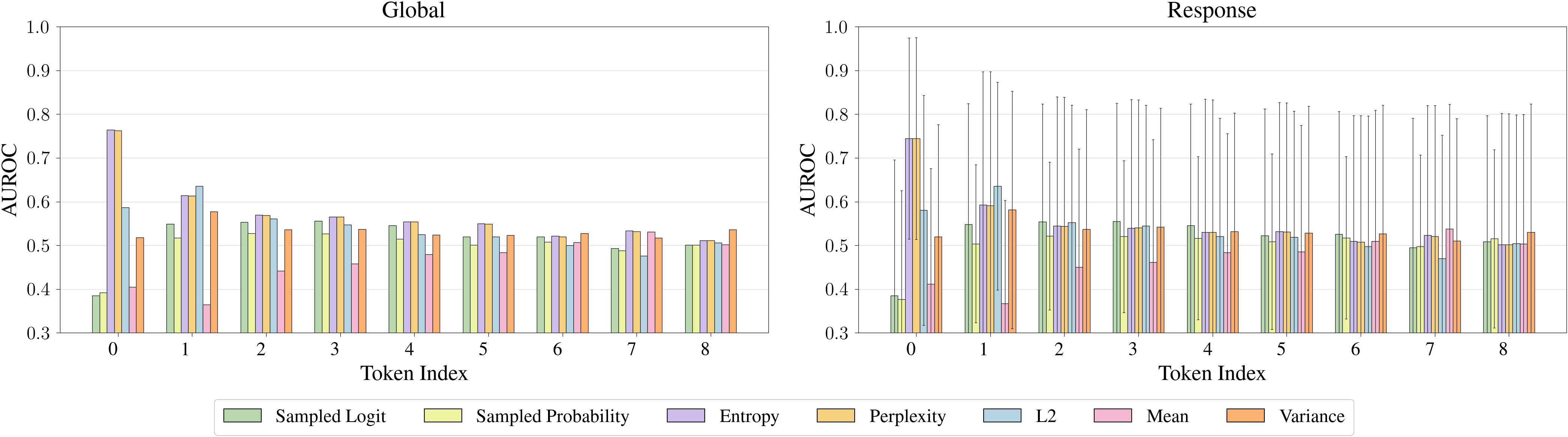}
        \subcaption{Mistral-7B-instruct}
        \label{det:first:mistral-7b}
    \end{subfigure}
    \caption{\textbf{[first]} AUROC per signal and in-span hallucination token indices from first hallucination spans at both global and response level.}
    \label{det:first}
    \vspace*{\fill}
\end{figure}
\begin{figure}[p]
    \vspace*{\fill}
    \centering
    \begin{subfigure}{0.8\textwidth}
        \centering
        \includegraphics[width=\linewidth]{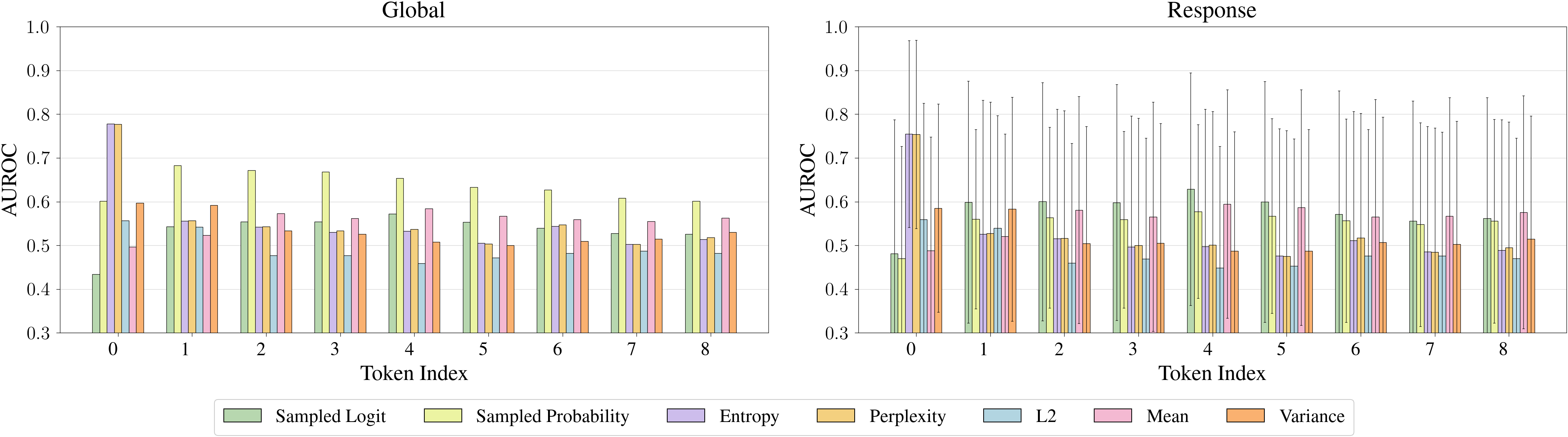}
        \subcaption{LLaMA-2-7B-chat}
        \label{det:second:llama-7b}
    \end{subfigure}
    \medskip
    \begin{subfigure}{0.8\textwidth}
        \centering
        \includegraphics[width=\linewidth]{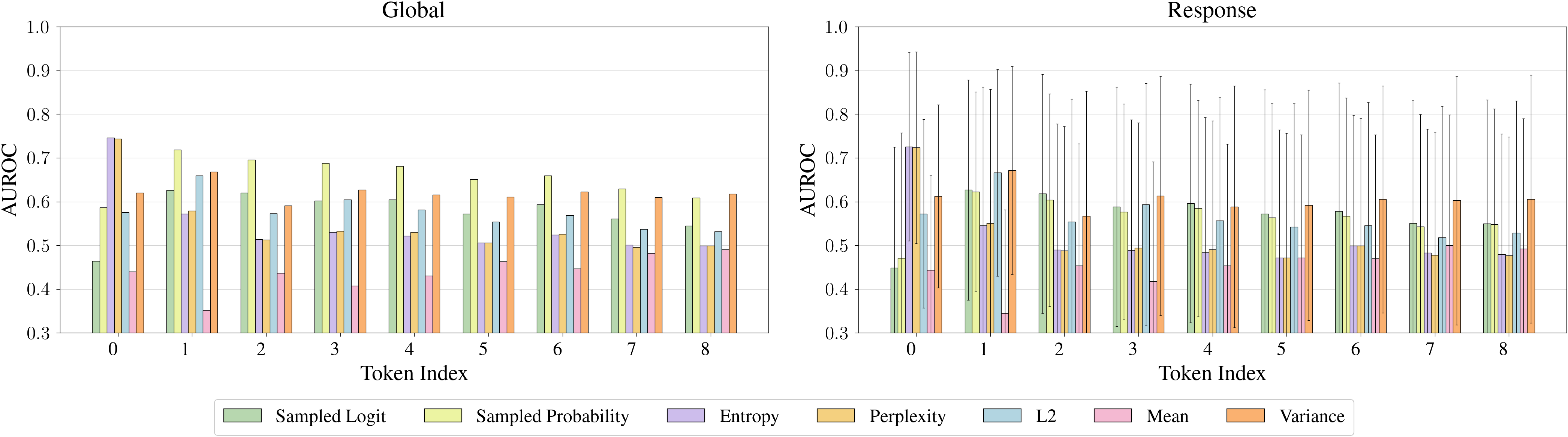}
        \subcaption{LLaMA-2-13B-chat}
        \label{det:second:llama-13b}
    \end{subfigure}
    \medskip
    \begin{subfigure}{0.8\textwidth}
        \centering
        \includegraphics[width=\linewidth]{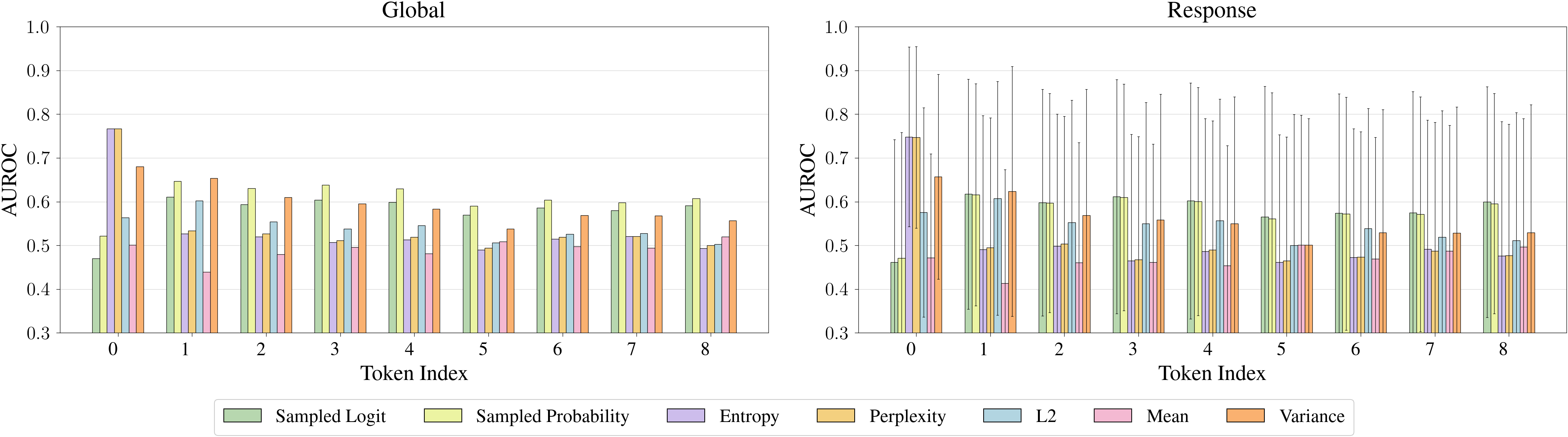}
        \subcaption{LLaMA-2-70B-chat}
        \label{det:second:llama-70b}
    \end{subfigure}
    \medskip
    \begin{subfigure}{0.8\textwidth}
        \centering
        \includegraphics[width=\linewidth]{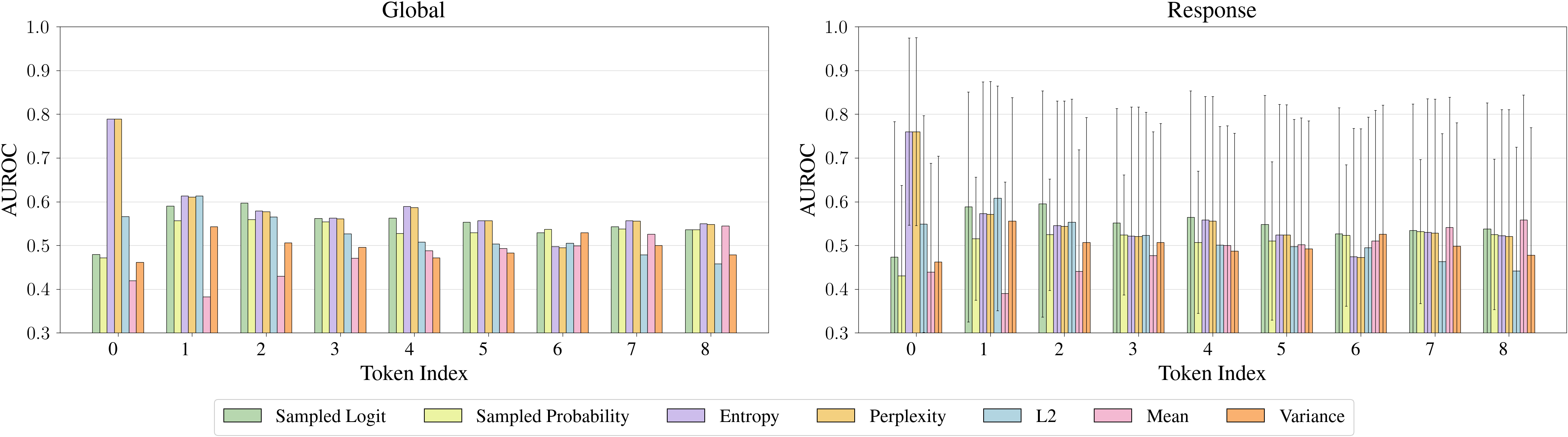}
        \subcaption{Mistral-7B-instruct}
        \label{det:second:mistral-7b}
    \end{subfigure}
    \caption{\textbf{[second]} AUROC per signal and in-span hallucination token indices from second hallucination spans at both global and response level.}
    \label{det:second}
    \vspace*{\fill}
\end{figure}
\begin{figure}[p]
    \vspace*{\fill}
    \centering
    \begin{subfigure}{0.8\textwidth}
        \centering
        \includegraphics[width=\linewidth]{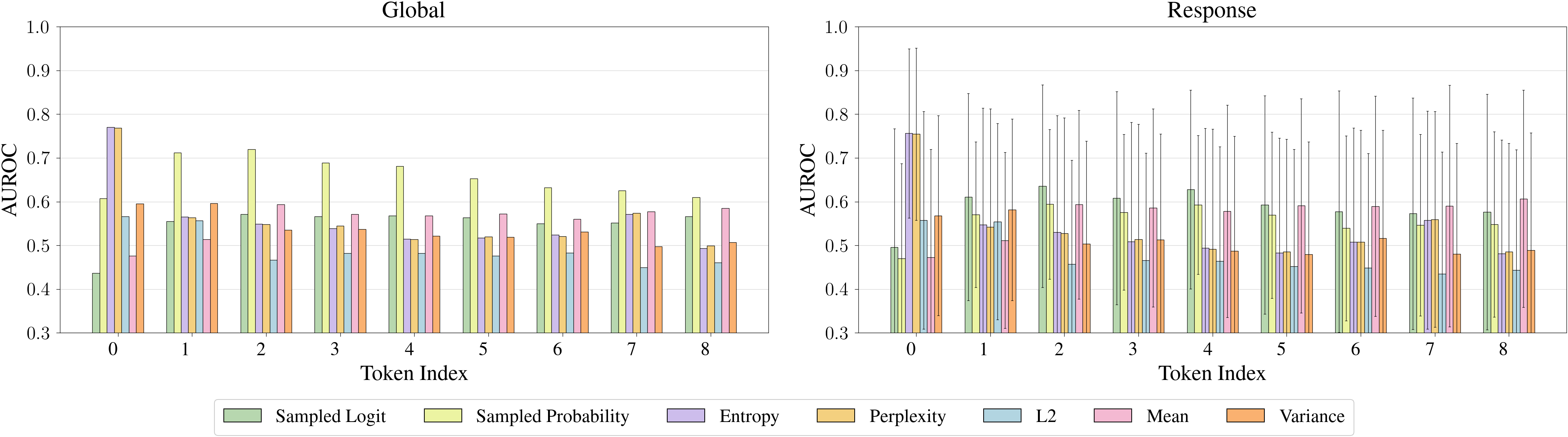}
        \subcaption{LLaMA-2-7B-chat}
        \label{det:third+:llama-7b}
    \end{subfigure}
    \medskip
    \begin{subfigure}{0.8\textwidth}
        \centering
        \includegraphics[width=\linewidth]{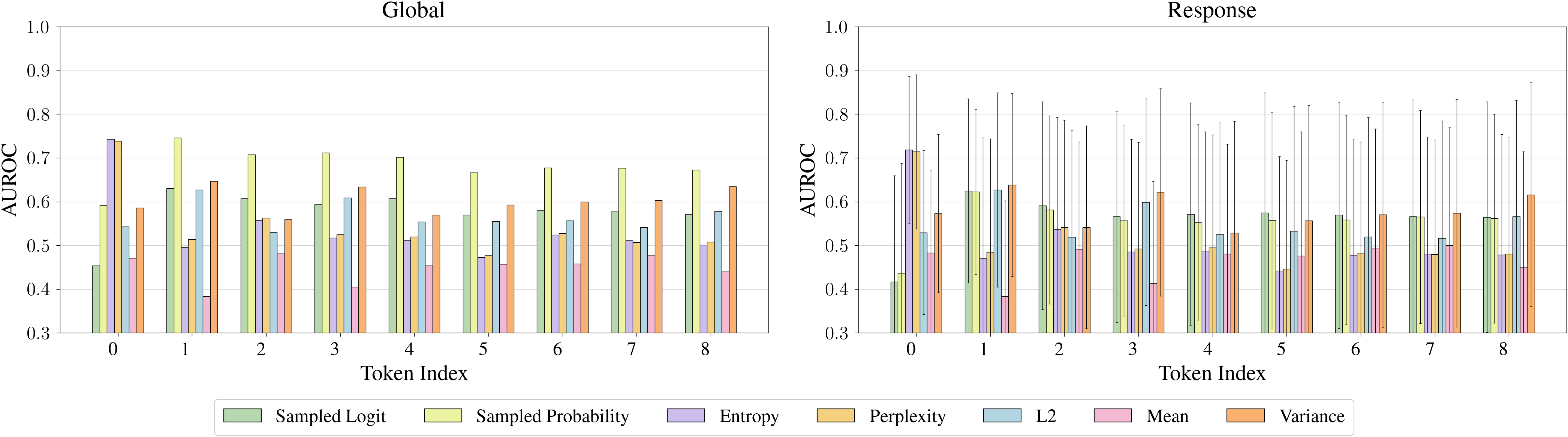}
        \subcaption{LLaMA-2-13B-chat}
        \label{det:third+:llama-13b}
    \end{subfigure}
    \medskip
    \begin{subfigure}{0.8\textwidth}
        \centering
        \includegraphics[width=\linewidth]{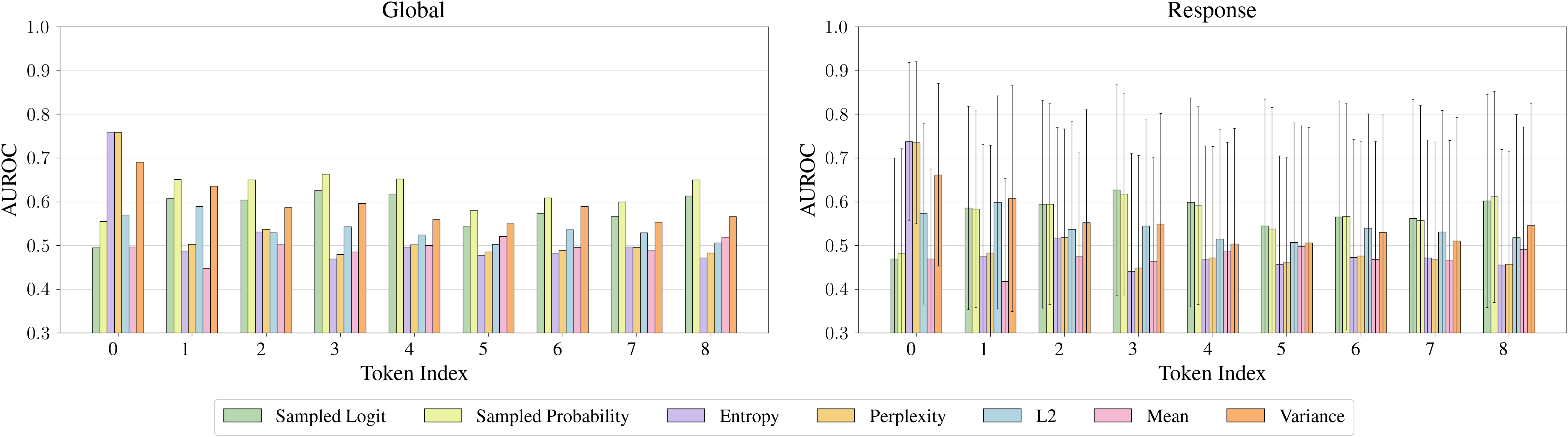}
        \subcaption{LLaMA-2-70B-chat}
        \label{det:third+:llama-70b}
    \end{subfigure}
    \medskip
    \begin{subfigure}{0.8\textwidth}
        \centering
        \includegraphics[width=\linewidth]{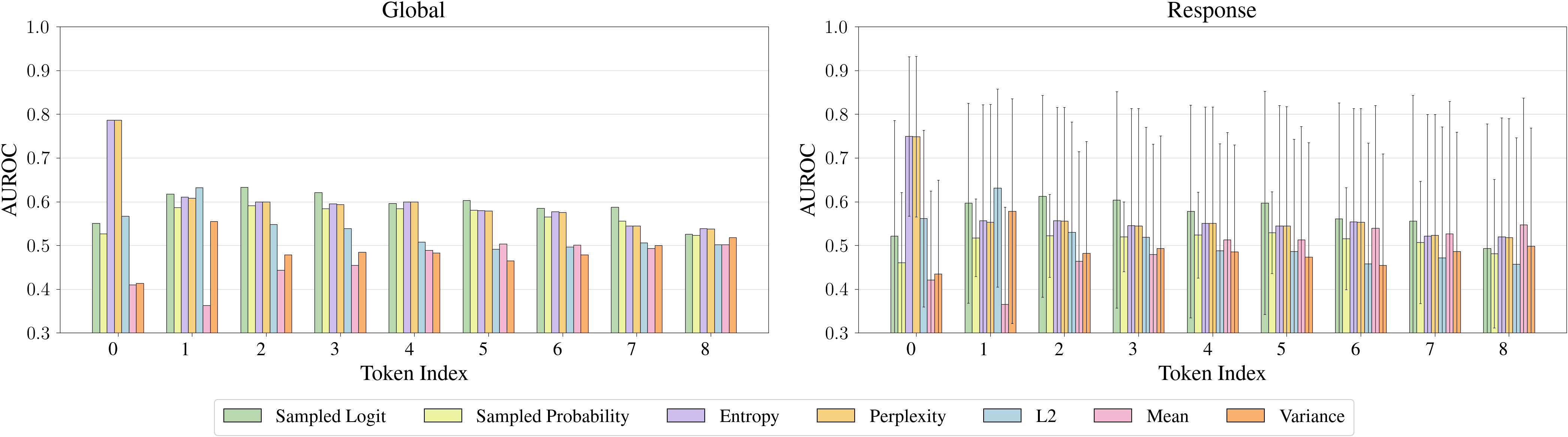}
        \subcaption{Mistral-7B-instruct}
        \label{det:third+:mistral-7b}
    \end{subfigure}
    \caption{\textbf{[third+]} AUROC per signal and in-span hallucination token indices from third+ hallucination spans at both global and response level.}
    \label{det:third+}
    \vspace*{\fill}
\end{figure}
\FloatBarrier

\subsection{Separability}
\label{plt:mink:sep}
\subsubsection{Min-K Probability}
\label{plt:mink:sep:prob}
\begin{figure}[!hp]
    \vspace*{\fill}
    \centering
    \begin{subfigure}{0.49\textwidth}
        \centering
        \includegraphics[width=\linewidth]{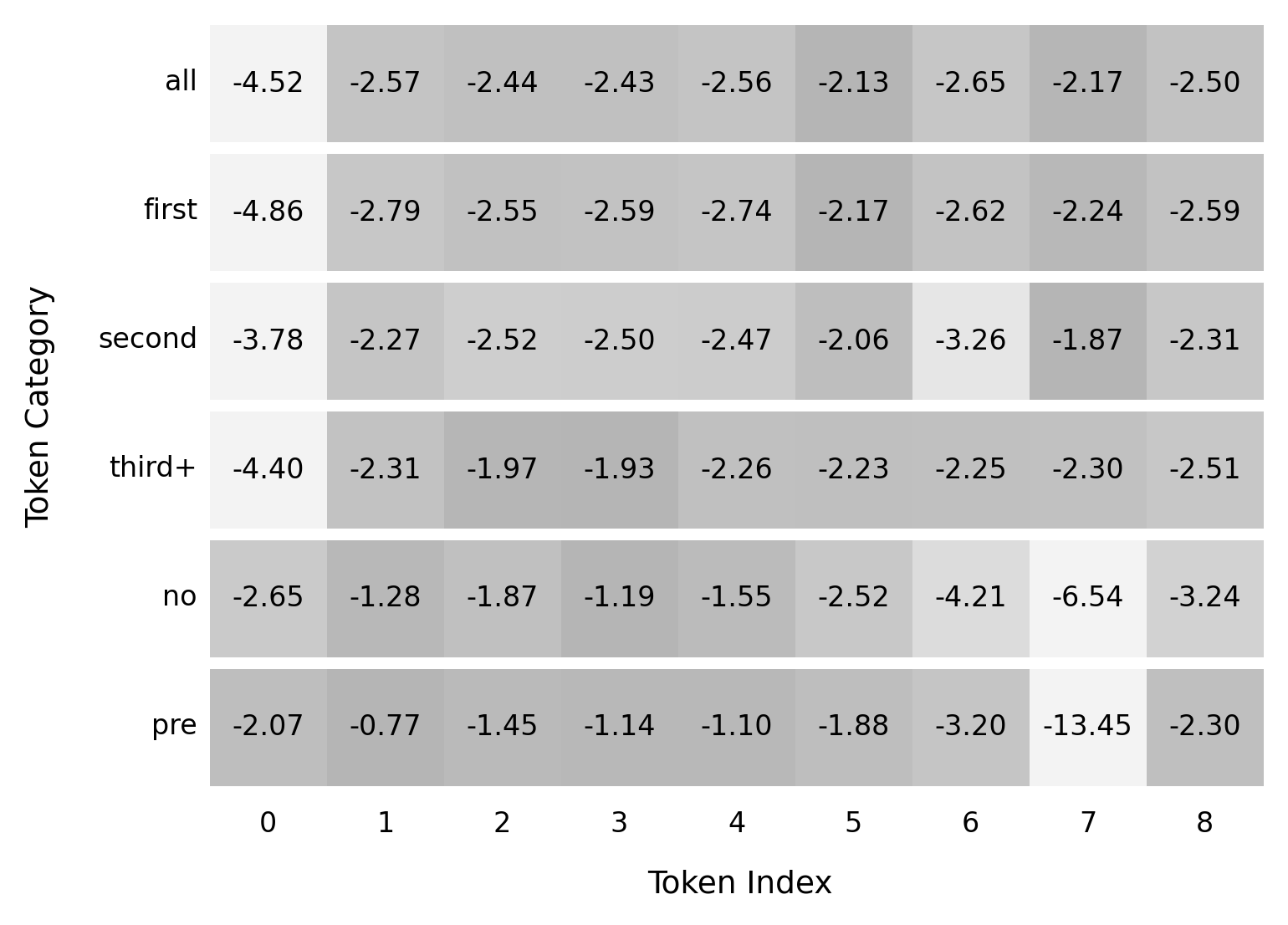}
        \caption{LLaMA-2-7B-chat}
        \label{mink:cat:p:10:llama-7b}
    \end{subfigure}
    \hfill
    \begin{subfigure}{0.49\textwidth}
        \centering
        \includegraphics[width=\linewidth]{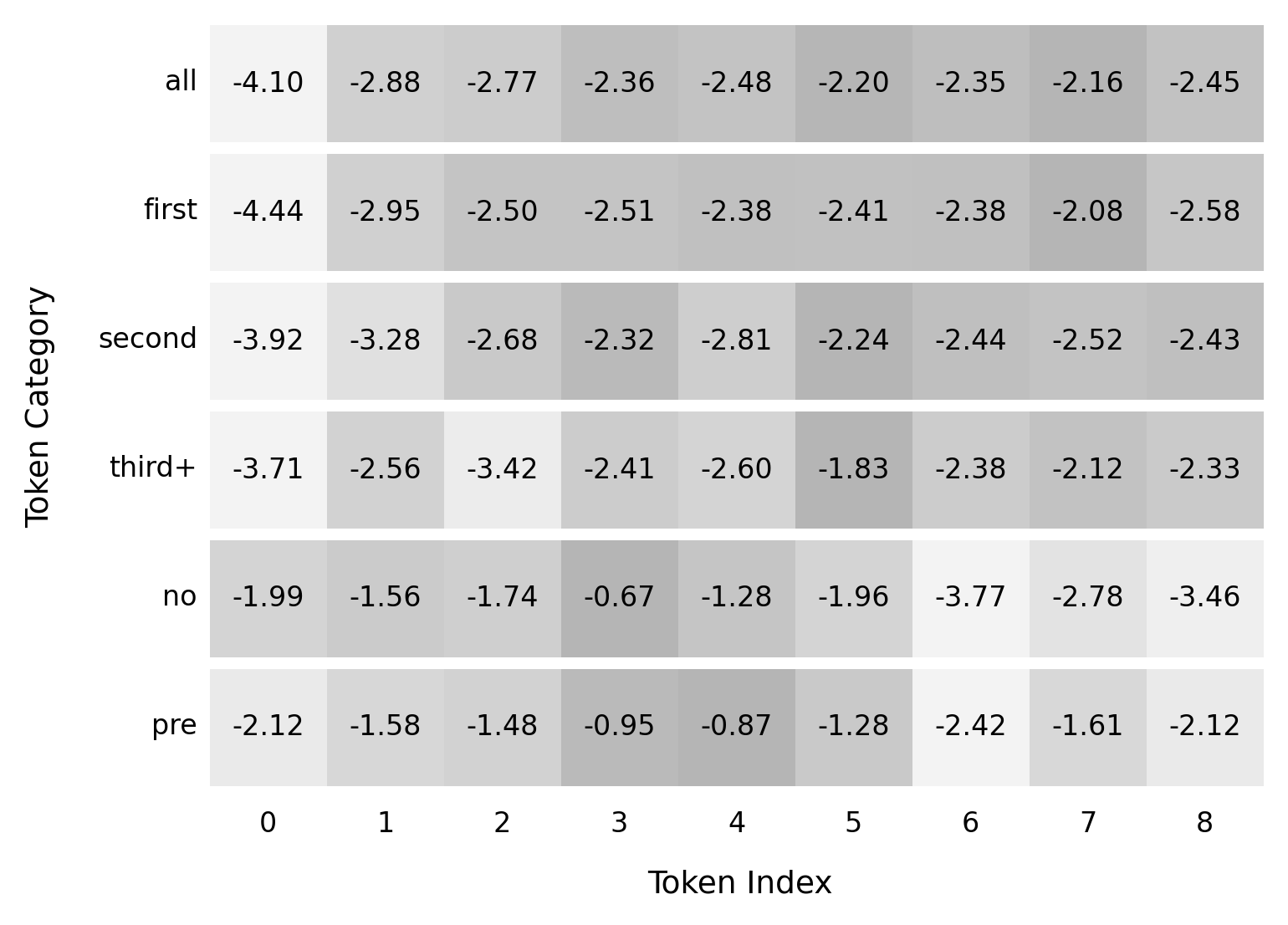}
        \caption{LLaMA-2-13B-chat}
        \label{mink:cat:p:10:llama-13b}
    \end{subfigure}
    \medskip
    \begin{subfigure}{0.49\textwidth}
        \centering
        \includegraphics[width=\linewidth]{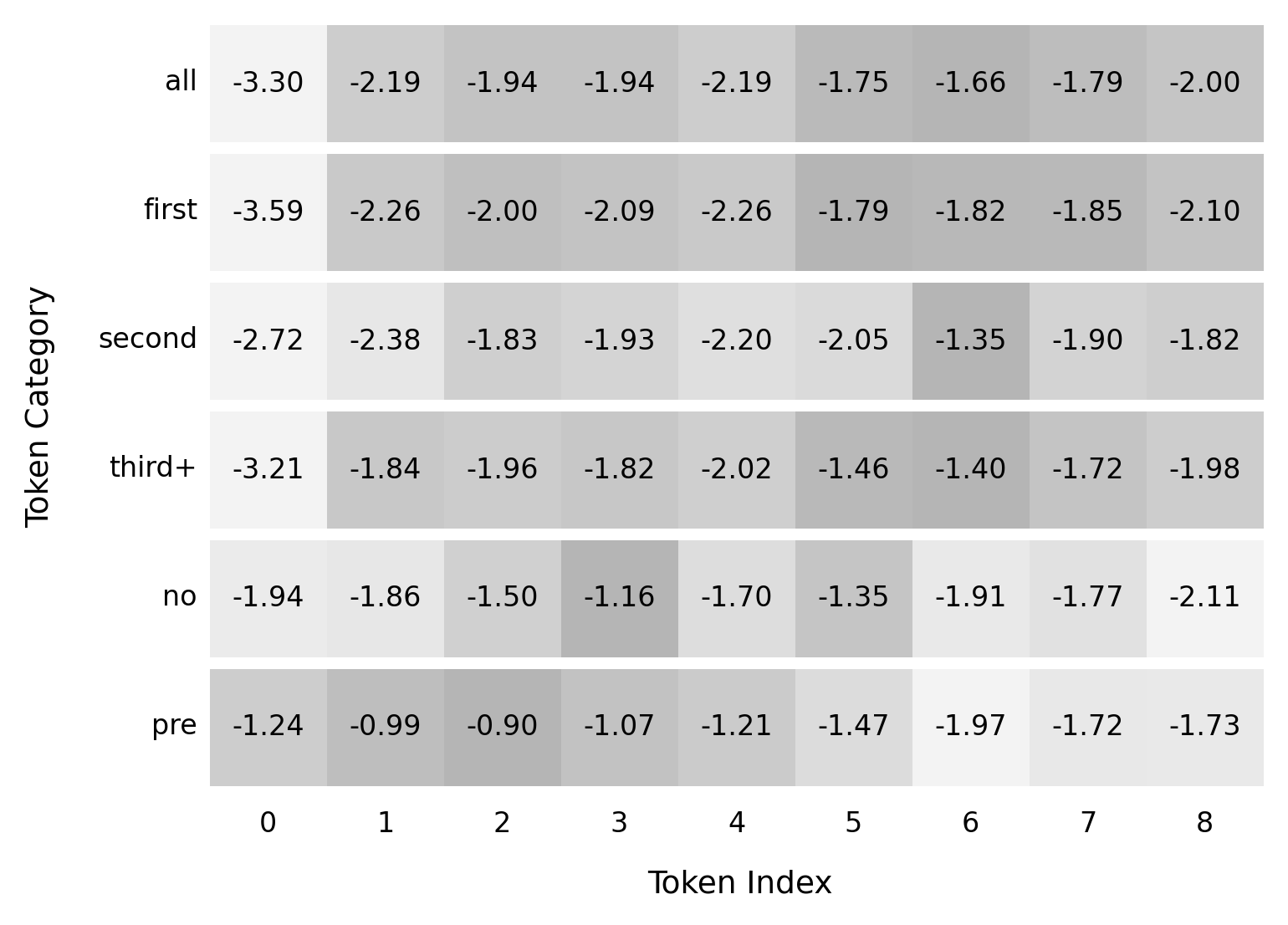}
        \caption{LLaMA-2-70B-chat}
        \label{mink:cat:p:10:llama-70b}
    \end{subfigure}
    \hfill
    \begin{subfigure}{0.49\textwidth}
        \centering
        \includegraphics[width=\linewidth]{plots/mink/categories/llama-2-70b-chat/mink_categories_10.png}
        \caption{Mistral-7B-instruct}
        \label{mink:cat:p:10:mistral-7b}
    \end{subfigure}
    \caption{\textbf{[10th percentile]} Min-K Probability scores per token category and index, over the first 9 tokens at global level.}
    \label{mink:cat:p:10}
    \vspace*{\fill}
\end{figure}

\begin{figure}[p]
    \centering
    \begin{subfigure}{0.49\textwidth}
        \centering
        \includegraphics[width=\linewidth]{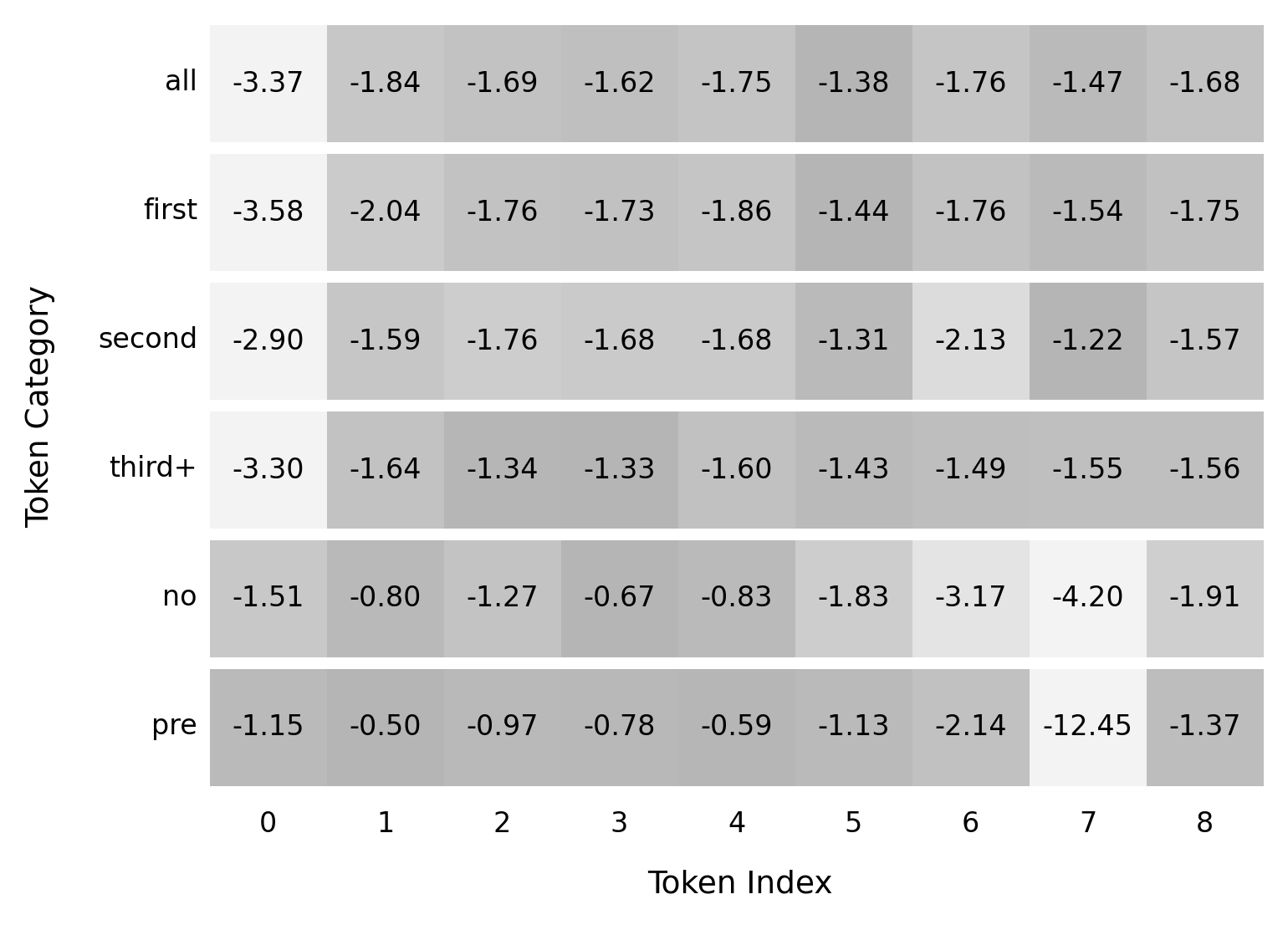}
        \caption{LLaMA-2-7B-chat}
        \label{mink:cat:p:20:llama-7b}
    \end{subfigure}
    \hfill
    \begin{subfigure}{0.49\textwidth}
        \centering
        \includegraphics[width=\linewidth]{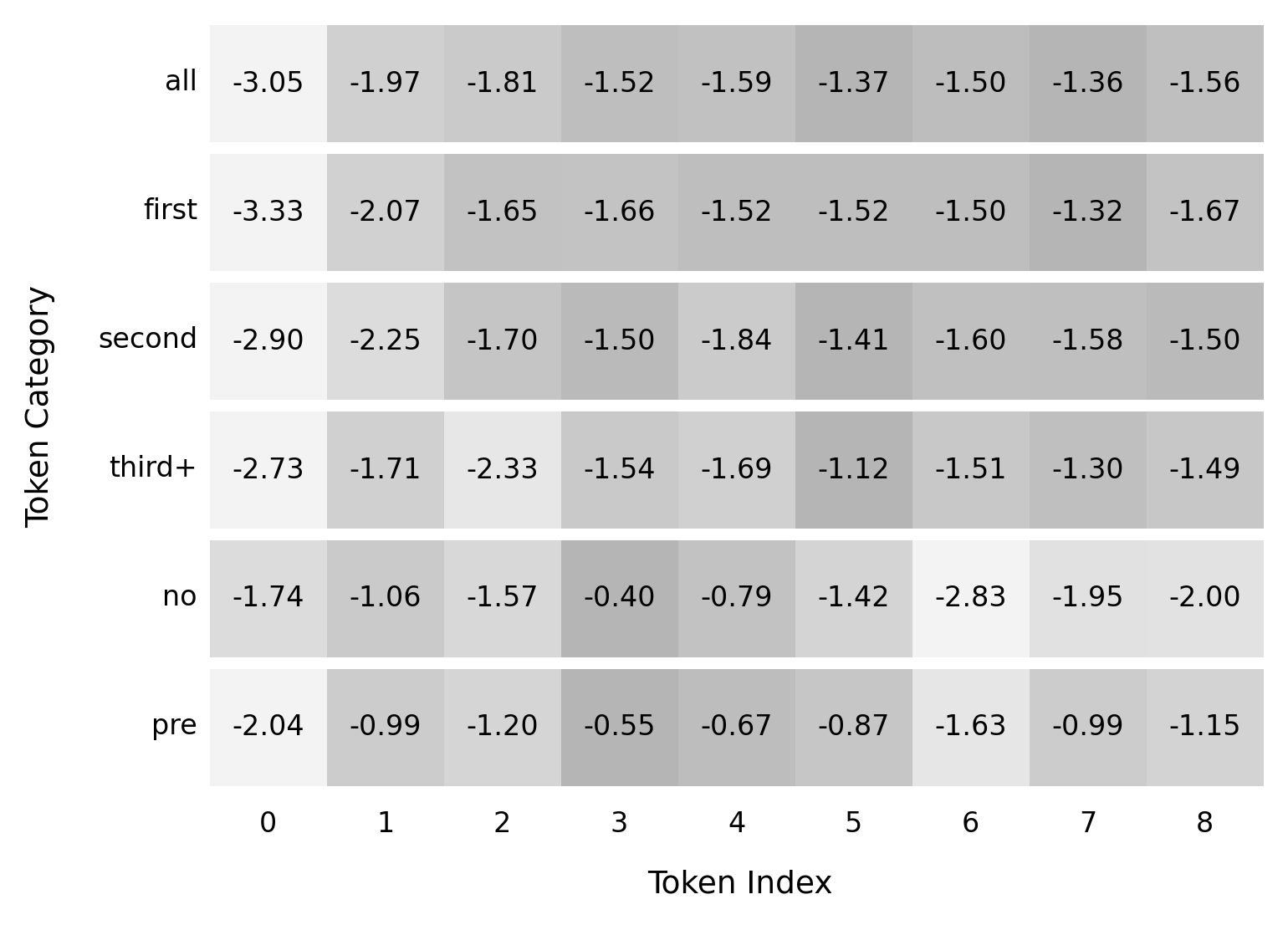}
        \caption{LLaMA-2-13B-chat}
        \label{mink:cat:p:20:llama-13b}
    \end{subfigure}
    \medskip
    \begin{subfigure}{0.49\textwidth}
        \centering
        \includegraphics[width=\linewidth]{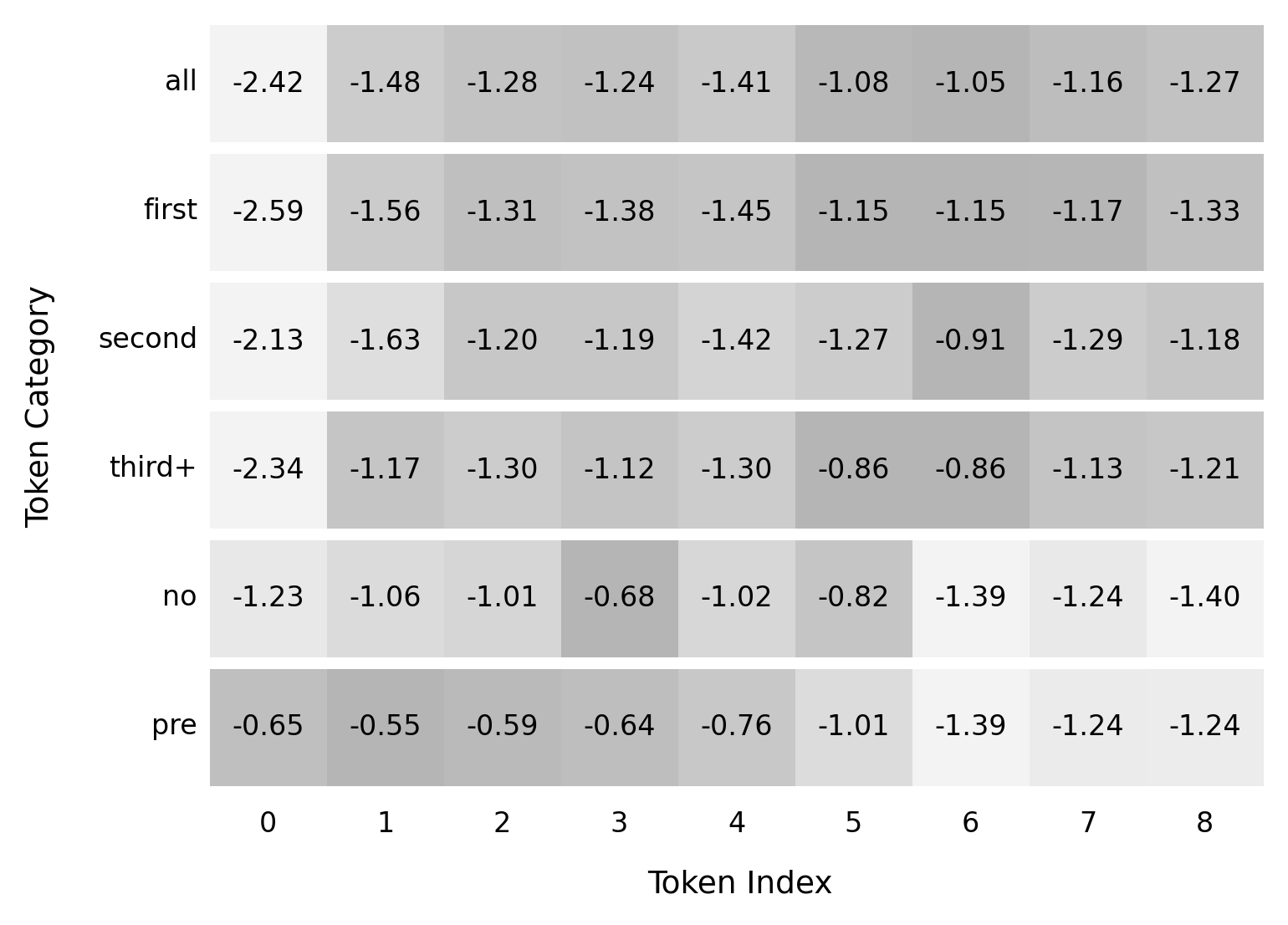}
        \caption{LLaMA-2-70B-chat}
        \label{mink:cat:p:20:llama-70b}
    \end{subfigure}
    \hfill
    \begin{subfigure}{0.49\textwidth}
        \centering
        \includegraphics[width=\linewidth]{plots/mink/categories/llama-2-70b-chat/mink_categories_20.png}
        \caption{Mistral-7B-instruct}
        \label{mink:cat:p:20:mistral-7b}
    \end{subfigure}
    \caption{\textbf{[20th percentile]} Min-K Probability scores per token category and index, over the first 9 tokens at global level.}
    \label{mink:cat:p:20}
\end{figure}

\begin{figure}[p]
    \centering
    \begin{subfigure}{0.49\textwidth}
        \centering
        \includegraphics[width=\linewidth]{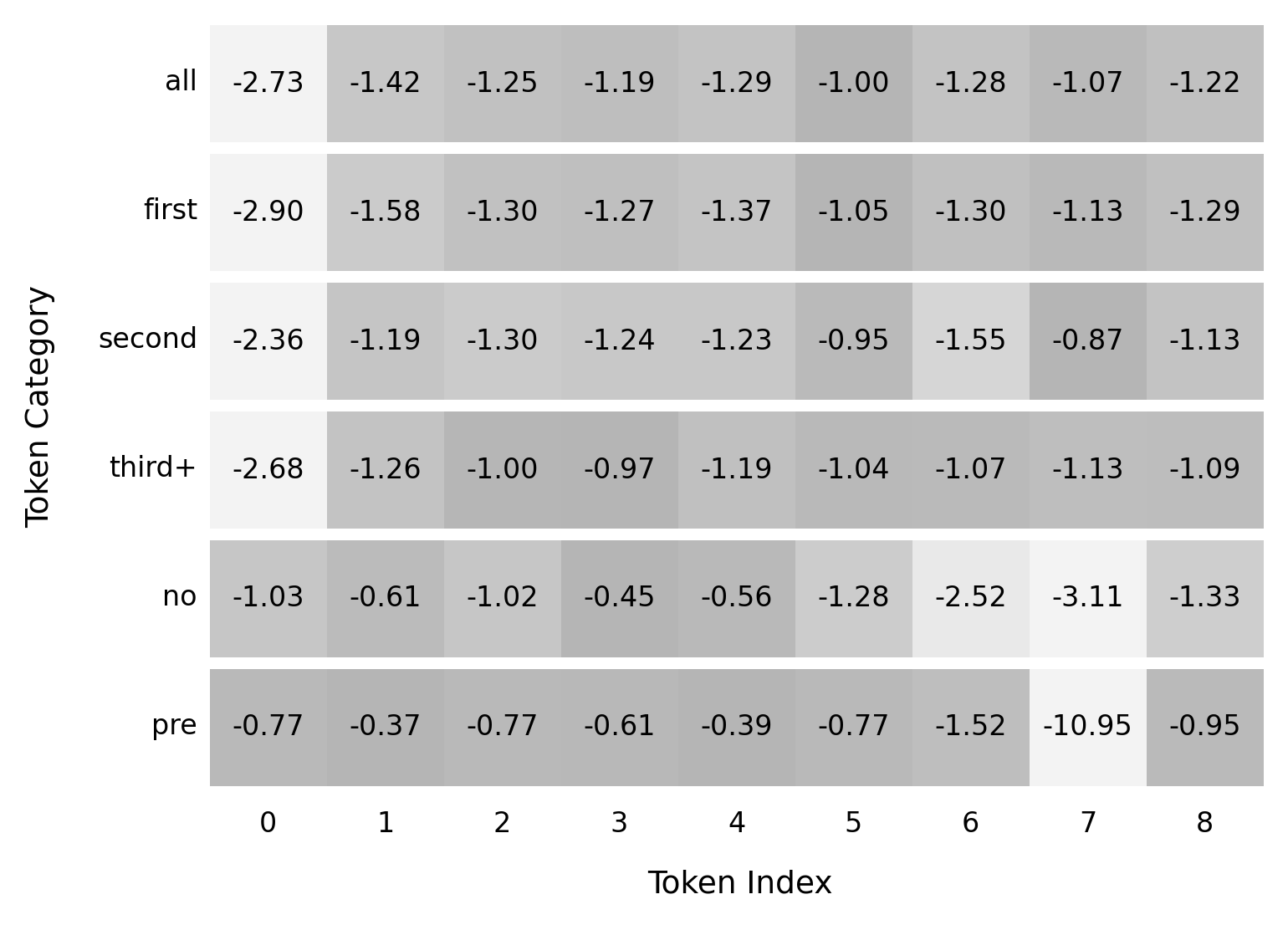}
        \caption{LLaMA-2-7B-chat}
        \label{mink:cat:p:30:llama-7b}
    \end{subfigure}
    \hfill
    \begin{subfigure}{0.49\textwidth}
        \centering
        \includegraphics[width=\linewidth]{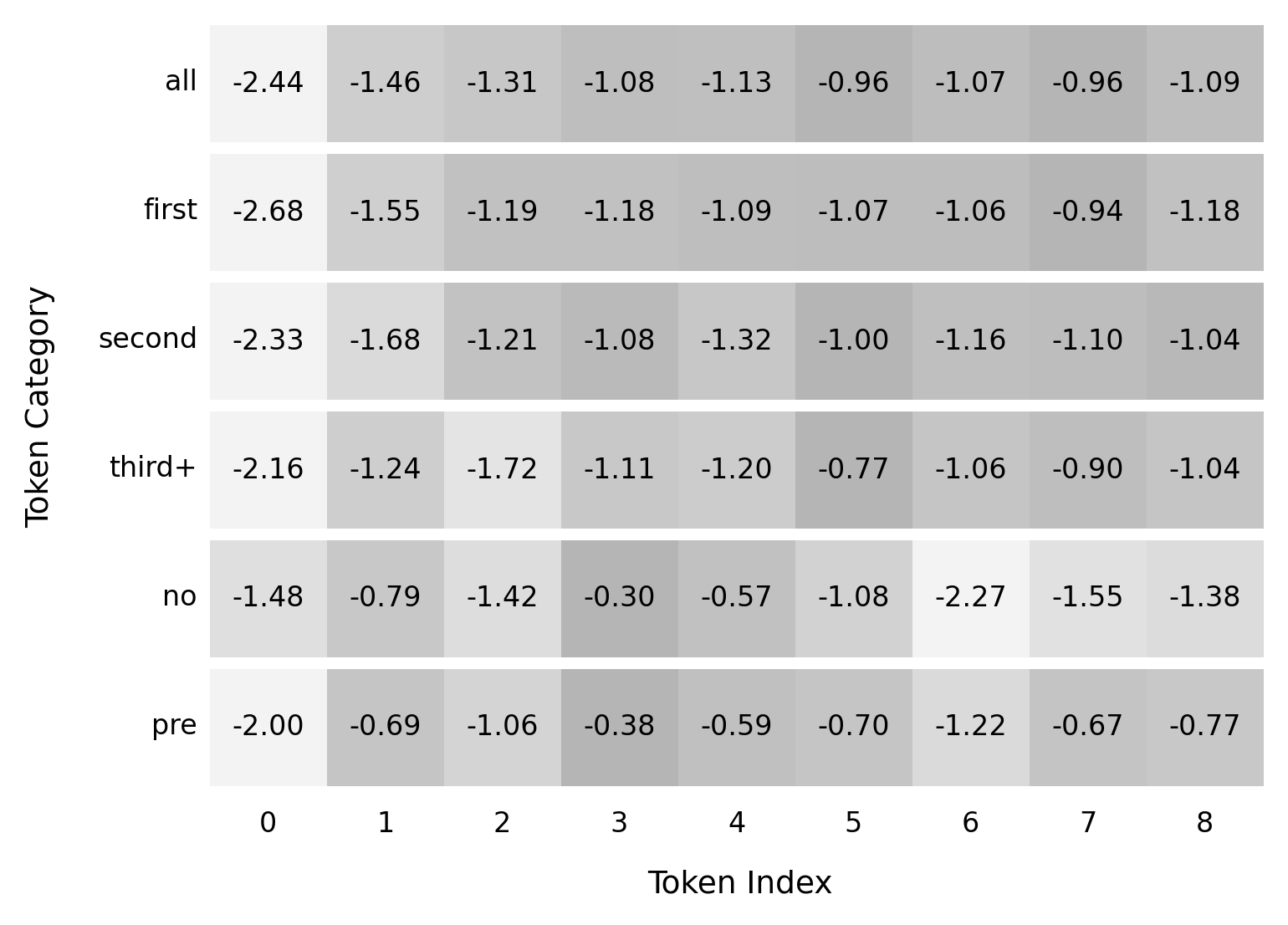}
        \caption{LLaMA-2-13B-chat}
        \label{mink:cat:p:30:llama-13b}
    \end{subfigure}
    \medskip
    \begin{subfigure}{0.49\textwidth}
        \centering
        \includegraphics[width=\linewidth]{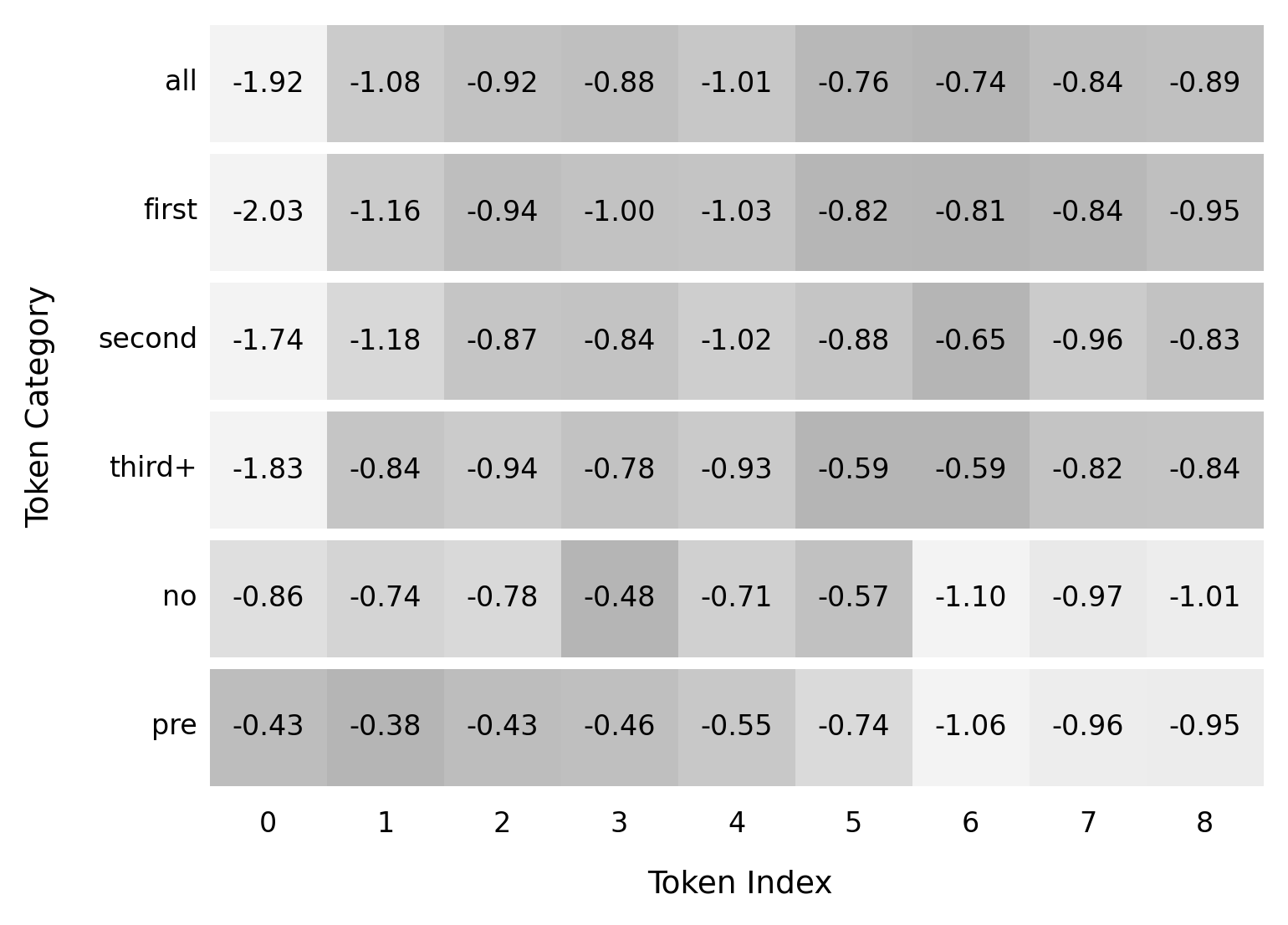}
        \caption{LLaMA-2-70B-chat}
        \label{mink:cat:p:30:llama-70b}
    \end{subfigure}
    \hfill
    \begin{subfigure}{0.49\textwidth}
        \centering
        \includegraphics[width=\linewidth]{plots/mink/categories/llama-2-70b-chat/mink_categories_30.png}
        \caption{Mistral-7B-instruct}
        \label{mink:cat:p:30:mistral-7b}
    \end{subfigure}
    \caption{\textbf{[30th percentile]} Min-K Probability scores per token category and index, over the first 9 tokens at global level.}
    \label{mink:cat:p:20}
\end{figure}

\begin{figure}[p]
    \centering
    \begin{subfigure}{0.49\textwidth}
        \centering
        \includegraphics[width=\linewidth]{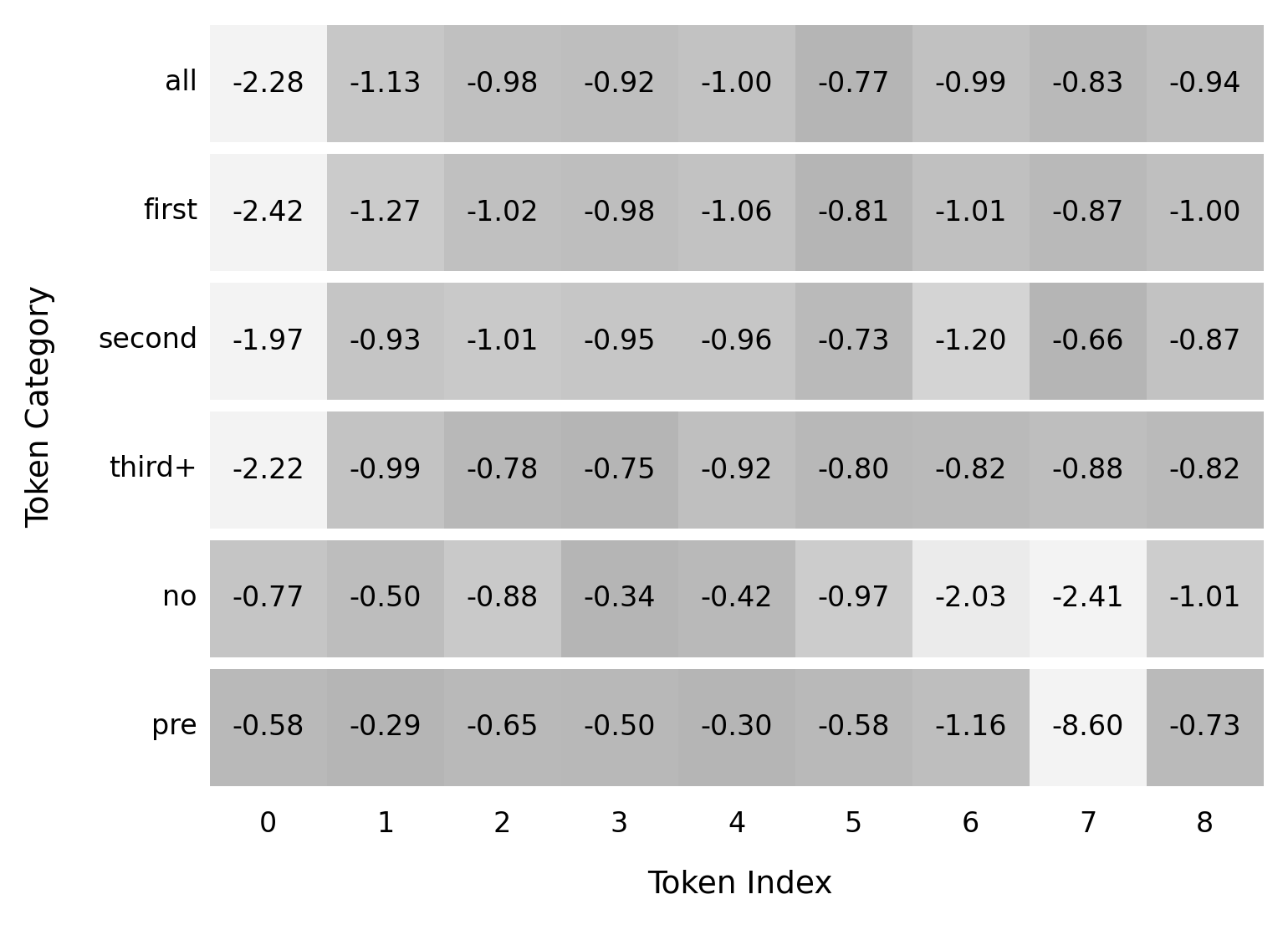}
        \caption{LLaMA-2-7B-chat}
        \label{mink:cat:p:40:llama-7b}
    \end{subfigure}
    \hfill
    \begin{subfigure}{0.49\textwidth}
        \centering
        \includegraphics[width=\linewidth]{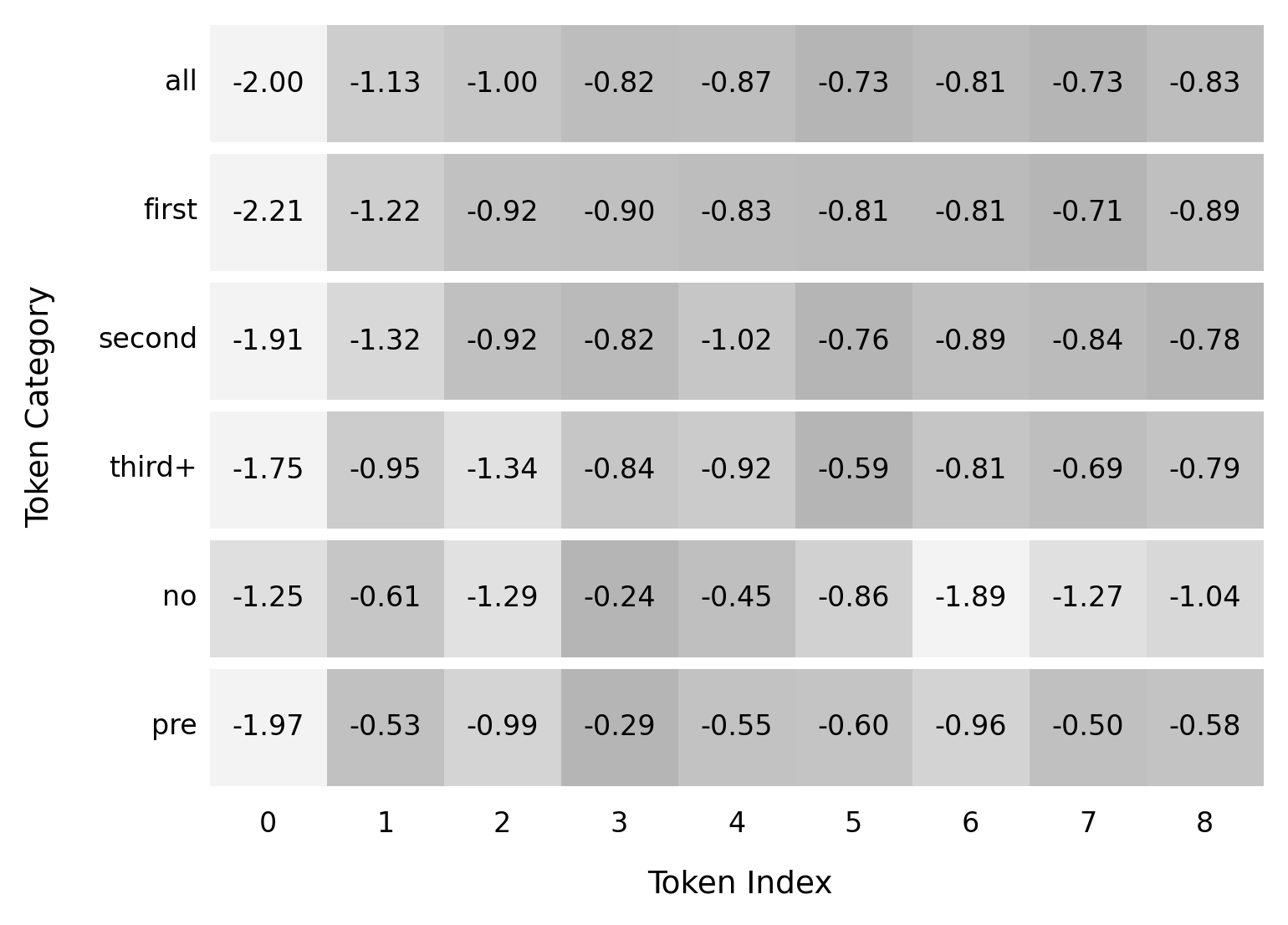}
        \caption{LLaMA-2-13B-chat}
        \label{mink:cat:p:40:llama-13b}
    \end{subfigure}
    \medskip
    \begin{subfigure}{0.49\textwidth}
        \centering
        \includegraphics[width=\linewidth]{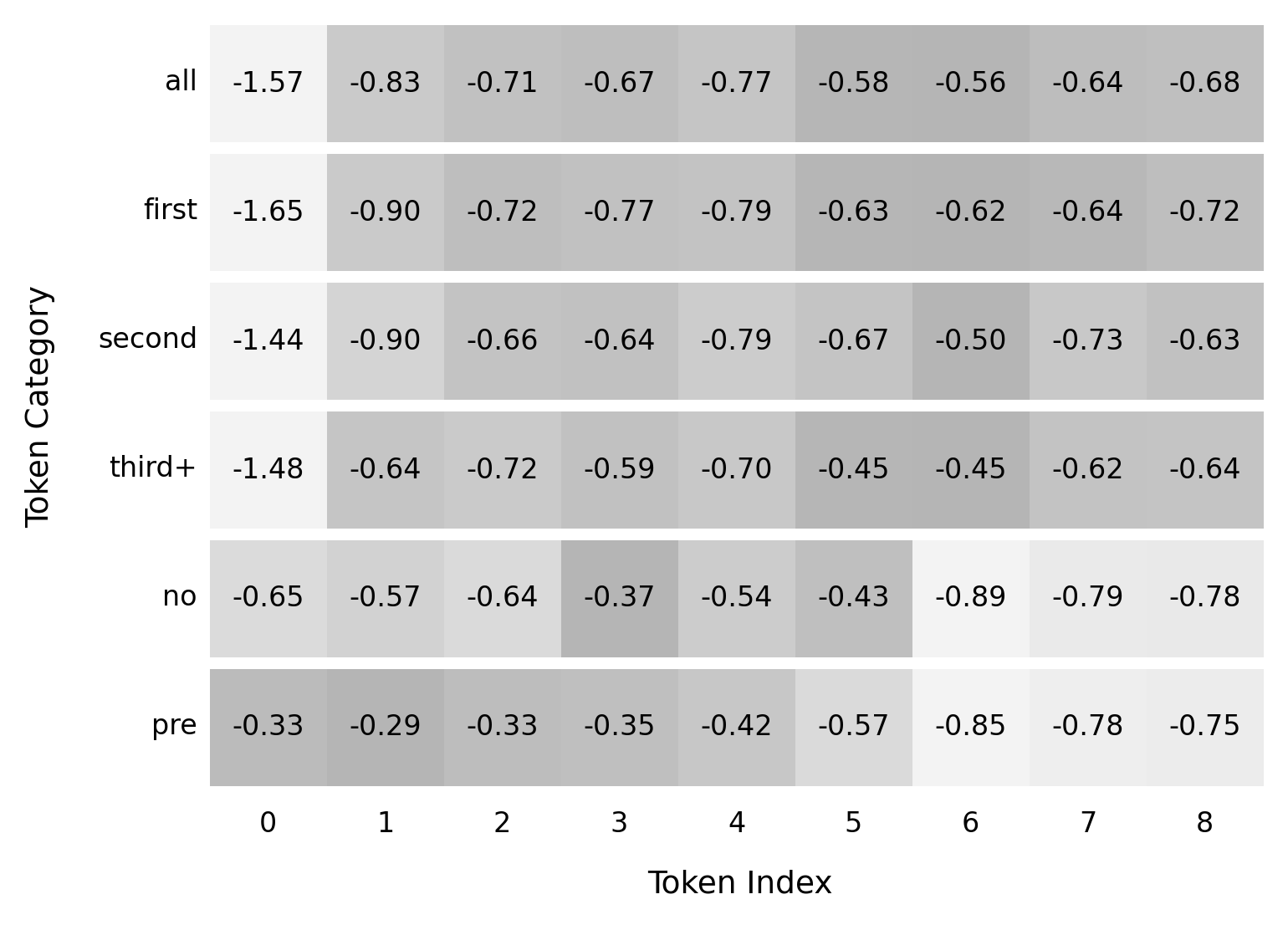}
        \caption{LLaMA-2-70B-chat}
        \label{mink:cat:p:40:llama-70b}
    \end{subfigure}
    \hfill
    \begin{subfigure}{0.49\textwidth}
        \centering
        \includegraphics[width=\linewidth]{plots/mink/categories/llama-2-70b-chat/mink_categories_40.png}
        \caption{Mistral-7B-instruct}
        \label{mink:cat:p:40:mistral-7b}
    \end{subfigure}
    \caption{\textbf{[40th percentile]} Min-K Probability scores per token category and index, over the first 9 tokens at global level.}
    \label{mink:cat:p:40}
\end{figure}

\begin{figure}[p]
    \centering
    \begin{subfigure}{0.49\textwidth}
        \centering
        \includegraphics[width=\linewidth]{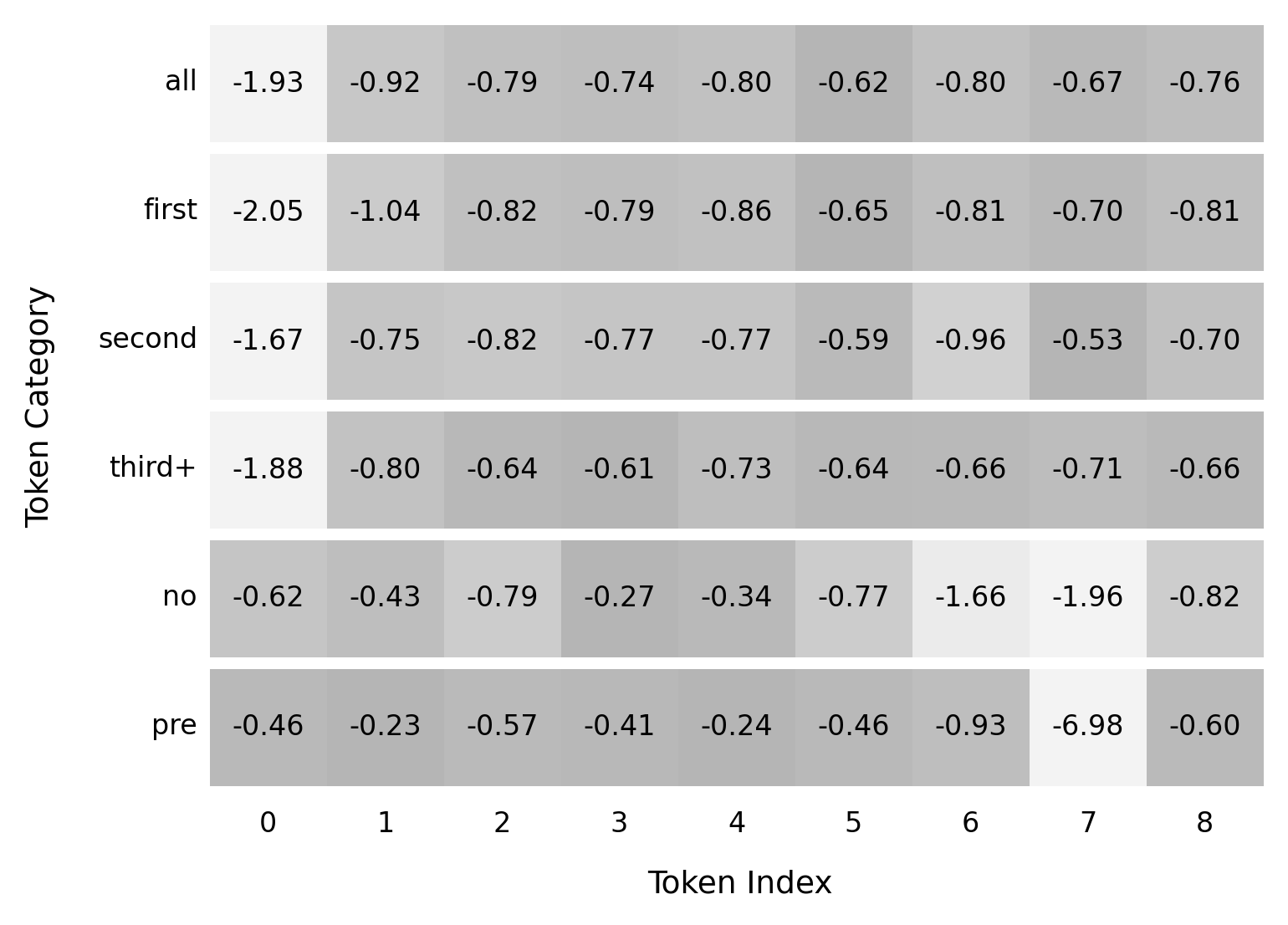}
        \caption{LLaMA-2-7B-chat}
        \label{mink:cat:p:50:llama-7b}
    \end{subfigure}
    \hfill
    \begin{subfigure}{0.49\textwidth}
        \centering
        \includegraphics[width=\linewidth]{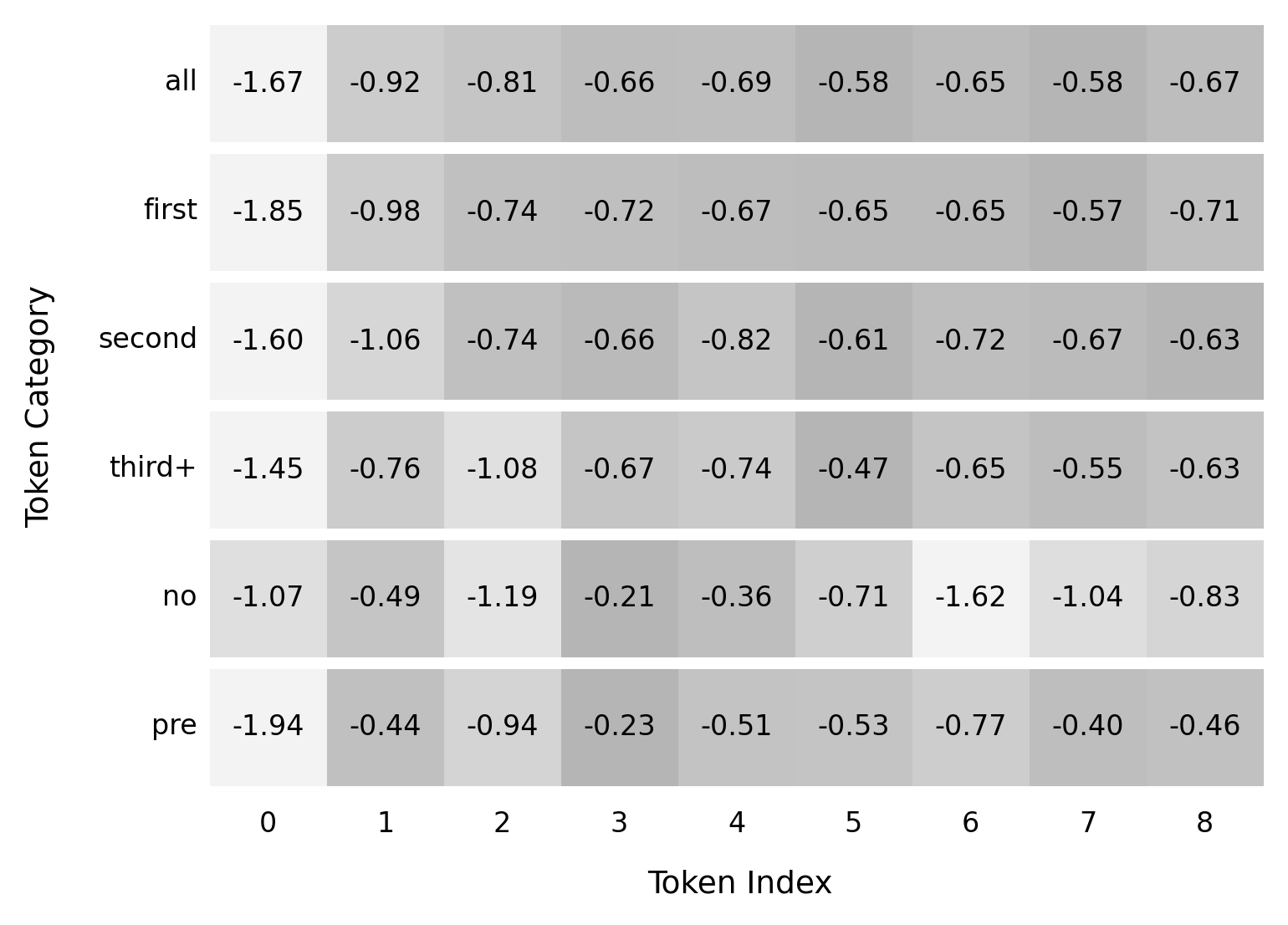}
        \caption{LLaMA-2-13B-chat}
        \label{mink:cat:p:50:llama-13b}
    \end{subfigure}
    \medskip
    \begin{subfigure}{0.49\textwidth}
        \centering
        \includegraphics[width=\linewidth]{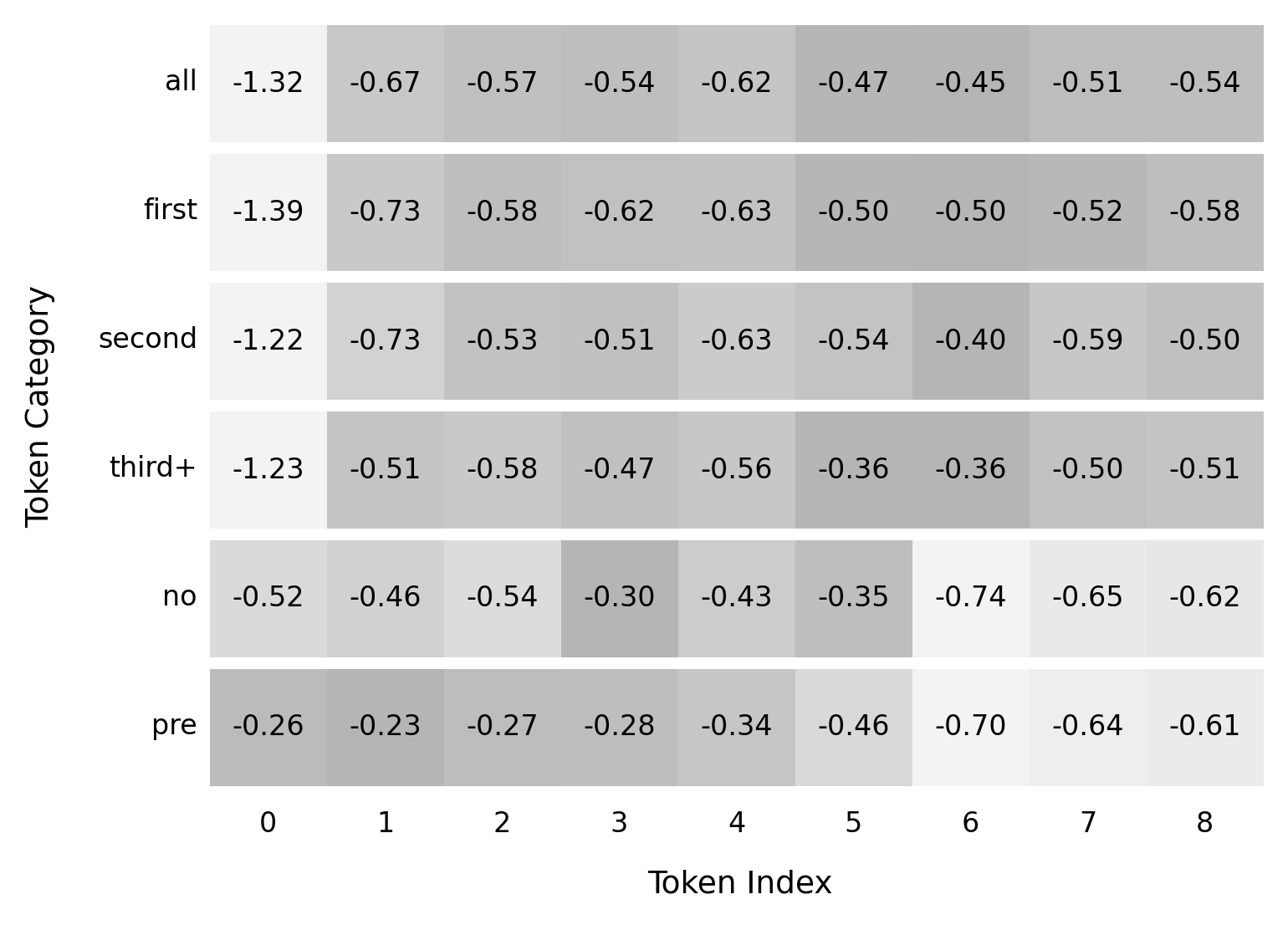}
        \caption{LLaMA-2-70B-chat}
        \label{mink:cat:p:50:llama-70b}
    \end{subfigure}
    \hfill
    \begin{subfigure}{0.49\textwidth}
        \centering
        \includegraphics[width=\linewidth]{plots/mink/categories/llama-2-70b-chat/mink_categories_50.png}
        \caption{Mistral-7B-instruct}
        \label{mink:cat:p:50:mistral-7b}
    \end{subfigure}
    \caption{\textbf{[50th percentile]} Min-K Probability scores per token category and index, over the first 9 tokens at global level.}
    \label{mink:cat:p:50}
\end{figure}

\begin{figure}[p]
    \centering
    \begin{subfigure}{0.49\textwidth}
        \centering
        \includegraphics[width=\linewidth]{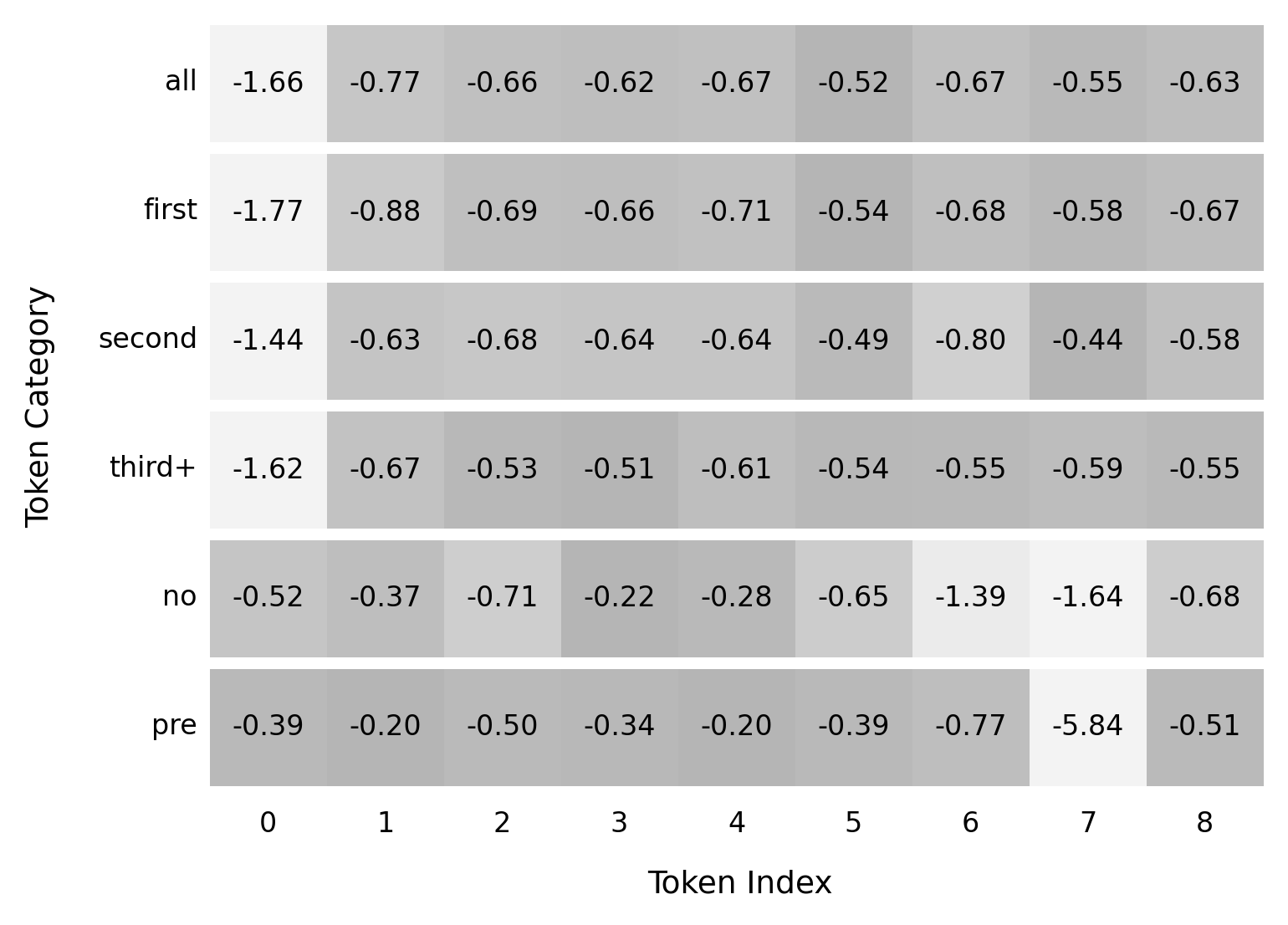}
        \caption{LLaMA-2-7B-chat}
        \label{mink:cat:p:60:llama-7b}
    \end{subfigure}
    \hfill
    \begin{subfigure}{0.49\textwidth}
        \centering
        \includegraphics[width=\linewidth]{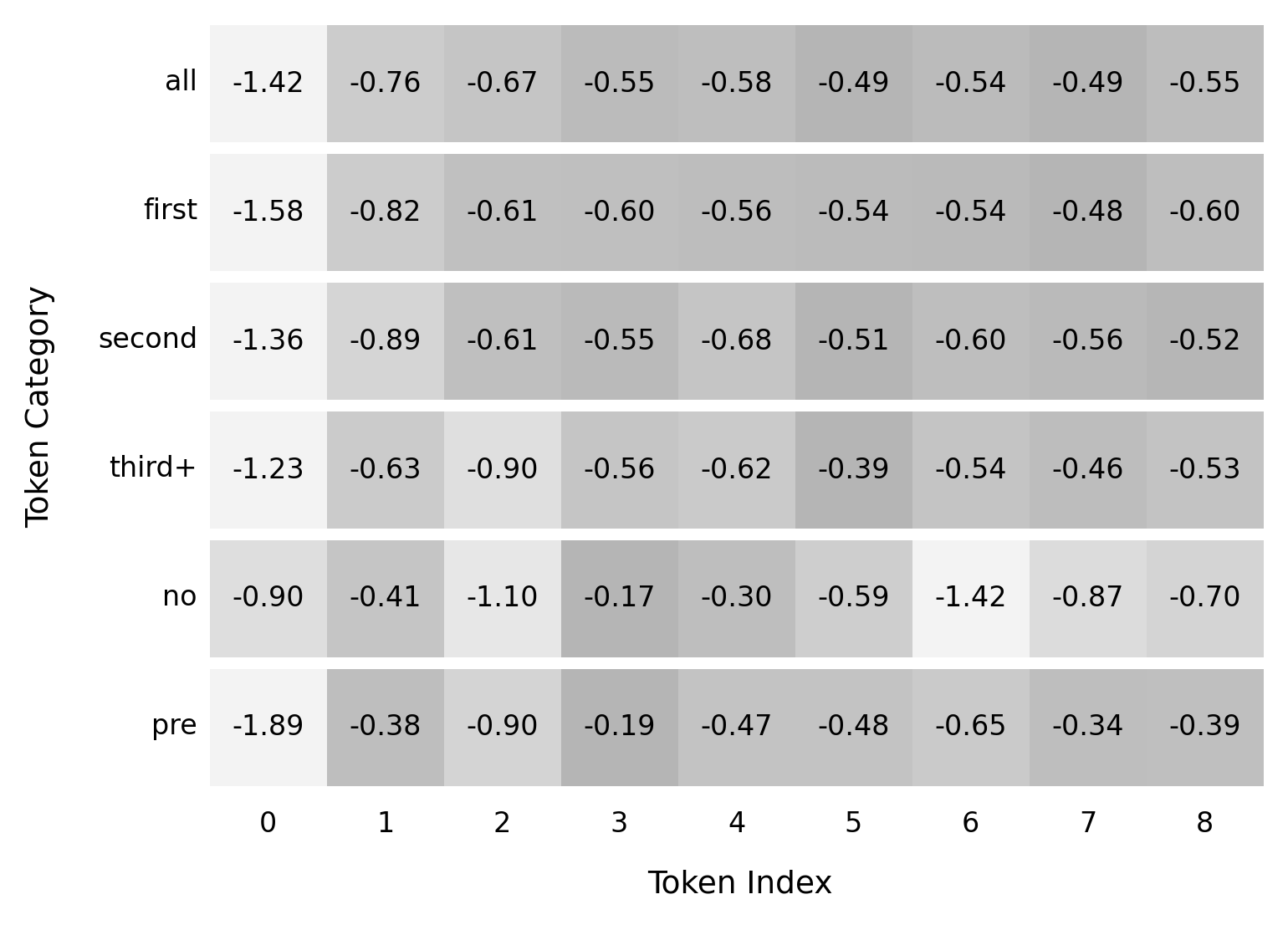}
        \caption{LLaMA-2-13B-chat}
        \label{mink:cat:p:60:llama-13b}
    \end{subfigure}
    \medskip
    \begin{subfigure}{0.49\textwidth}
        \centering
        \includegraphics[width=\linewidth]{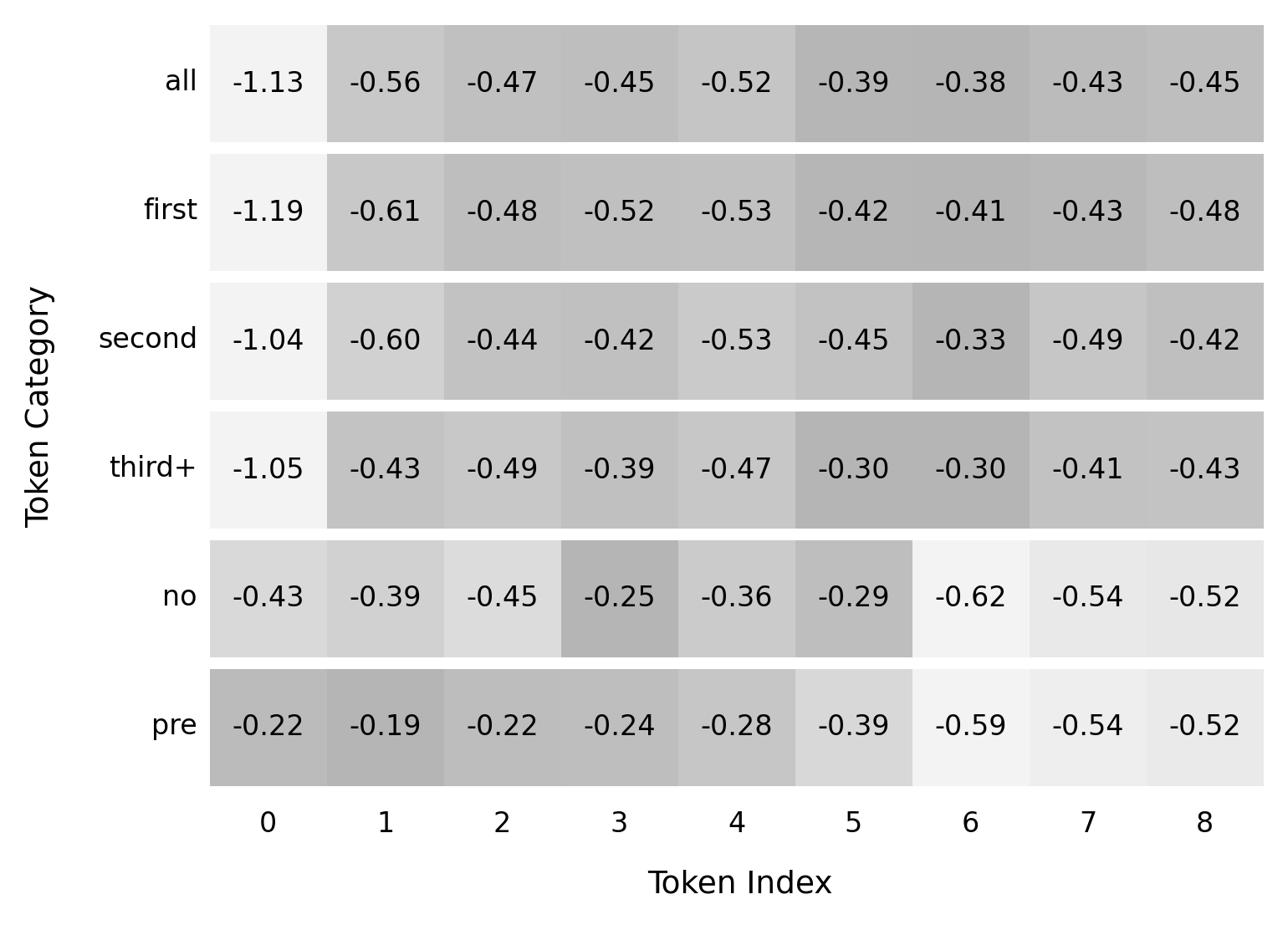}
        \caption{LLaMA-2-70B-chat}
        \label{mink:cat:p:60:llama-70b}
    \end{subfigure}
    \hfill
    \begin{subfigure}{0.49\textwidth}
        \centering
        \includegraphics[width=\linewidth]{plots/mink/categories/llama-2-70b-chat/mink_categories_60.png}
        \caption{Mistral-7B-instruct}
        \label{mink:cat:p:60:mistral-7b}
    \end{subfigure}
    \caption{\textbf{[60th percentile]} Min-K Probability scores per token category and index, over the first 9 tokens at global level.}
    \label{mink:cat:p:60}
\end{figure}

\begin{figure}[p]
    \centering
    \begin{subfigure}{0.49\textwidth}
        \centering
        \includegraphics[width=\linewidth]{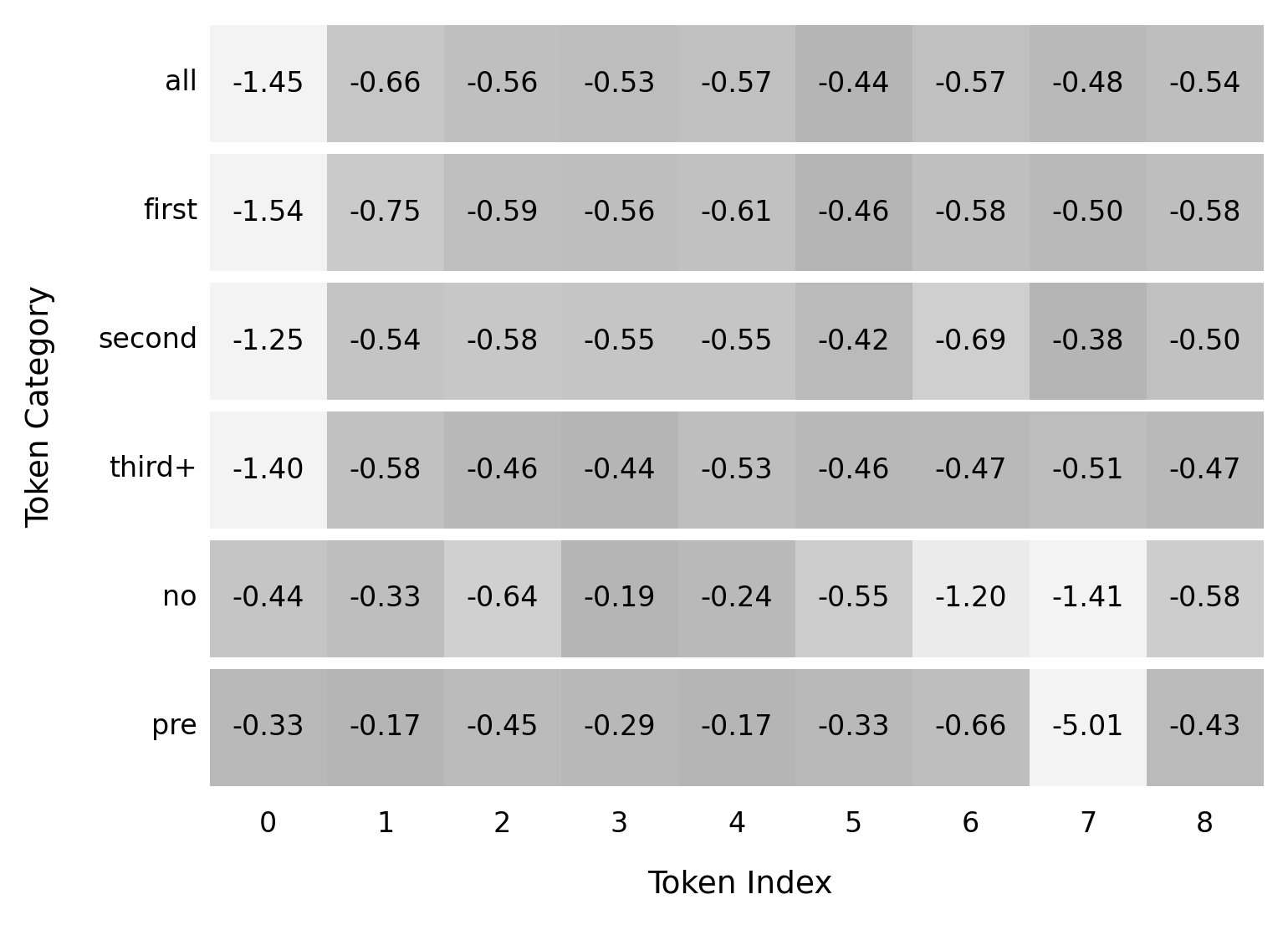}
        \caption{LLaMA-2-7B-chat}
        \label{mink:cat:p:70:llama-7b}
    \end{subfigure}
    \hfill
    \begin{subfigure}{0.49\textwidth}
        \centering
        \includegraphics[width=\linewidth]{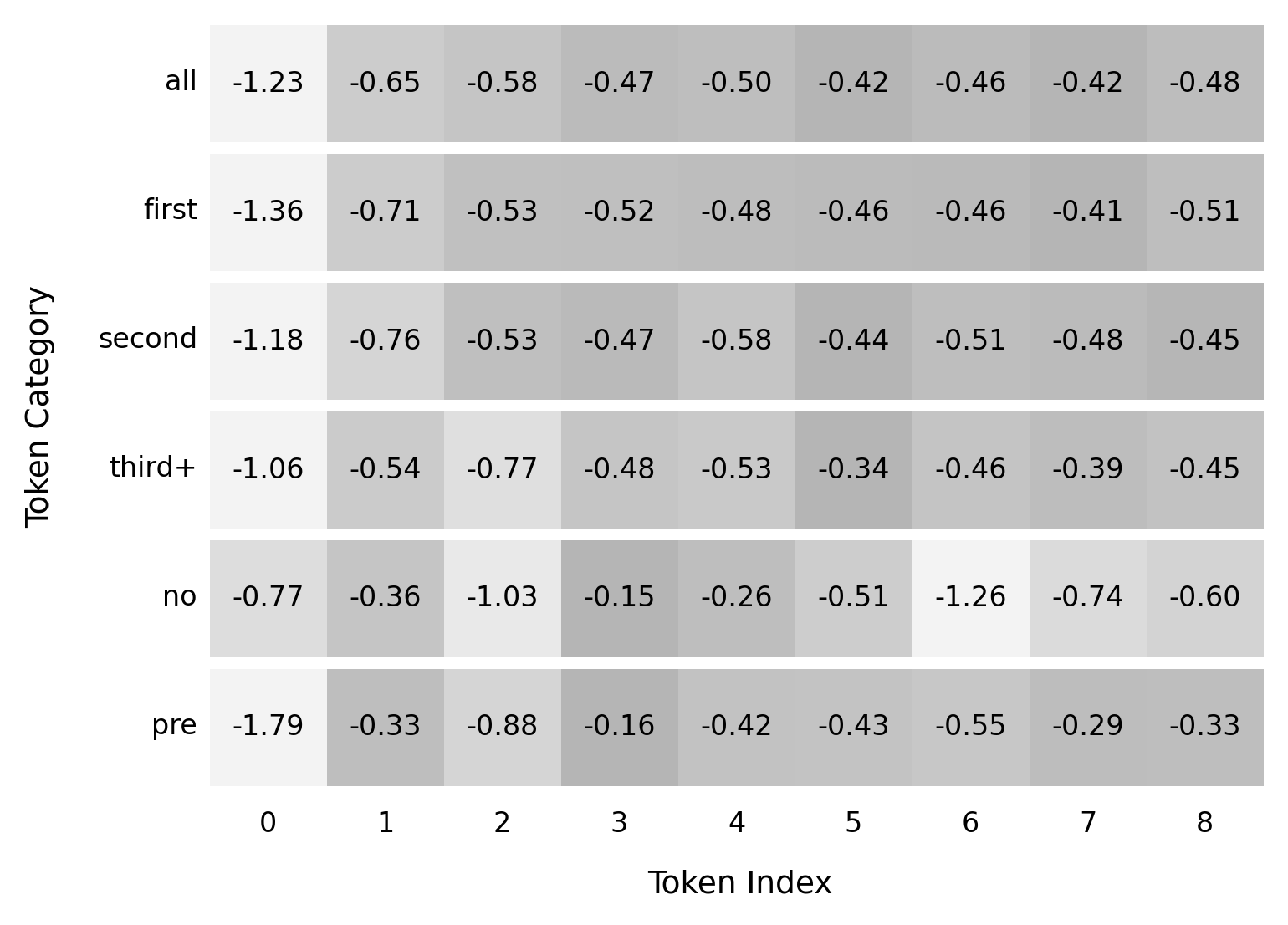}
        \caption{LLaMA-2-13B-chat}
        \label{mink:cat:p:70:llama-13b}
    \end{subfigure}
    \medskip
    \begin{subfigure}{0.49\textwidth}
        \centering
        \includegraphics[width=\linewidth]{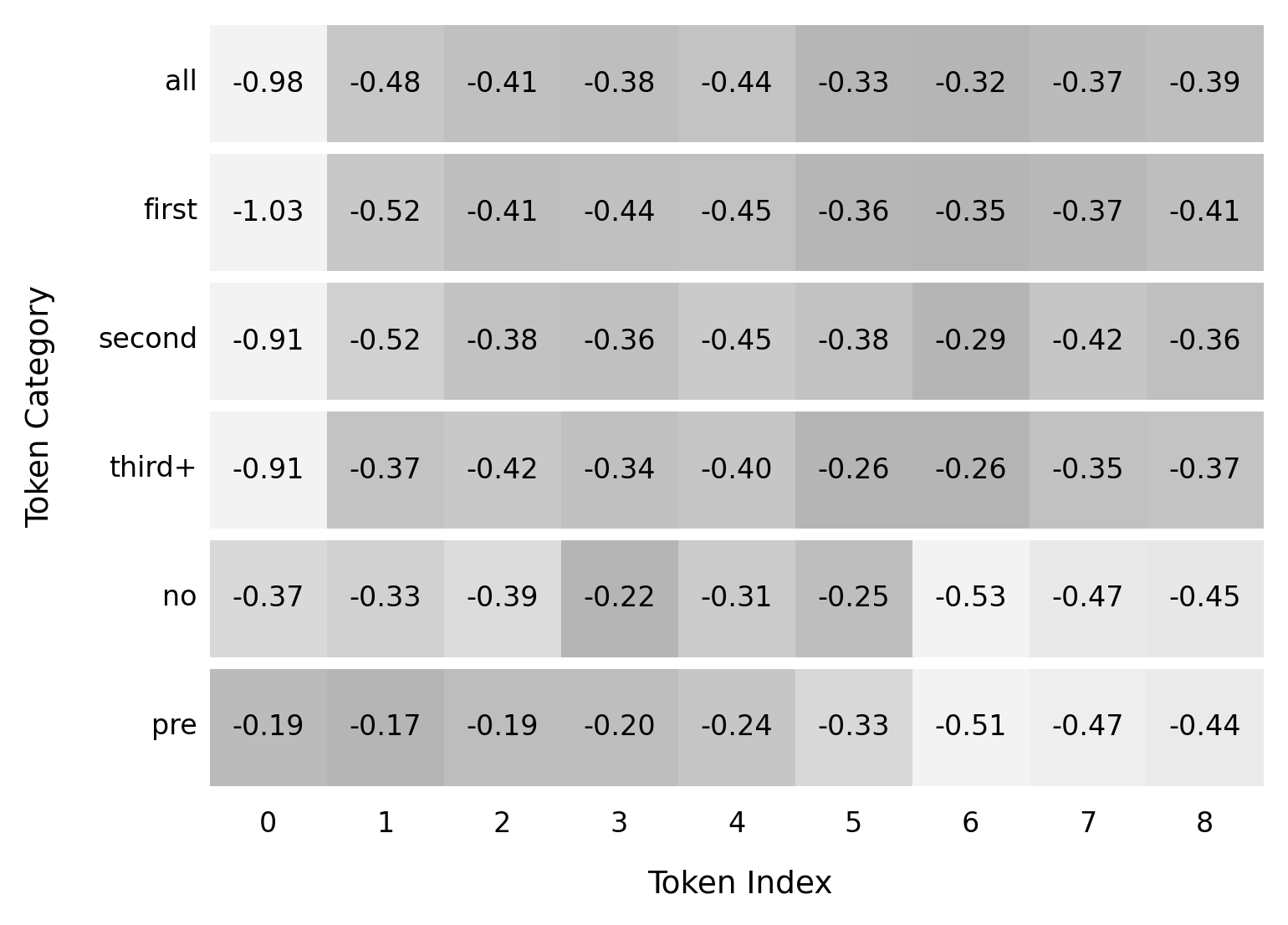}
        \caption{LLaMA-2-70B-chat}
        \label{mink:cat:p:70:llama-70b}
    \end{subfigure}
    \hfill
    \begin{subfigure}{0.49\textwidth}
        \centering
        \includegraphics[width=\linewidth]{plots/mink/categories/llama-2-70b-chat/mink_categories_70.png}
        \caption{Mistral-7B-instruct}
        \label{mink:cat:p:70:mistral-7b}
    \end{subfigure}
    \caption{\textbf{[70th percentile]} Min-K Probability scores per token category and index, over the first 9 tokens at global level.}
    \label{mink:cat:p:70}
\end{figure}

\begin{figure}[p]
    \centering
    \begin{subfigure}{0.49\textwidth}
        \centering
        \includegraphics[width=\linewidth]{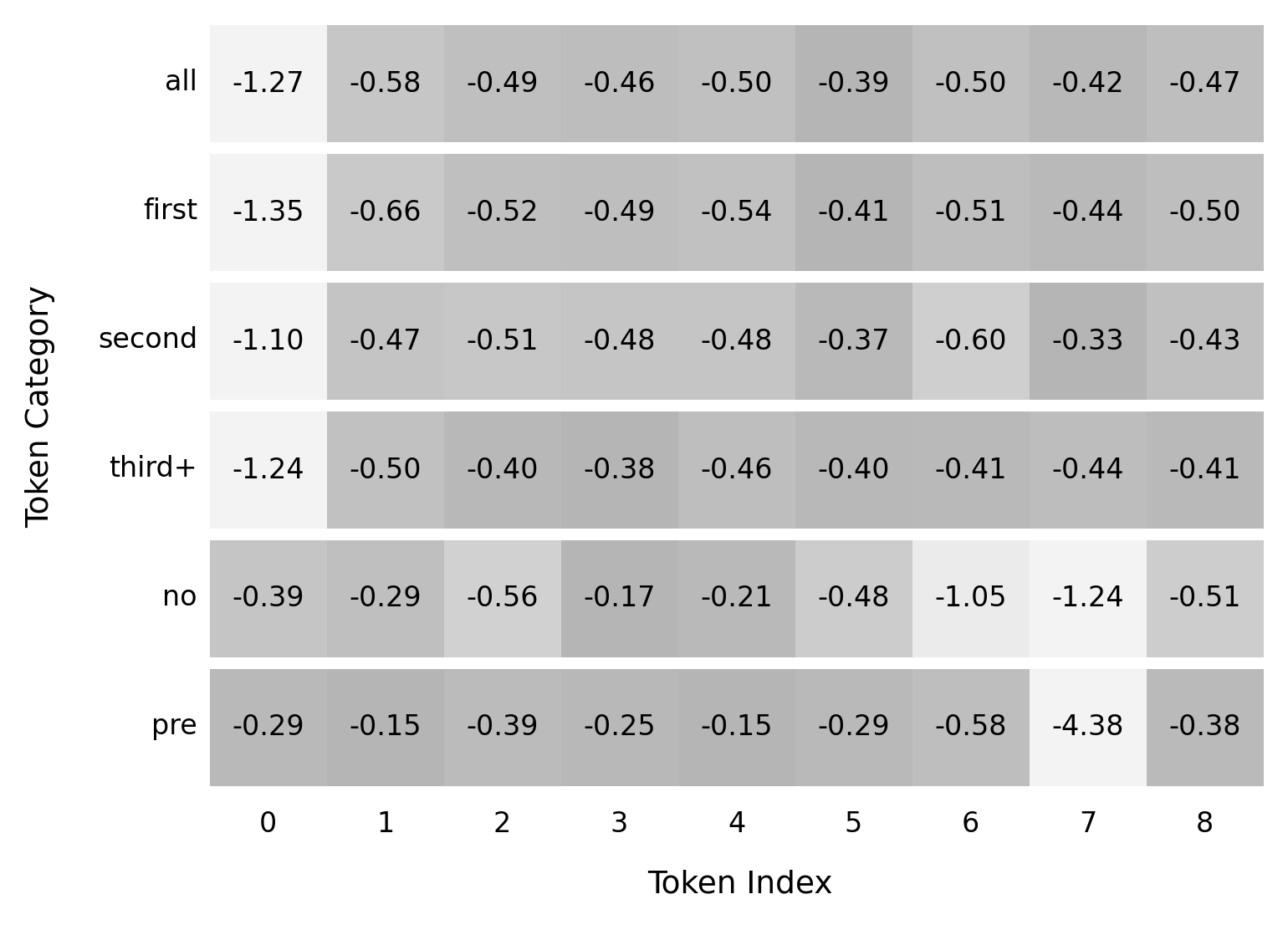}
        \caption{LLaMA-2-7B-chat}
        \label{mink:cat:p:80:llama-7b}
    \end{subfigure}
    \hfill
    \begin{subfigure}{0.49\textwidth}
        \centering
        \includegraphics[width=\linewidth]{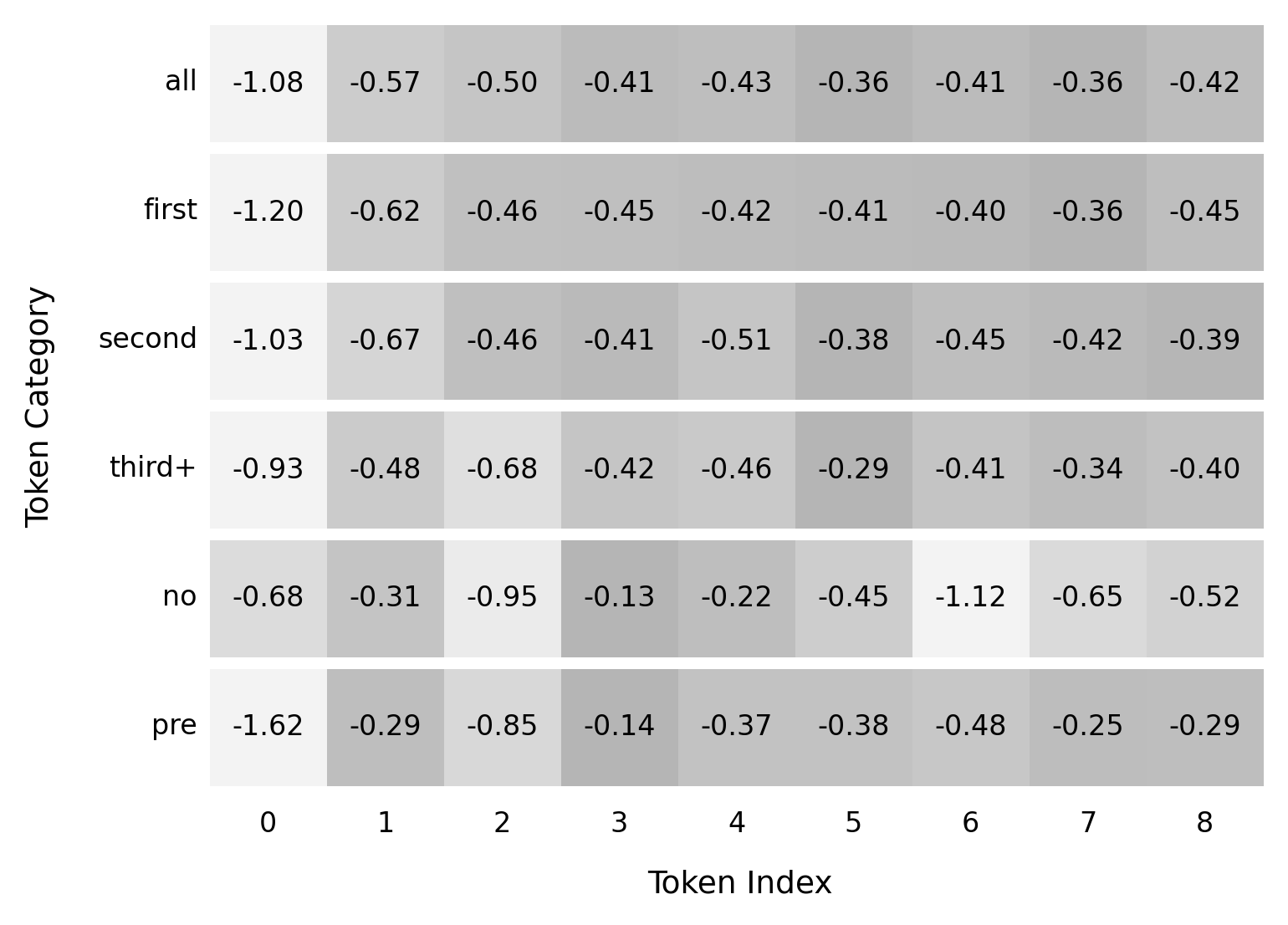}
        \caption{LLaMA-2-13B-chat}
        \label{mink:cat:p:80:llama-13b}
    \end{subfigure}
    \medskip
    \begin{subfigure}{0.49\textwidth}
        \centering
        \includegraphics[width=\linewidth]{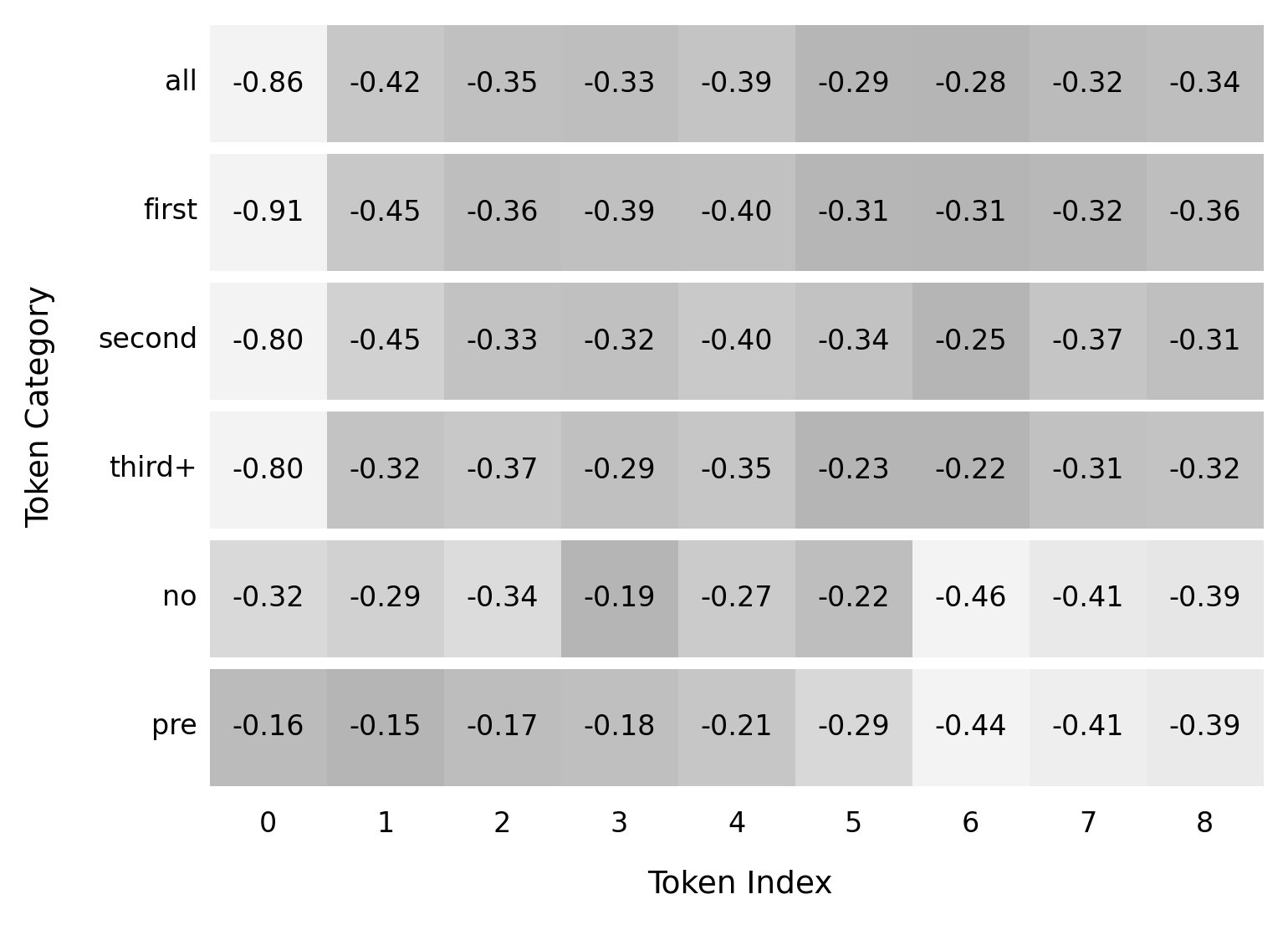}
        \caption{LLaMA-2-70B-chat}
        \label{mink:cat:p:80:llama-70b}
    \end{subfigure}
    \hfill
    \begin{subfigure}{0.49\textwidth}
        \centering
        \includegraphics[width=\linewidth]{plots/mink/categories/llama-2-70b-chat/mink_categories_80.png}
        \caption{Mistral-7B-instruct}
        \label{mink:cat:p:80:mistral-7b}
    \end{subfigure}
    \caption{\textbf{[80th percentile]} Min-K Probability scores per token category and index, over the first 9 tokens at global level.}
    \label{mink:cat:p:80}
\end{figure}

\begin{figure}[p]
    \centering
    \begin{subfigure}{0.49\textwidth}
        \centering
        \includegraphics[width=\linewidth]{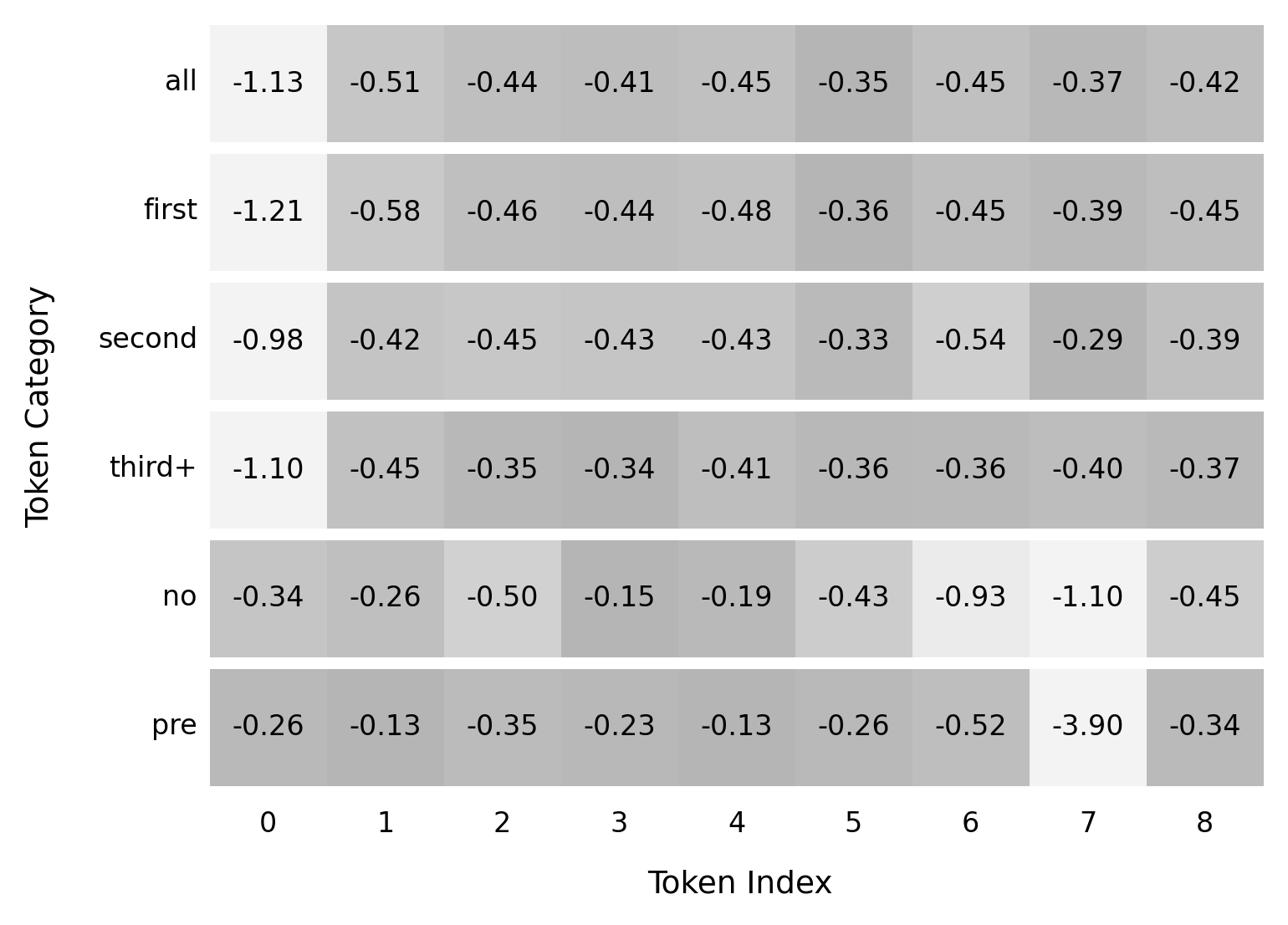}
        \caption{LLaMA-2-7B-chat}
        \label{mink:cat:p:90:llama-7b}
    \end{subfigure}
    \hfill
    \begin{subfigure}{0.49\textwidth}
        \centering
        \includegraphics[width=\linewidth]{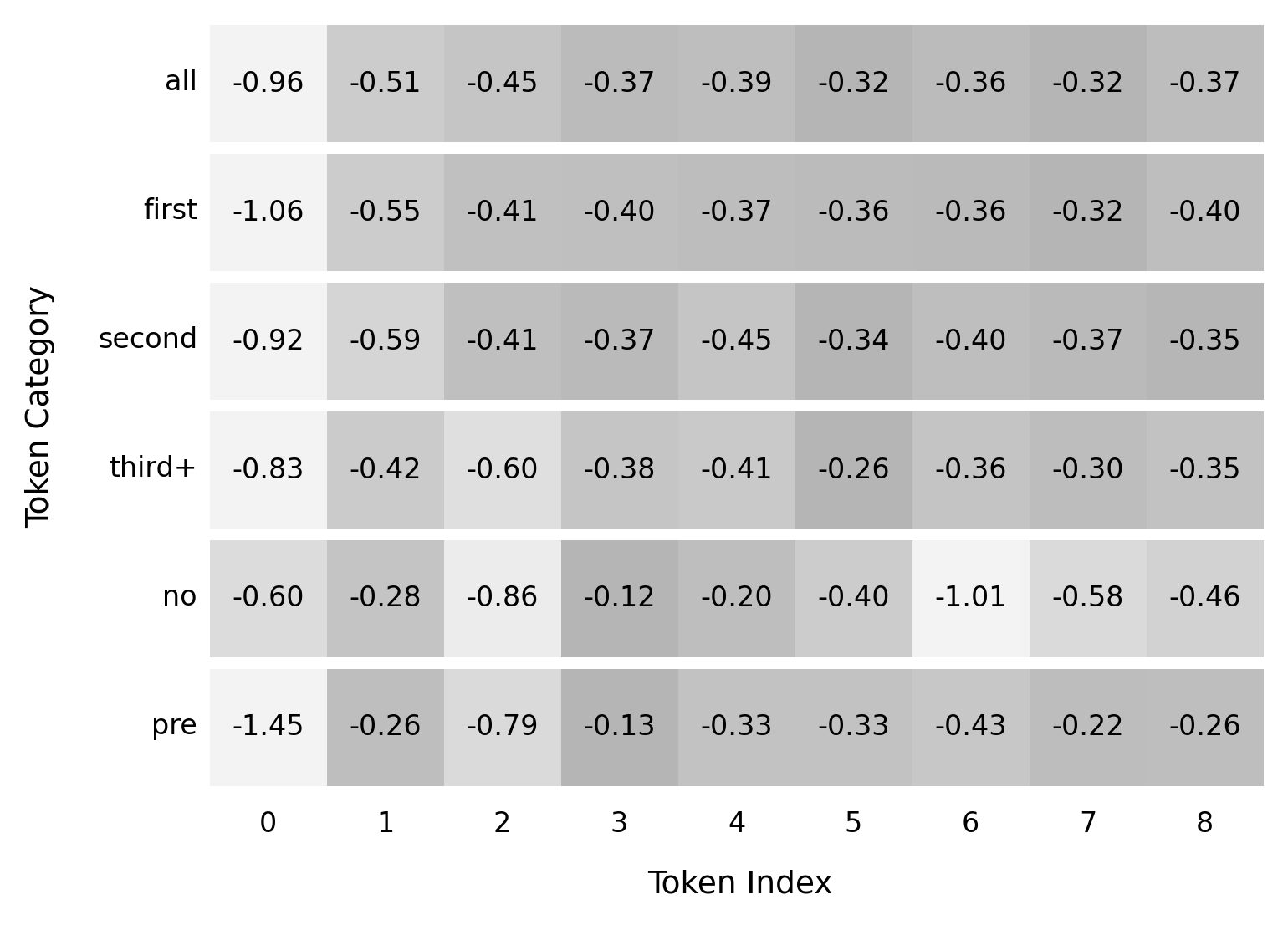}
        \caption{LLaMA-2-13B-chat}
        \label{mink:cat:p:90:llama-13b}
    \end{subfigure}
    \medskip
    \begin{subfigure}{0.49\textwidth}
        \centering
        \includegraphics[width=\linewidth]{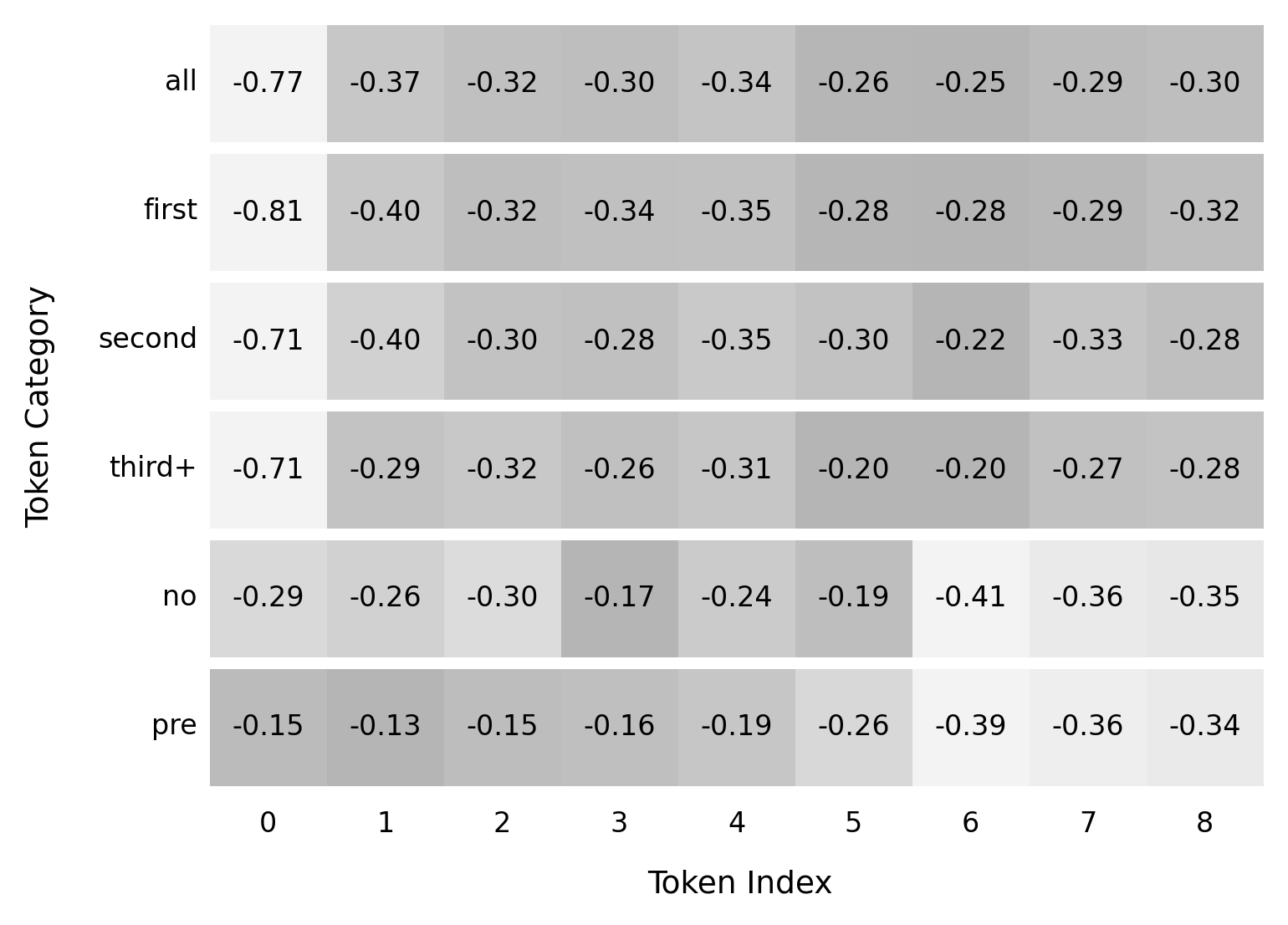}
        \caption{LLaMA-2-70B-chat}
        \label{mink:cat:p:90:llama-70b}
    \end{subfigure}
    \hfill
    \begin{subfigure}{0.49\textwidth}
        \centering
        \includegraphics[width=\linewidth]{plots/mink/categories/llama-2-70b-chat/mink_categories_90.png}
        \caption{Mistral-7B-instruct}
        \label{mink:cat:p:90:mistral-7b}
    \end{subfigure}
    \caption{\textbf{[90th percentile]} Min-K Probability scores per token category and index, over the first 9 tokens at global level.}
    \label{mink:cat:p:90}
\end{figure}

\begin{figure}[p]
    \centering
    \begin{subfigure}{0.49\textwidth}
        \centering
        \includegraphics[width=\linewidth]{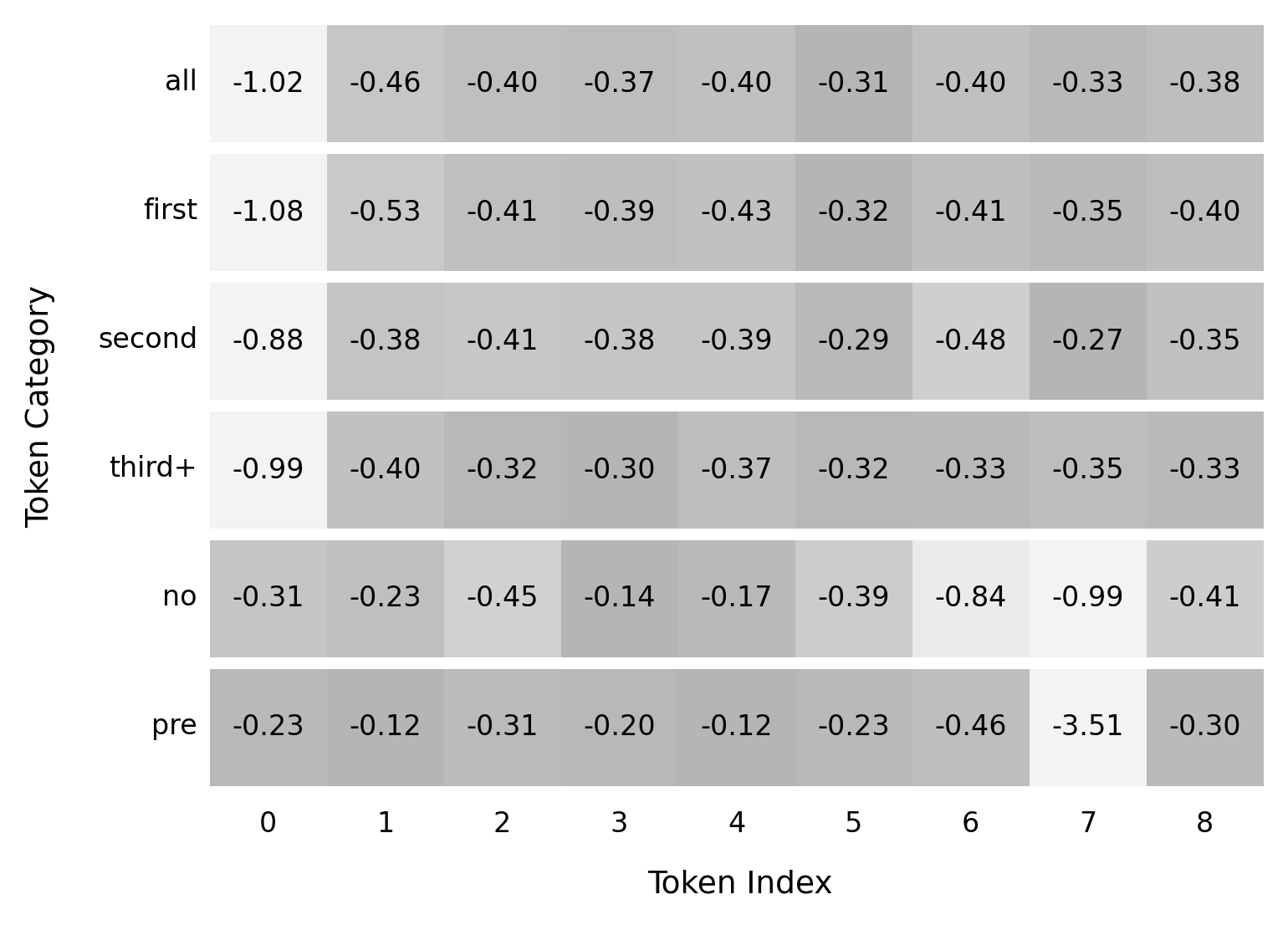}
        \caption{LLaMA-2-7B-chat}
        \label{mink:cat:p:100:llama-7b}
    \end{subfigure}
    \hfill
    \begin{subfigure}{0.49\textwidth}
        \centering
        \includegraphics[width=\linewidth]{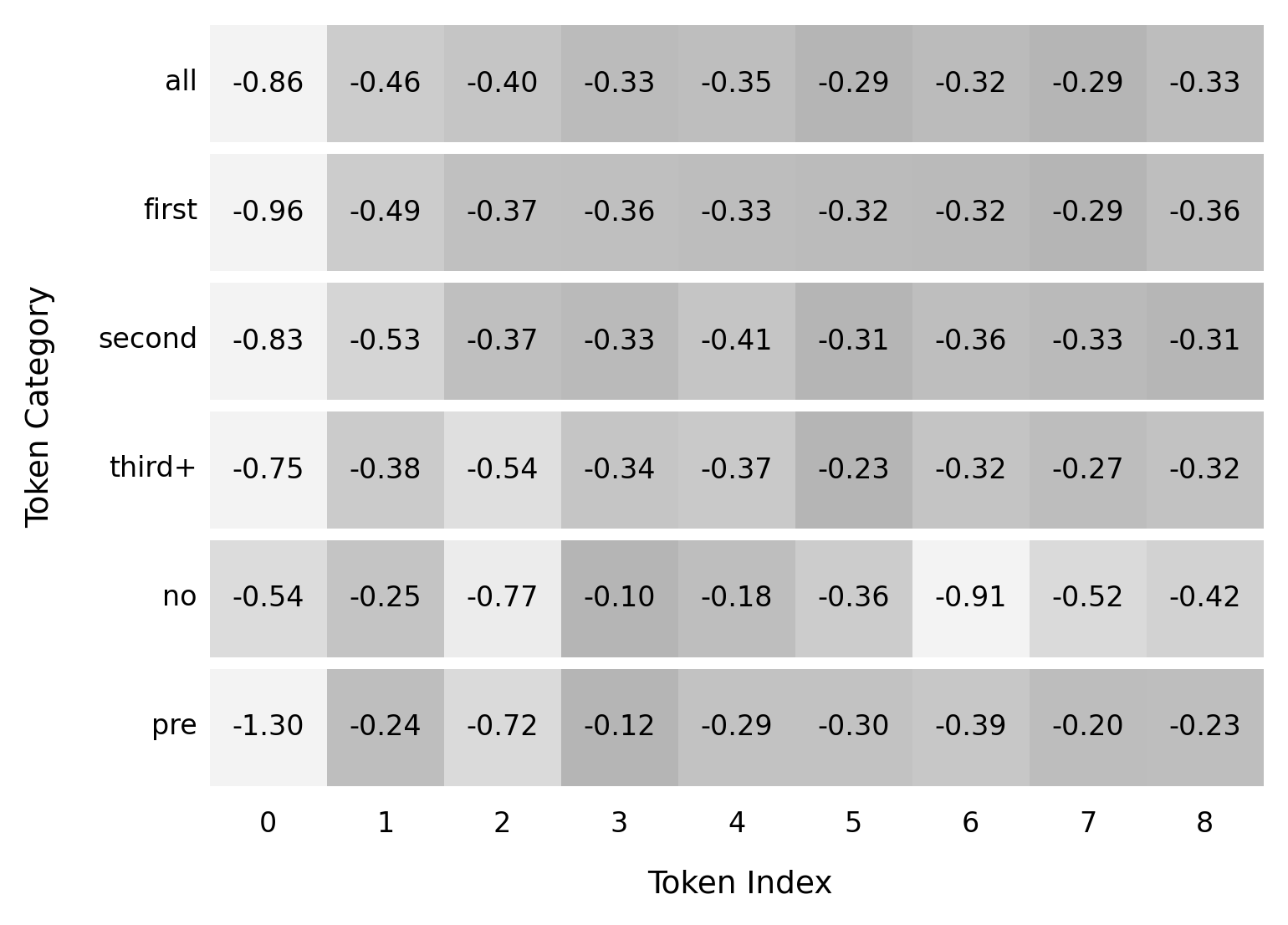}
        \caption{LLaMA-2-13B-chat}
        \label{mink:cat:p:100:llama-13b}
    \end{subfigure}
    \medskip
    \begin{subfigure}{0.49\textwidth}
        \centering
        \includegraphics[width=\linewidth]{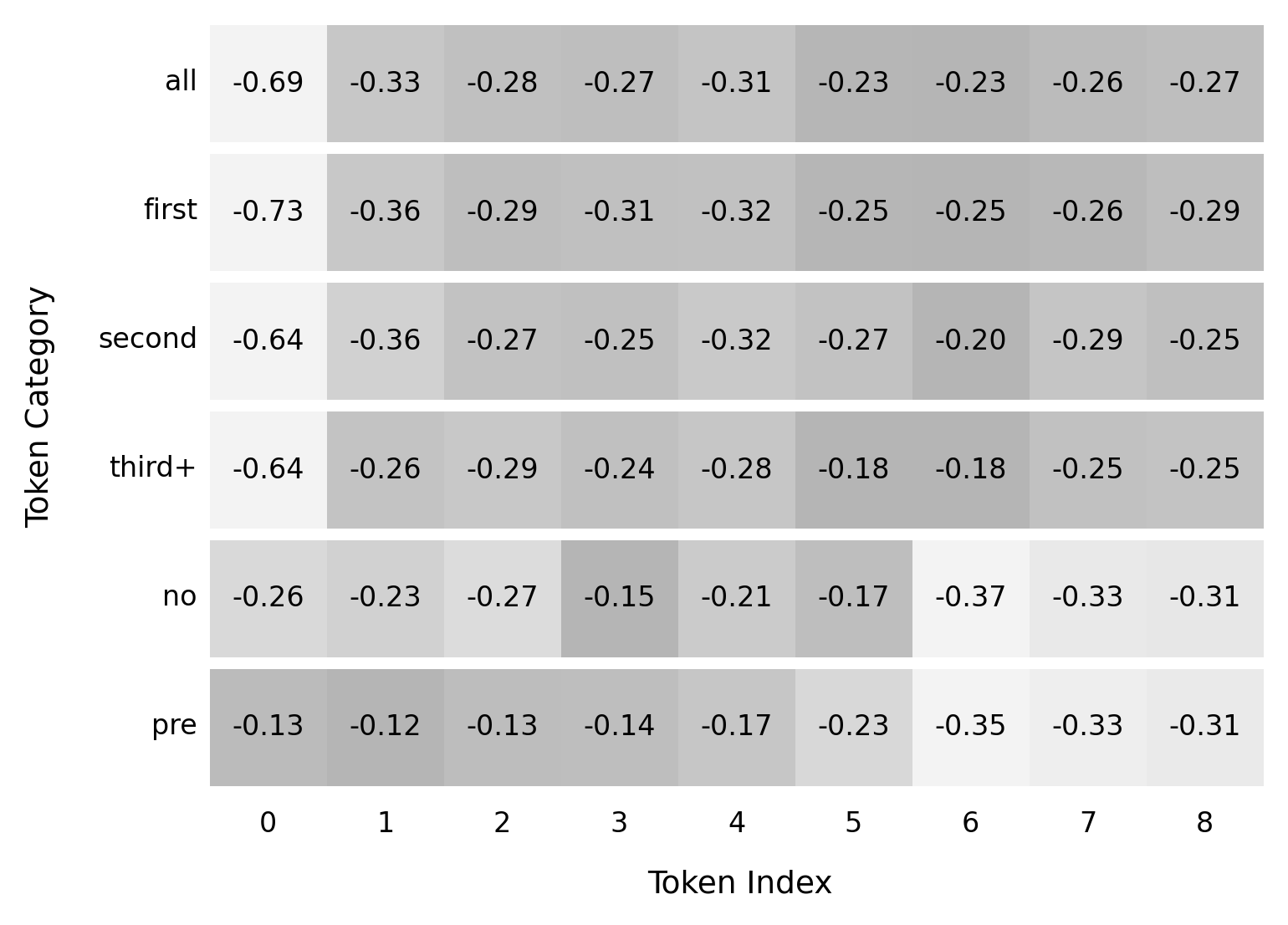}
        \caption{LLaMA-2-70B-chat}
        \label{mink:cat:p:100:llama-70b}
    \end{subfigure}
    \hfill
    \begin{subfigure}{0.49\textwidth}
        \centering
        \includegraphics[width=\linewidth]{plots/mink/categories/llama-2-70b-chat/mink_categories_100.png}
        \caption{Mistral-7B-instruct}
        \label{mink:cat:p:100:mistral-7b}
    \end{subfigure}
    \caption{\textbf{[100th percentile]} Min-K Probability scores per token category and index, over the first 9 tokens at global level.}
    \label{mink:cat:p:100}
\end{figure}

\FloatBarrier
\subsubsection{Min-K Entropy}
\label{plt:mink:sep:entropy}
\begin{figure}[!hp]
    \vspace*{\fill}
    \centering
    \begin{subfigure}{0.49\textwidth}
        \centering
        \includegraphics[width=\linewidth]{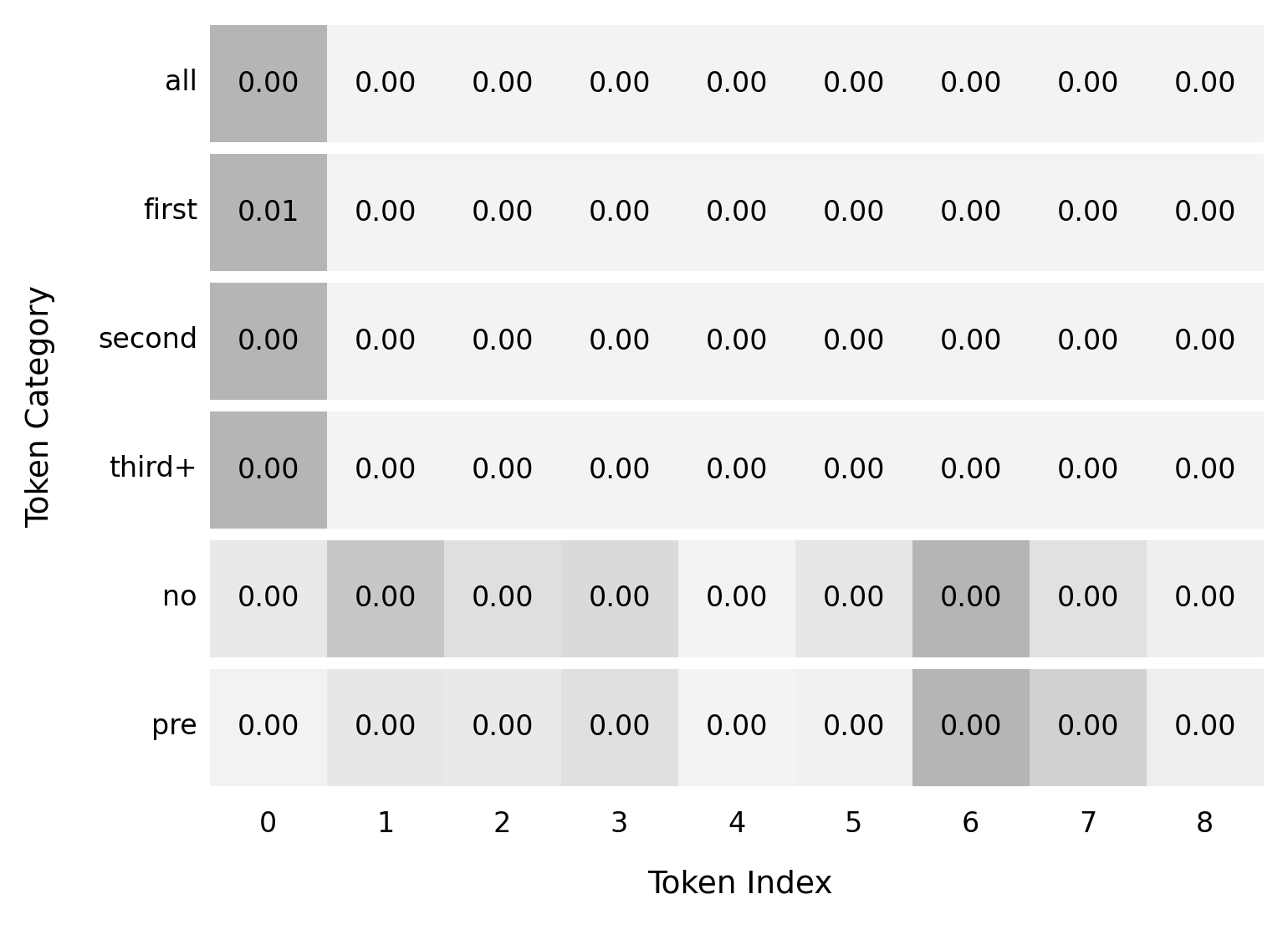}
        \caption{LLaMA-2-7B-chat}
        \label{mink:cat:e:10:llama-7b}
    \end{subfigure}
    \hfill
    \begin{subfigure}{0.49\textwidth}
        \centering
        \includegraphics[width=\linewidth]{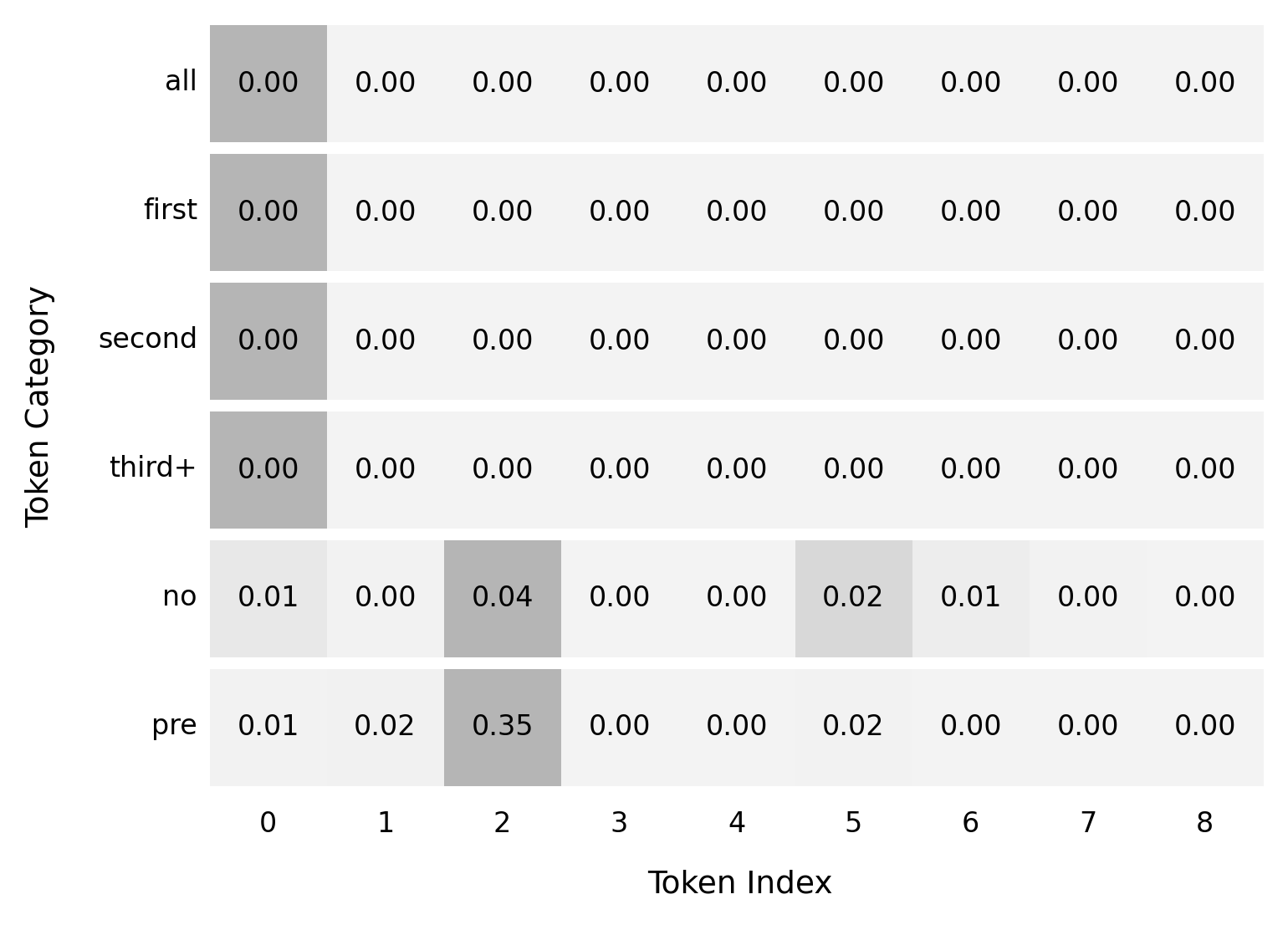}
        \caption{LLaMA-2-13B-chat}
        \label{mink:cat:e:10:llama-13b}
    \end{subfigure}
    \medskip
    \begin{subfigure}{0.49\textwidth}
        \centering
        \includegraphics[width=\linewidth]{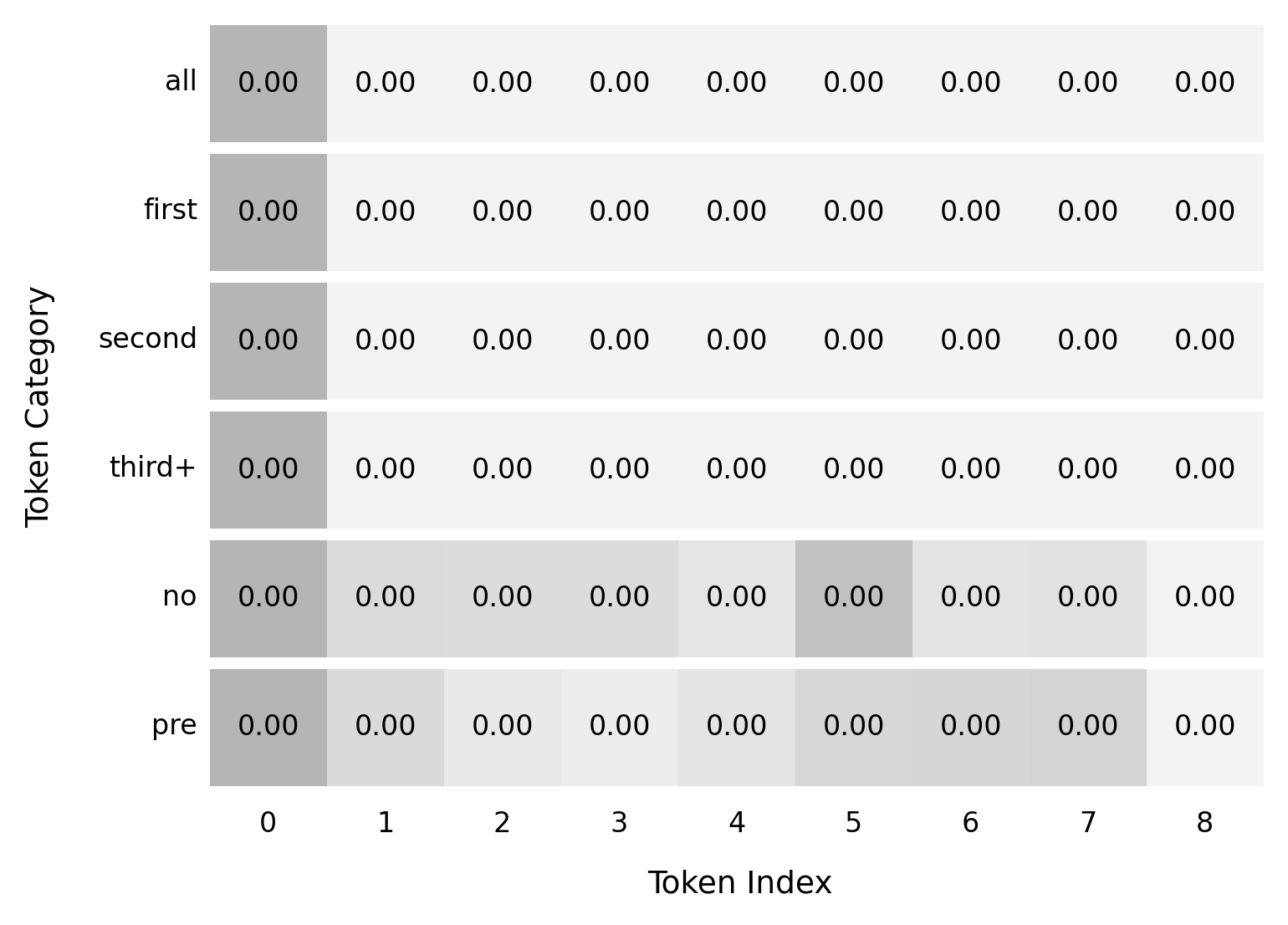}
        \caption{LLaMA-2-70B-chat}
        \label{mink:cat:e:10:llama-70b}
    \end{subfigure}
    \hfill
    \begin{subfigure}{0.49\textwidth}
        \centering
        \includegraphics[width=\linewidth]{plots/mink/categories/llama-2-70b-chat/mink-e_categories_10.png}
        \caption{Mistral-7B-instruct}
        \label{mink:cat:e:10:mistral-7b}
    \end{subfigure}
    \caption{\textbf{[10th percentile]} Min-K Entropy scores per token category and index, over the first 9 tokens at global level.}
    \label{mink:cat:e:10}
    \vspace*{\fill}
\end{figure}

\begin{figure}[htbp]
    \centering
    \begin{subfigure}{0.49\textwidth}
        \centering
        \includegraphics[width=\linewidth]{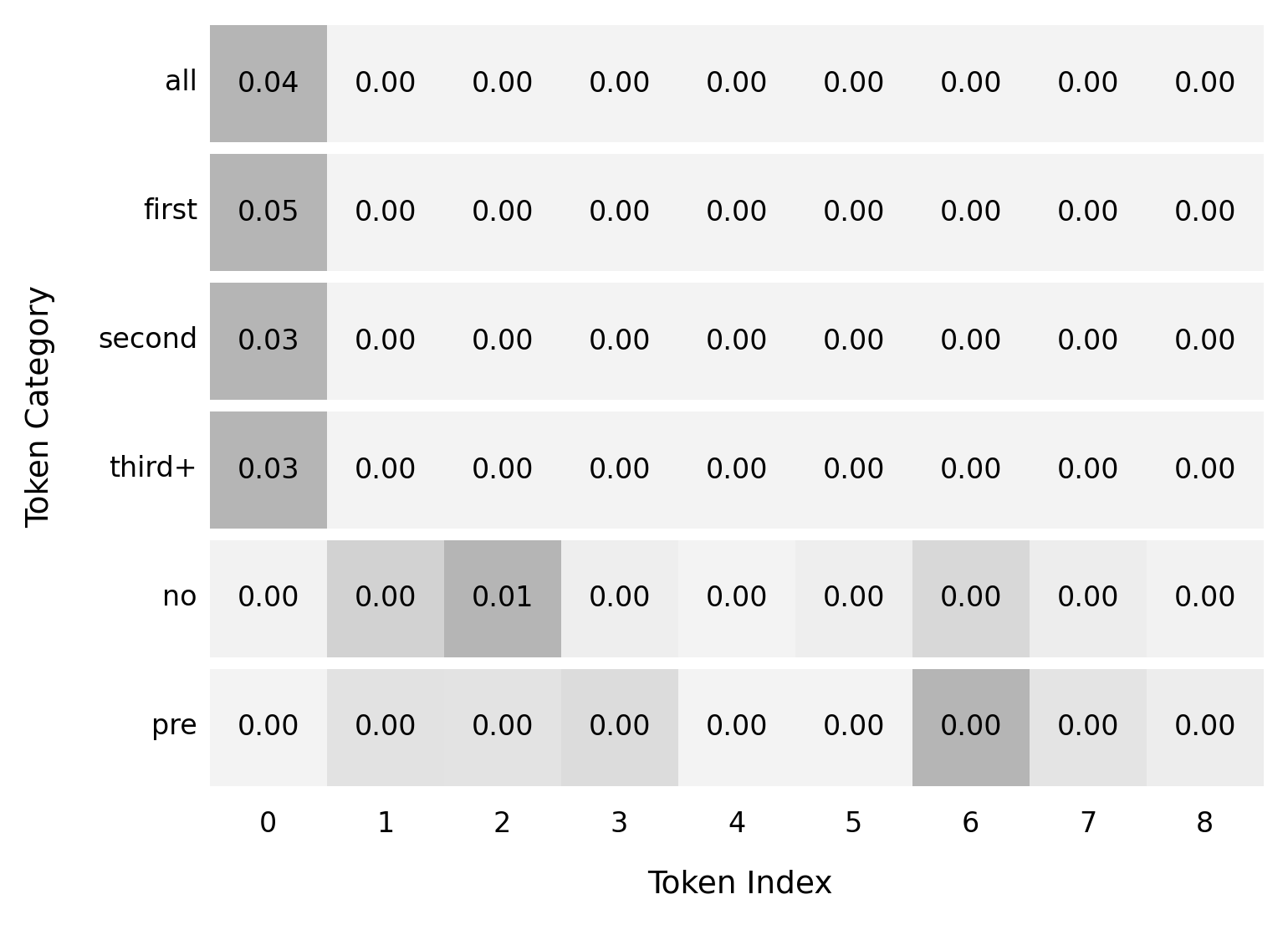}
        \caption{LLaMA-2-7B-chat}
        \label{mink:cat:e:20:llama-7b}
    \end{subfigure}
    \hfill
    \begin{subfigure}{0.49\textwidth}
        \centering
        \includegraphics[width=\linewidth]{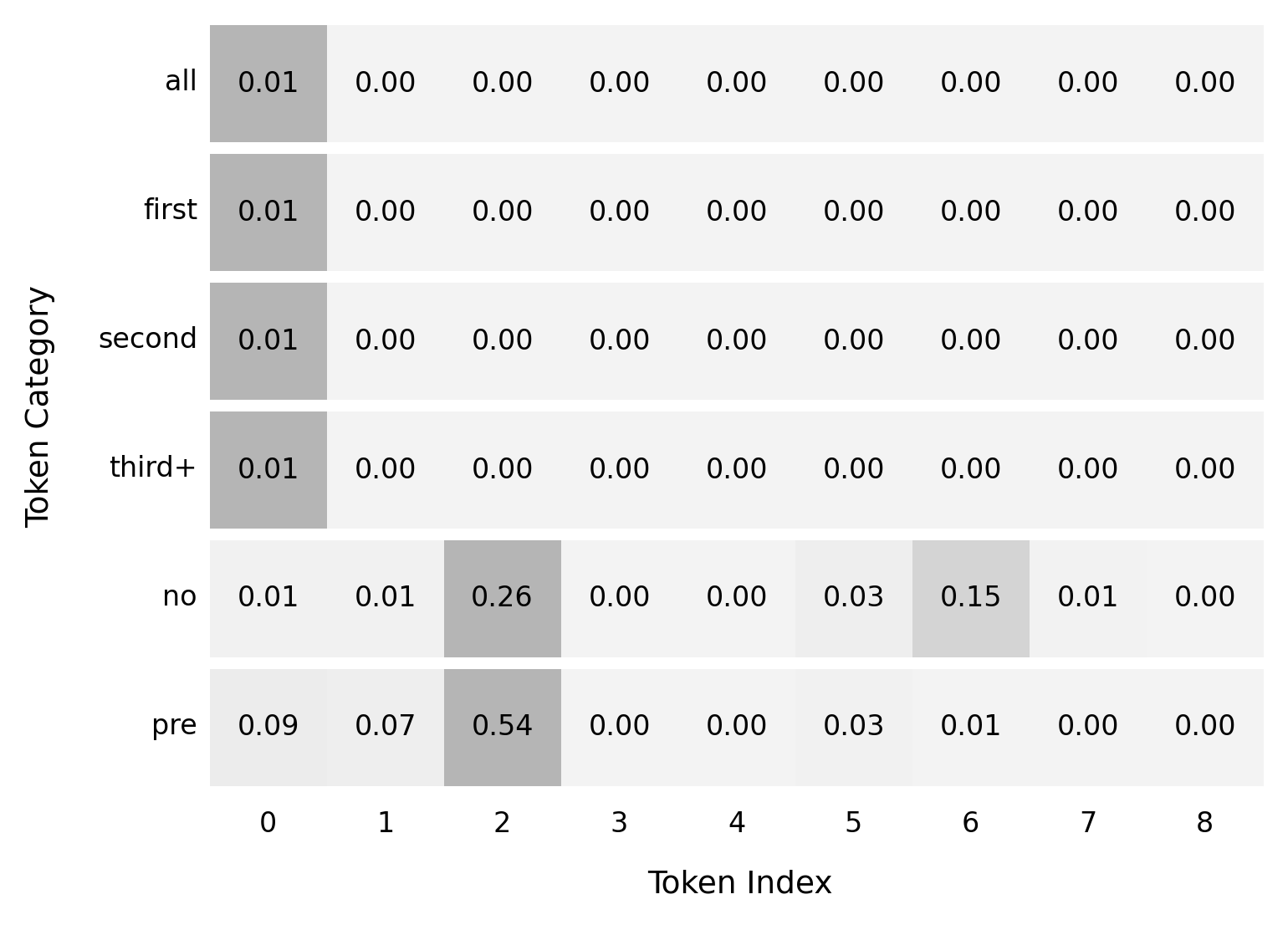}
        \caption{LLaMA-2-13B-chat}
        \label{mink:cat:e:20:llama-13b}
    \end{subfigure}
    \medskip
    \begin{subfigure}{0.49\textwidth}
        \centering
        \includegraphics[width=\linewidth]{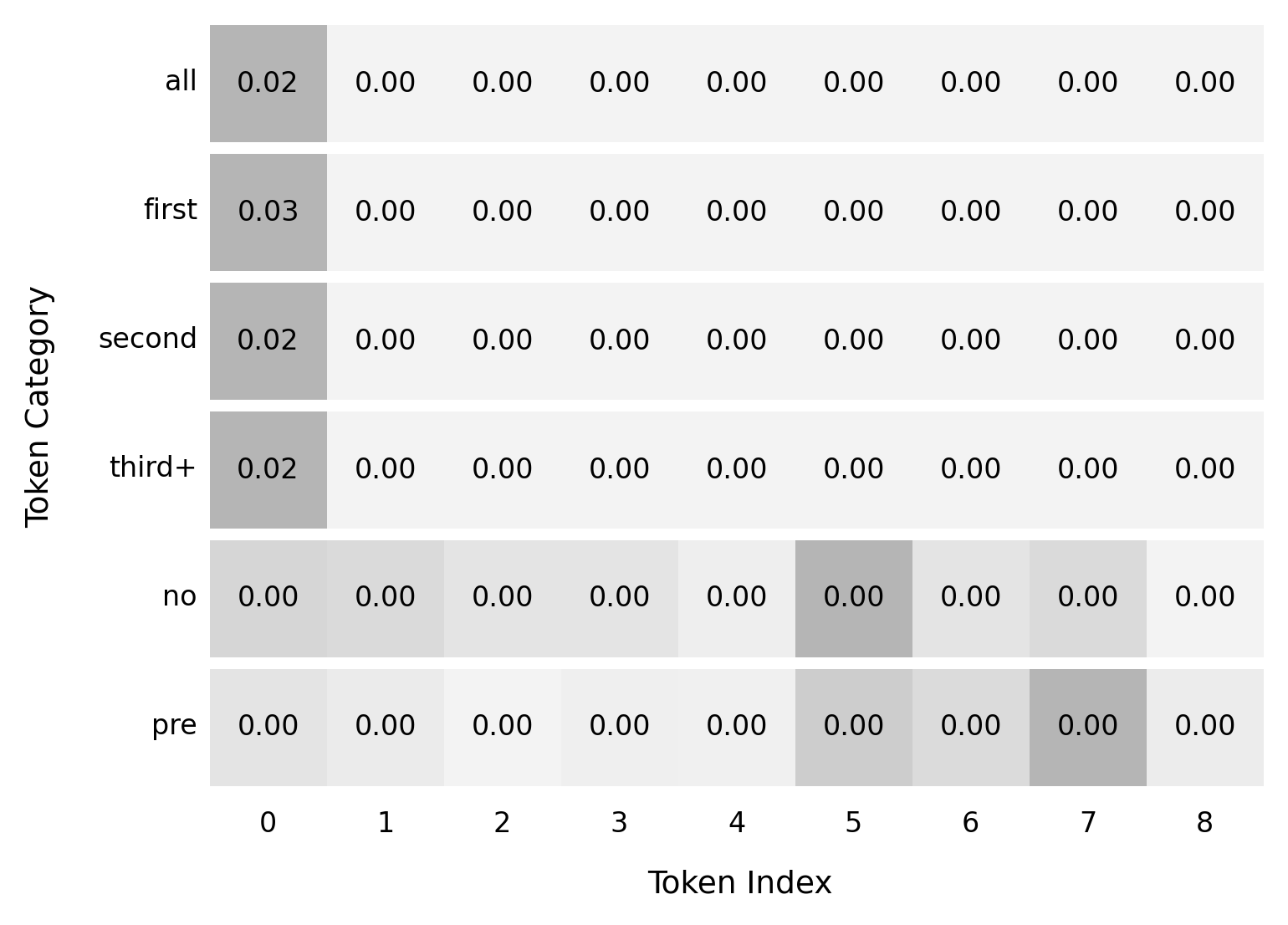}
        \caption{LLaMA-2-70B-chat}
        \label{mink:cat:e:20:llama-70b}
    \end{subfigure}
    \hfill
    \begin{subfigure}{0.49\textwidth}
        \centering
        \includegraphics[width=\linewidth]{plots/mink/categories/llama-2-70b-chat/mink-e_categories_20.png}
        \caption{Mistral-7B-instruct}
        \label{mink:cat:e:20:mistral-7b}
    \end{subfigure}
    \caption{\textbf{[20th percentile]} Min-K Entropy scores per token category and index, over the first 9 tokens at global level.}
    \label{mink:cat:e:20}
\end{figure}

\begin{figure}[htbp]
    \centering
    \begin{subfigure}{0.49\textwidth}
        \centering
        \includegraphics[width=\linewidth]{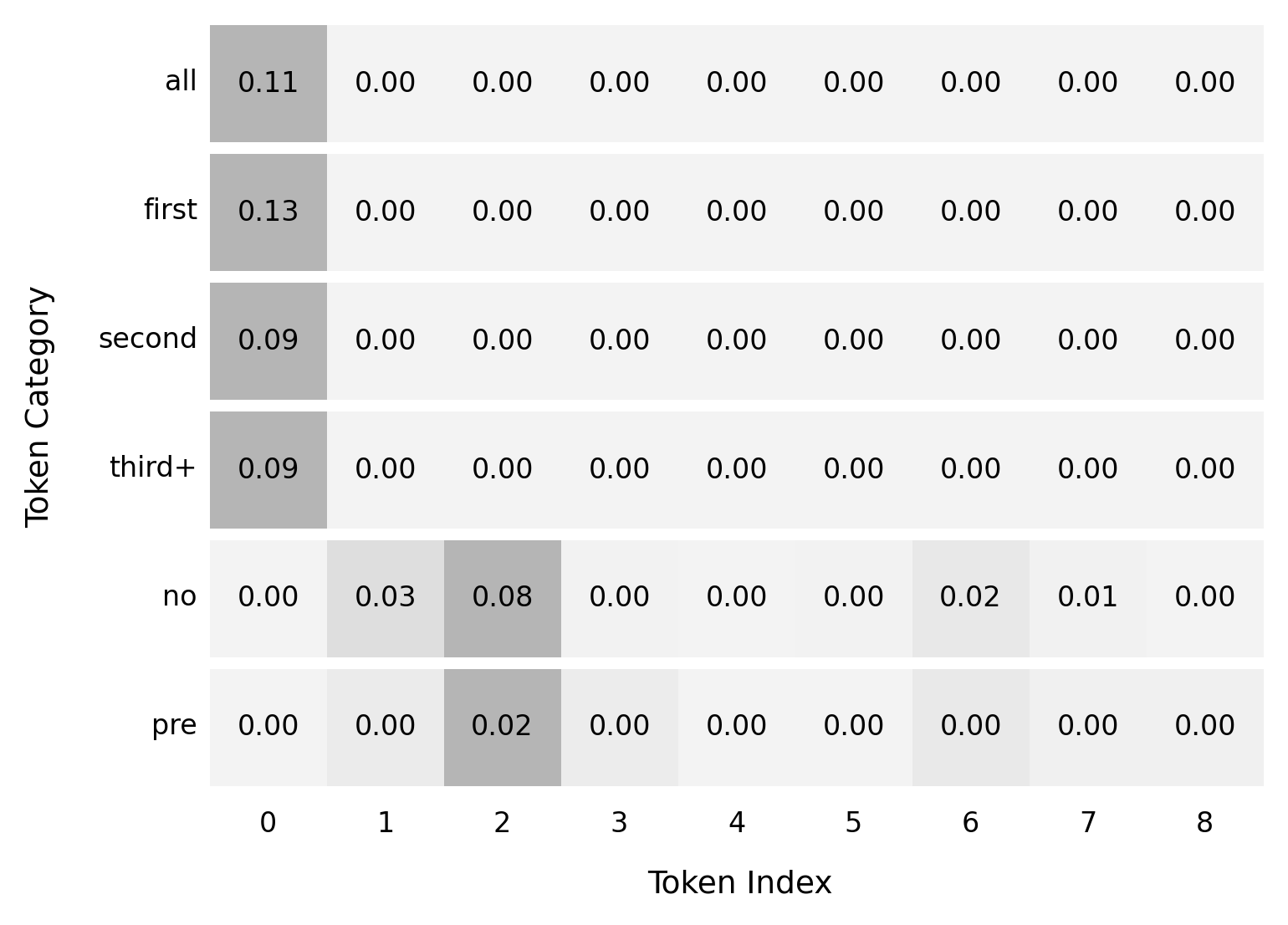}
        \caption{LLaMA-2-7B-chat}
        \label{mink:cat:e:30:llama-7b}
    \end{subfigure}
    \hfill
    \begin{subfigure}{0.49\textwidth}
        \centering
        \includegraphics[width=\linewidth]{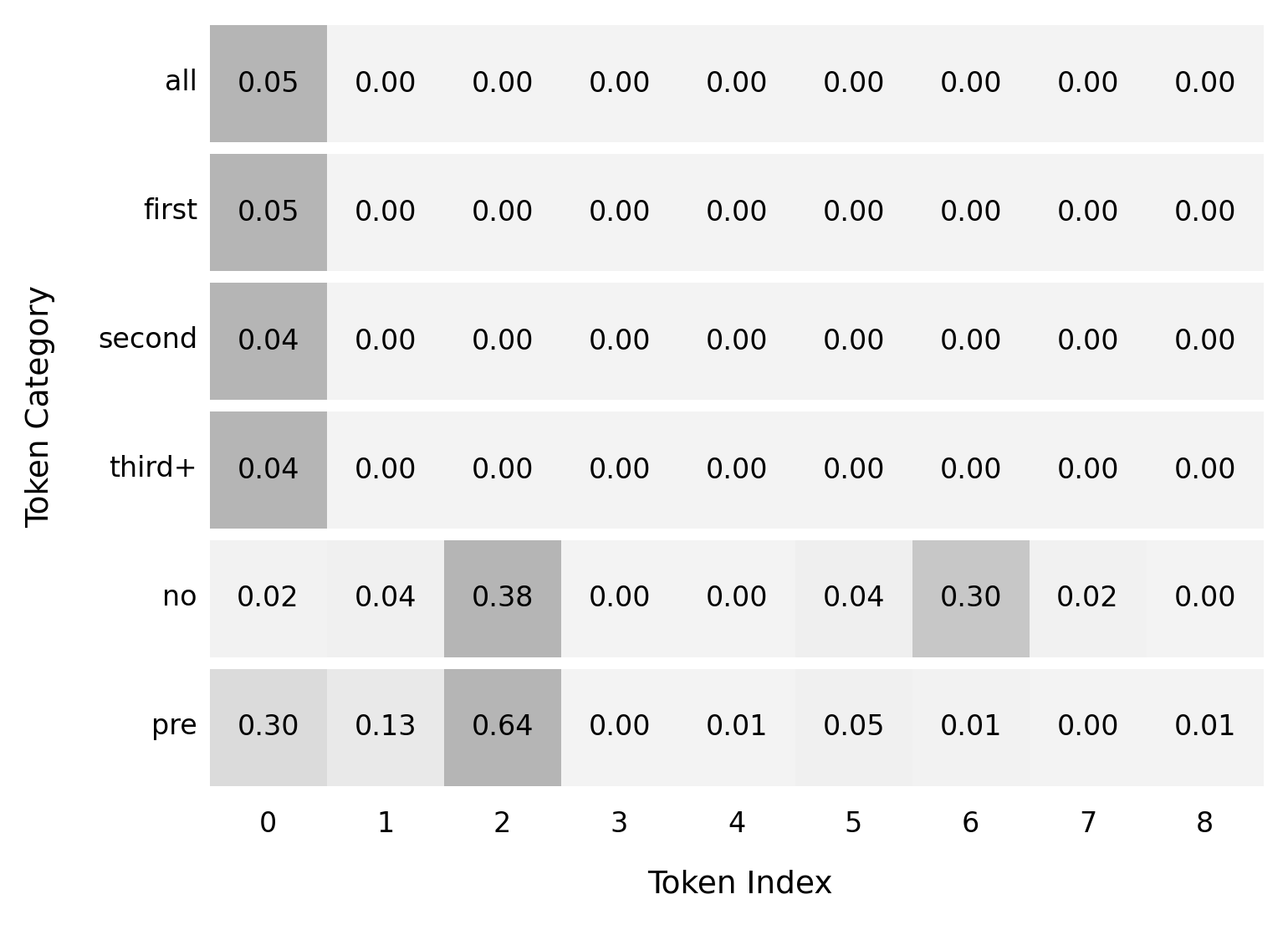}
        \caption{LLaMA-2-13B-chat}
        \label{mink:cat:e:30:llama-13b}
    \end{subfigure}
    \medskip
    \begin{subfigure}{0.49\textwidth}
        \centering
        \includegraphics[width=\linewidth]{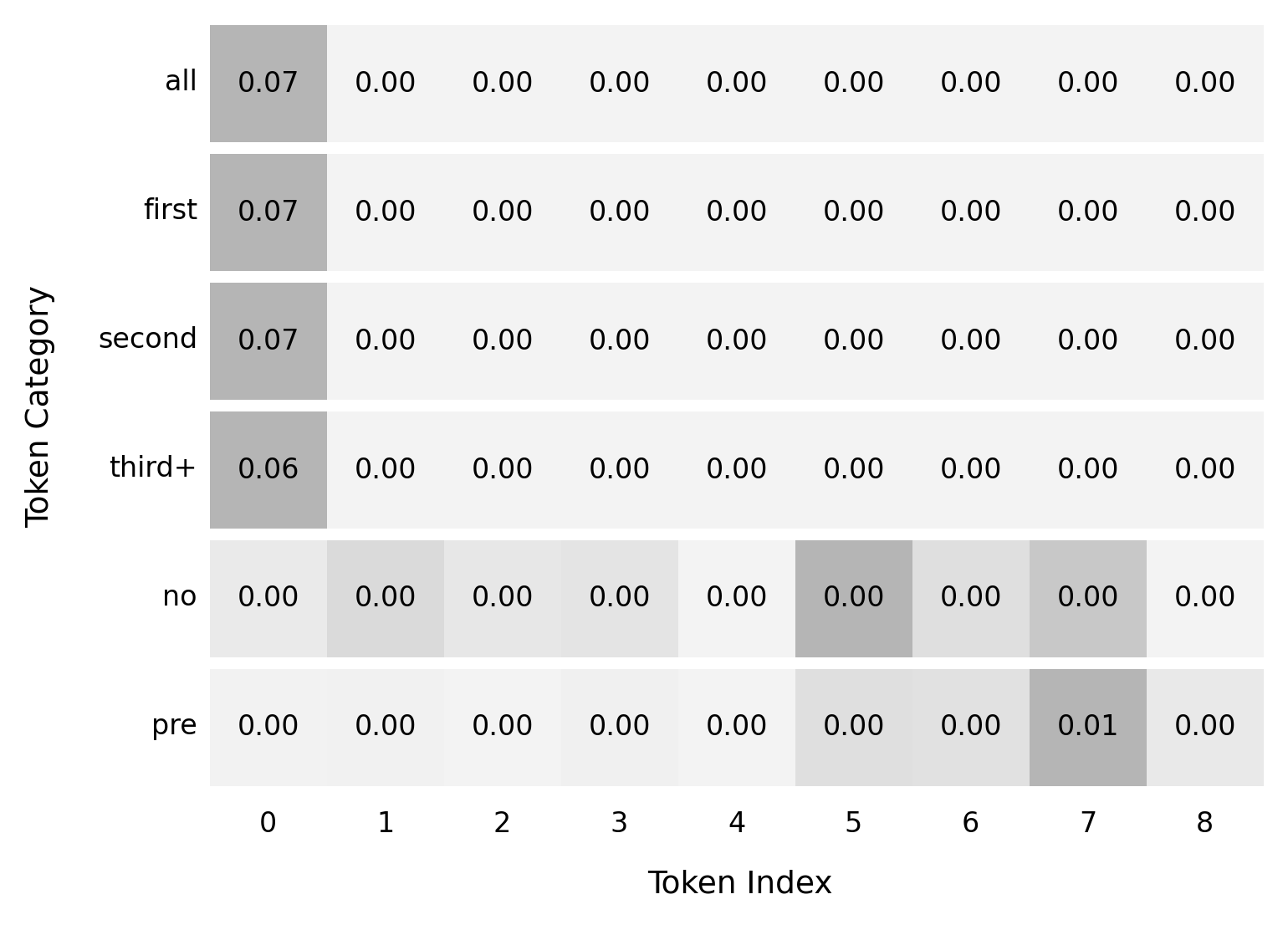}
        \caption{LLaMA-2-70B-chat}
        \label{mink:cat:e:30:llama-70b}
    \end{subfigure}
    \hfill
    \begin{subfigure}{0.49\textwidth}
        \centering
        \includegraphics[width=\linewidth]{plots/mink/categories/llama-2-70b-chat/mink-e_categories_30.png}
        \caption{Mistral-7B-instruct}
        \label{mink:cat:e:30:mistral-7b}
    \end{subfigure}
    \caption{\textbf{[30th percentile]} Min-K Entropy scores per token category and index, over the first 9 tokens at global level.}
    \label{mink:cat:e:30}
\end{figure}

\begin{figure}[htbp]
    \centering
    \begin{subfigure}{0.49\textwidth}
        \centering
        \includegraphics[width=\linewidth]{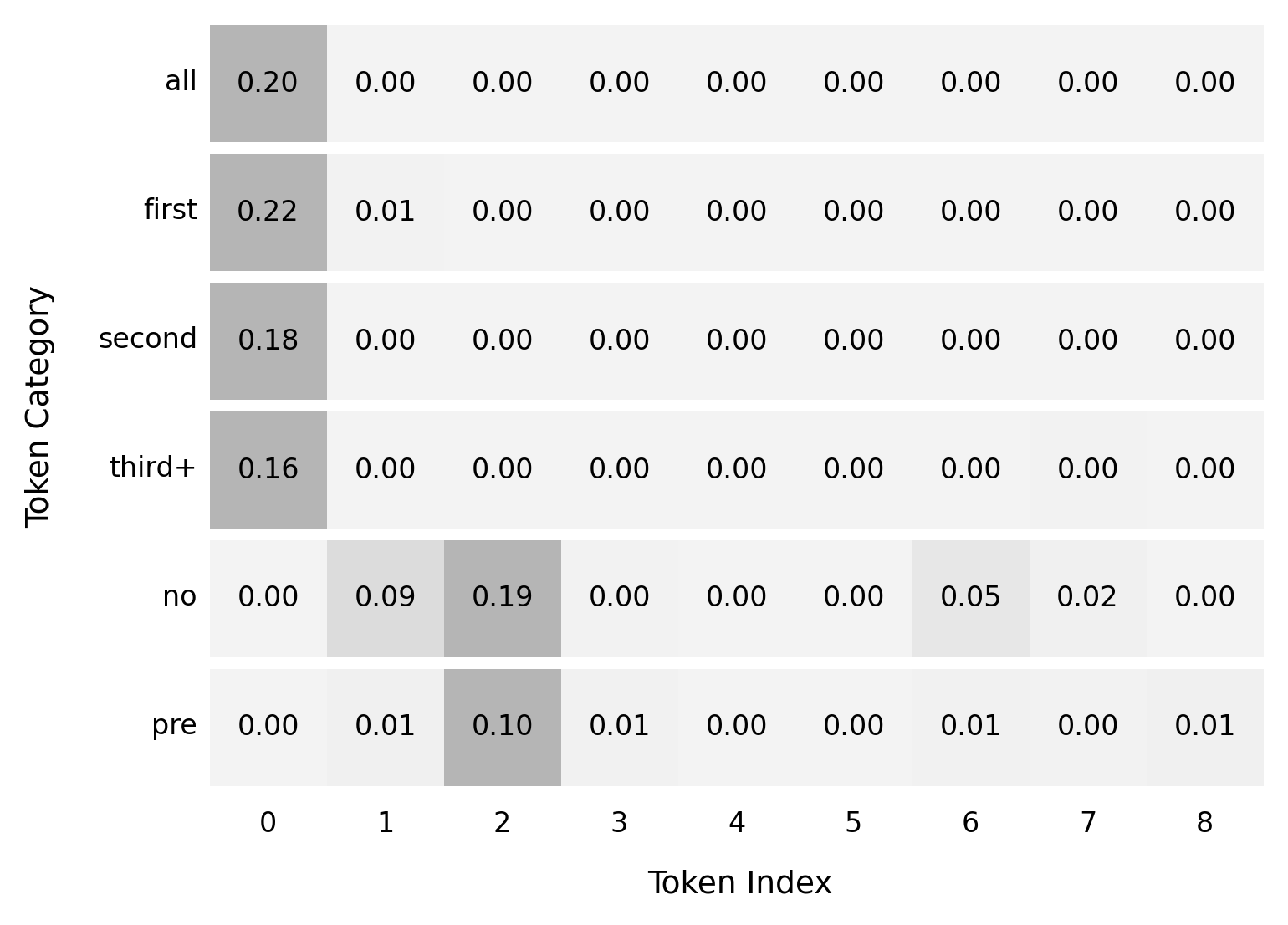}
        \caption{LLaMA-2-7B-chat}
        \label{mink:cat:e:40:llama-7b}
    \end{subfigure}
    \hfill
    \begin{subfigure}{0.49\textwidth}
        \centering
        \includegraphics[width=\linewidth]{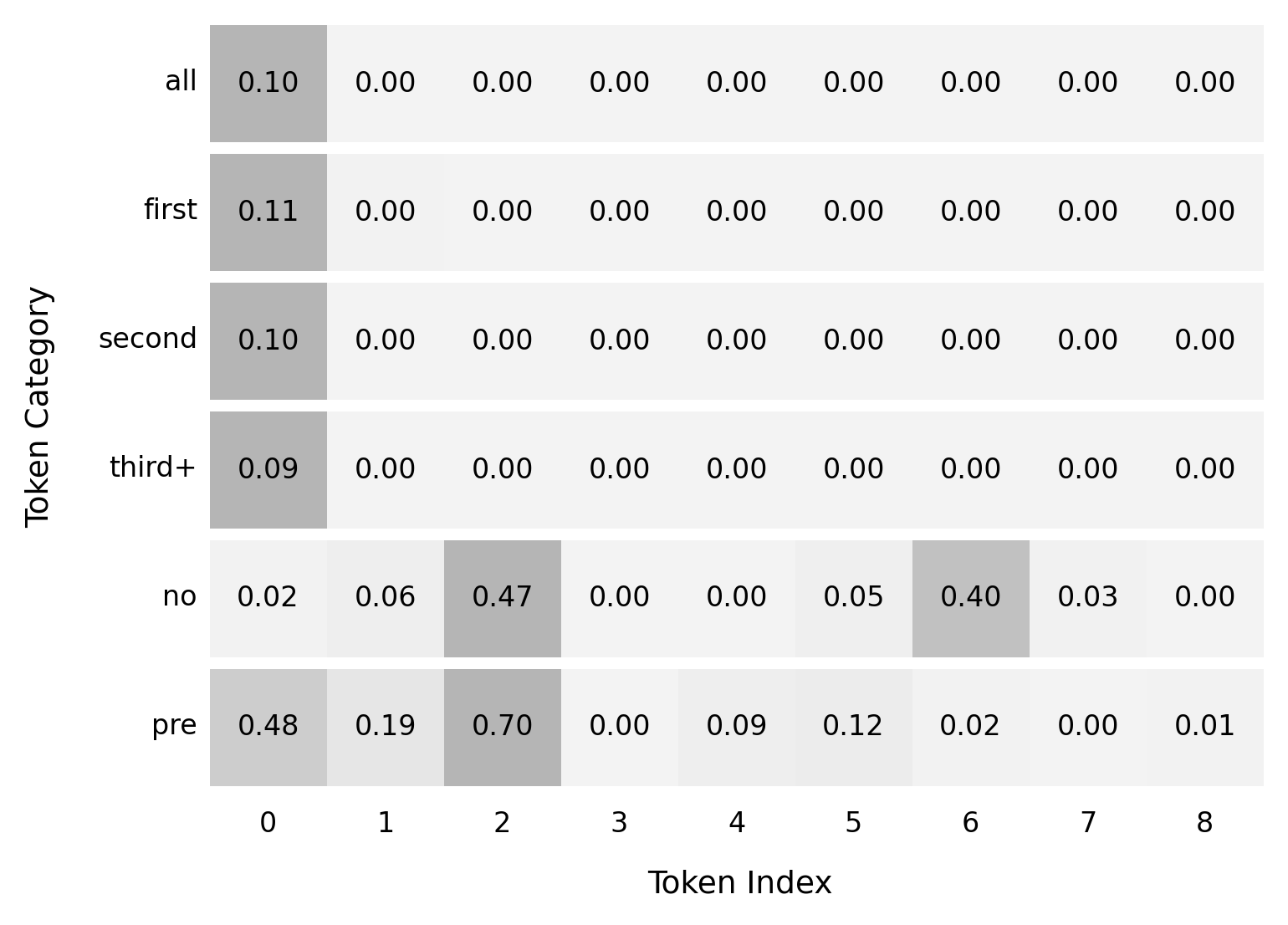}
        \caption{LLaMA-2-13B-chat}
        \label{mink:cat:e:40:llama-13b}
    \end{subfigure}
    \medskip
    \begin{subfigure}{0.49\textwidth}
        \centering
        \includegraphics[width=\linewidth]{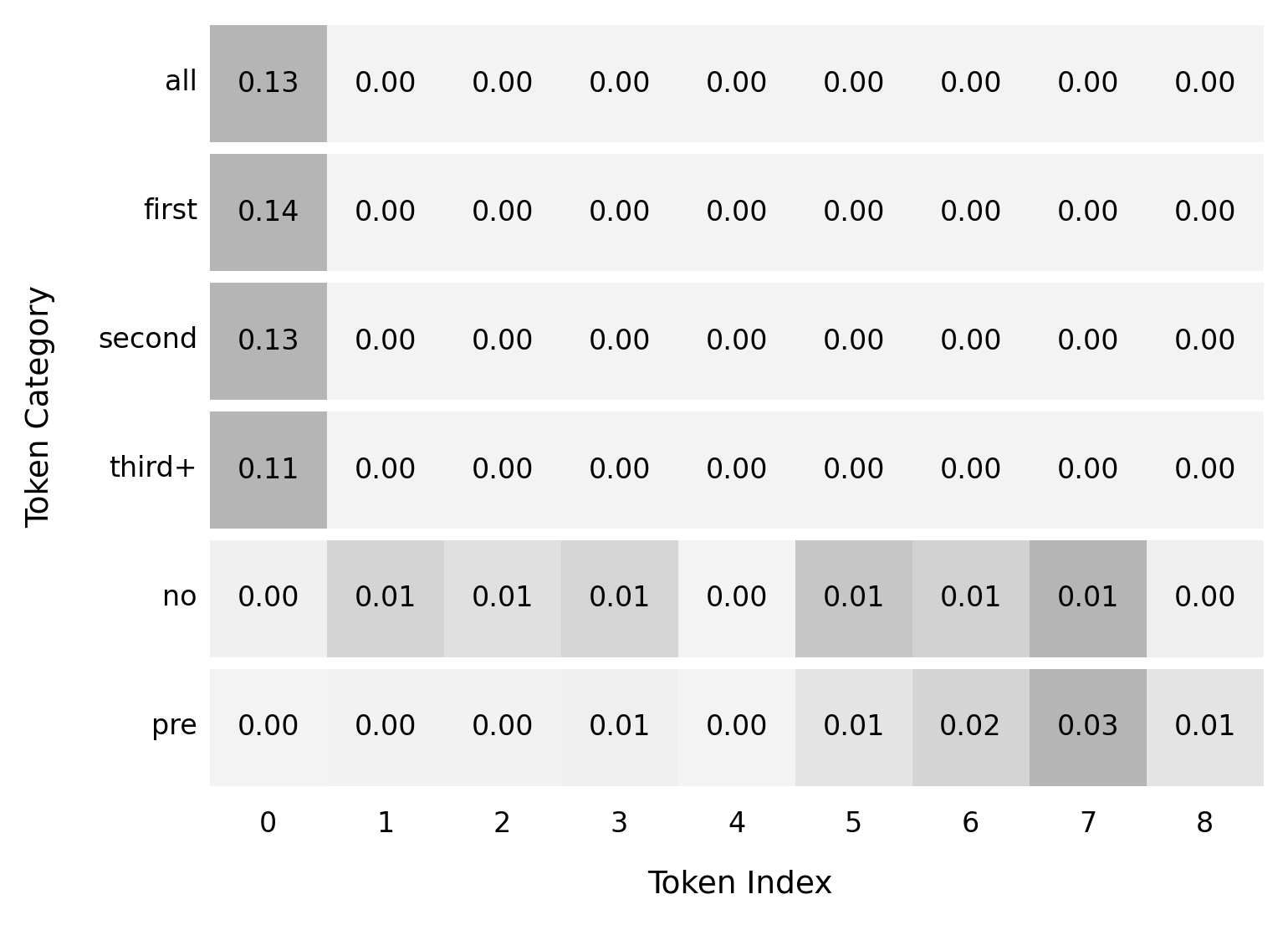}
        \caption{LLaMA-2-70B-chat}
        \label{mink:cat:e:40:llama-70b}
    \end{subfigure}
    \hfill
    \begin{subfigure}{0.49\textwidth}
        \centering
        \includegraphics[width=\linewidth]{plots/mink/categories/llama-2-70b-chat/mink-e_categories_40.png}
        \caption{Mistral-7B-instruct}
        \label{mink:cat:e:40:mistral-7b}
    \end{subfigure}
    \caption{\textbf{[40th percentile]} Min-K Entropy scores per token category and index, over the first 9 tokens at global level.}
    \label{mink:cat:e:40}
\end{figure}

\begin{figure}[htbp]
    \centering
    \begin{subfigure}{0.49\textwidth}
        \centering
        \includegraphics[width=\linewidth]{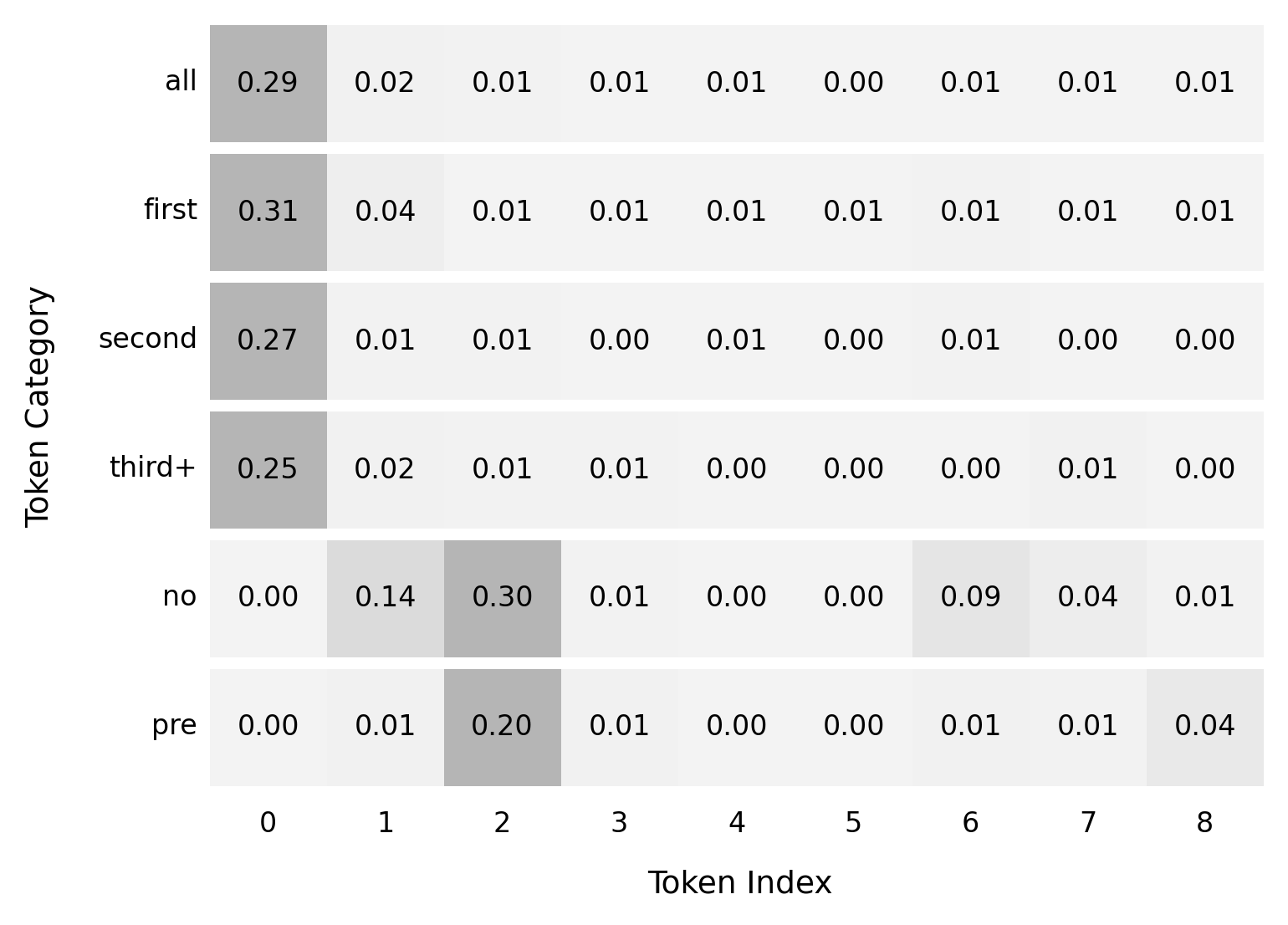}
        \caption{LLaMA-2-7B-chat}
        \label{mink:cat:e:50:llama-7b}
    \end{subfigure}
    \hfill
    \begin{subfigure}{0.49\textwidth}
        \centering
        \includegraphics[width=\linewidth]{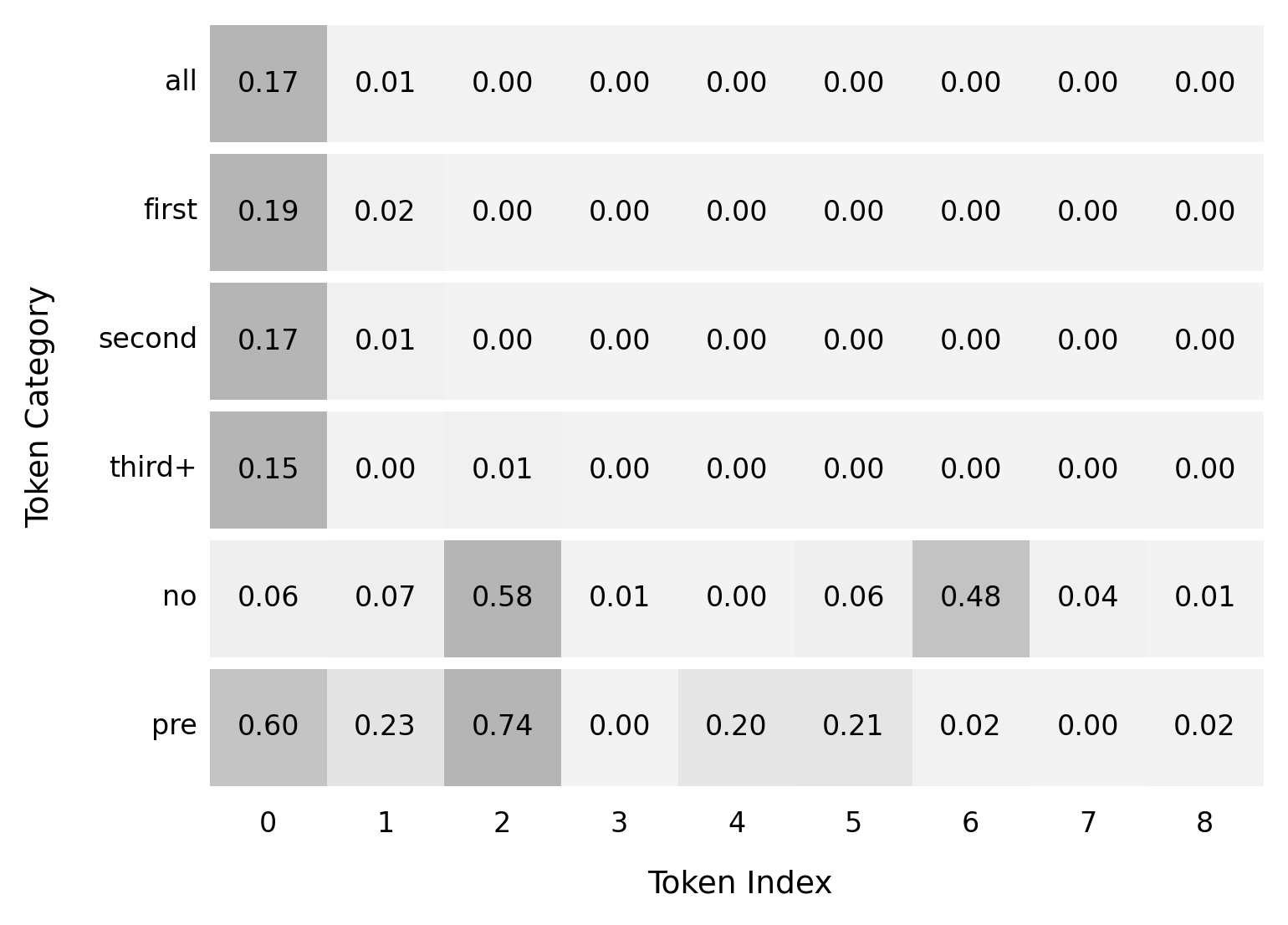}
        \caption{LLaMA-2-13B-chat}
        \label{mink:cat:e:50:llama-13b}
    \end{subfigure}
    \medskip
    \begin{subfigure}{0.49\textwidth}
        \centering
        \includegraphics[width=\linewidth]{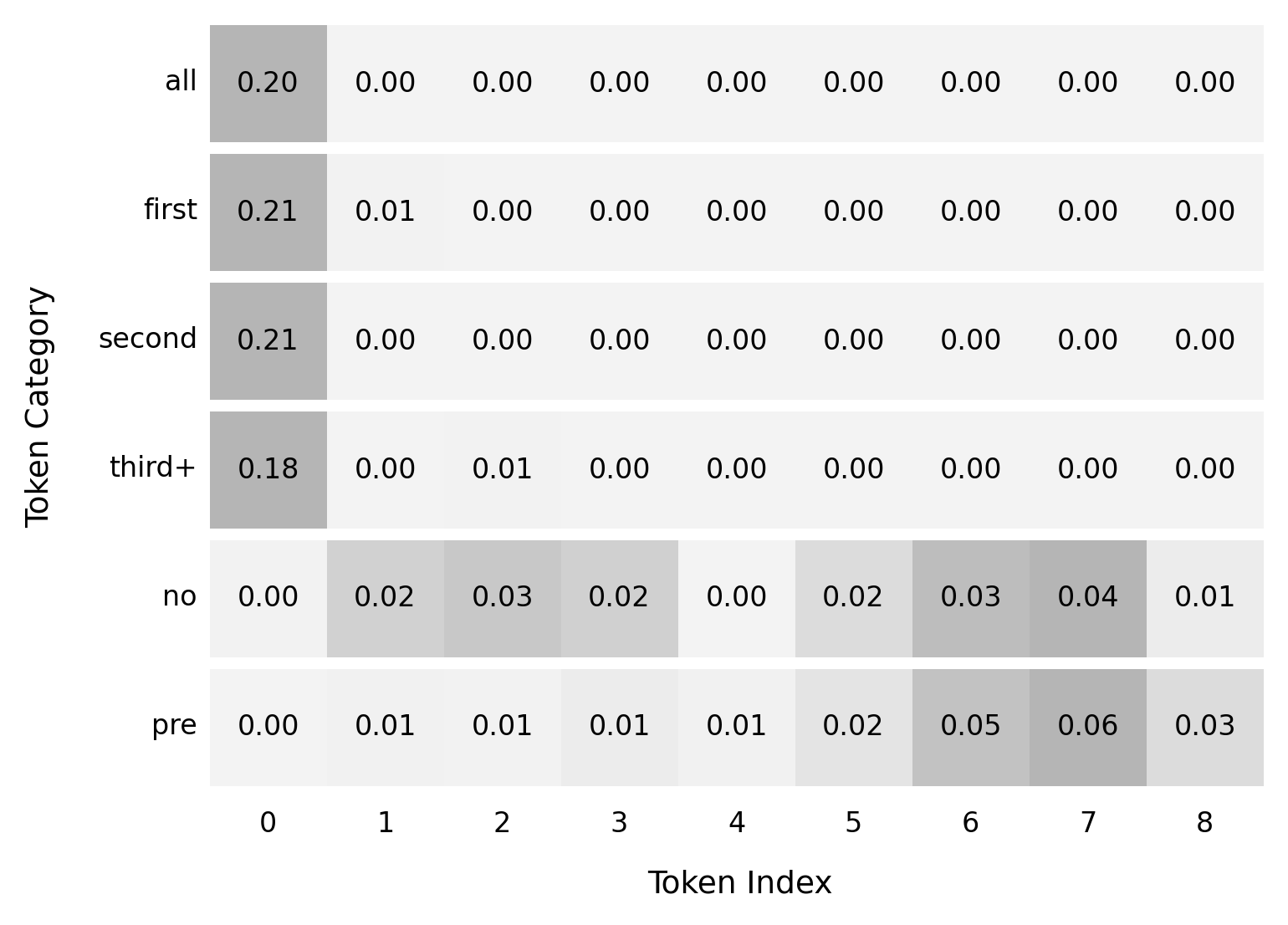}
        \caption{LLaMA-2-70B-chat}
        \label{mink:cat:e:50:llama-70b}
    \end{subfigure}
    \hfill
    \begin{subfigure}{0.49\textwidth}
        \centering
        \includegraphics[width=\linewidth]{plots/mink/categories/llama-2-70b-chat/mink-e_categories_50.png}
        \caption{Mistral-7B-instruct}
        \label{mink:cat:e:50:mistral-7b}
    \end{subfigure}
    \caption{\textbf{[50th percentile]} Min-K Entropy scores per token category and index, over the first 9 tokens at global level.}
    \label{mink:cat:e:50}
\end{figure}

\begin{figure}[htbp]
    \centering
    \begin{subfigure}{0.49\textwidth}
        \centering
        \includegraphics[width=\linewidth]{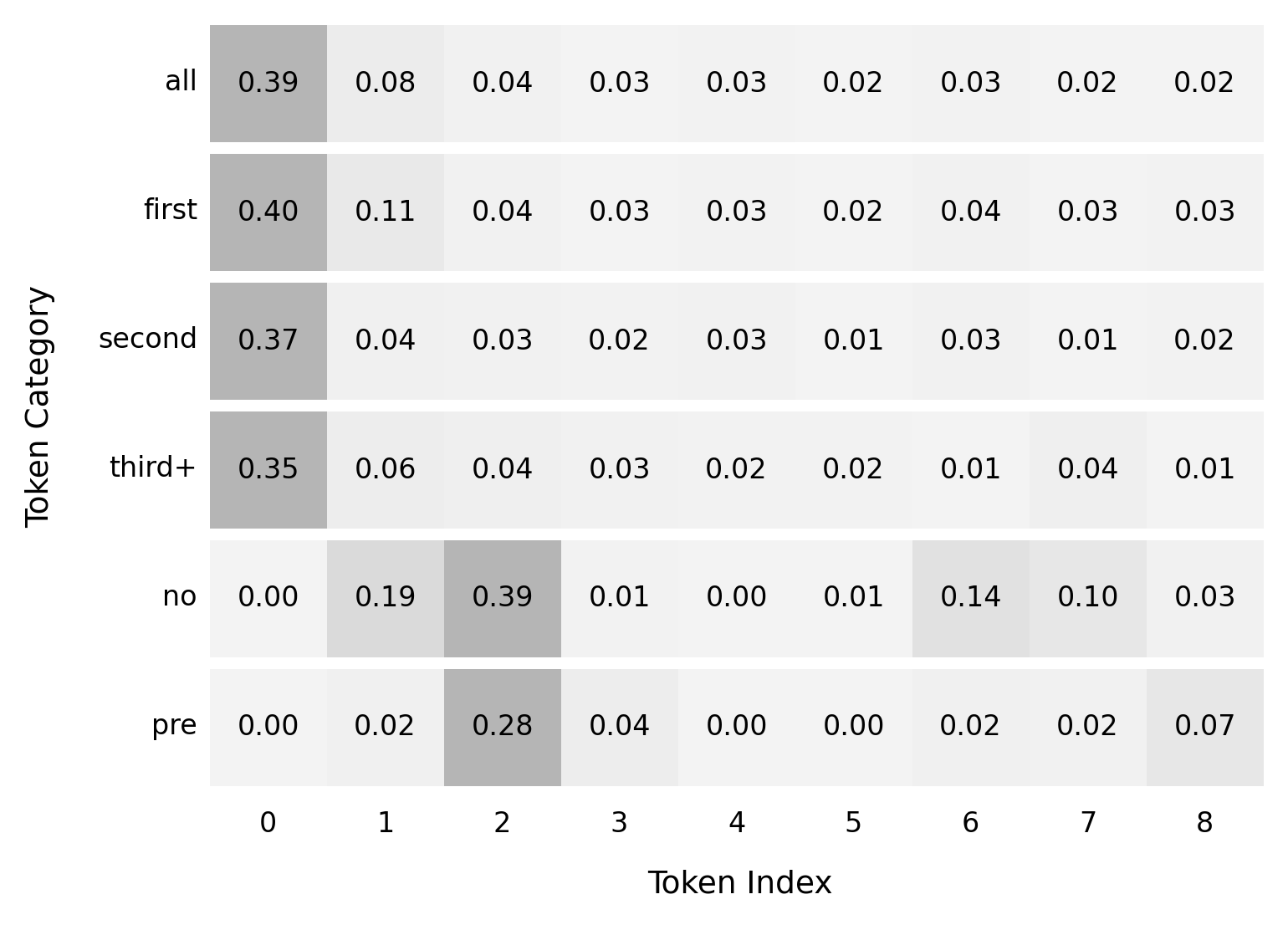}
        \caption{LLaMA-2-7B-chat}
        \label{mink:cat:e:60:llama-7b}
    \end{subfigure}
    \hfill
    \begin{subfigure}{0.49\textwidth}
        \centering
        \includegraphics[width=\linewidth]{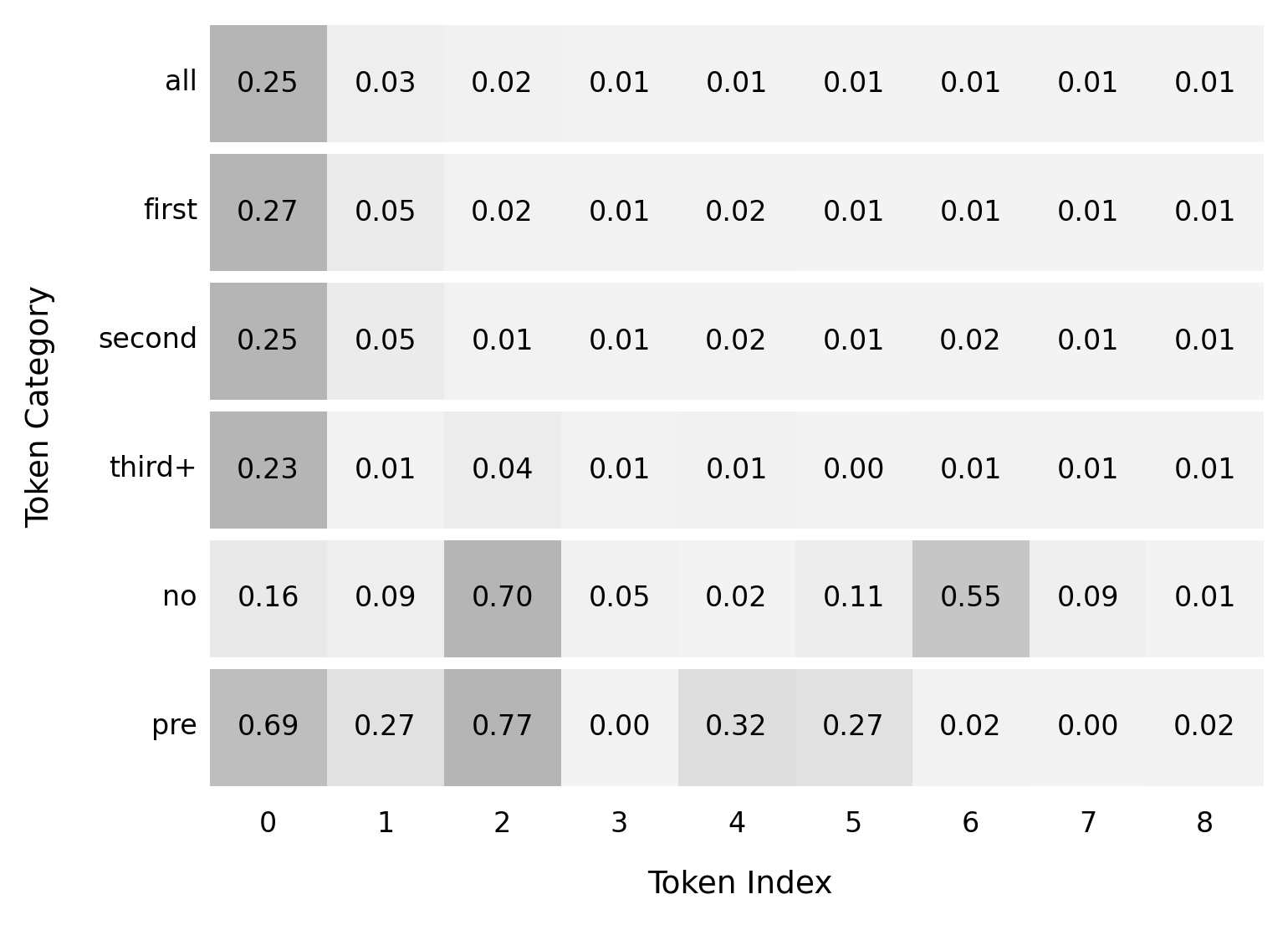}
        \caption{LLaMA-2-13B-chat}
        \label{mink:cat:e:60:llama-13b}
    \end{subfigure}
    \medskip
    \begin{subfigure}{0.49\textwidth}
        \centering
        \includegraphics[width=\linewidth]{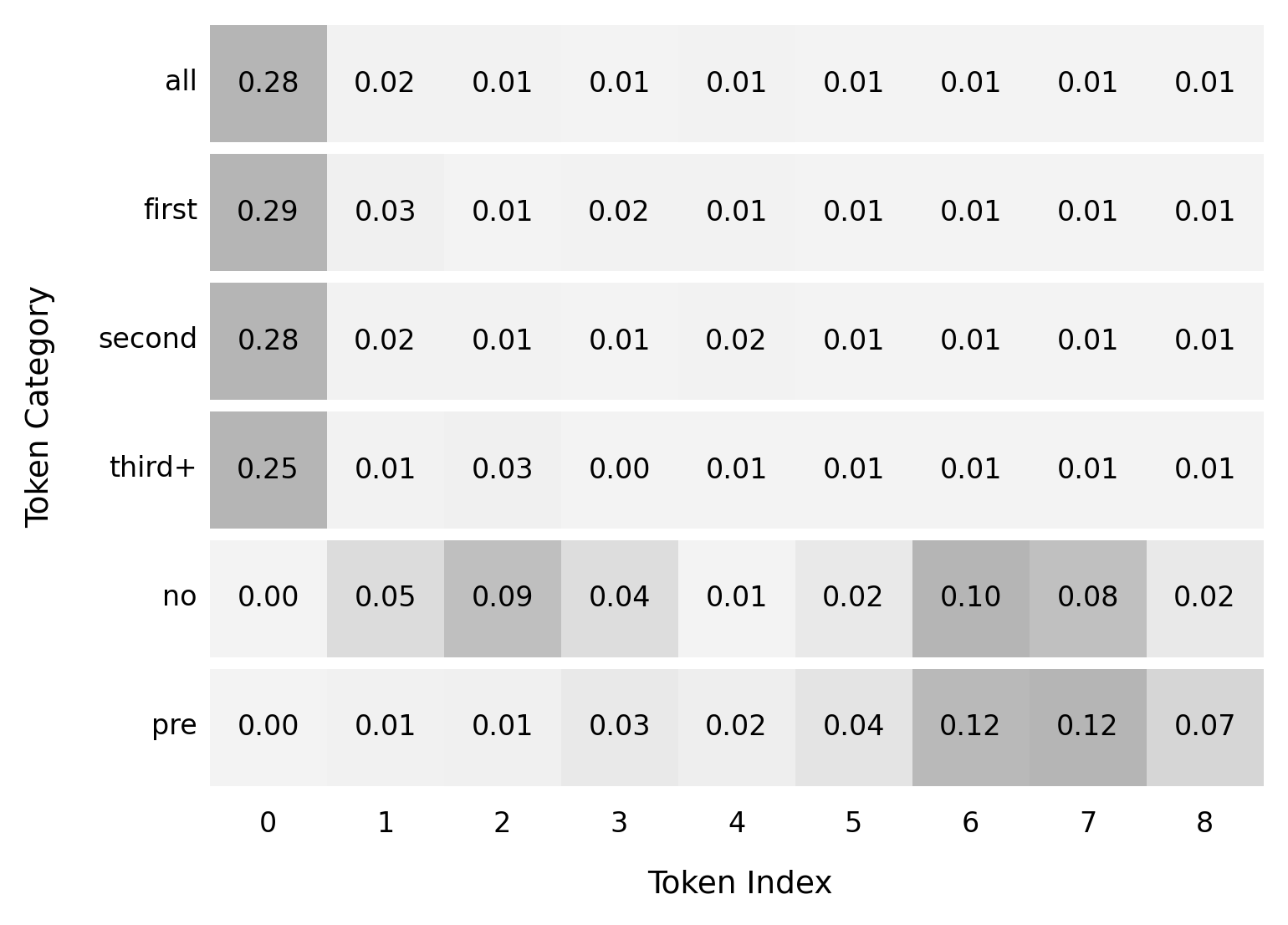}
        \caption{LLaMA-2-70B-chat}
        \label{mink:cat:e:60:llama-70b}
    \end{subfigure}
    \hfill
    \begin{subfigure}{0.49\textwidth}
        \centering
        \includegraphics[width=\linewidth]{plots/mink/categories/llama-2-70b-chat/mink-e_categories_60.png}
        \caption{Mistral-7B-instruct}
        \label{mink:cat:e:60:mistral-7b}
    \end{subfigure}
    \caption{\textbf{[60th percentile]} Min-K Entropy scores per token category and index, over the first 9 tokens at global level.}
    \label{mink:cat:e:60}
\end{figure}

\begin{figure}[htbp]
    \centering
    \begin{subfigure}{0.49\textwidth}
        \centering
        \includegraphics[width=\linewidth]{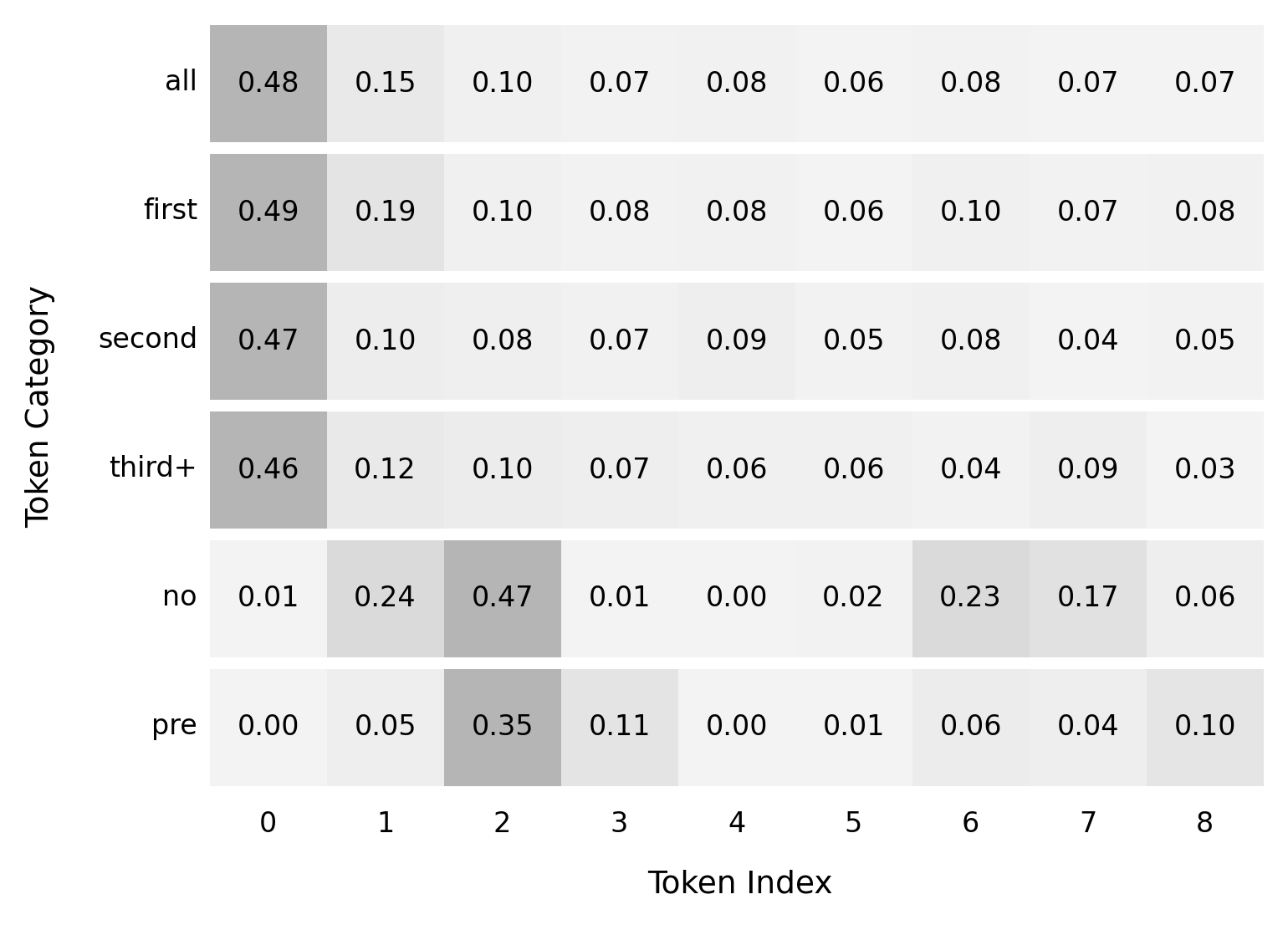}
        \caption{LLaMA-2-7B-chat}
        \label{mink:cat:e:70:llama-7b}
    \end{subfigure}
    \hfill
    \begin{subfigure}{0.49\textwidth}
        \centering
        \includegraphics[width=\linewidth]{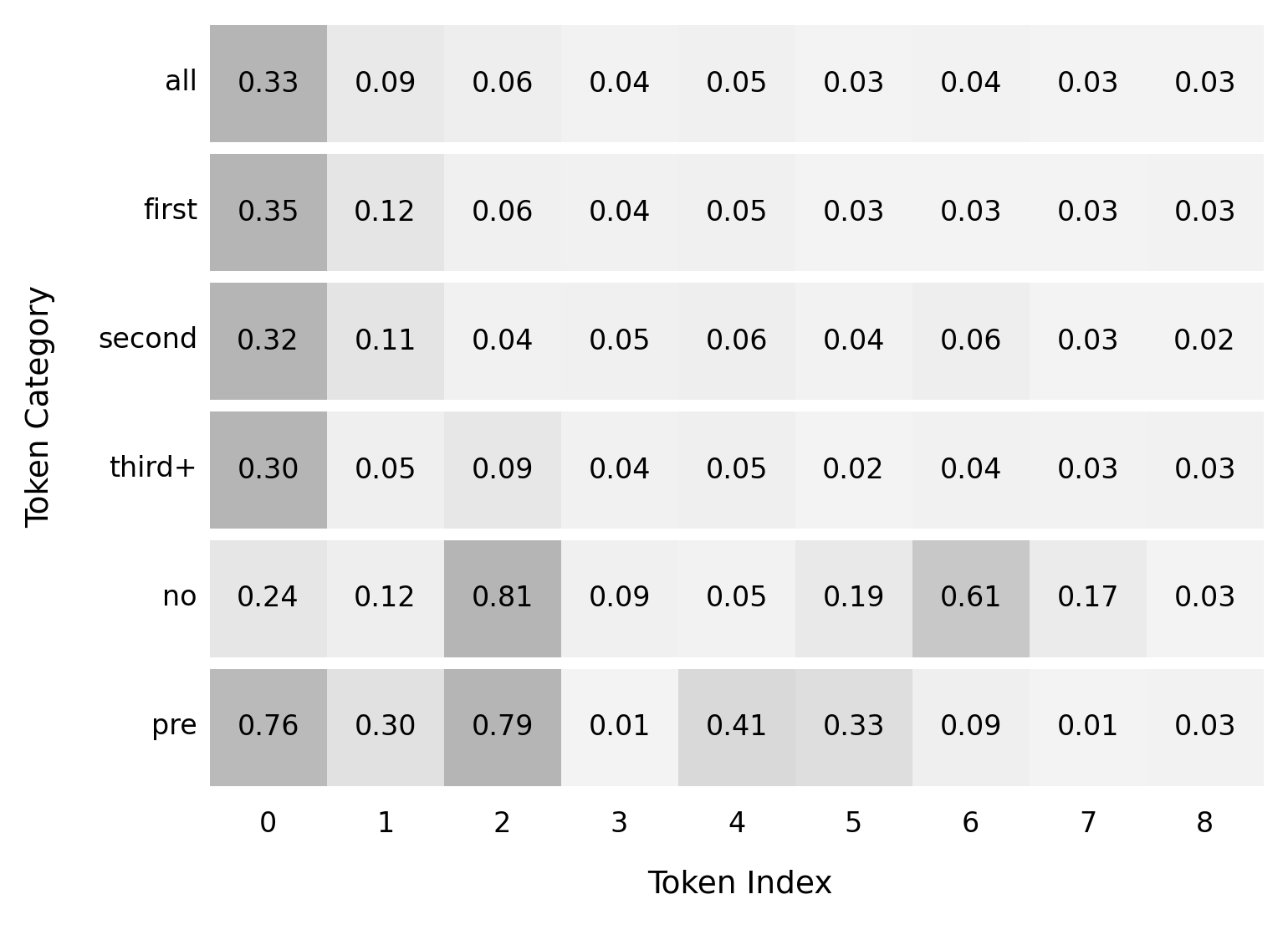}
        \caption{LLaMA-2-13B-chat}
        \label{mink:cat:e:70:llama-13b}
    \end{subfigure}
    \medskip
    \begin{subfigure}{0.49\textwidth}
        \centering
        \includegraphics[width=\linewidth]{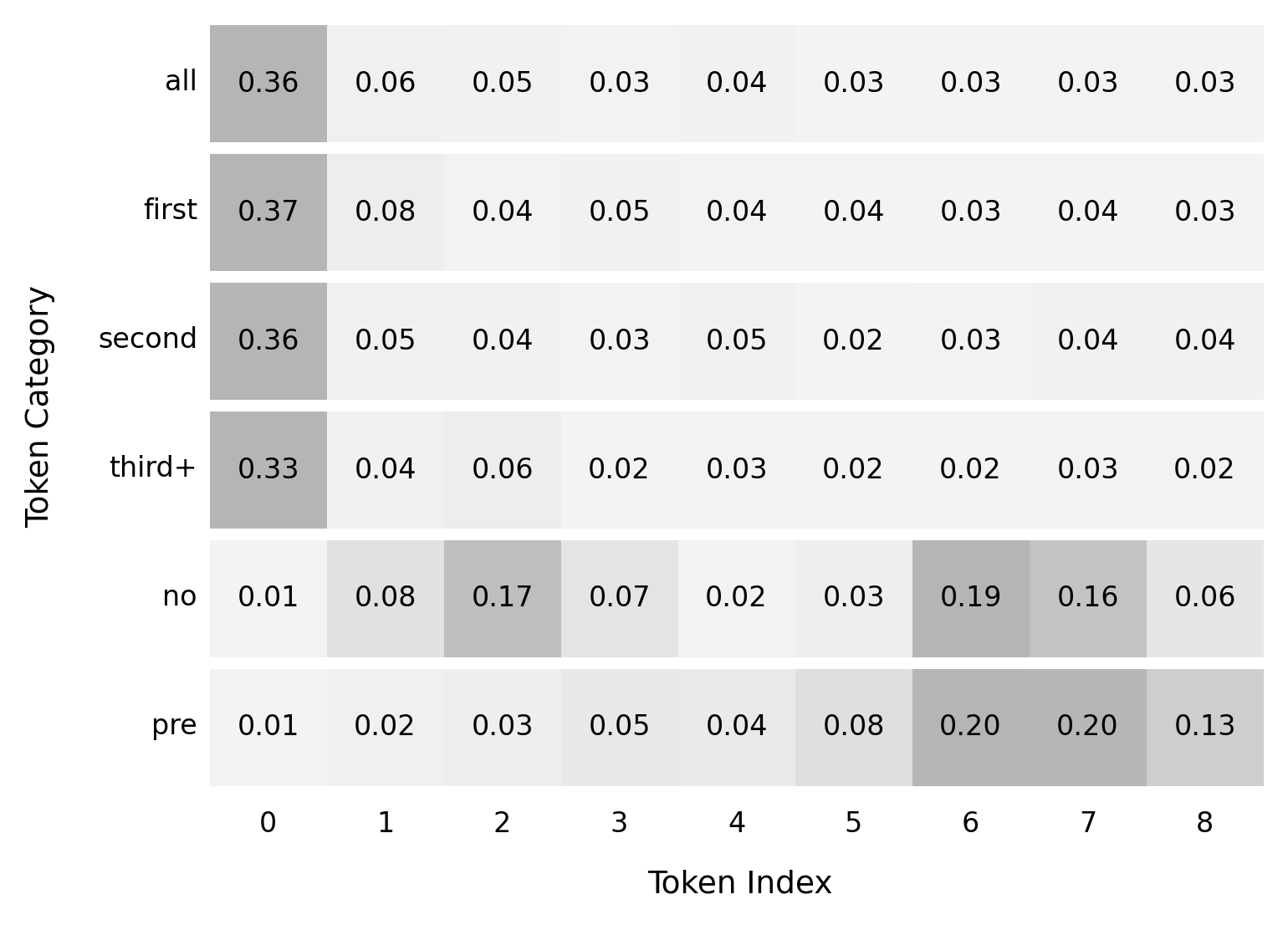}
        \caption{LLaMA-2-70B-chat}
        \label{mink:cat:e:70:llama-70b}
    \end{subfigure}
    \hfill
    \begin{subfigure}{0.49\textwidth}
        \centering
        \includegraphics[width=\linewidth]{plots/mink/categories/llama-2-70b-chat/mink-e_categories_70.png}
        \caption{Mistral-7B-instruct}
        \label{mink:cat:e:70:mistral-7b}
    \end{subfigure}
    \caption{\textbf{[70th percentile]} Min-K Entropy scores per token category and index, over the first 9 tokens at global level.}
    \label{mink:cat:e:70}
\end{figure}

\begin{figure}[htbp]
    \centering
    \begin{subfigure}{0.49\textwidth}
        \centering
        \includegraphics[width=\linewidth]{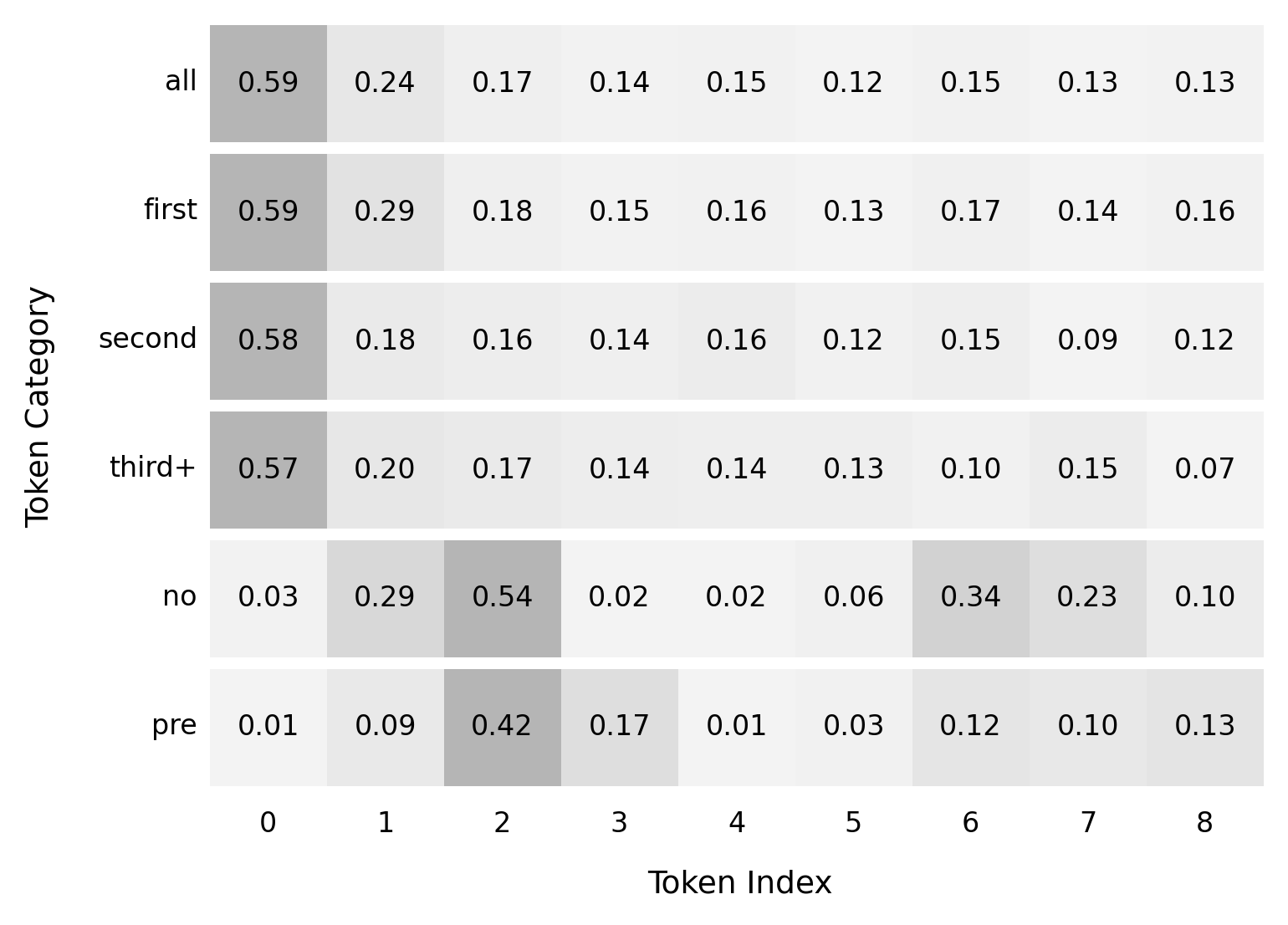}
        \caption{LLaMA-2-7B-chat}
        \label{mink:cat:e:80:llama-7b}
    \end{subfigure}
    \hfill
    \begin{subfigure}{0.49\textwidth}
        \centering
        \includegraphics[width=\linewidth]{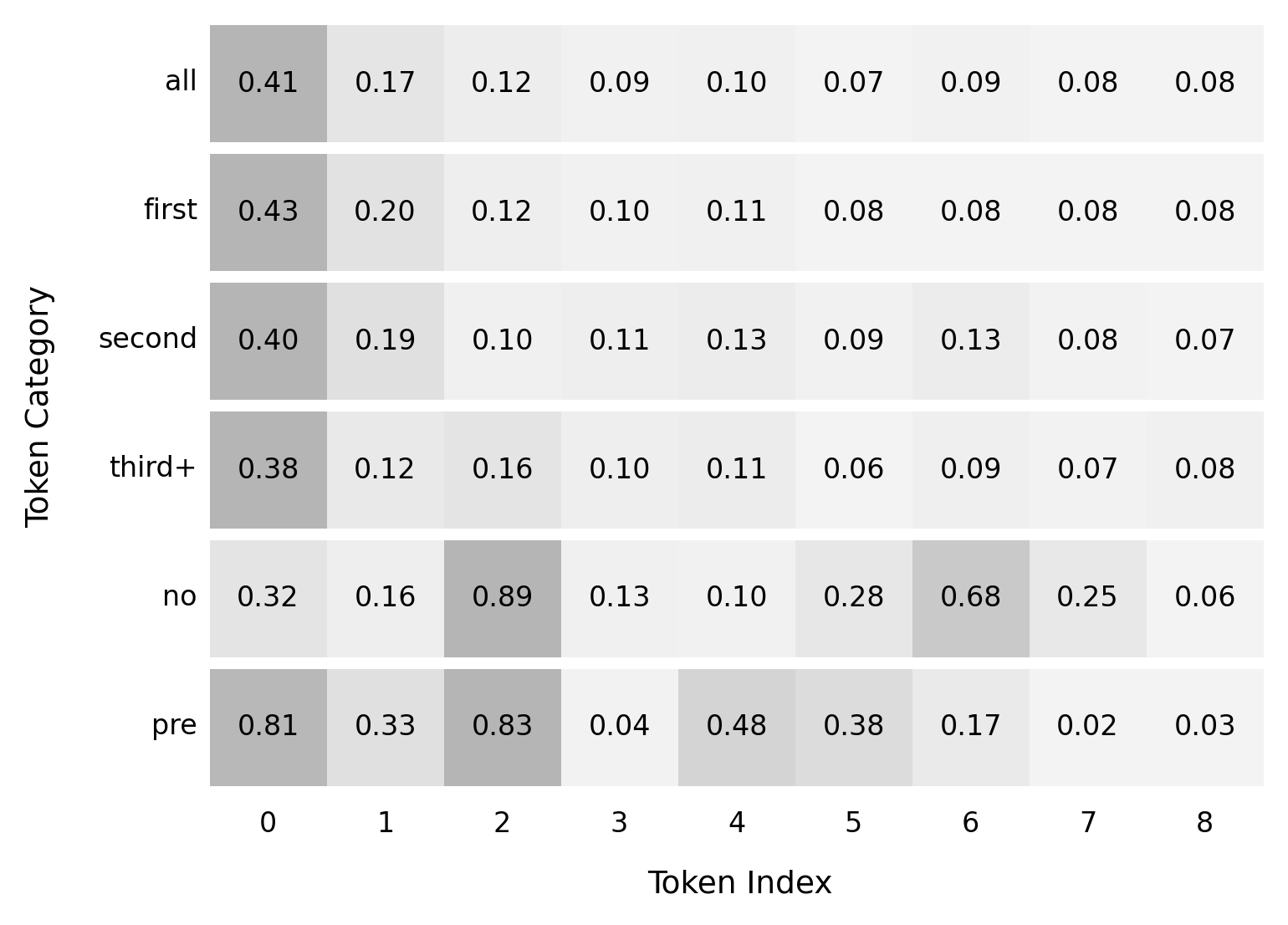}
        \caption{LLaMA-2-13B-chat}
        \label{mink:cat:e:80:llama-13b}
    \end{subfigure}
    \medskip
    \begin{subfigure}{0.49\textwidth}
        \centering
        \includegraphics[width=\linewidth]{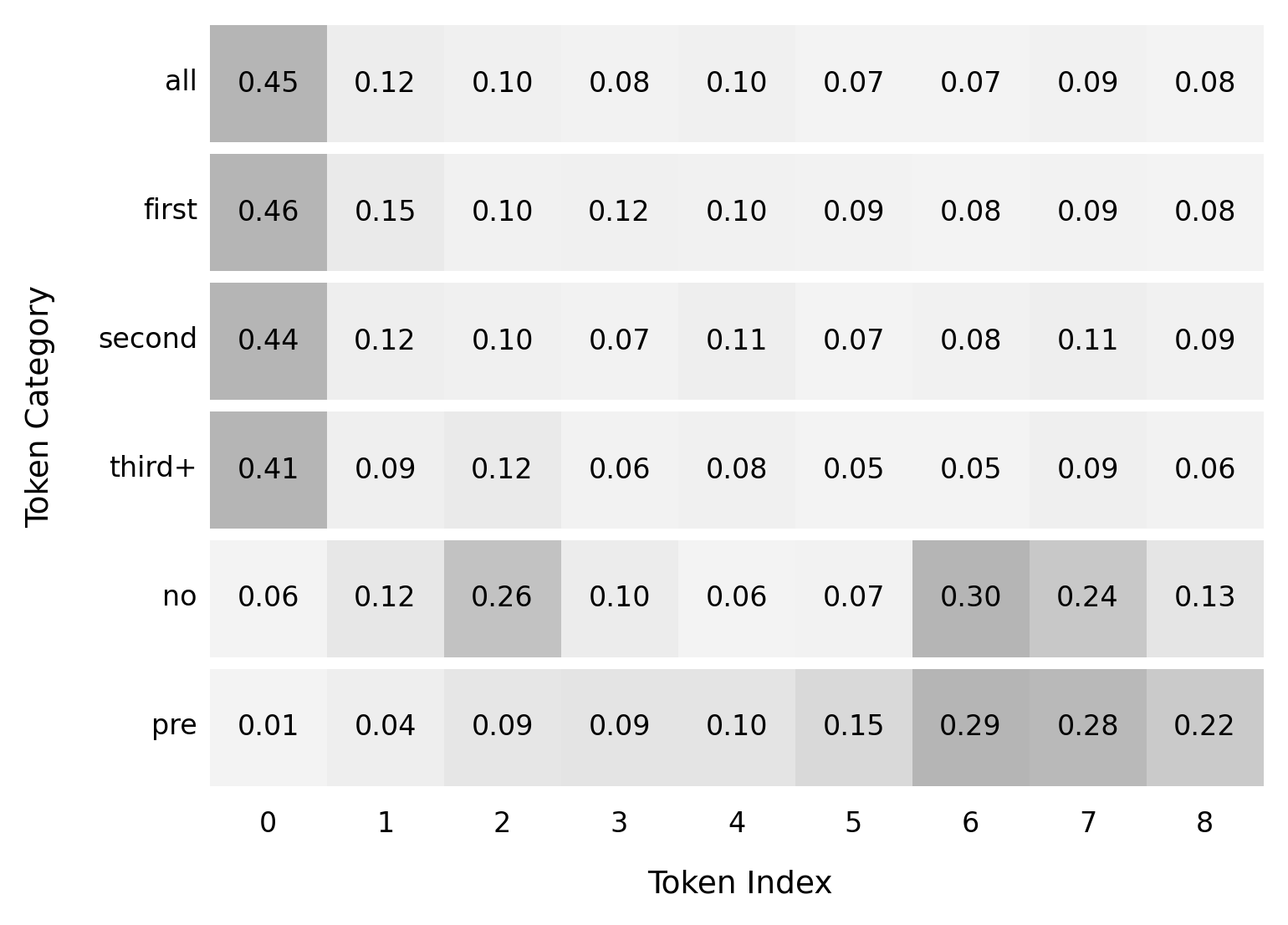}
        \caption{LLaMA-2-70B-chat}
        \label{mink:cat:e:80:llama-70b}
    \end{subfigure}
    \hfill
    \begin{subfigure}{0.49\textwidth}
        \centering
        \includegraphics[width=\linewidth]{plots/mink/categories/llama-2-70b-chat/mink-e_categories_80.png}
        \caption{Mistral-7B-instruct}
        \label{mink:cat:e:80:mistral-7b}
    \end{subfigure}
    \caption{\textbf{[80th percentile]} Min-K Entropy scores per token category and index, over the first 9 tokens at global level.}
    \label{mink:cat:e:80}
\end{figure}

\begin{figure}[htbp]
    \centering
    \begin{subfigure}{0.49\textwidth}
        \centering
        \includegraphics[width=\linewidth]{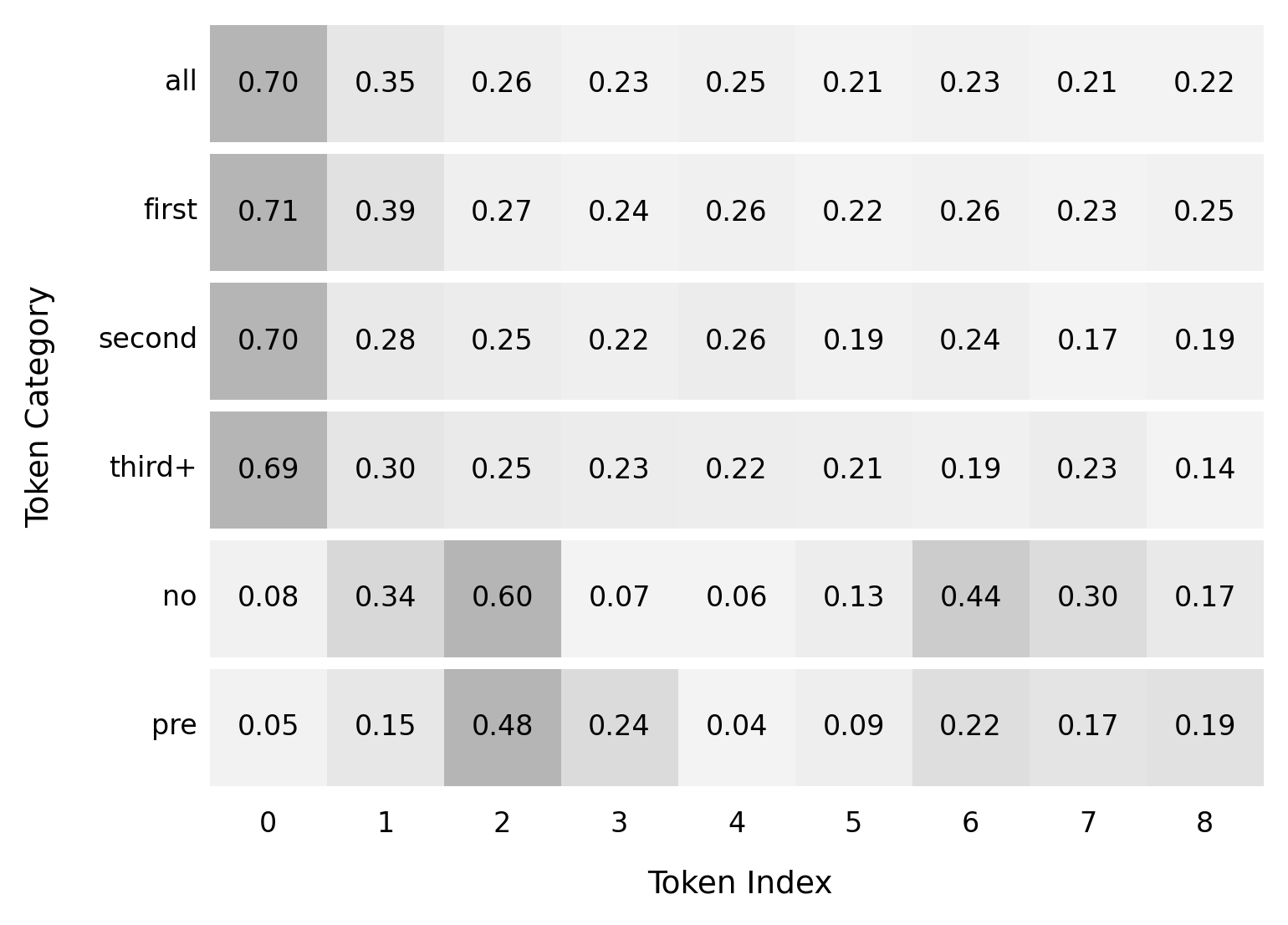}
        \caption{LLaMA-2-7B-chat}
        \label{mink:cat:e:90:llama-7b}
    \end{subfigure}
    \hfill
    \begin{subfigure}{0.49\textwidth}
        \centering
        \includegraphics[width=\linewidth]{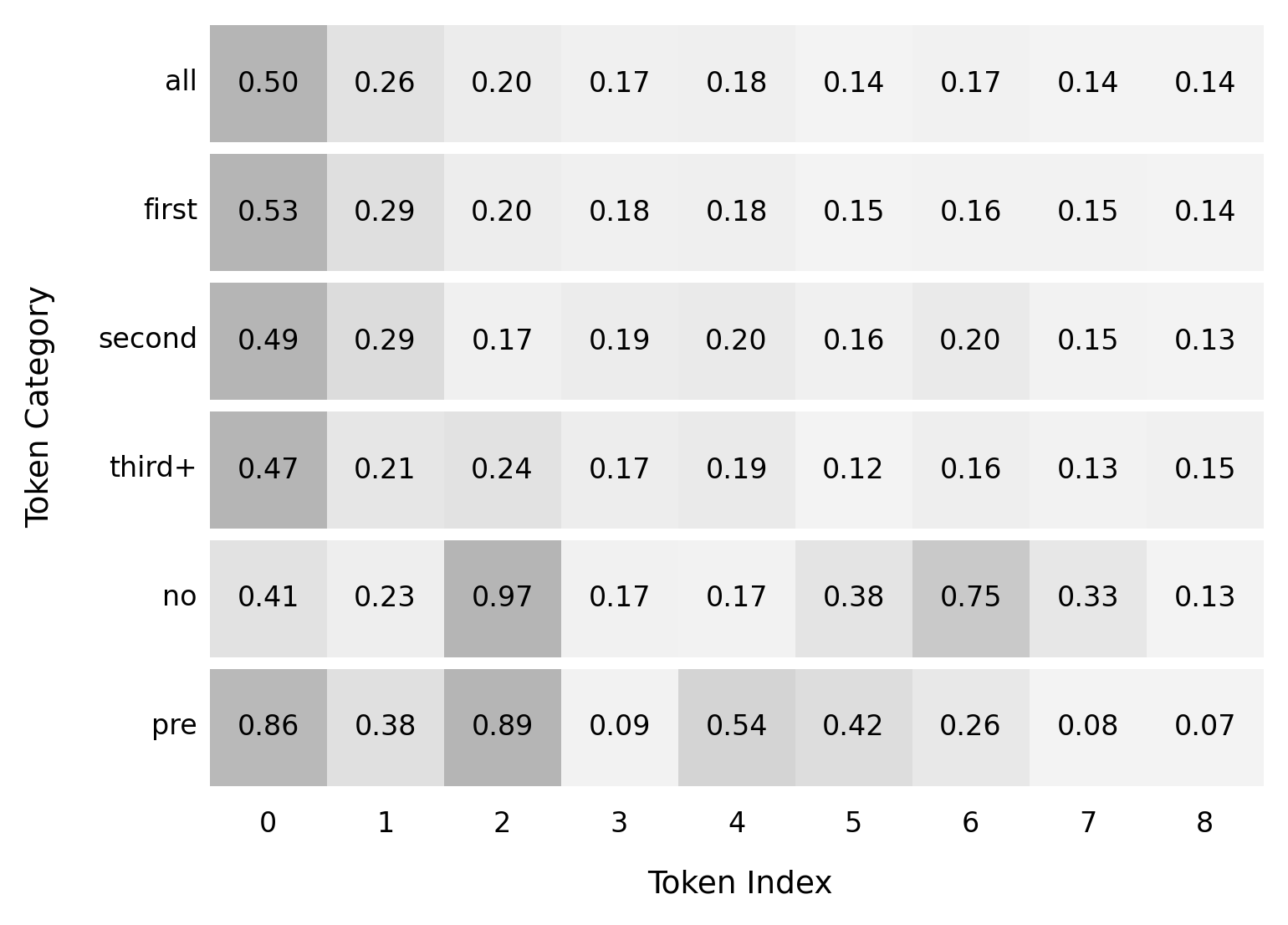}
        \caption{LLaMA-2-13B-chat}
        \label{mink:cat:e:90:llama-13b}
    \end{subfigure}
    \medskip
    \begin{subfigure}{0.49\textwidth}
        \centering
        \includegraphics[width=\linewidth]{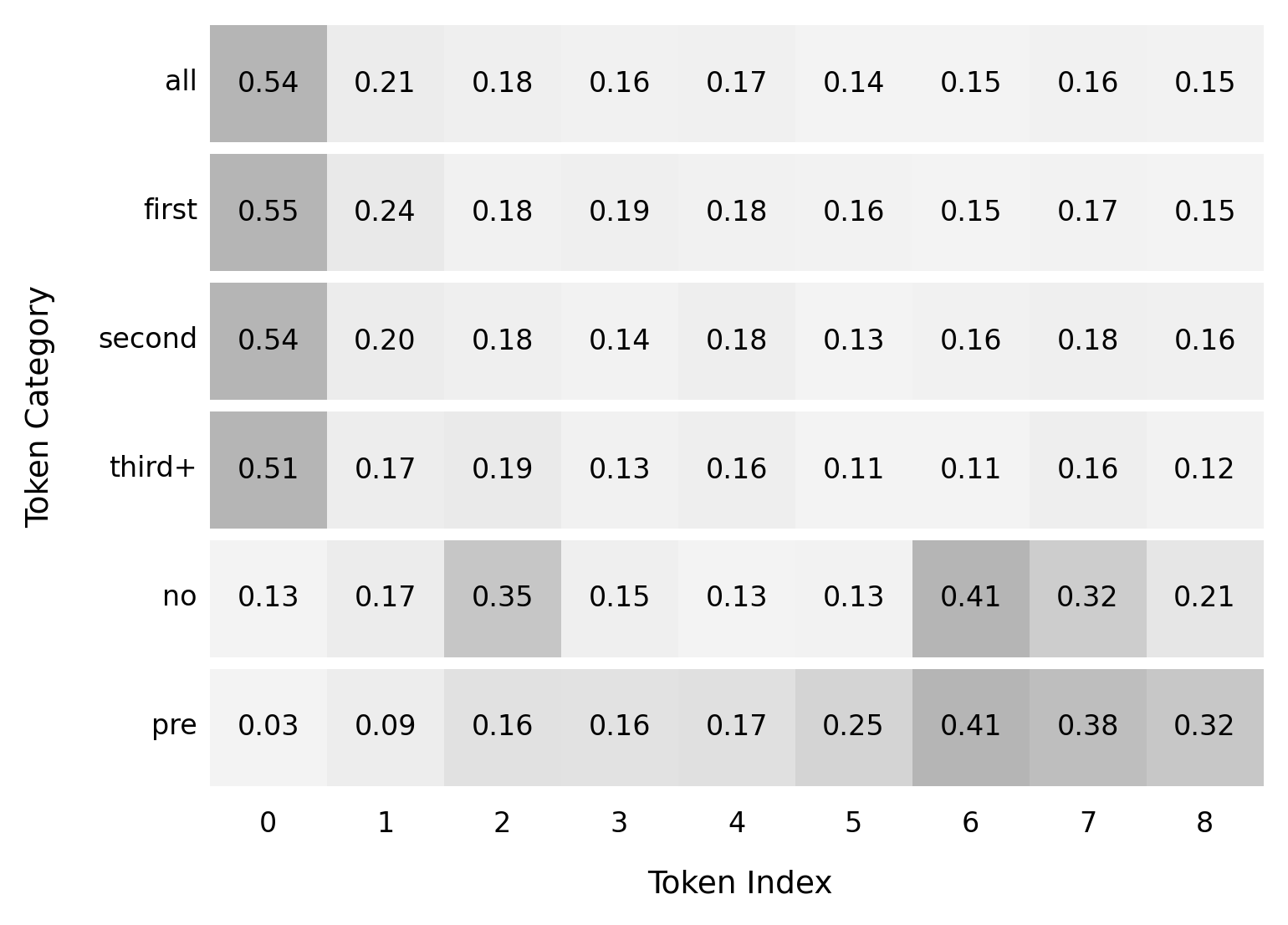}
        \caption{LLaMA-2-70B-chat}
        \label{mink:cat:e:90:llama-70b}
    \end{subfigure}
    \hfill
    \begin{subfigure}{0.49\textwidth}
        \centering
        \includegraphics[width=\linewidth]{plots/mink/categories/llama-2-70b-chat/mink-e_categories_90.png}
        \caption{Mistral-7B-instruct}
        \label{mink:cat:e:90:mistral-7b}
    \end{subfigure}
    \caption{\textbf{[90th percentile]} Min-K Entropy scores per token category and index, over the first 9 tokens at global level.}
    \label{mink:cat:e:90}
\end{figure}

\begin{figure}[htbp]
    \centering
    \begin{subfigure}{0.49\textwidth}
        \centering
        \includegraphics[width=\linewidth]{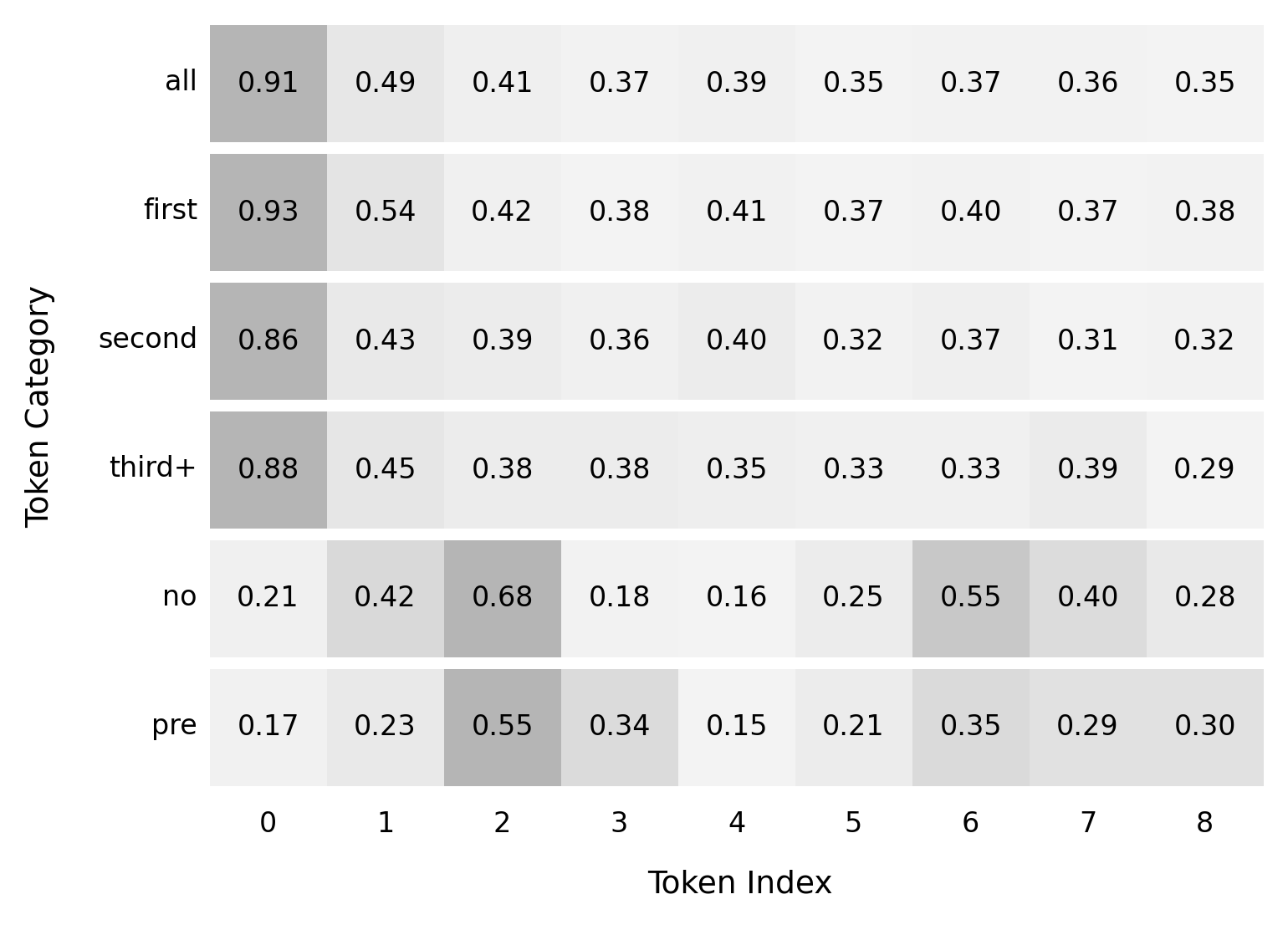}
        \caption{LLaMA-2-7B-chat}
        \label{mink:cat:e:100:llama-7b}
    \end{subfigure}
    \hfill
    \begin{subfigure}{0.49\textwidth}
        \centering
        \includegraphics[width=\linewidth]{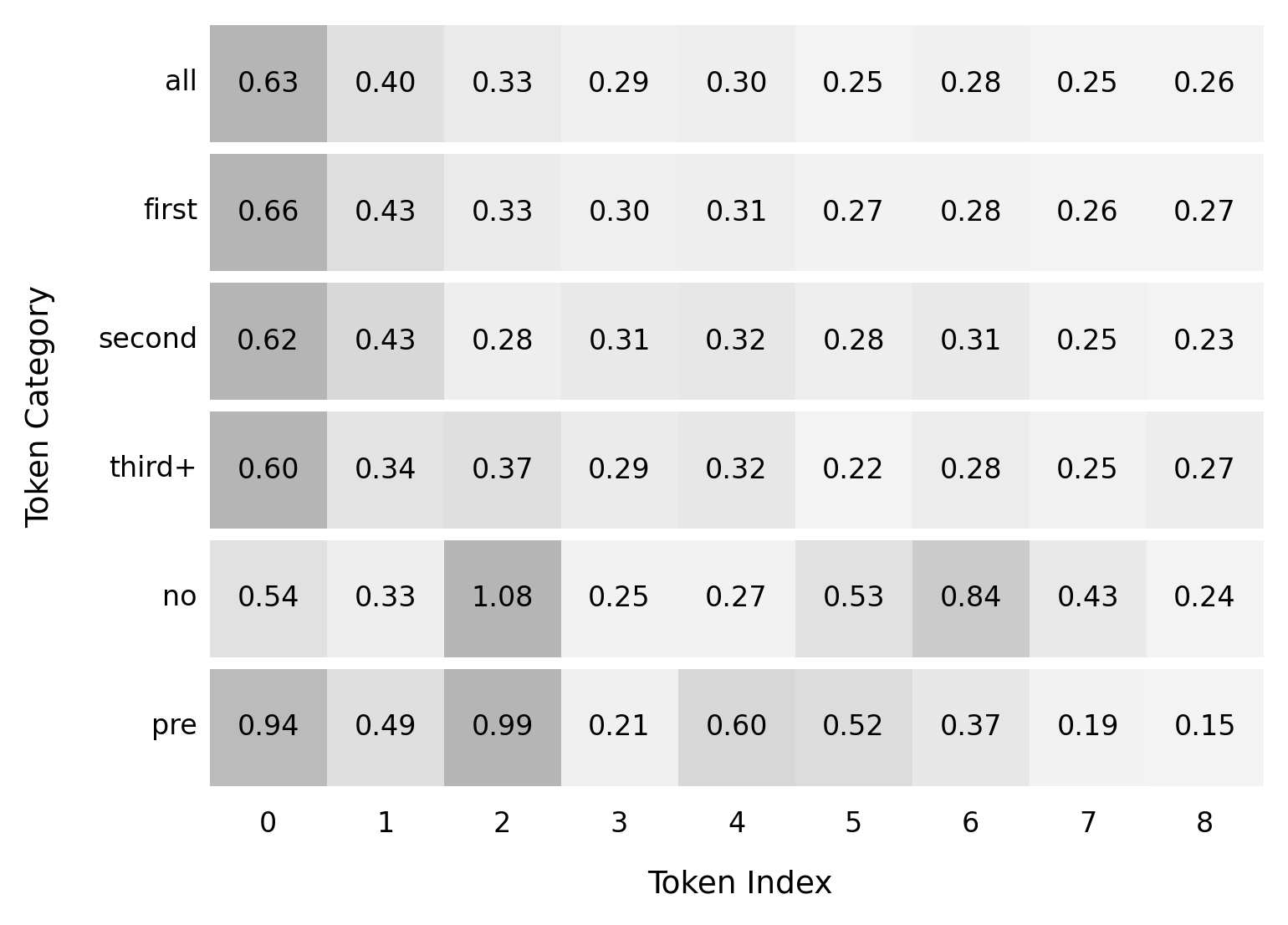}
        \caption{LLaMA-2-13B-chat}
        \label{mink:cat:e:100:llama-13b}
    \end{subfigure}
    \medskip
    \begin{subfigure}{0.49\textwidth}
        \centering
        \includegraphics[width=\linewidth]{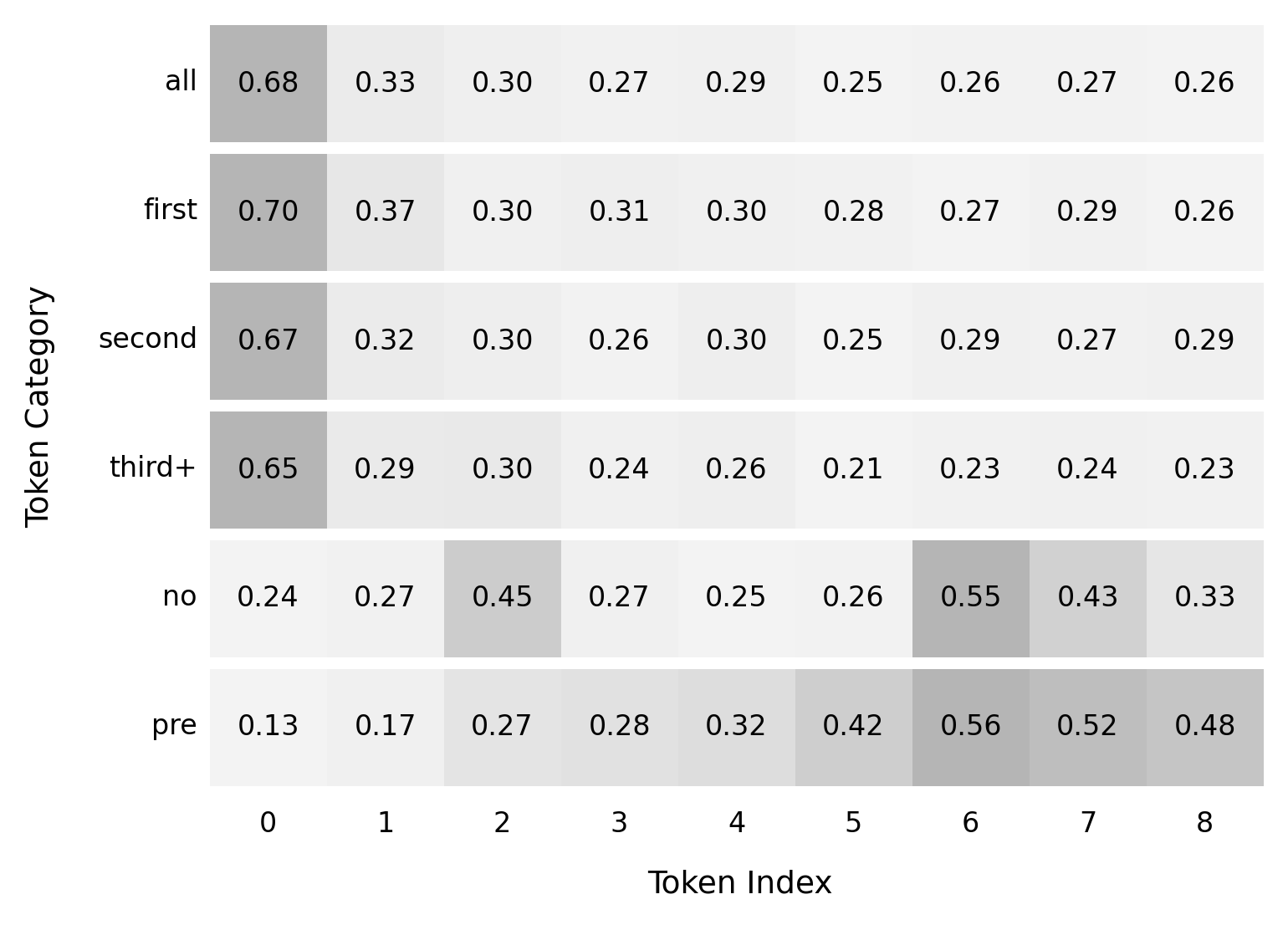}
        \caption{LLaMA-2-70B-chat}
        \label{mink:cat:e:100:llama-70b}
    \end{subfigure}
    \hfill
    \begin{subfigure}{0.49\textwidth}
        \centering
        \includegraphics[width=\linewidth]{plots/mink/categories/llama-2-70b-chat/mink-e_categories_100.png}
        \caption{Mistral-7B-instruct}
        \label{mink:cat:e:100:mistral-7b}
    \end{subfigure}
    \caption{\textbf{[100th percentile]} Min-K Entropy scores per token category and index, over the first 9 tokens at global level.}
    \label{mink:cat:e:100}
\end{figure}

\FloatBarrier

\subsection{Min-K Across All Percentiles}
\label{plt:mink:perc}

\subsubsection{Min-K Probability}
\label{plt:mink:perc:prob}
\begin{figure}[!hp]
    \centering
    \begin{subfigure}{0.45\textwidth}
        \centering
        \includegraphics[width=\linewidth]{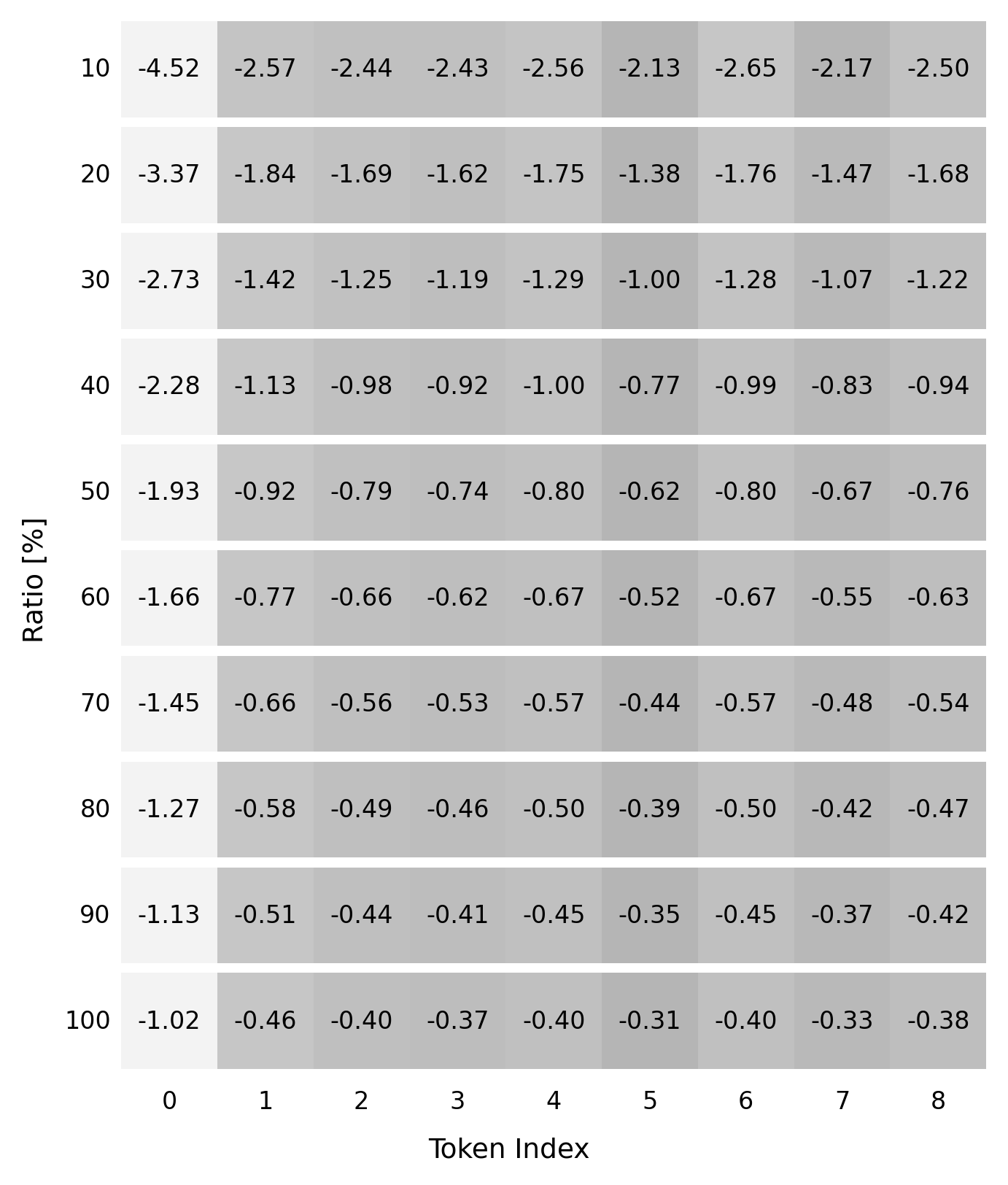}
        \caption{LLaMA-2-7B-chat}
        \label{mink:princ:p:all:llama-7b}
    \end{subfigure}
    \hfill
    \begin{subfigure}{0.45\textwidth}
        \centering
        \includegraphics[width=\linewidth]{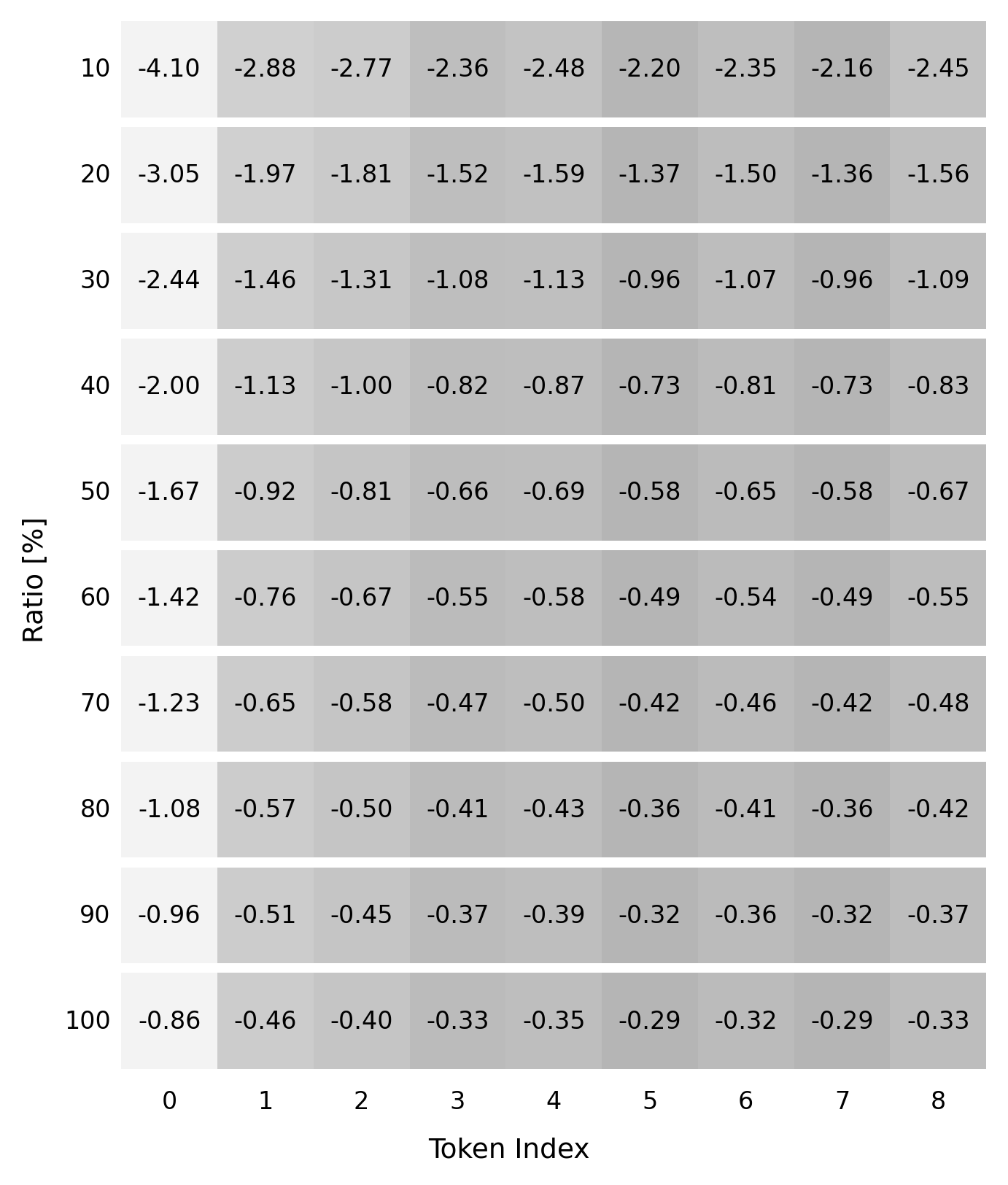}
        \caption{LLaMA-2-13B-chat}
        \label{mink:princ:p:all:llama-13b}
    \end{subfigure}
    \medskip
    \begin{subfigure}{0.45\textwidth}
        \centering
        \includegraphics[width=\linewidth]{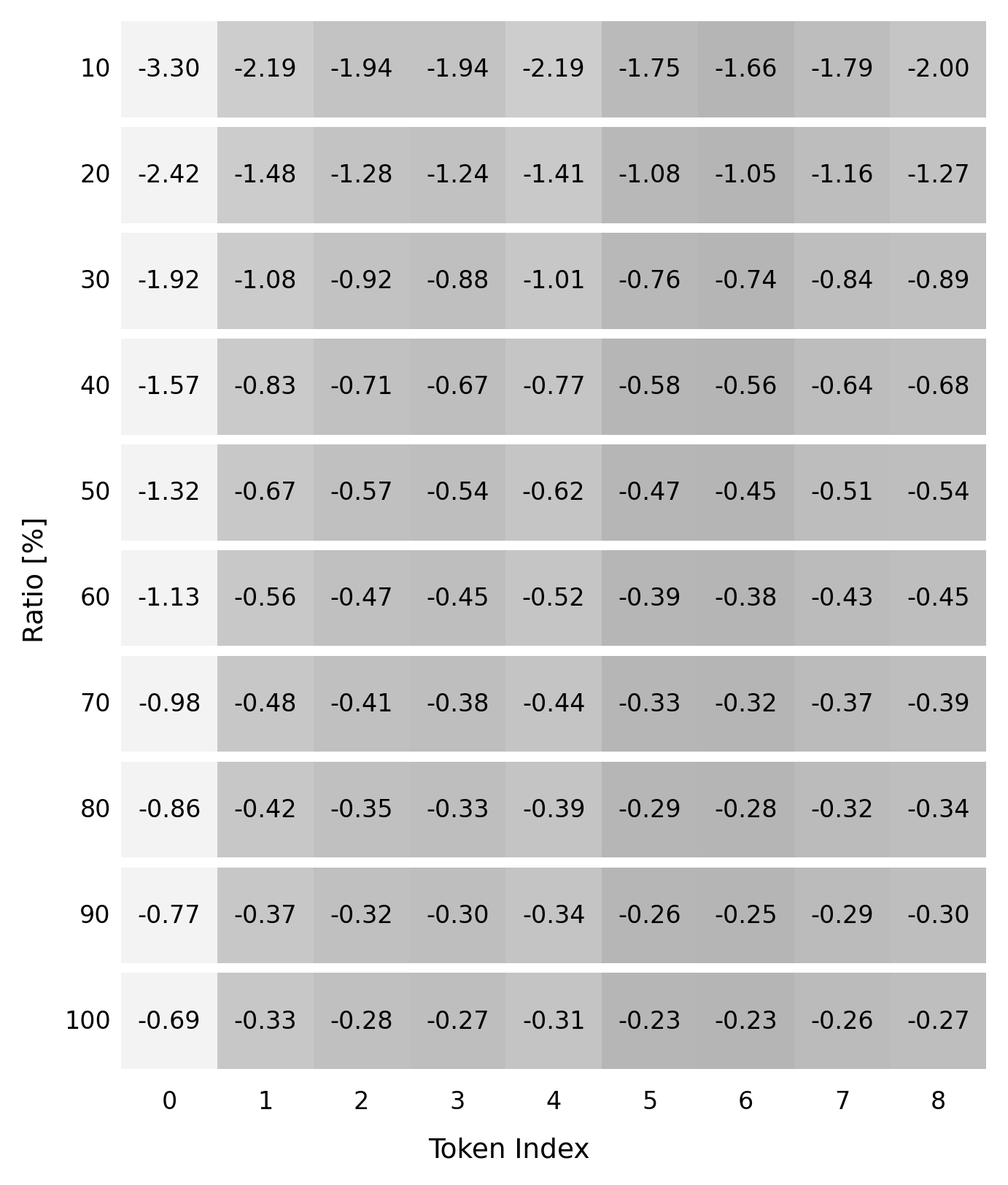}
        \caption{LLaMA-2-70B-chat}
        \label{mink:princ:p:all:llama-70b}
    \end{subfigure}
    \hfill
    \begin{subfigure}{0.45\textwidth}
        \centering
        \includegraphics[width=\linewidth]{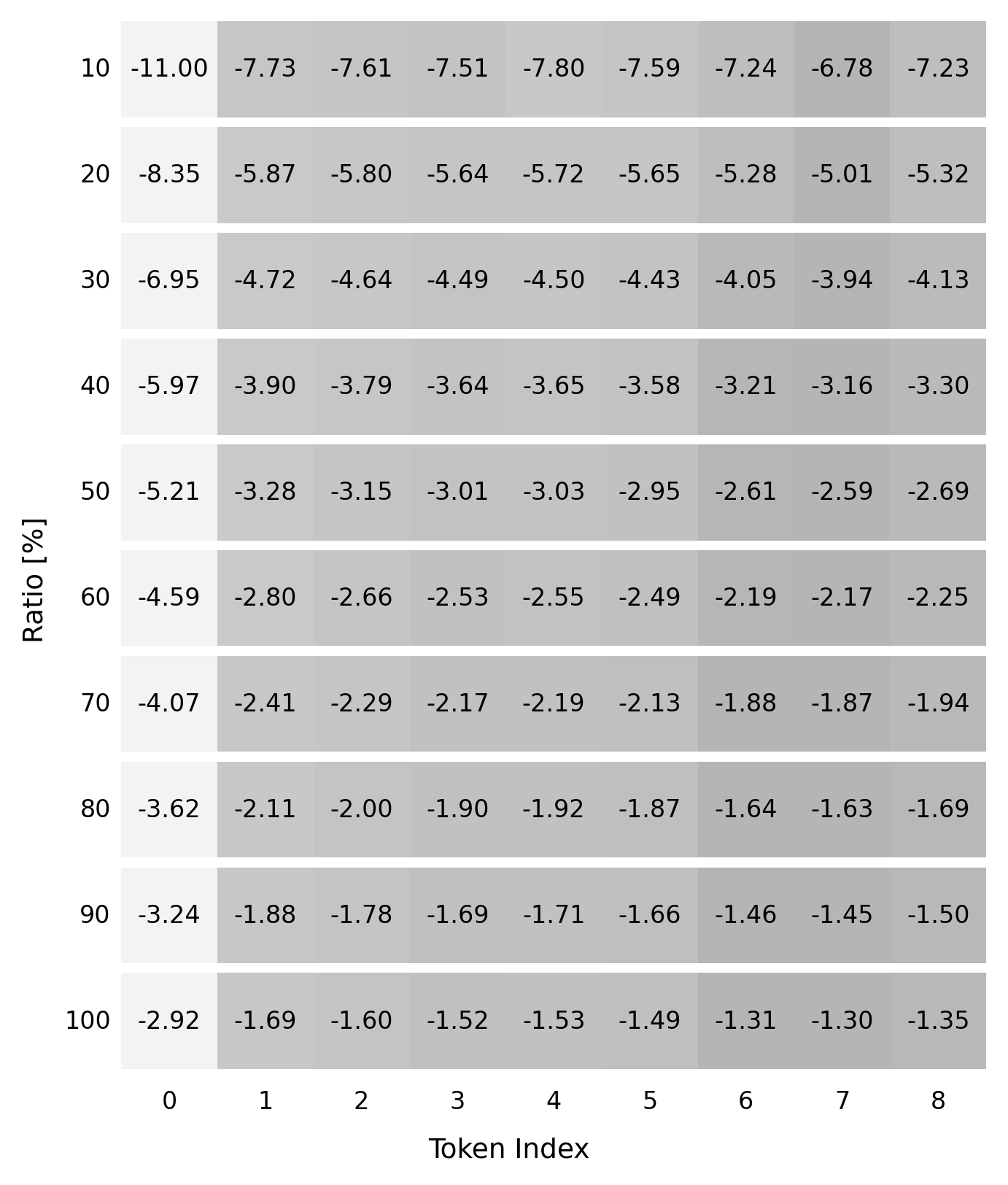}
        \caption{Mistral-7B-instruct}
        \label{mink:princ:p:all:mistral-7b}
    \end{subfigure}
    \caption{\textbf{[all]} Min-K Probability scores across all percentiles over the first 9 tokens from all hallucination spans at global level.}
    \label{mink:princ:p:all}
\end{figure}

\begin{figure}[htbp]
    \centering
    \begin{subfigure}{0.49\textwidth}
        \centering
        \includegraphics[width=\linewidth]{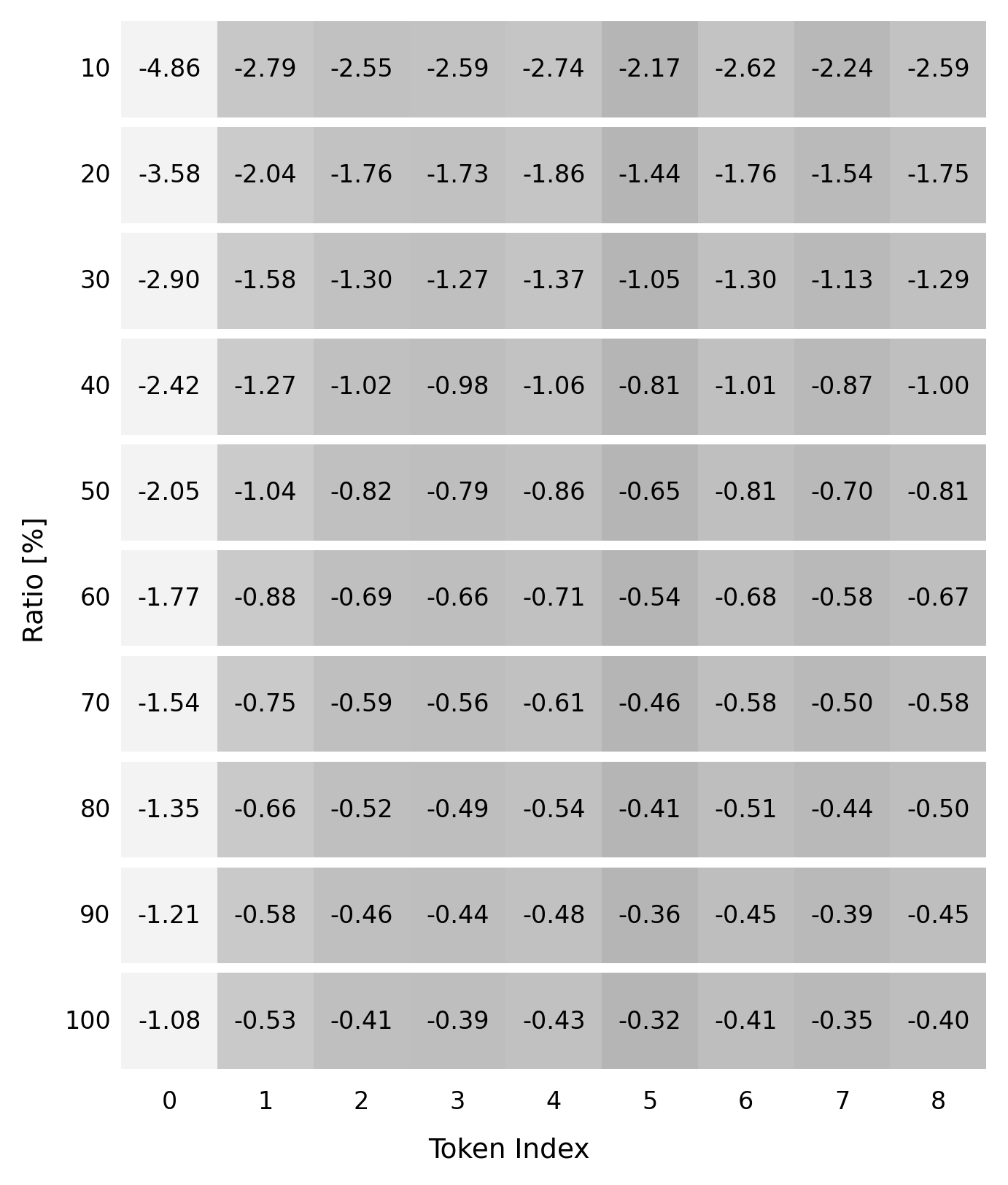}
        \caption{LLaMA-2-7B-chat}
        \label{mink:princ:p:first:llama-7b}
    \end{subfigure}
    \hfill
    \begin{subfigure}{0.49\textwidth}
        \centering
        \includegraphics[width=\linewidth]{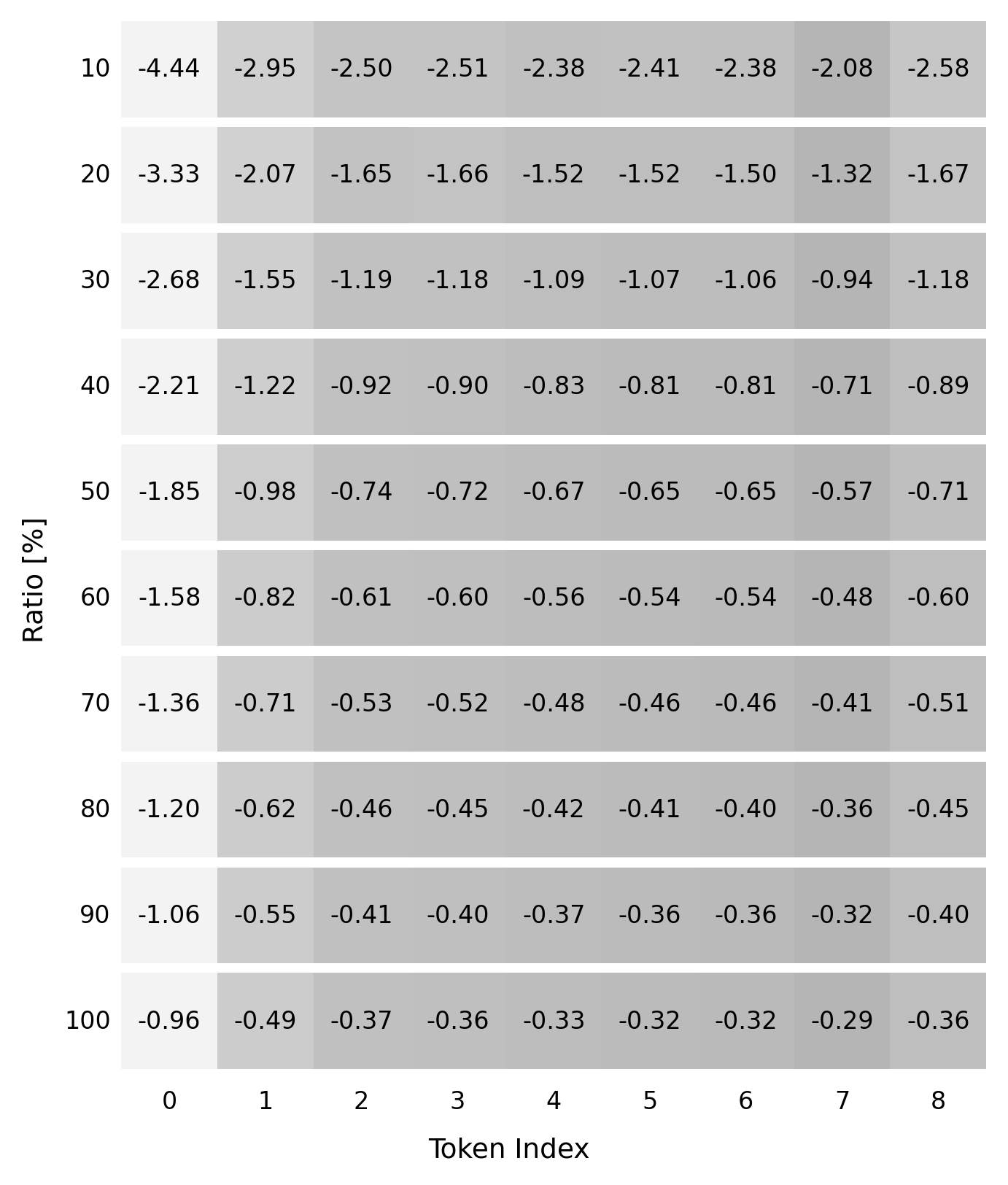}
        \caption{LLaMA-2-13B-chat}
        \label{mink:princ:p:first:llama-13b}
    \end{subfigure}
    \medskip
    \begin{subfigure}{0.49\textwidth}
        \centering
        \includegraphics[width=\linewidth]{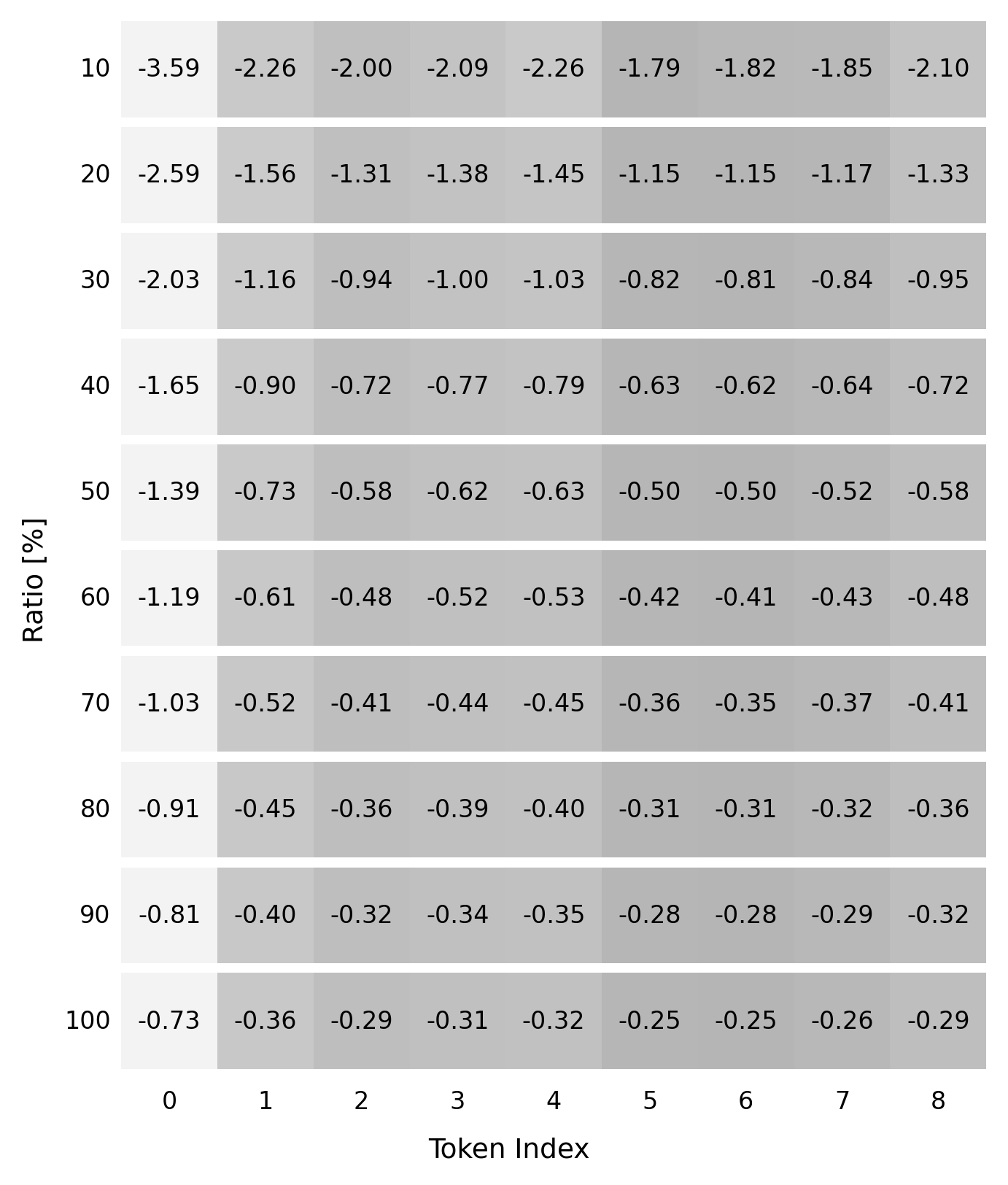}
        \caption{LLaMA-2-70B-chat}
        \label{mink:princ:p:first:llama-70b}
    \end{subfigure}
    \hfill
    \begin{subfigure}{0.49\textwidth}
        \centering
        \includegraphics[width=\linewidth]{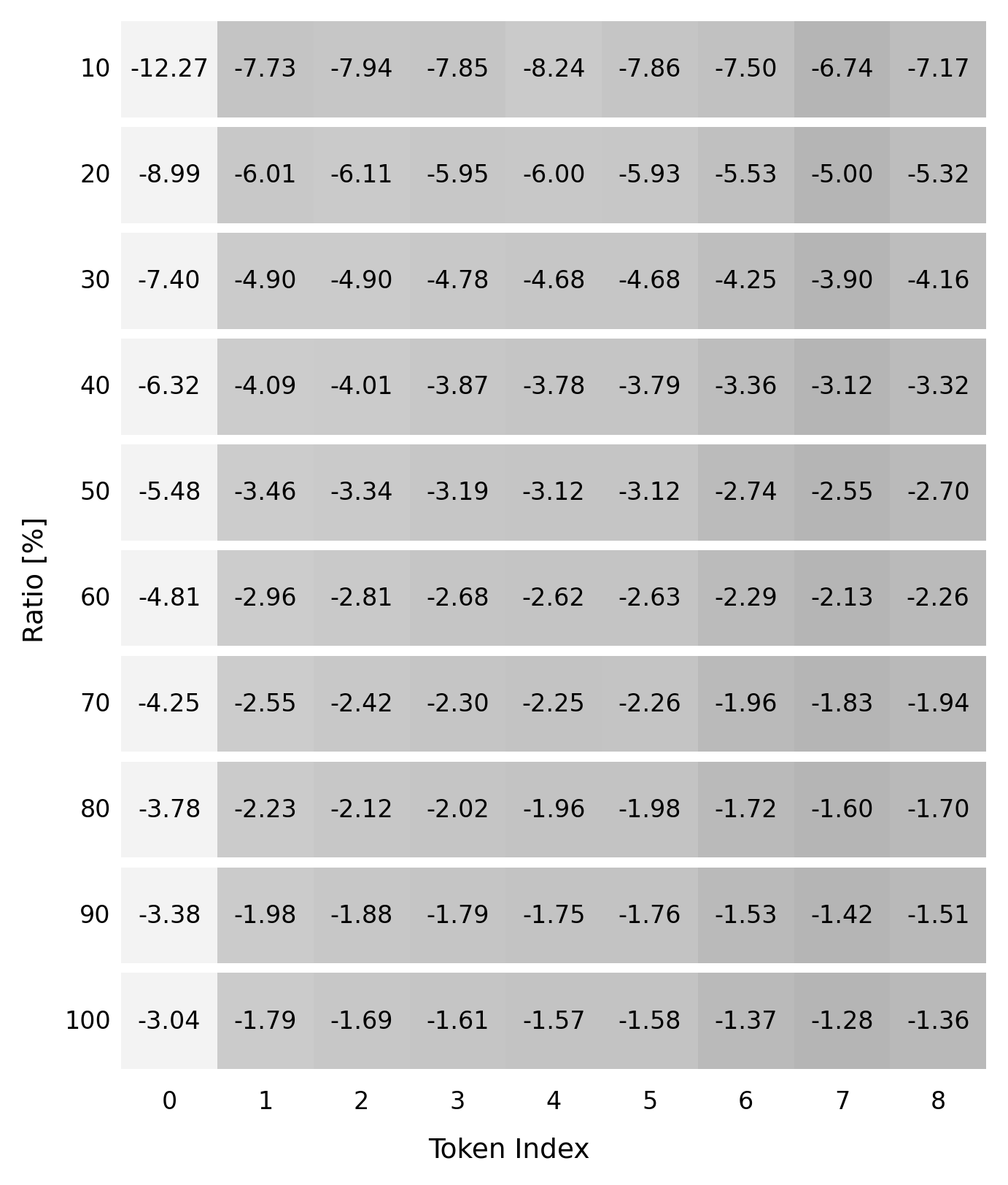}
        \caption{Mistral-7B-instruct}
        \label{mink:princ:p:first:mistral-7b}
    \end{subfigure}
    \caption{\textbf{[first]} Min-K Probability scores across all percentiles over the first 9 tokens from first hallucination spans at global level.}
    \label{mink:princ:p:first}
\end{figure}

\begin{figure}[htbp]
    \centering
    \begin{subfigure}{0.49\textwidth}
        \centering
        \includegraphics[width=\linewidth]{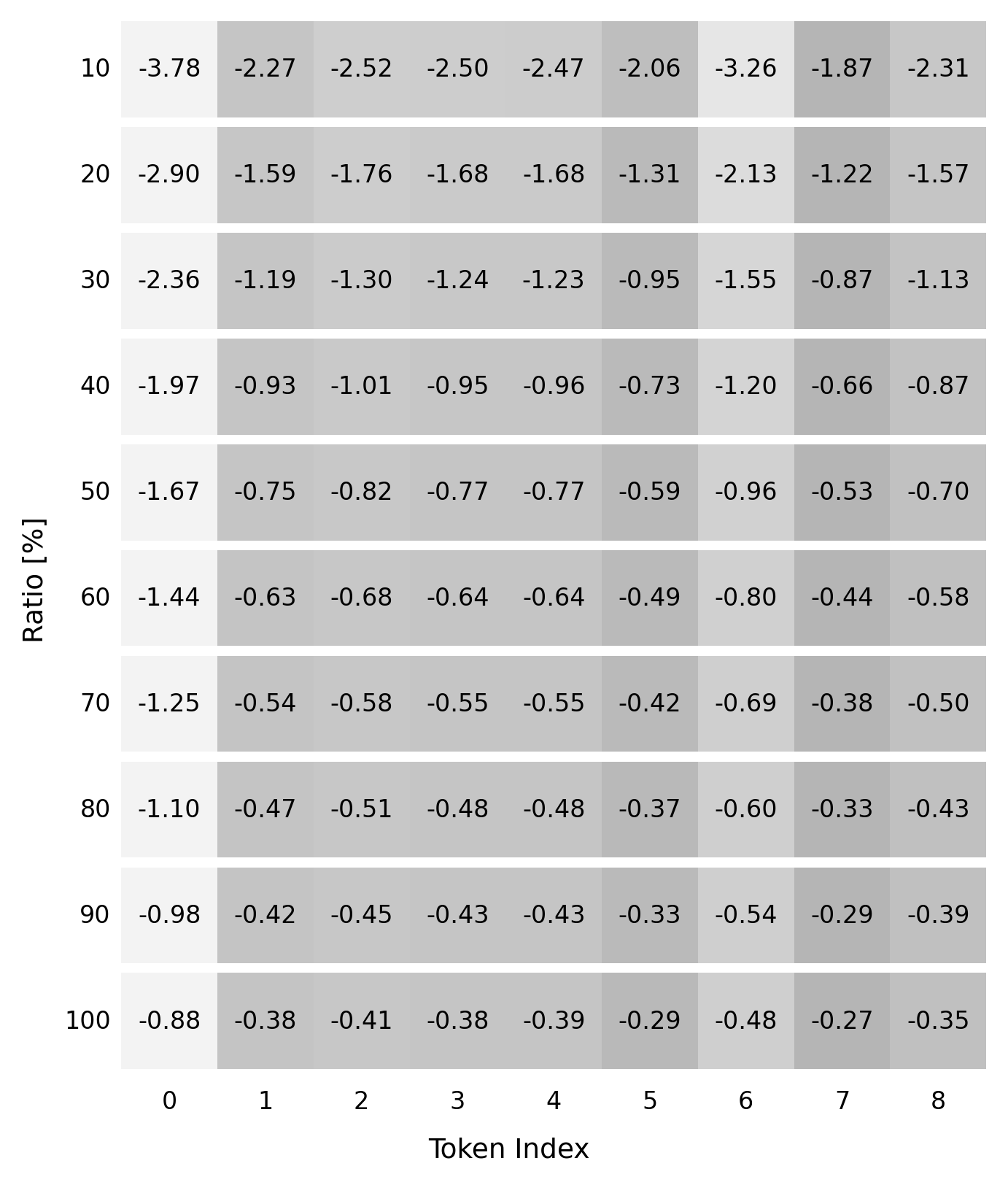}
        \caption{LLaMA-2-7B-chat}
        \label{mink:princ:p:second:llama-7b}
    \end{subfigure}
    \hfill
    \begin{subfigure}{0.49\textwidth}
        \centering
        \includegraphics[width=\linewidth]{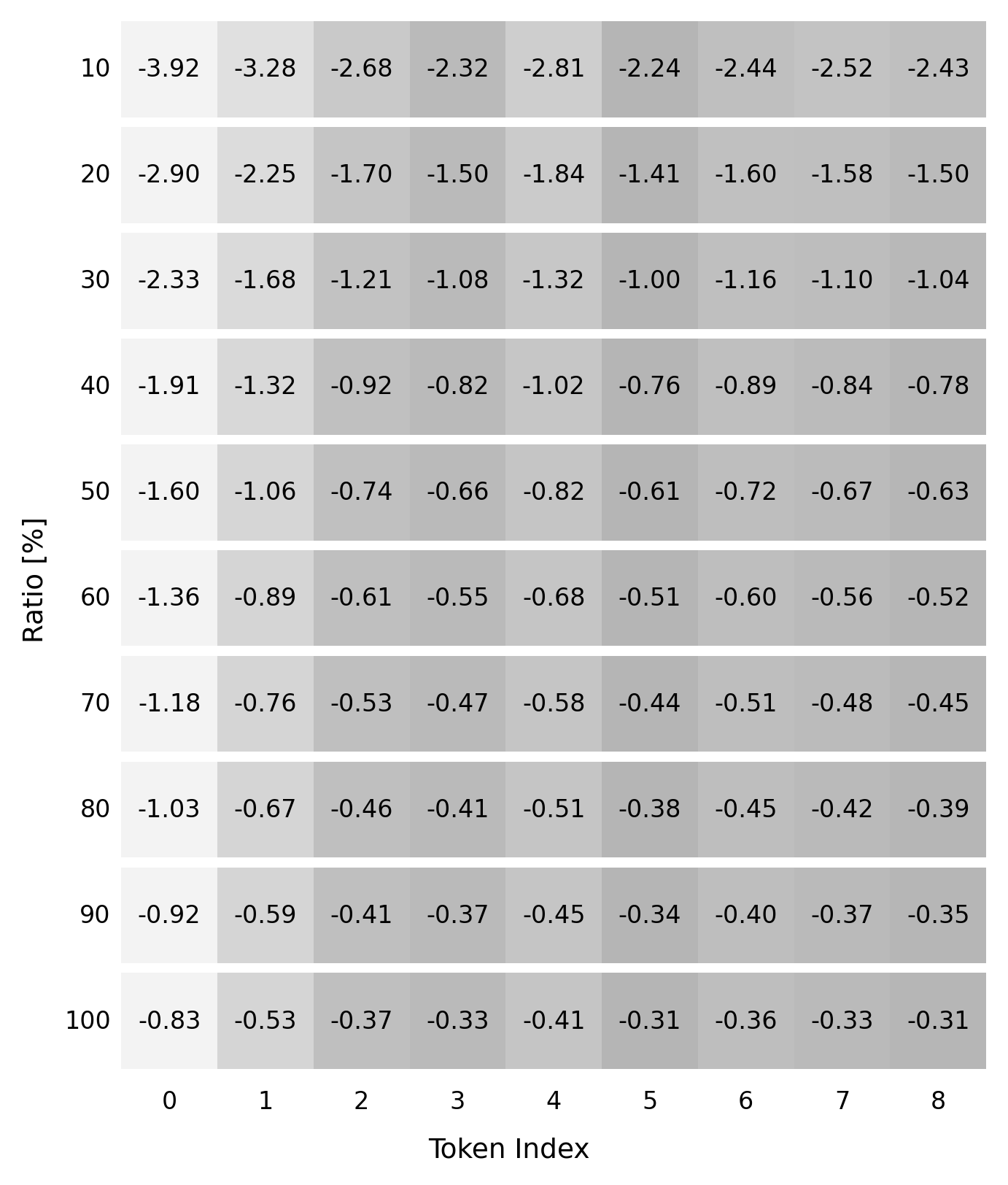}
        \caption{LLaMA-2-13B-chat}
        \label{mink:princ:p:second:llama-13b}
    \end{subfigure}
    \medskip
    \begin{subfigure}{0.49\textwidth}
        \centering
        \includegraphics[width=\linewidth]{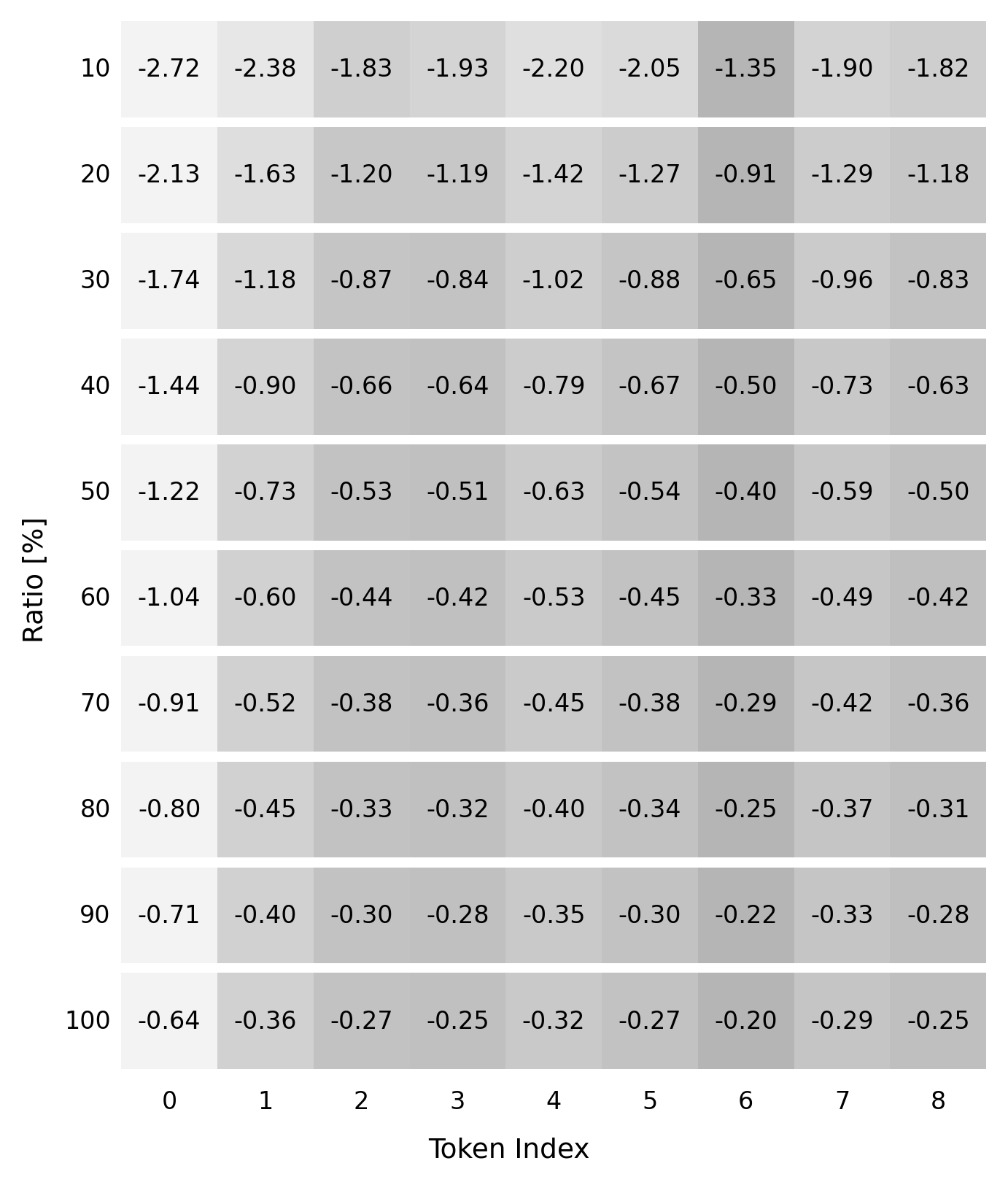}
        \caption{LLaMA-2-70B-chat}
        \label{mink:princ:p:second:llama-70b}
    \end{subfigure}
    \hfill
    \begin{subfigure}{0.49\textwidth}
        \centering
        \includegraphics[width=\linewidth]{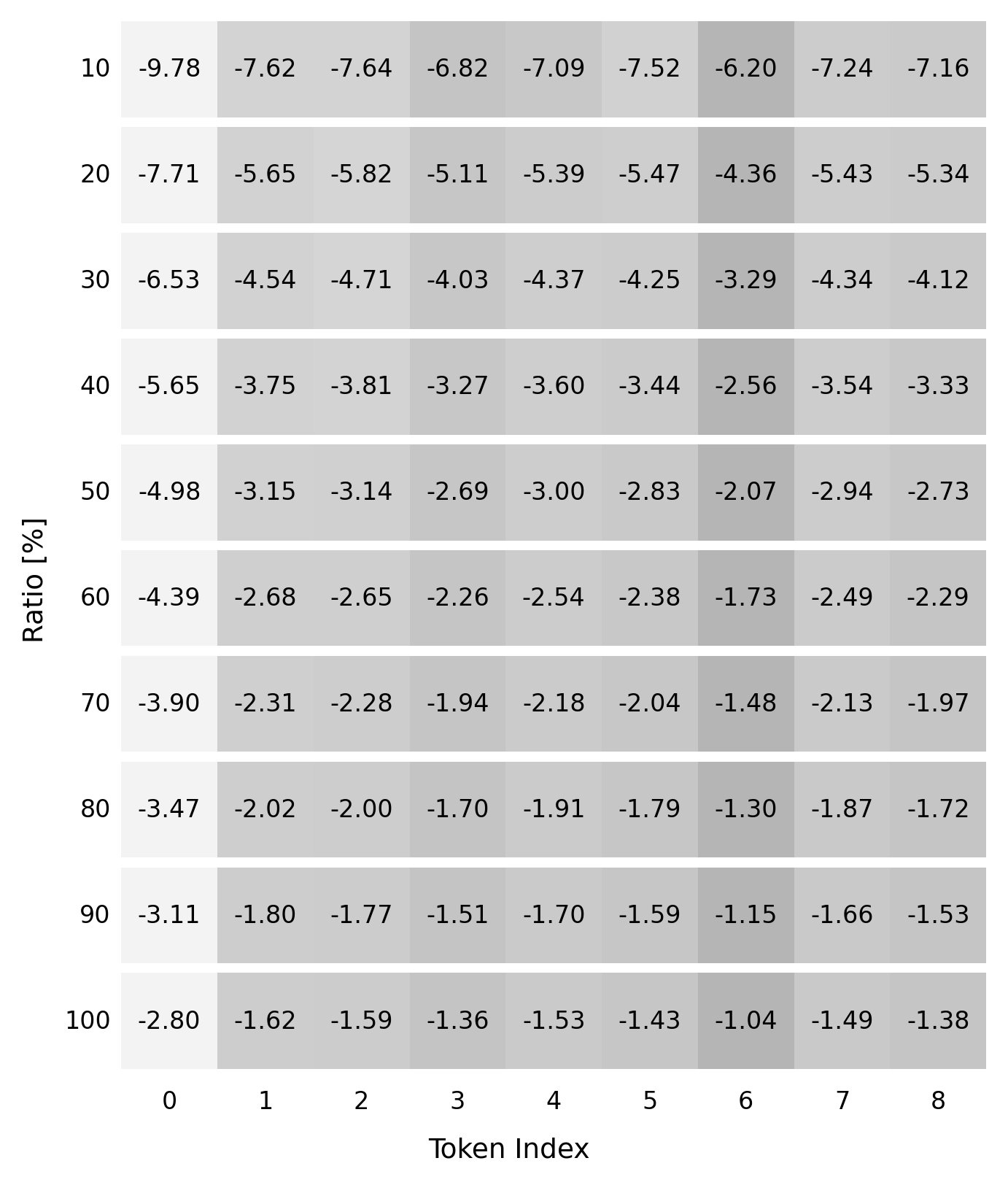}
        \caption{Mistral-7B-instruct}
        \label{mink:princ:p:second:mistral-7b}
    \end{subfigure}
    \caption{\textbf{[second]} Min-K Probability scores across all percentiles over the first 9 tokens from second hallucination spans at global level.}
    \label{mink:princ:p:second}
\end{figure}

\begin{figure}[htbp]
    \centering
    \begin{subfigure}{0.49\textwidth}
        \centering
        \includegraphics[width=\linewidth]{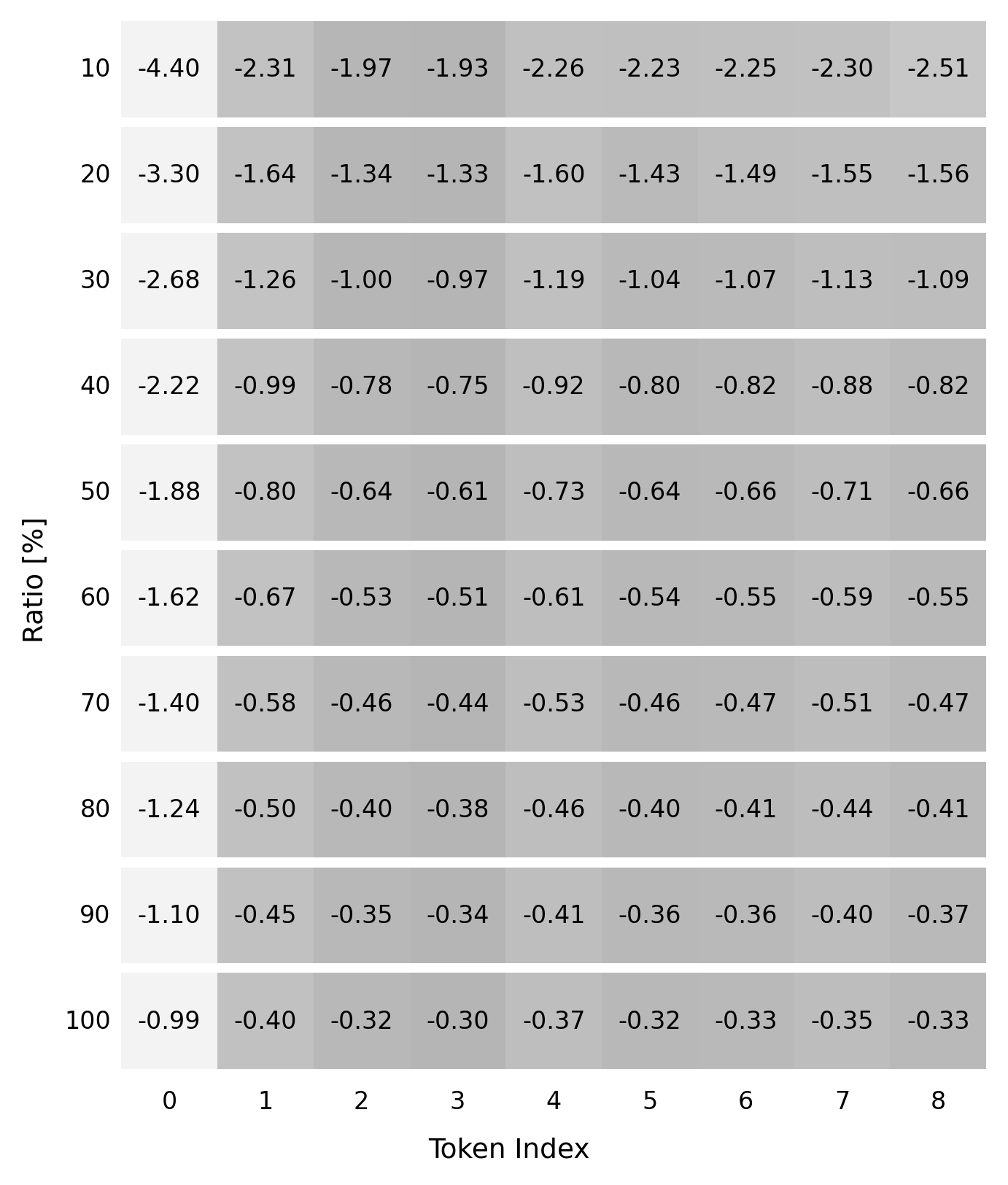}
        \caption{LLaMA-2-7B-chat}
        \label{mink:princ:p:third+:llama-7b}
    \end{subfigure}
    \hfill
    \begin{subfigure}{0.49\textwidth}
        \centering
        \includegraphics[width=\linewidth]{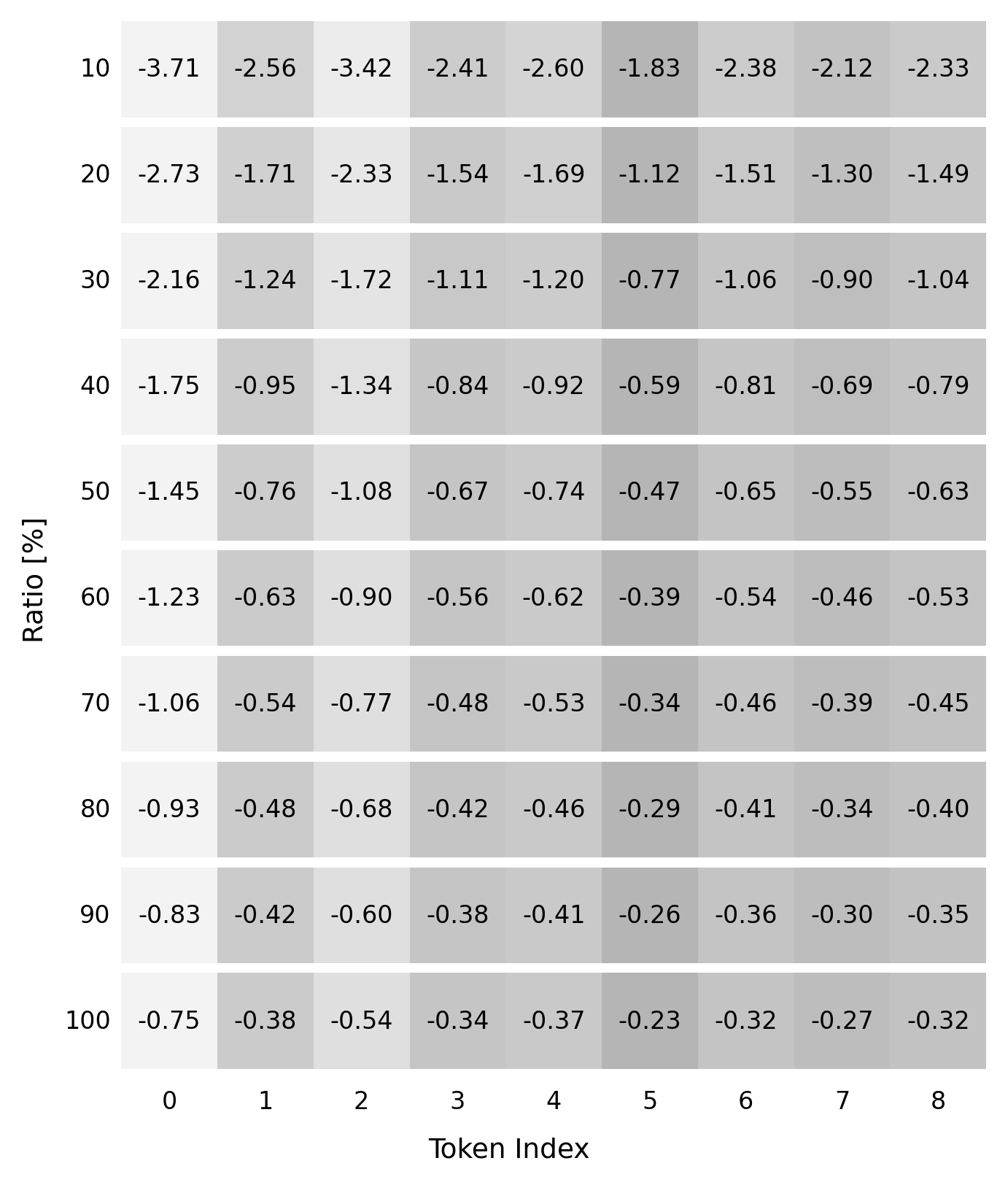}
        \caption{LLaMA-2-13B-chat}
        \label{mink:princ:p:third+:llama-13b}
    \end{subfigure}
    \medskip
    \begin{subfigure}{0.49\textwidth}
        \centering
        \includegraphics[width=\linewidth]{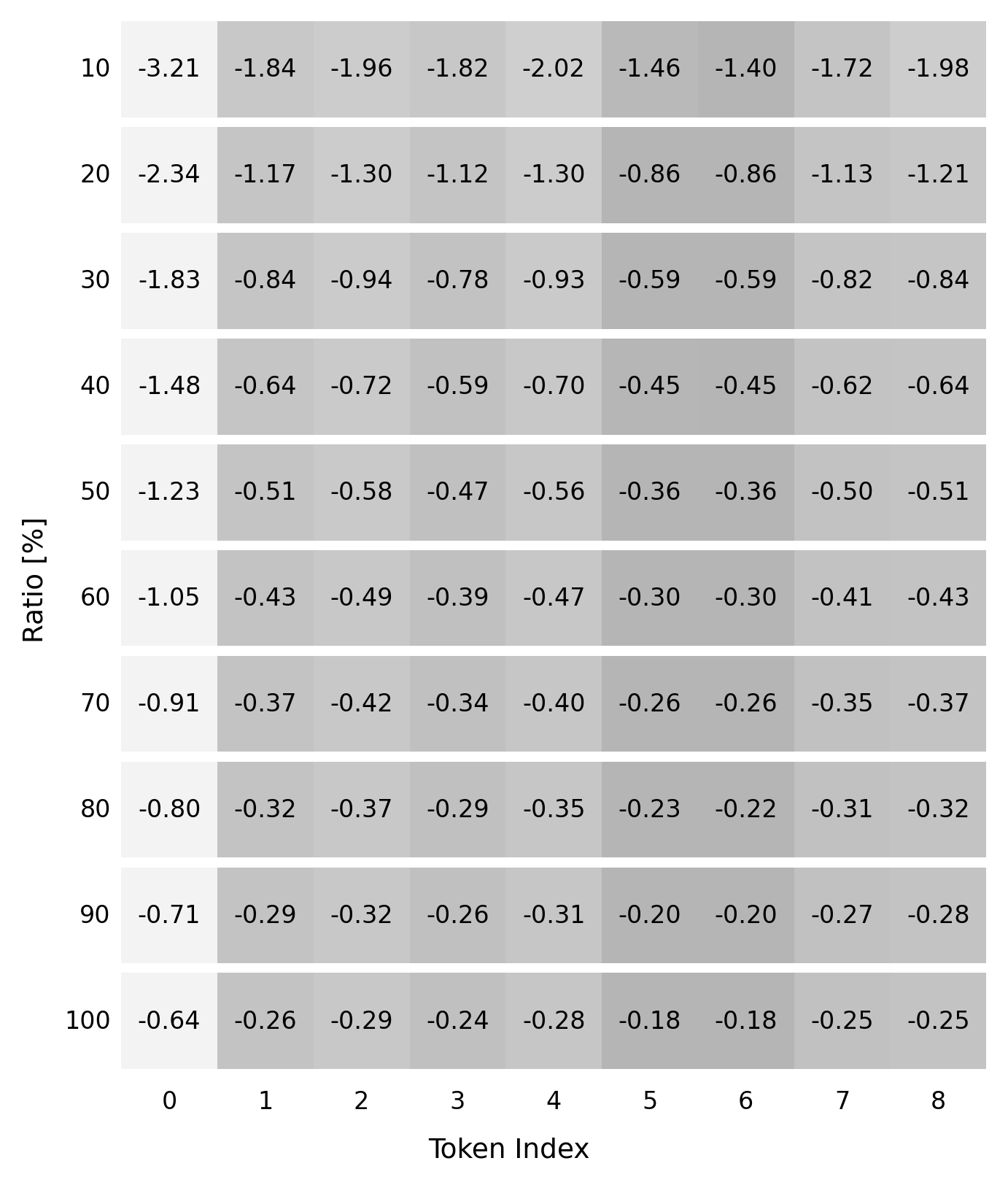}
        \caption{LLaMA-2-70B-chat}
        \label{mink:princ:p:third+:llama-70b}
    \end{subfigure}
    \hfill
    \begin{subfigure}{0.49\textwidth}
        \centering
        \includegraphics[width=\linewidth]{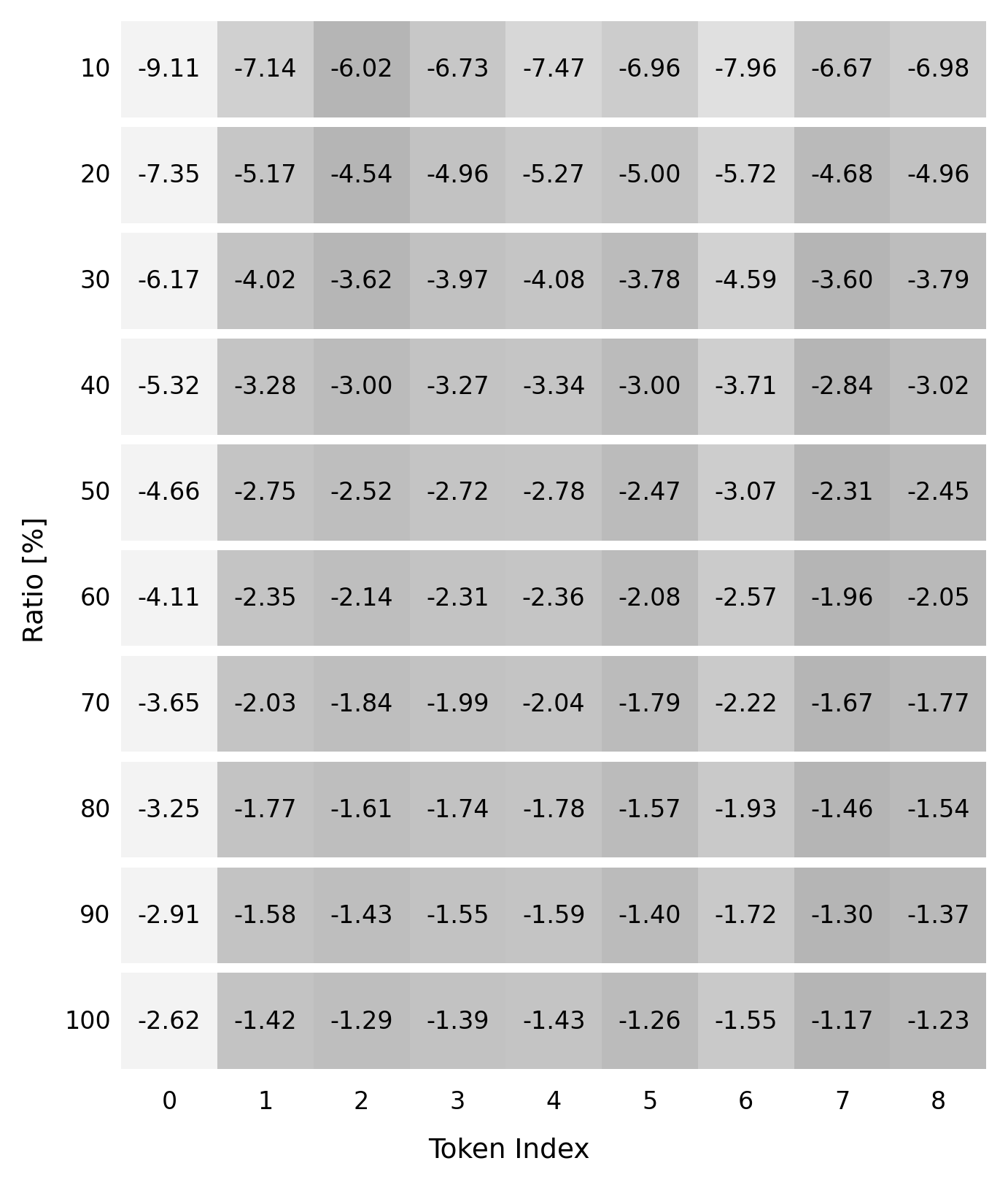}
        \caption{Mistral-7B-instruct}
        \label{mink:princ:p:third+:mistral-7b}
    \end{subfigure}
    \caption{\textbf{[third+]} Min-K Probability scores across all percentiles over the first 9 tokens from third+ hallucination spans at global level.}
    \label{mink:princ:p:third+}
\end{figure}

\begin{figure}[htbp]
    \centering
    \begin{subfigure}{0.49\textwidth}
        \centering
        \includegraphics[width=\linewidth]{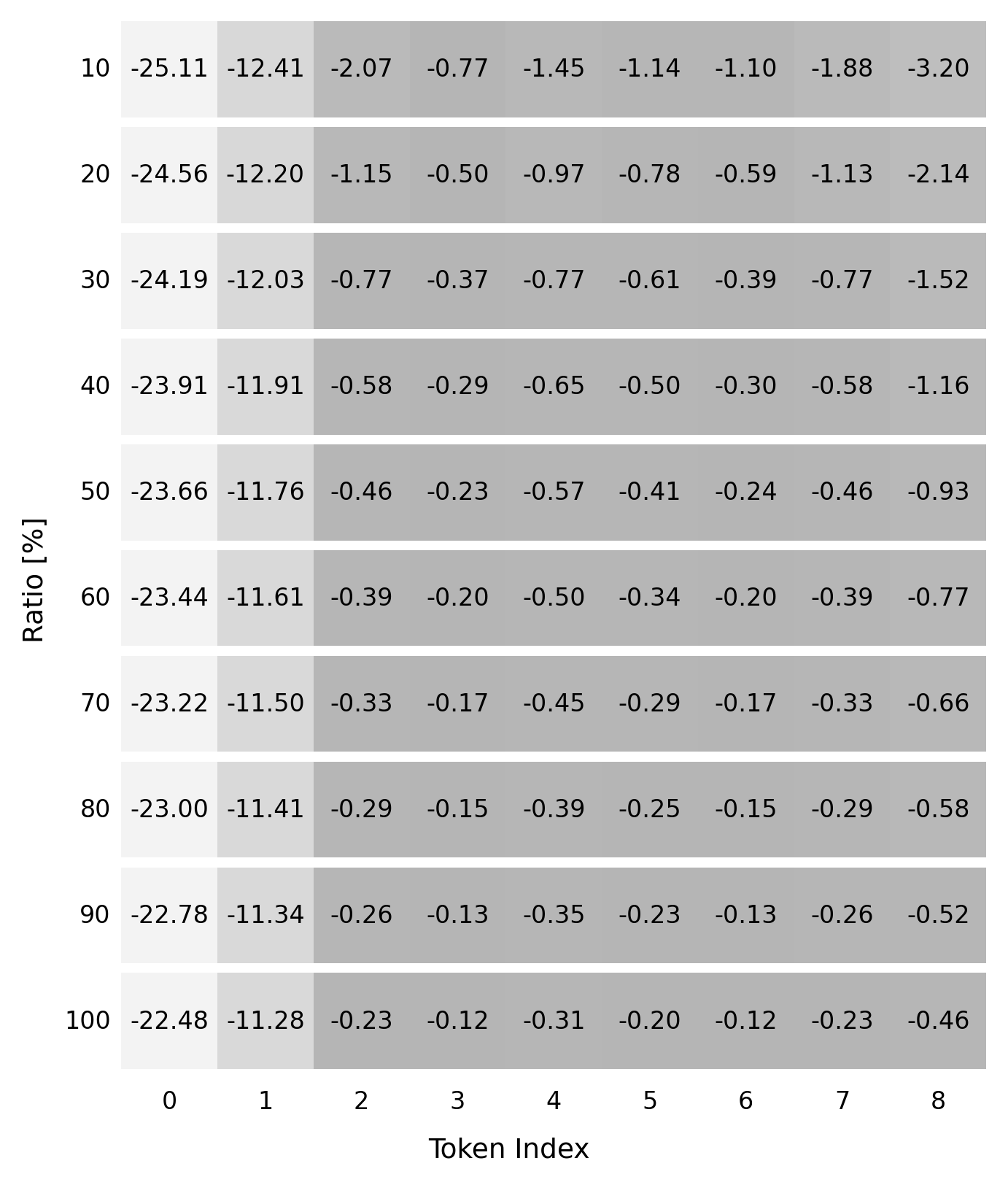}
        \caption{LLaMA-2-7B-chat}
        \label{mink:princ:p:pre:llama-7b}
    \end{subfigure}
    \hfill
    \begin{subfigure}{0.49\textwidth}
        \centering
        \includegraphics[width=\linewidth]{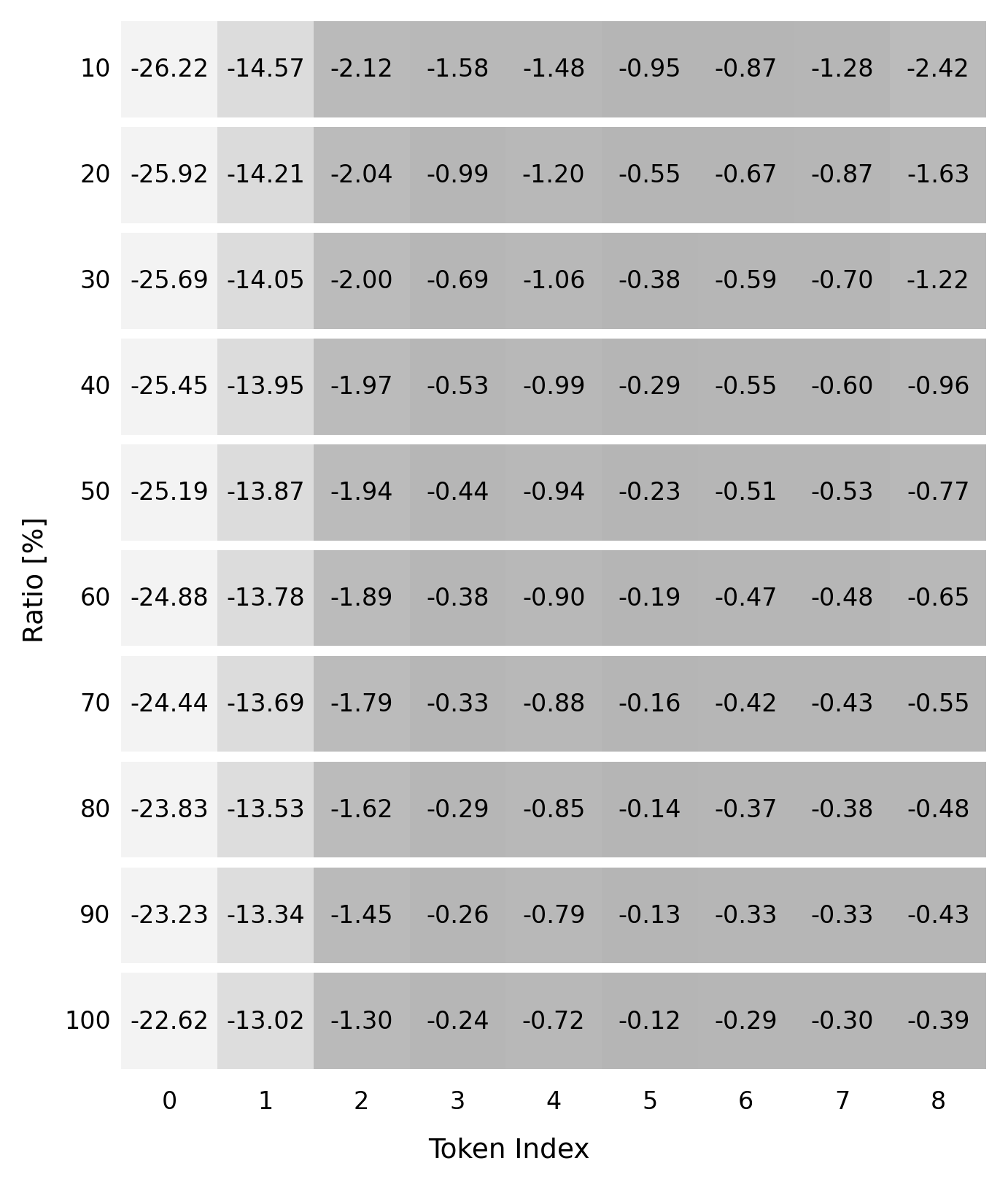}
        \caption{LLaMA-2-13B-chat}
        \label{mink:princ:p:pre:llama-13b}
    \end{subfigure}
    \medskip
    \begin{subfigure}{0.49\textwidth}
        \centering
        \includegraphics[width=\linewidth]{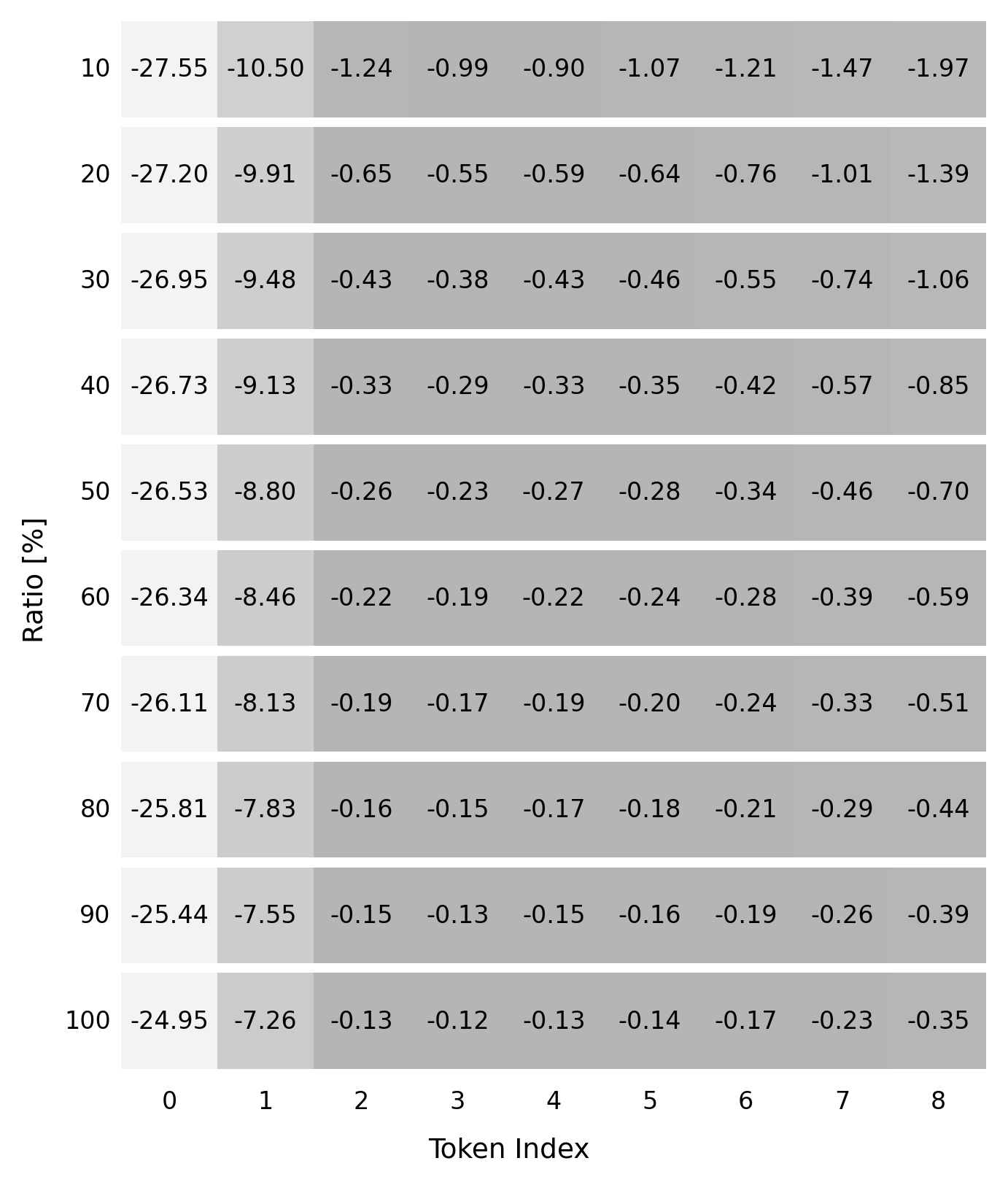}
        \caption{LLaMA-2-70B-chat}
        \label{mink:princ:p:pre:llama-70b}
    \end{subfigure}
    \hfill
    \begin{subfigure}{0.49\textwidth}
        \centering
        \includegraphics[width=\linewidth]{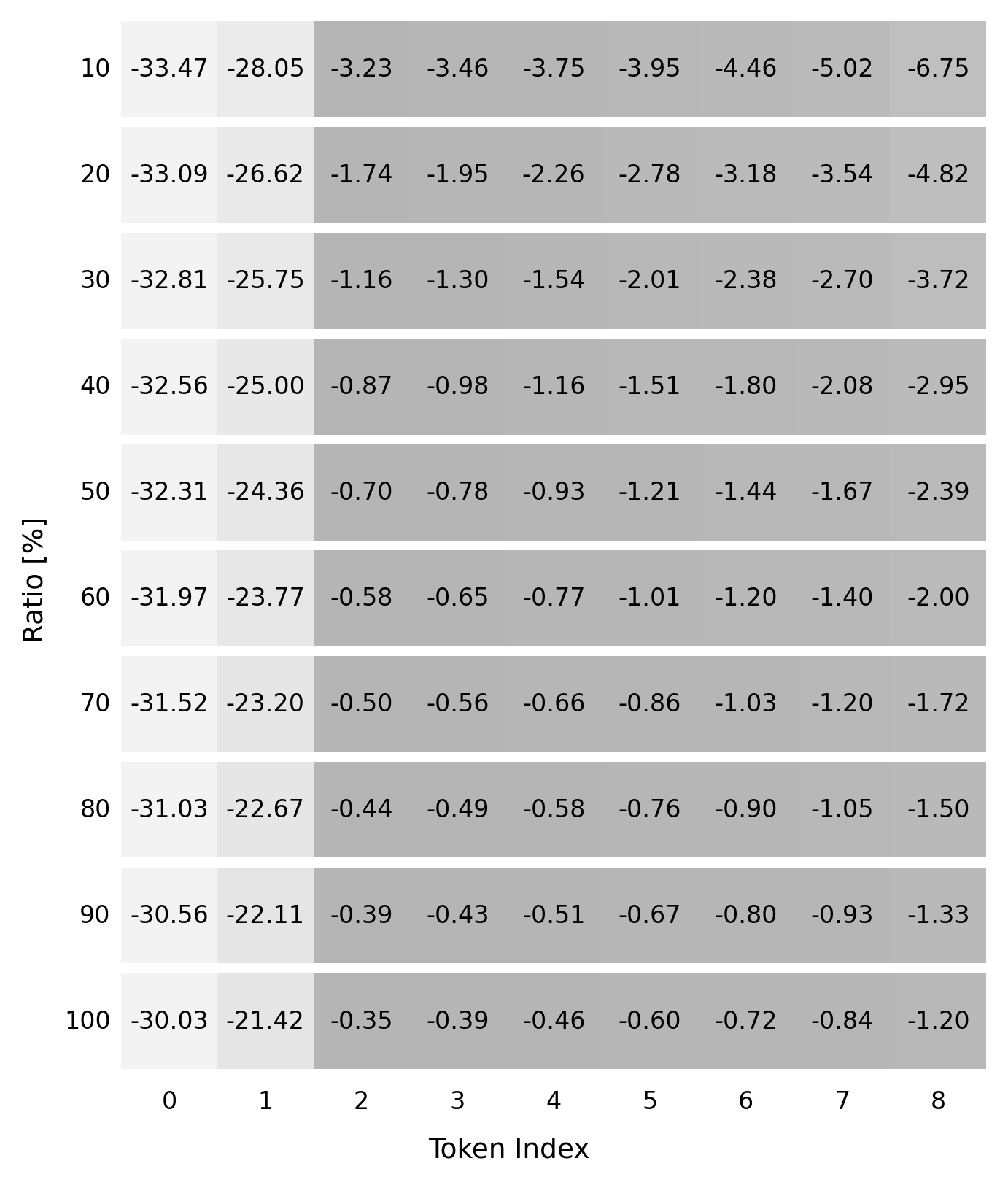}
        \caption{Mistral-7B-instruct}
        \label{mink:princ:p:pre:mistral-7b}
    \end{subfigure}
    \caption{\textbf{[pre]} Min-K Probability scores across all percentiles over the first 9 pre-hallucination tokens at global level.}
    \label{mink:princ:p:pre}
\end{figure}

\begin{figure}[htbp]
    \centering
    \begin{subfigure}{0.49\textwidth}
        \centering
        \includegraphics[width=\linewidth]{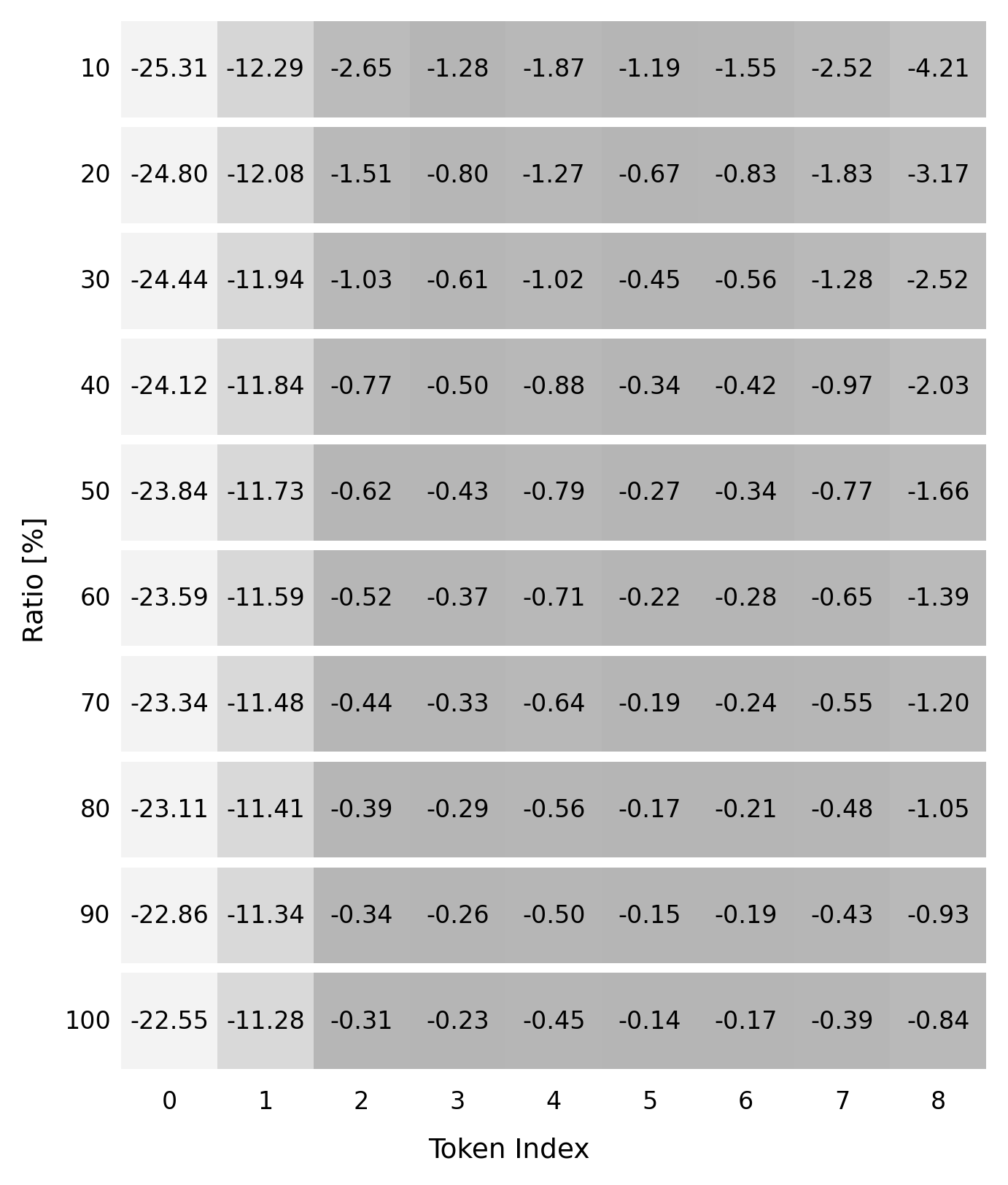}
        \caption{LLaMA-2-7B-chat}
        \label{mink:princ:p:no:llama-7b}
    \end{subfigure}
    \hfill
    \begin{subfigure}{0.49\textwidth}
        \centering
        \includegraphics[width=\linewidth]{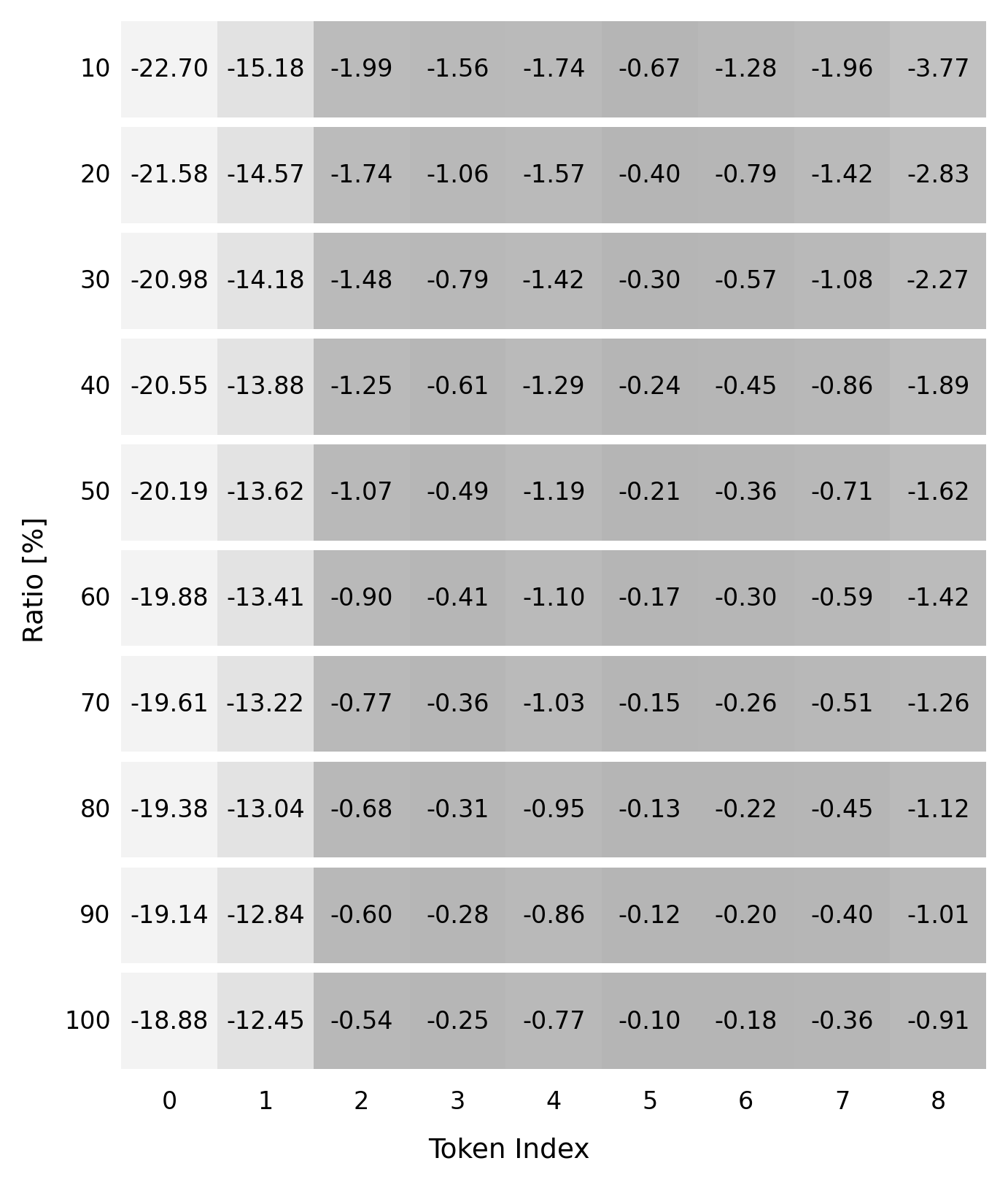}
        \caption{LLaMA-2-13B-chat}
        \label{mink:princ:p:no:llama-13b}
    \end{subfigure}
    \medskip
    \begin{subfigure}{0.49\textwidth}
        \centering
        \includegraphics[width=\linewidth]{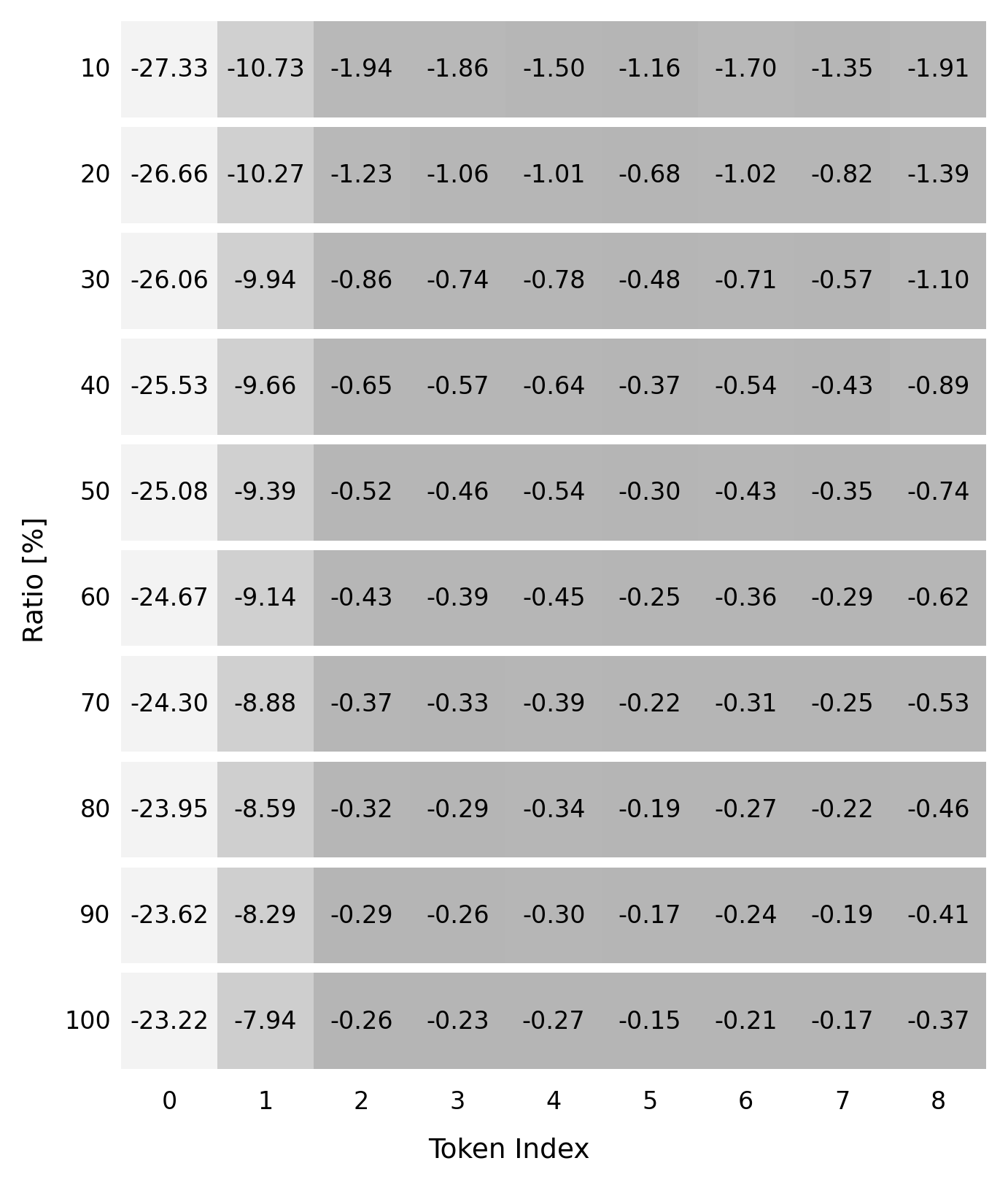}
        \caption{LLaMA-2-70B-chat}
        \label{mink:princ:p:no:llama-70b}
    \end{subfigure}
    \hfill
    \begin{subfigure}{0.49\textwidth}
        \centering
        \includegraphics[width=\linewidth]{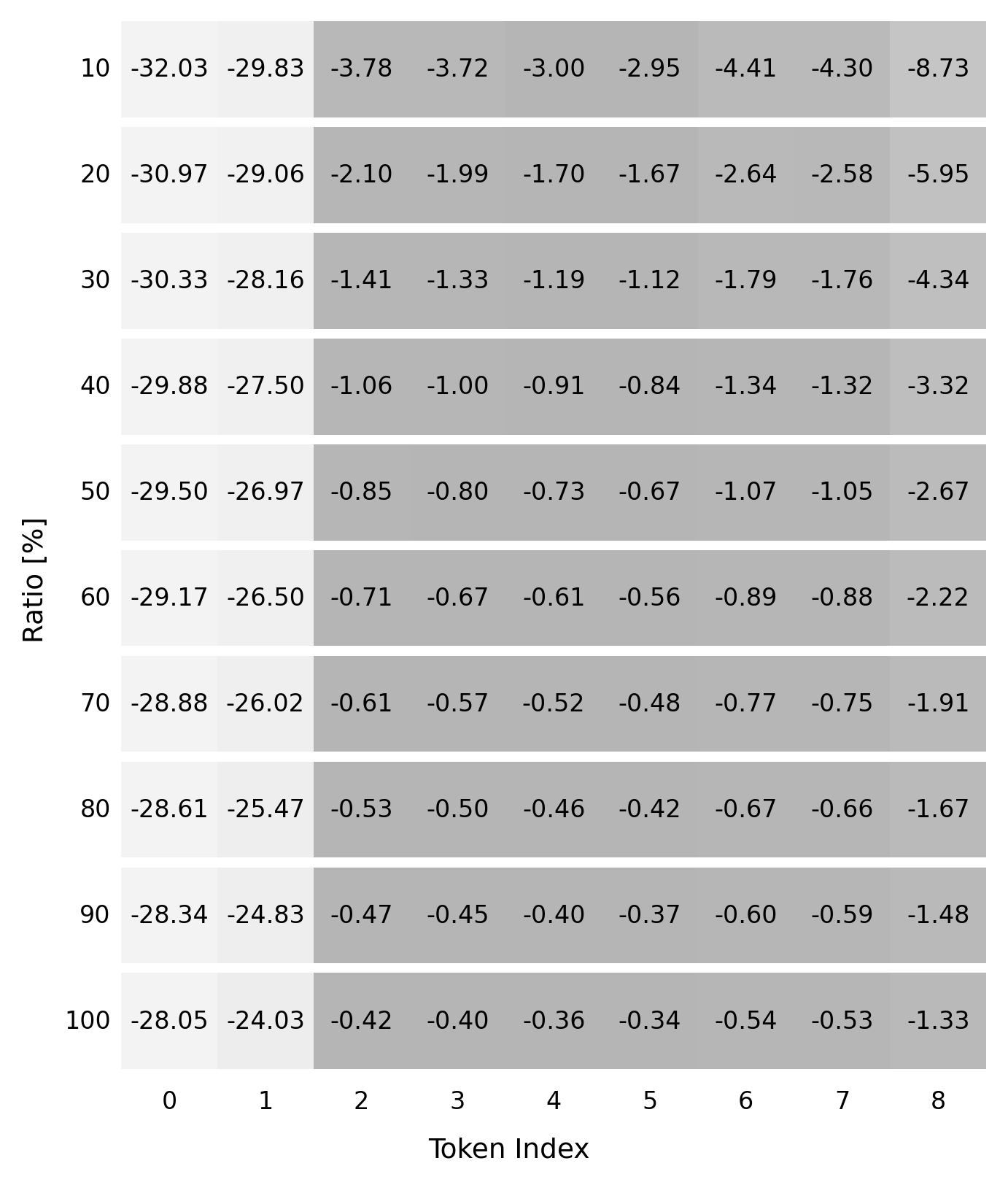}
        \caption{Mistral-7B-instruct}
        \label{mink:princ:p:no:mistral-7b}
    \end{subfigure}
    \caption{\textbf{[no]} Min-K Probability scores across all percentiles over the first 9 tokens from responses without hallucination at global level.}
    \label{mink:princ:p:no}
\end{figure}

\FloatBarrier
\subsubsection{Min-K Entropy}
\label{plt:mink:perc:entropy}
\begin{figure}[!hp]
    \centering
    \begin{subfigure}{0.45\textwidth}
        \centering
        \includegraphics[width=\linewidth]{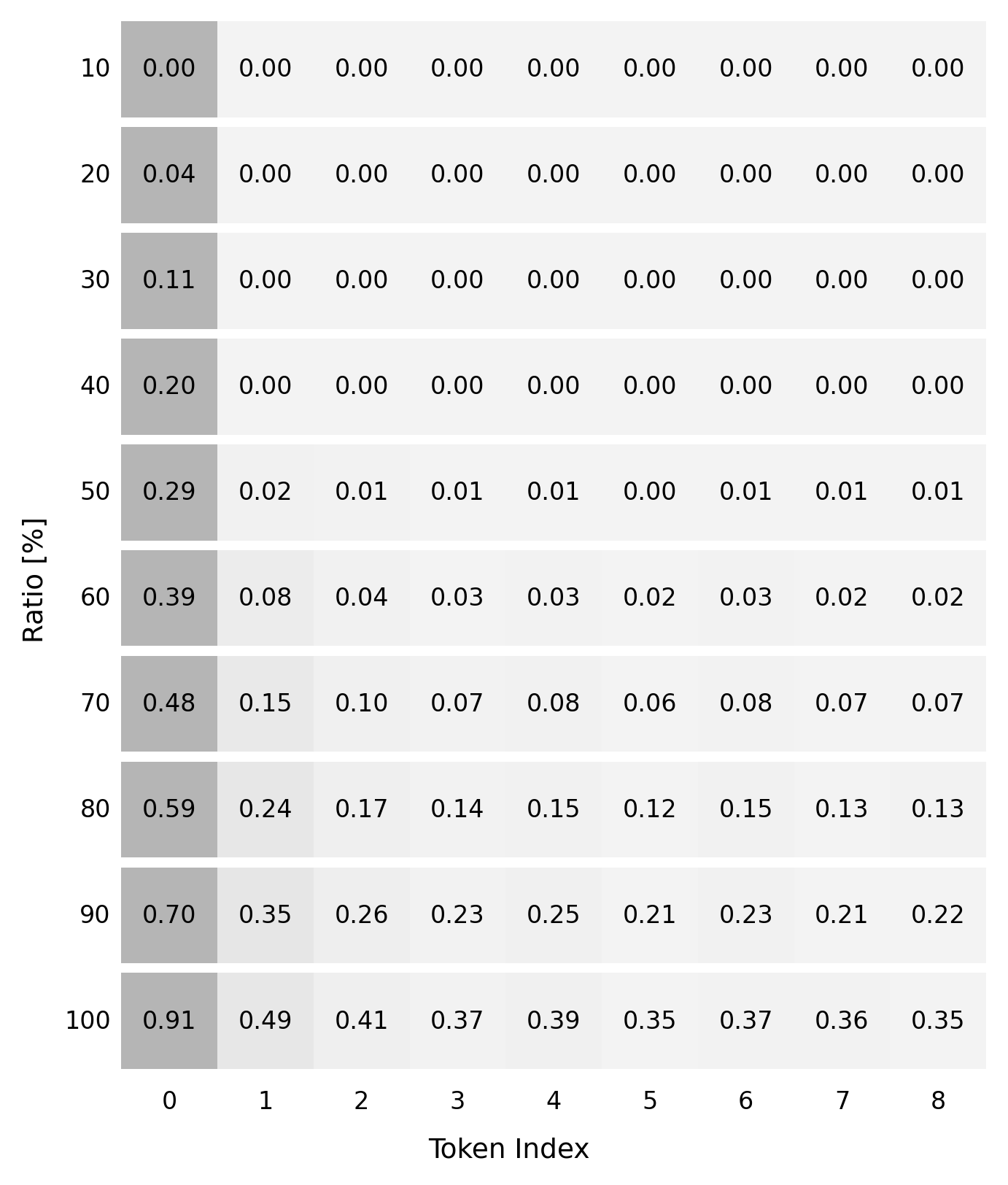}
        \caption{LLaMA-2-7B-chat}
        \label{mink:princ:e:all:llama-7b}
    \end{subfigure}
    \hfill
    \begin{subfigure}{0.45\textwidth}
        \centering
        \includegraphics[width=\linewidth]{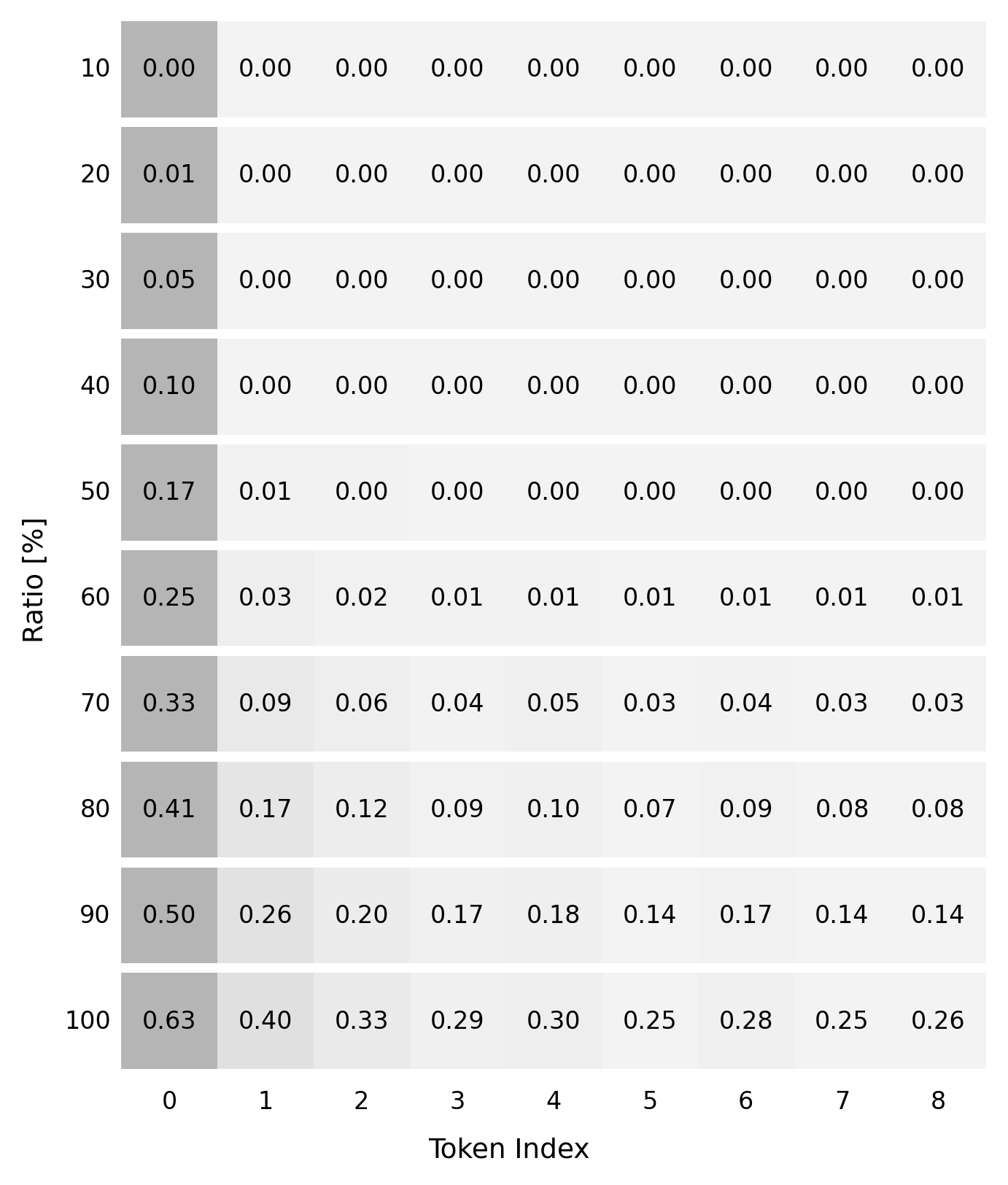}
        \caption{LLaMA-2-13B-chat}
        \label{mink:princ:e:all:llama-13b}
    \end{subfigure}
    \medskip
    \begin{subfigure}{0.45\textwidth}
        \centering
        \includegraphics[width=\linewidth]{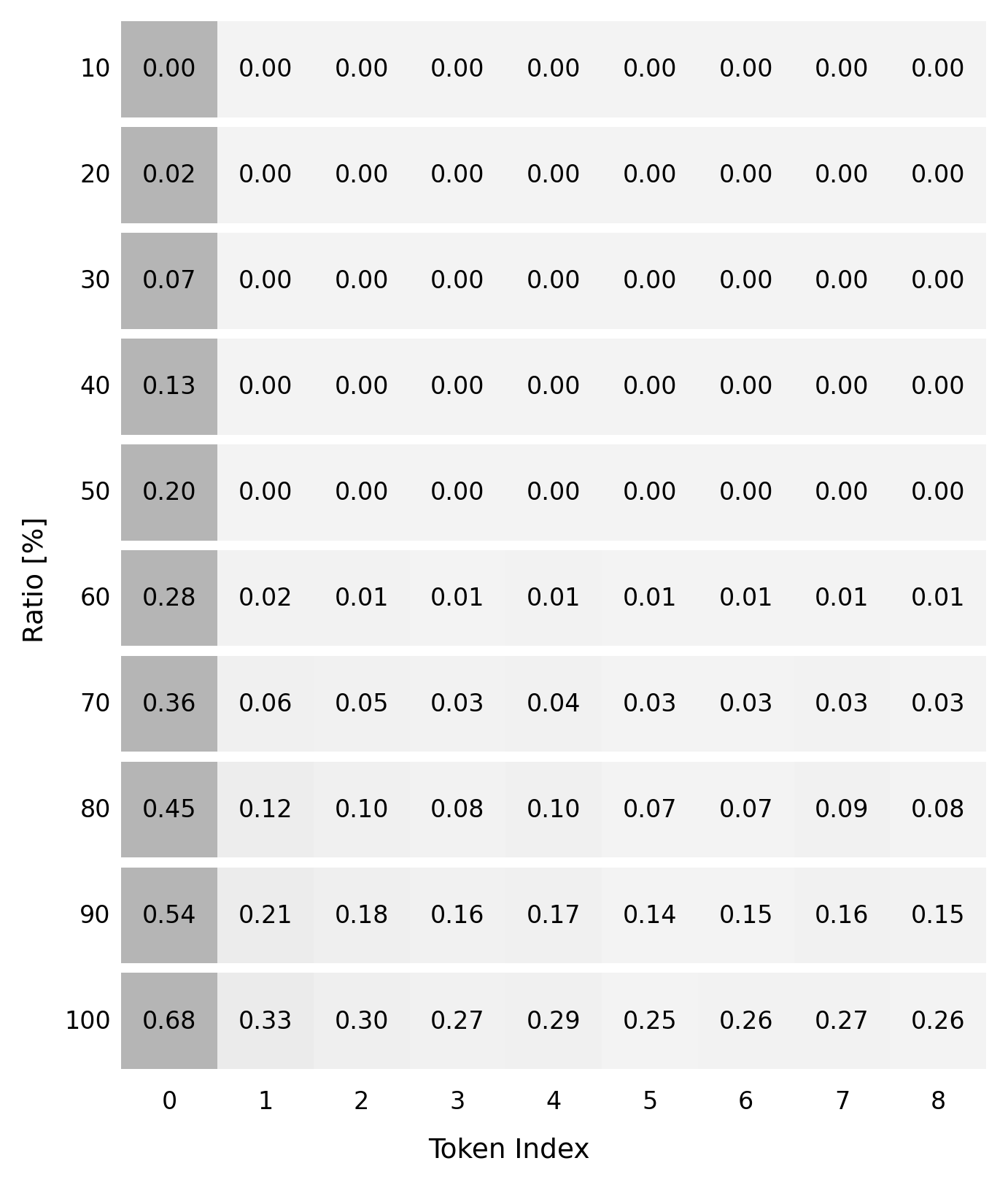}
        \caption{LLaMA-2-70B-chat}
        \label{mink:princ:e:all:llama-70b}
    \end{subfigure}
    \hfill
    \begin{subfigure}{0.45\textwidth}
        \centering
        \includegraphics[width=\linewidth]{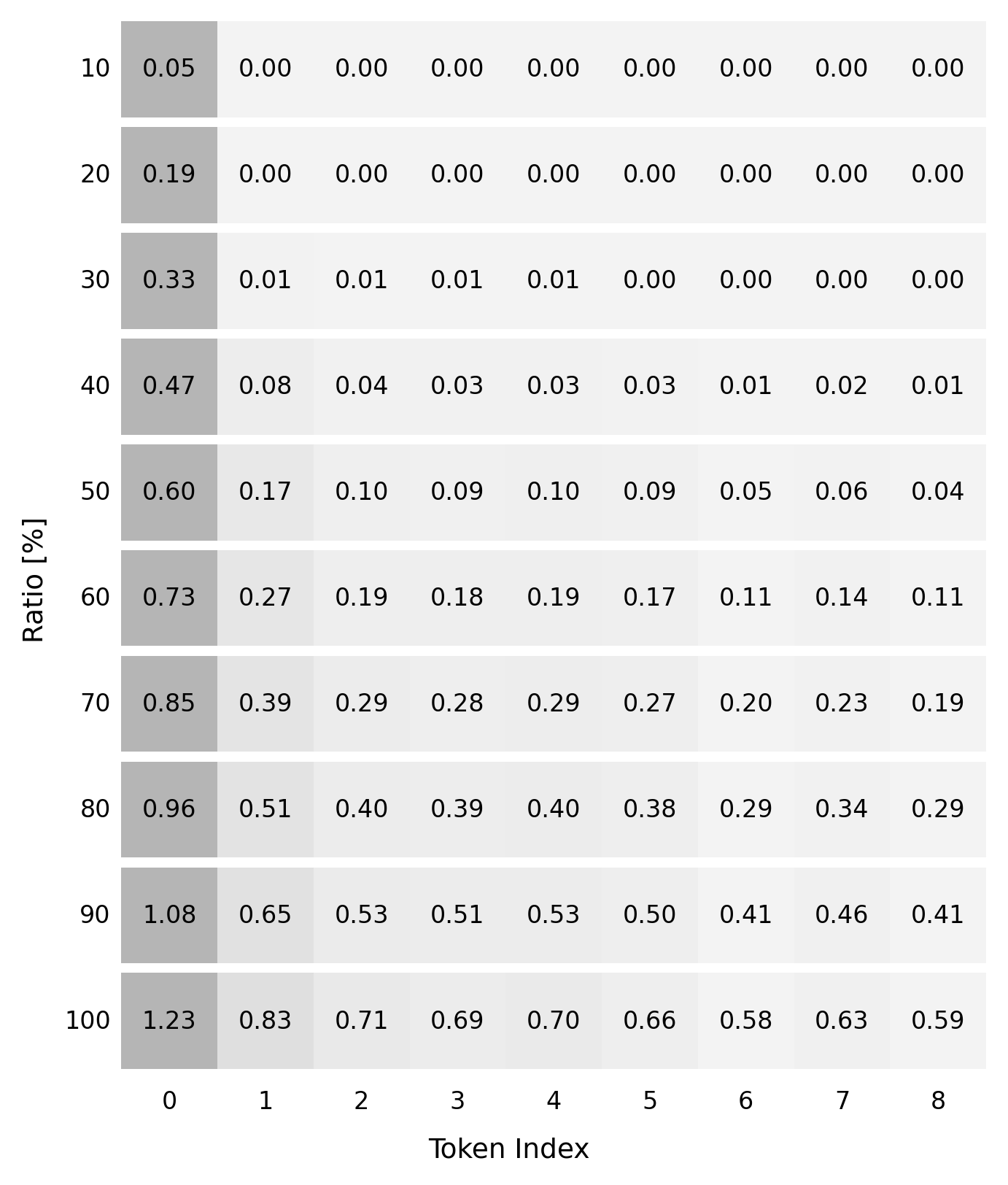}
        \caption{Mistral-7B-instruct}
        \label{mink:princ:e:all:mistral-7b}
    \end{subfigure}
    \caption{\textbf{[all]} Min-K Entropy scores across all percentiles over the first 9 tokens from all hallucination spans at global level.}
    \label{mink:princ:e:all}
\end{figure}

\begin{figure}[htbp]
    \centering
    \begin{subfigure}{0.45\textwidth}
        \centering
        \includegraphics[width=\linewidth]{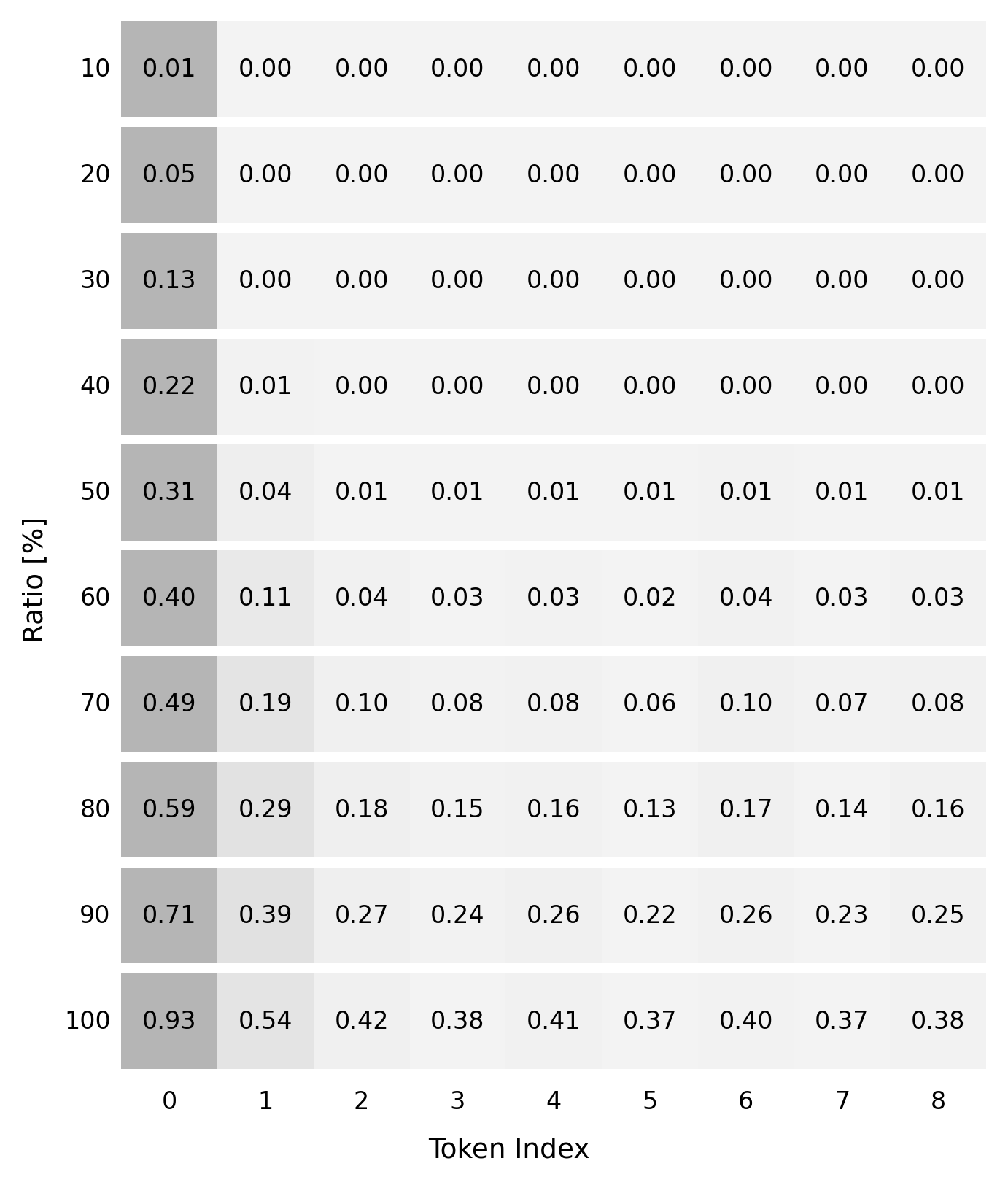}
        \caption{LLaMA-2-7B-chat}
        \label{mink:princ:e:first:llama-7b}
    \end{subfigure}
    \hfill
    \begin{subfigure}{0.45\textwidth}
        \centering
        \includegraphics[width=\linewidth]{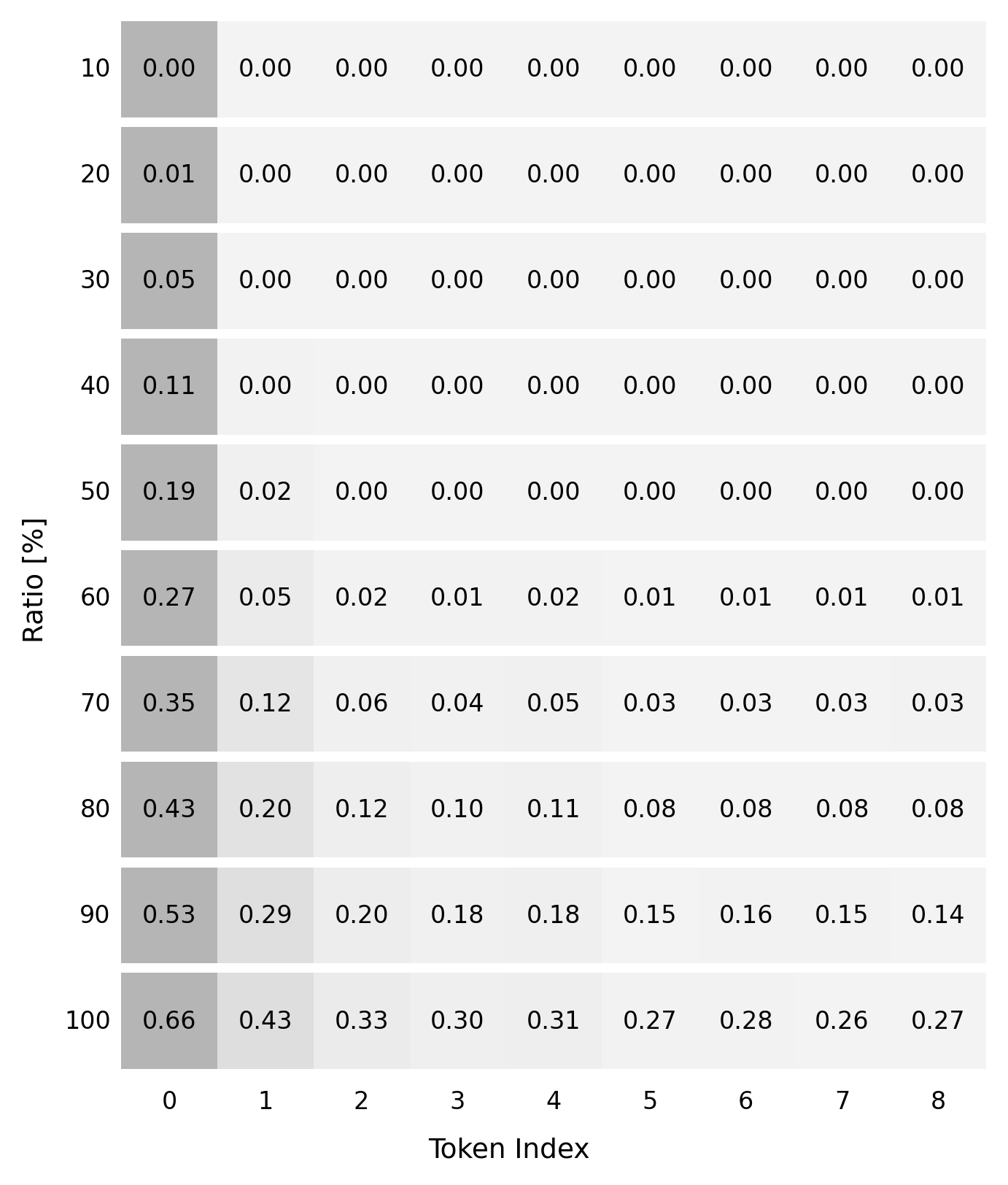}
        \caption{LLaMA-2-13B-chat}
        \label{mink:princ:e:first:llama-13b}
    \end{subfigure}
    \medskip
    \begin{subfigure}{0.45\textwidth}
        \centering
        \includegraphics[width=\linewidth]{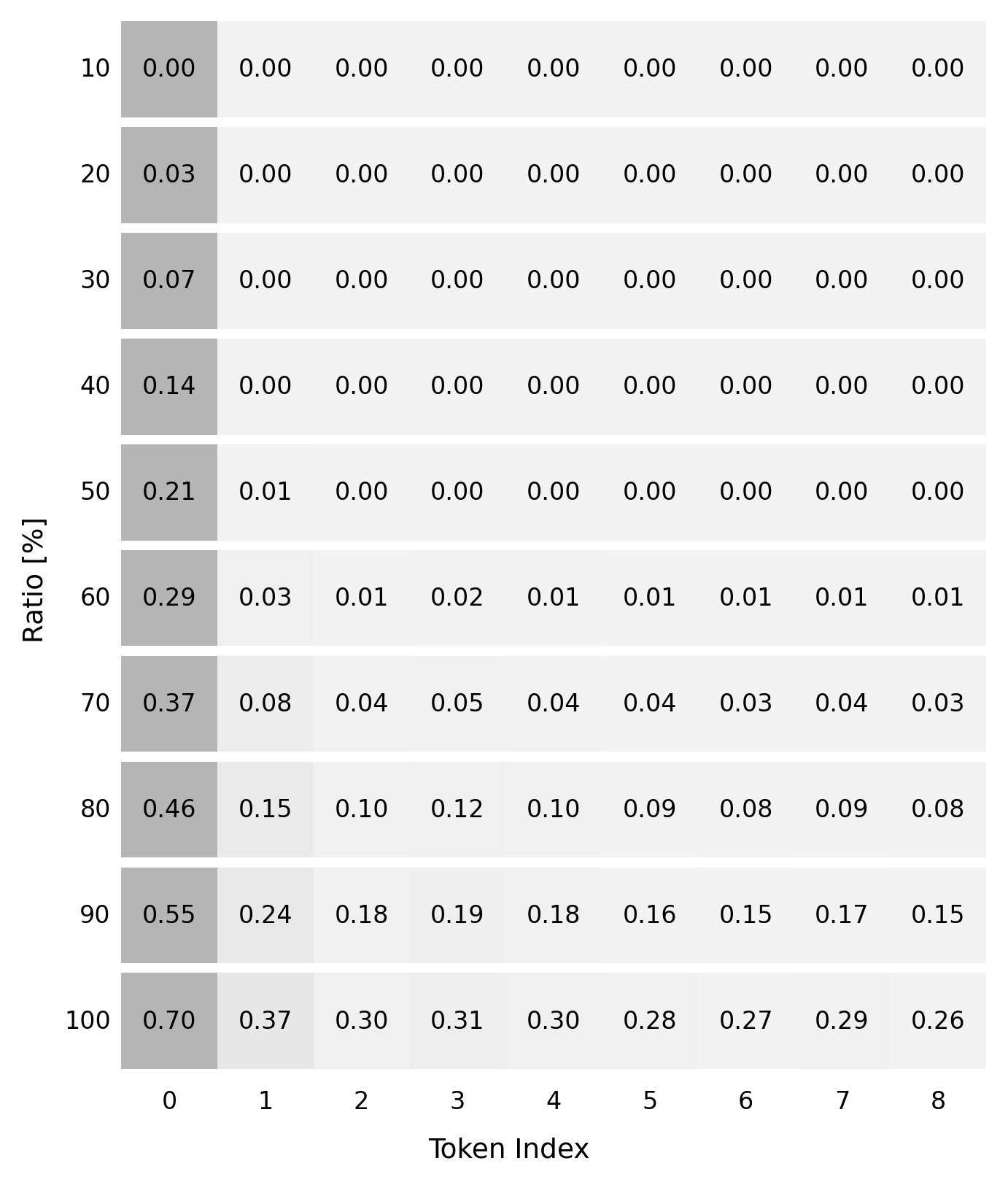}
        \caption{LLaMA-2-70B-chat}
        \label{mink:princ:e:first:llama-70b}
    \end{subfigure}
    \hfill
    \begin{subfigure}{0.45\textwidth}
        \centering
        \includegraphics[width=\linewidth]{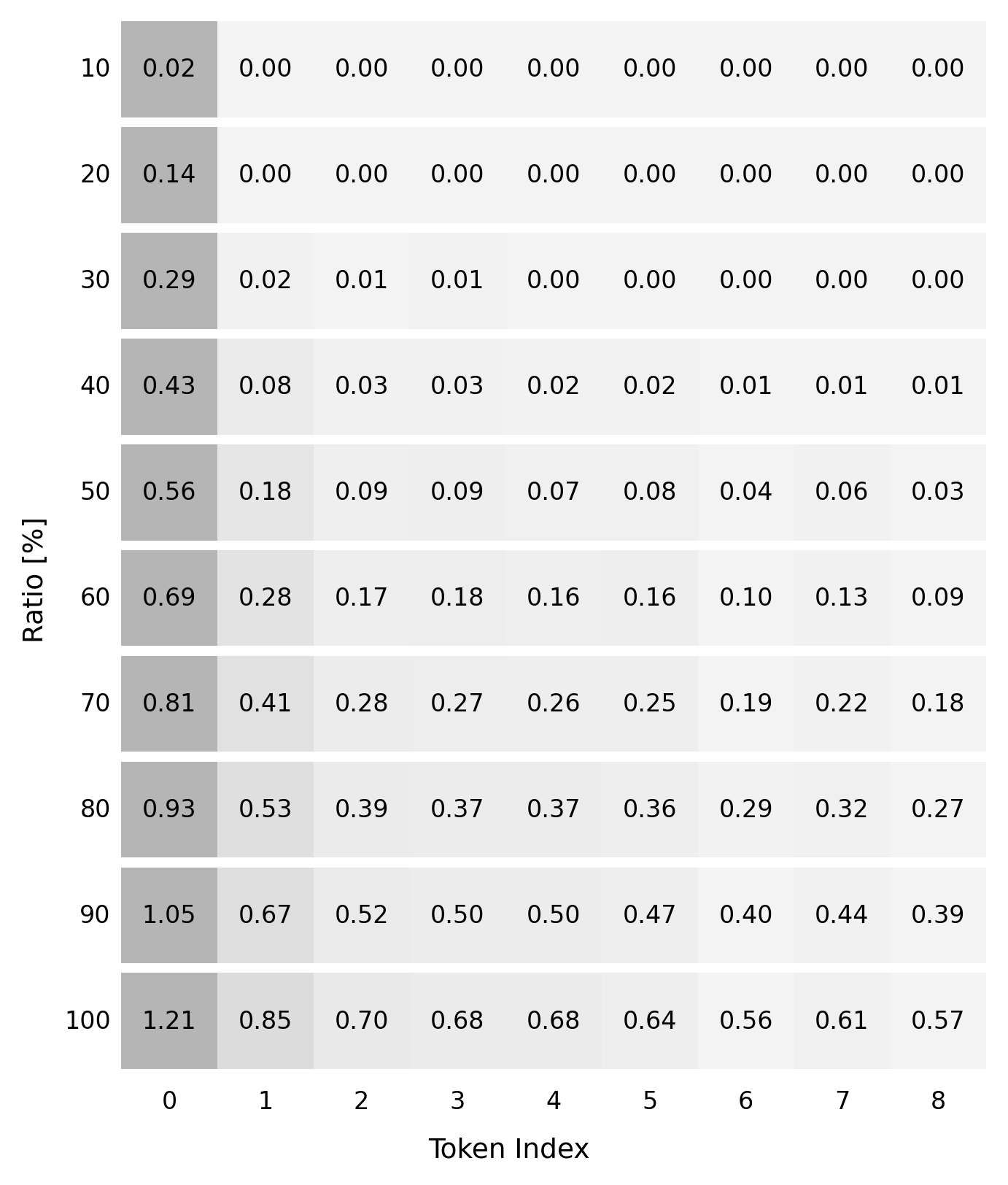}
        \caption{Mistral-7B-instruct}
        \label{mink:princ:e:first:mistral-7b}
    \end{subfigure}
    \caption{\textbf{[first]} Min-K Entropy scores across all percentiles over the first 9 tokens from first hallucination spans at global level.}
    \label{mink:princ:e:first}
\end{figure}

\begin{figure}[htbp]
    \centering
    \begin{subfigure}{0.45\textwidth}
        \centering
        \includegraphics[width=\linewidth]{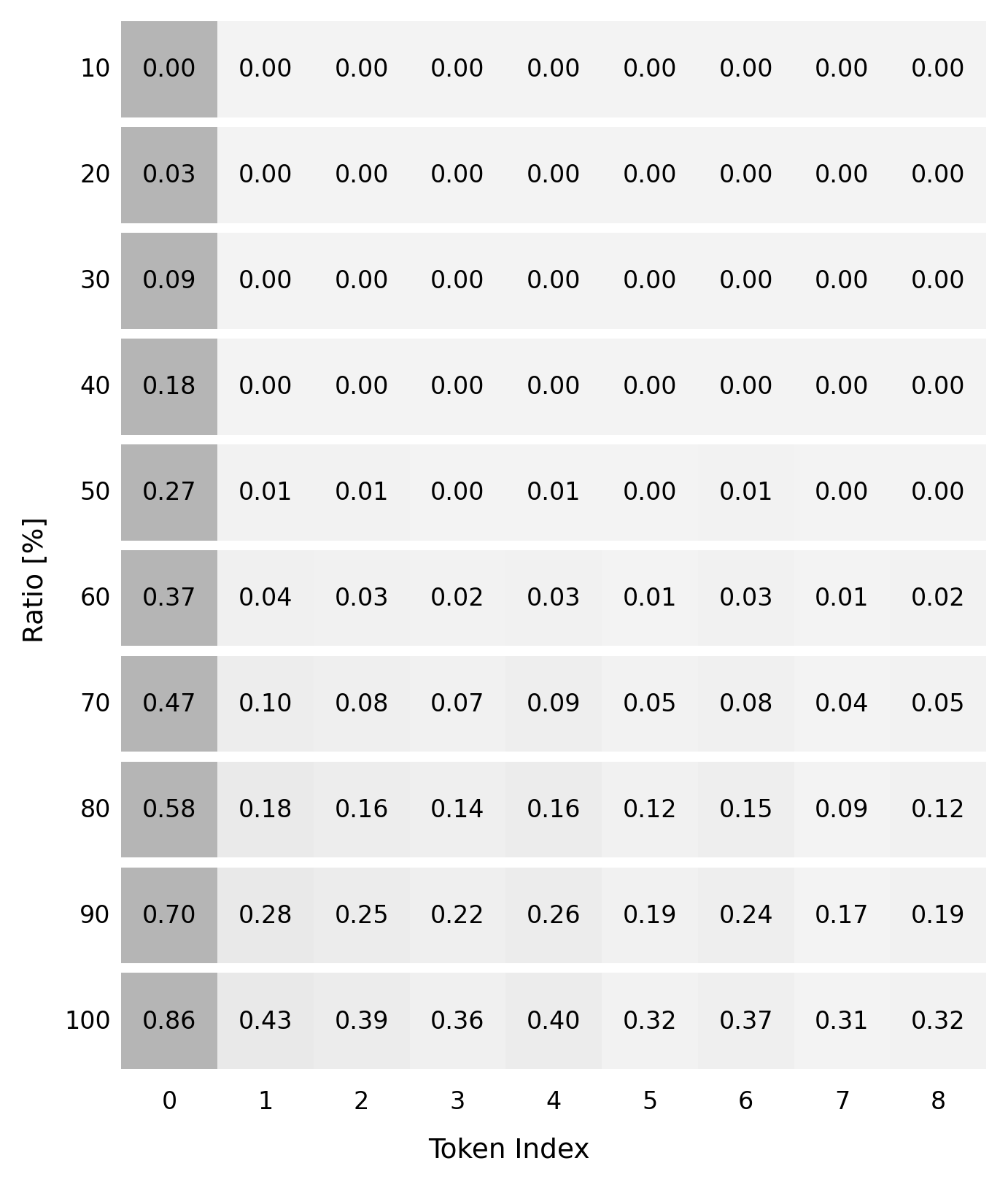}
        \caption{LLaMA-2-7B-chat}
        \label{mink:princ:e:second:llama-7b}
    \end{subfigure}
    \hfill
    \begin{subfigure}{0.45\textwidth}
        \centering
        \includegraphics[width=\linewidth]{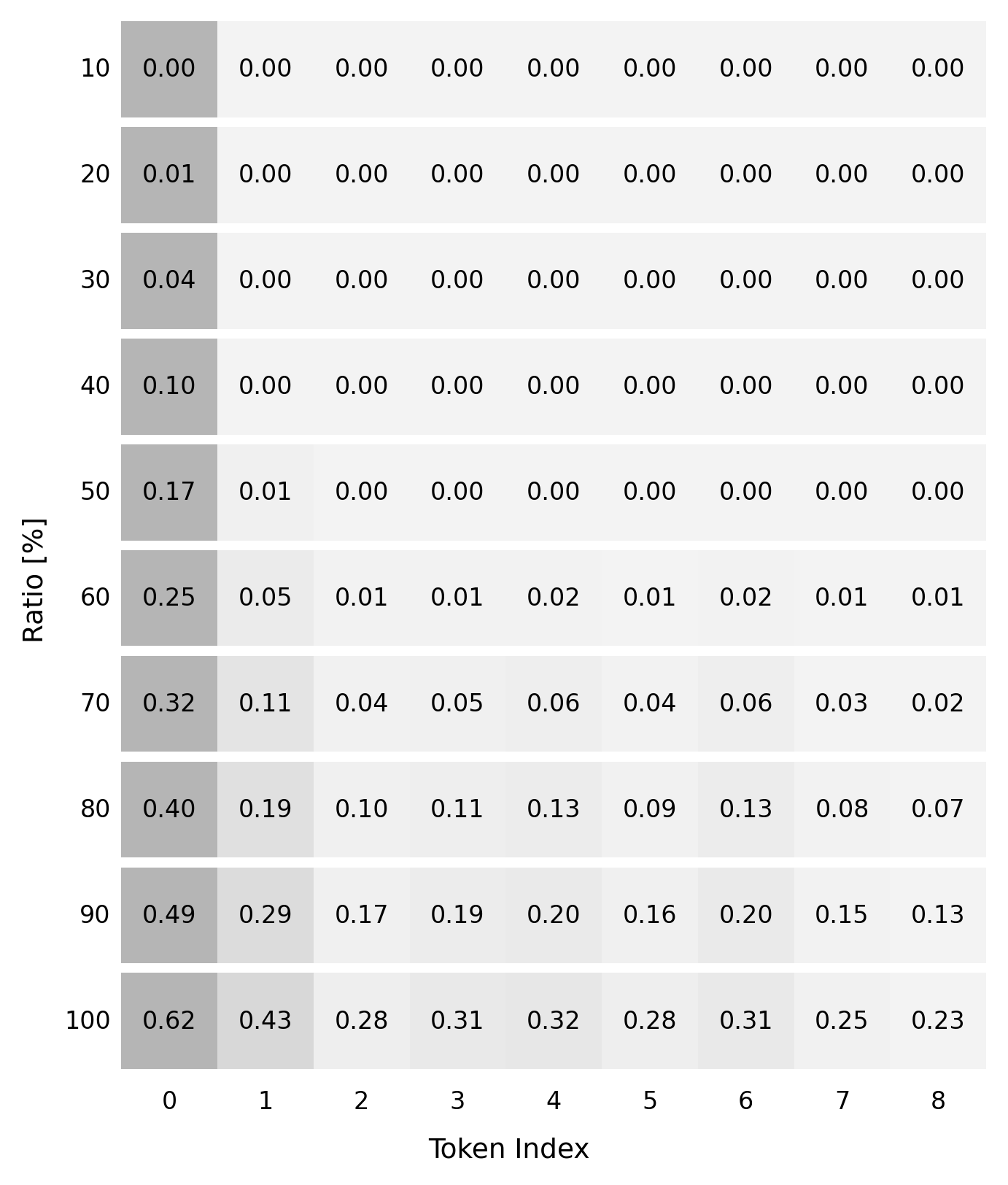}
        \caption{LLaMA-2-13B-chat}
        \label{mink:princ:e:second:llama-13b}
    \end{subfigure}
    \medskip
    \begin{subfigure}{0.45\textwidth}
        \centering
        \includegraphics[width=\linewidth]{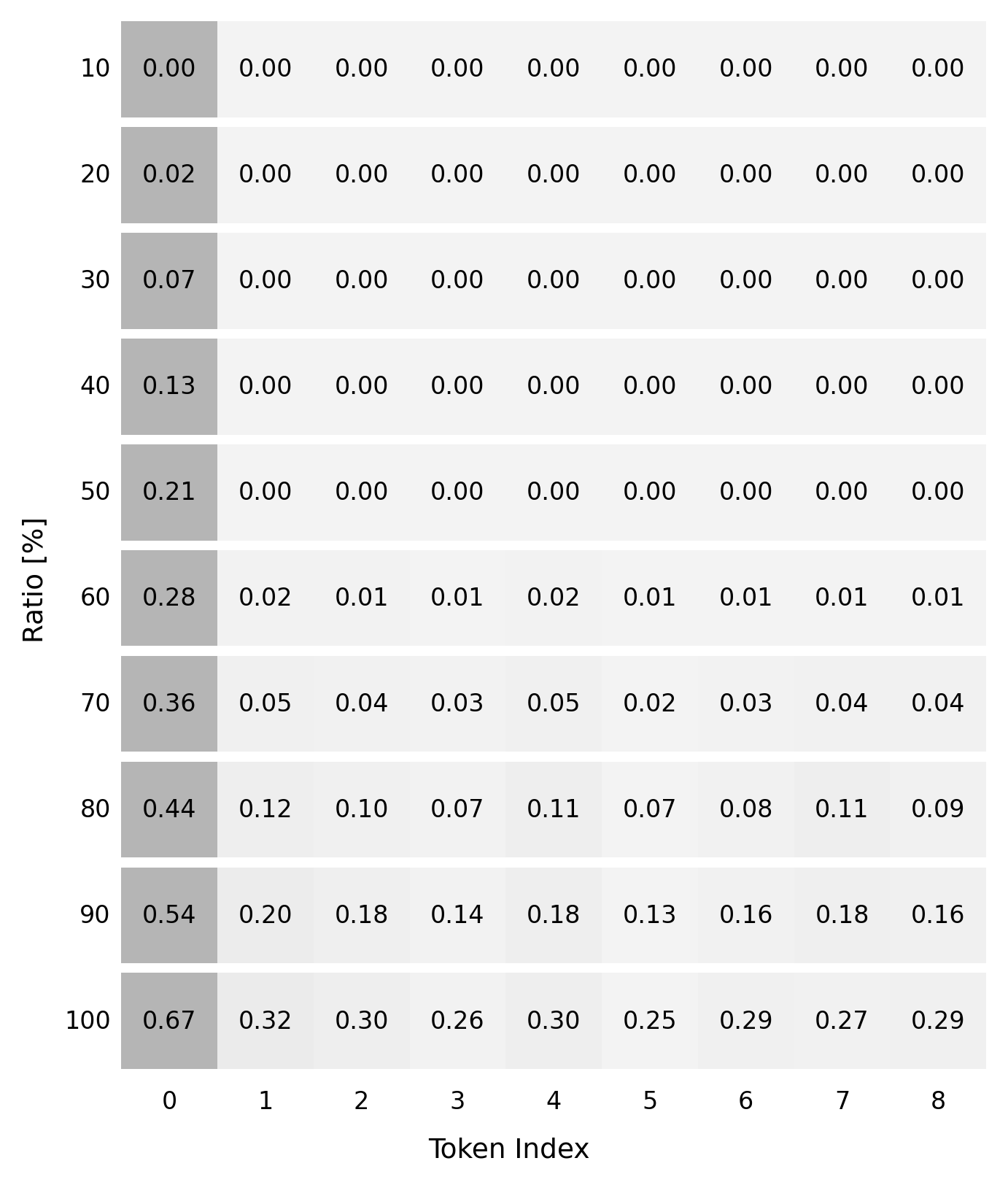}
        \caption{LLaMA-2-70B-chat}
        \label{mink:princ:e:second:llama-70b}
    \end{subfigure}
    \hfill
    \begin{subfigure}{0.45\textwidth}
        \centering
        \includegraphics[width=\linewidth]{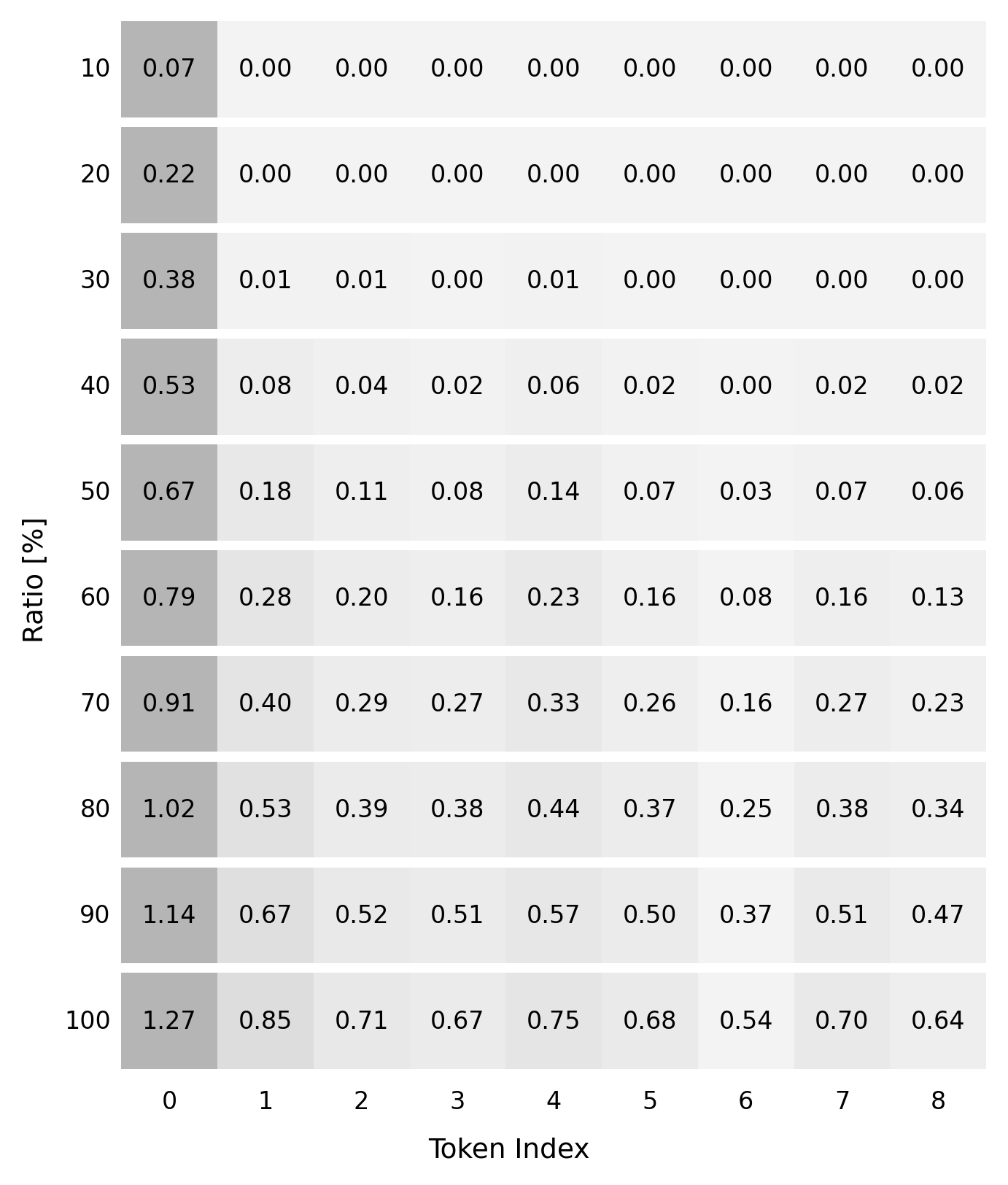}
        \caption{Mistral-7B-instruct}
        \label{mink:princ:e:second:mistral-7b}
    \end{subfigure}
    \caption{\textbf{[second]} Min-K Entropy scores across all percentiles over the first 9 tokens from second hallucination spans at global level.}
    \label{mink:princ:e:second}
\end{figure}

\begin{figure}[htbp]
    \centering
    \begin{subfigure}{0.45\textwidth}
        \centering
        \includegraphics[width=\linewidth]{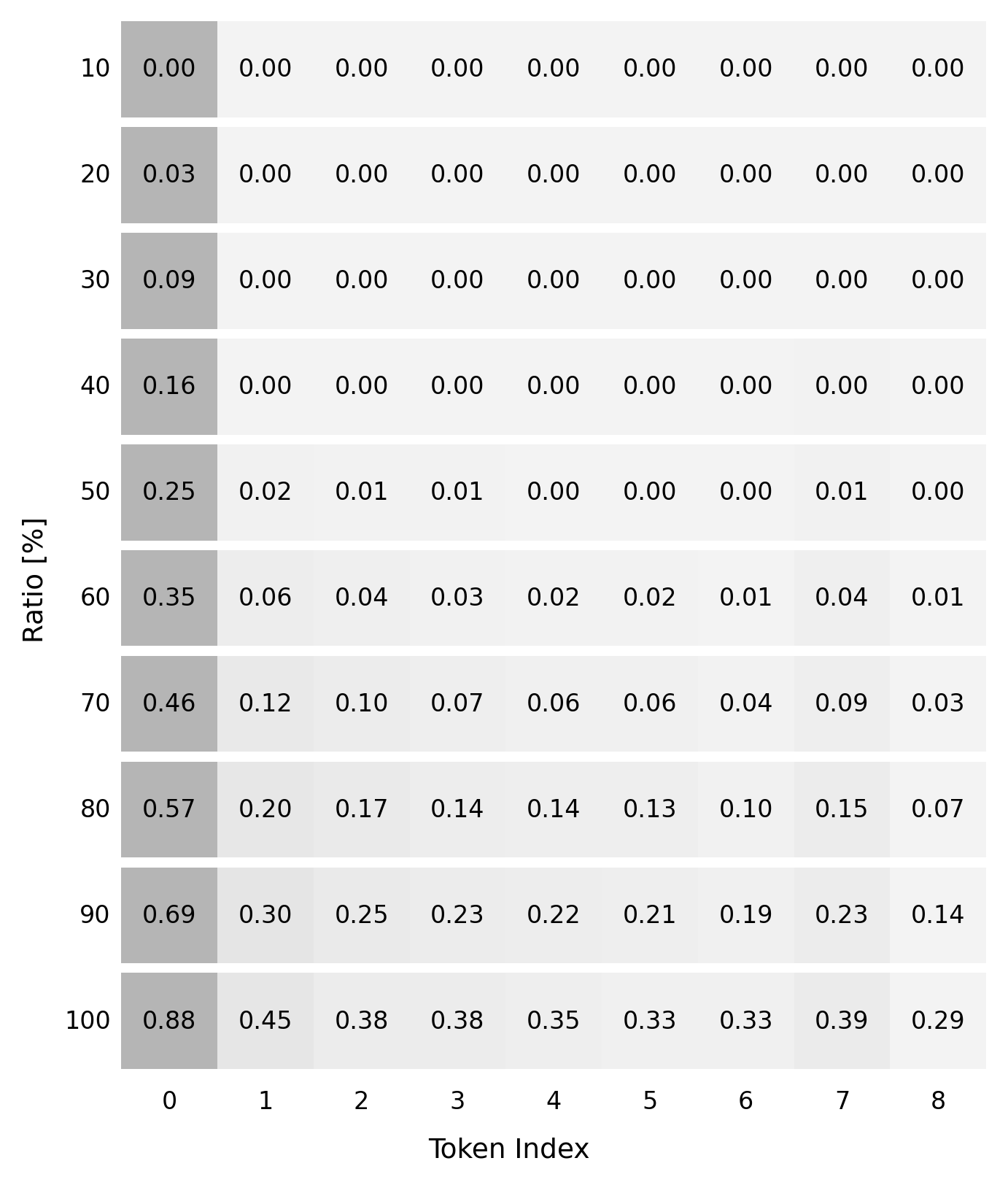}
        \caption{LLaMA-2-7B-chat}
        \label{mink:princ:e:third+:llama-7b}
    \end{subfigure}
    \hfill
    \begin{subfigure}{0.45\textwidth}
        \centering
        \includegraphics[width=\linewidth]{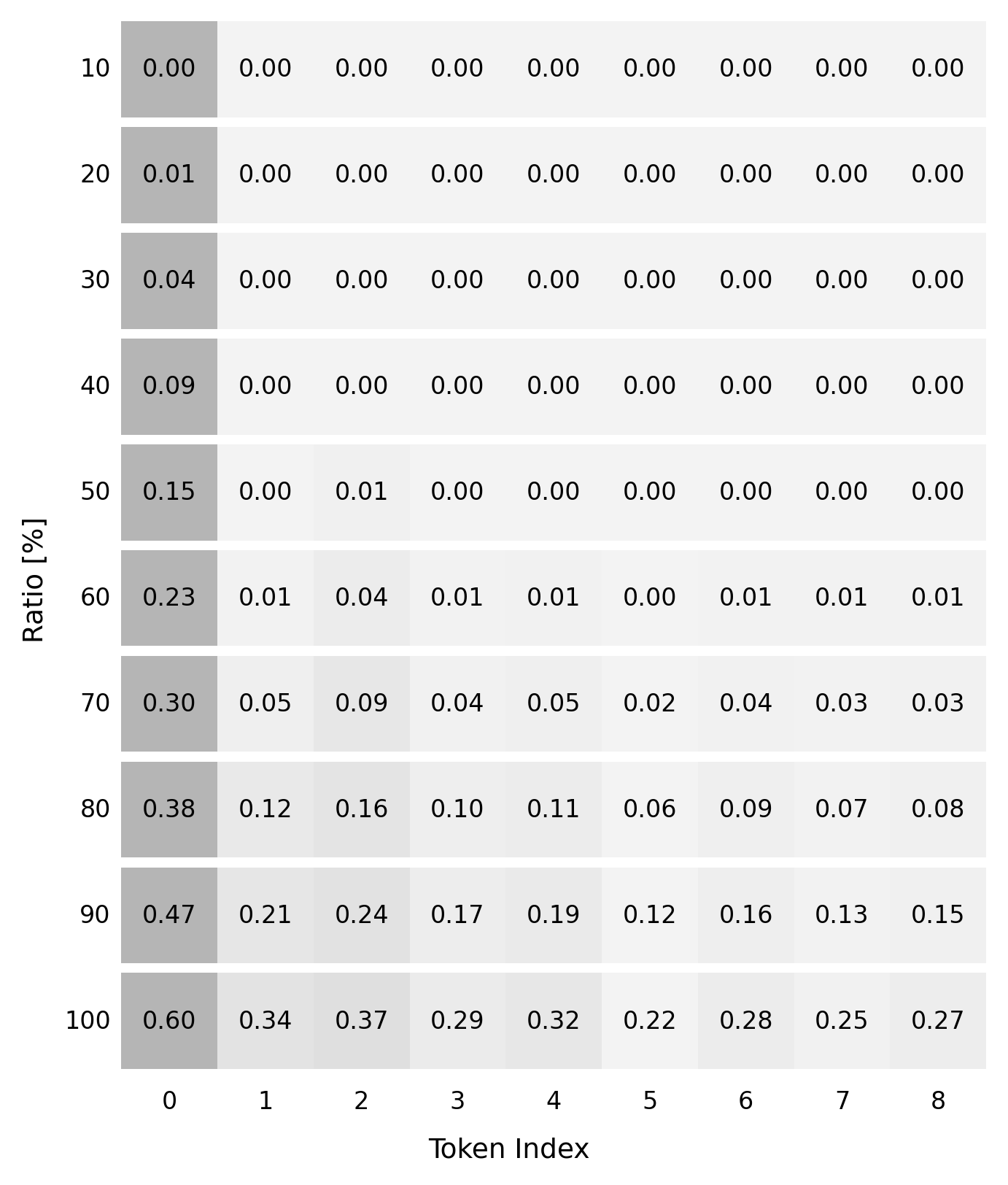}
        \caption{LLaMA-2-13B-chat}
        \label{mink:princ:e:third+:llama-13b}
    \end{subfigure}
    \medskip
    \begin{subfigure}{0.45\textwidth}
        \centering
        \includegraphics[width=\linewidth]{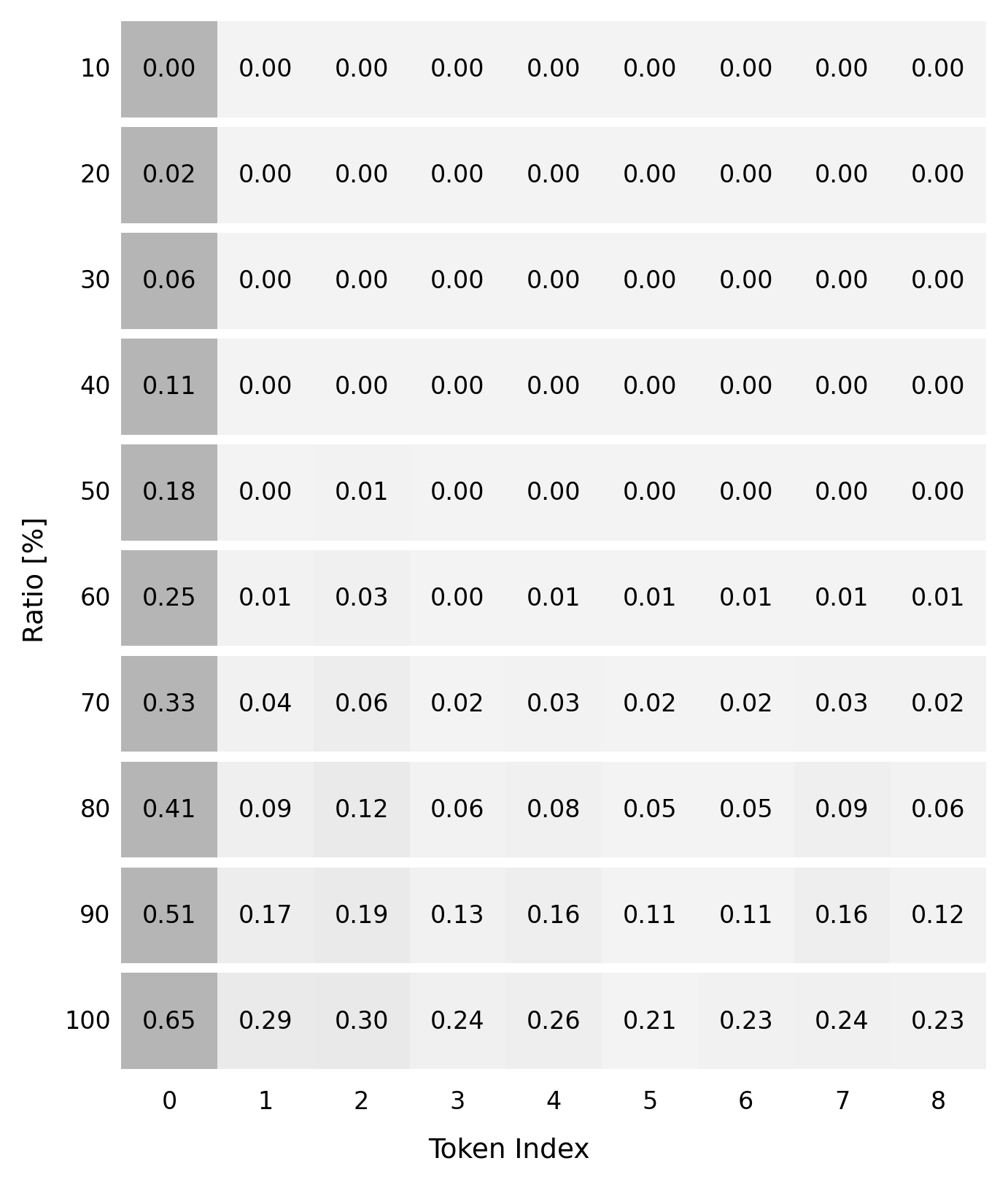}
        \caption{LLaMA-2-70B-chat}
        \label{mink:princ:e:third+:llama-70b}
    \end{subfigure}
    \hfill
    \begin{subfigure}{0.45\textwidth}
        \centering
        \includegraphics[width=\linewidth]{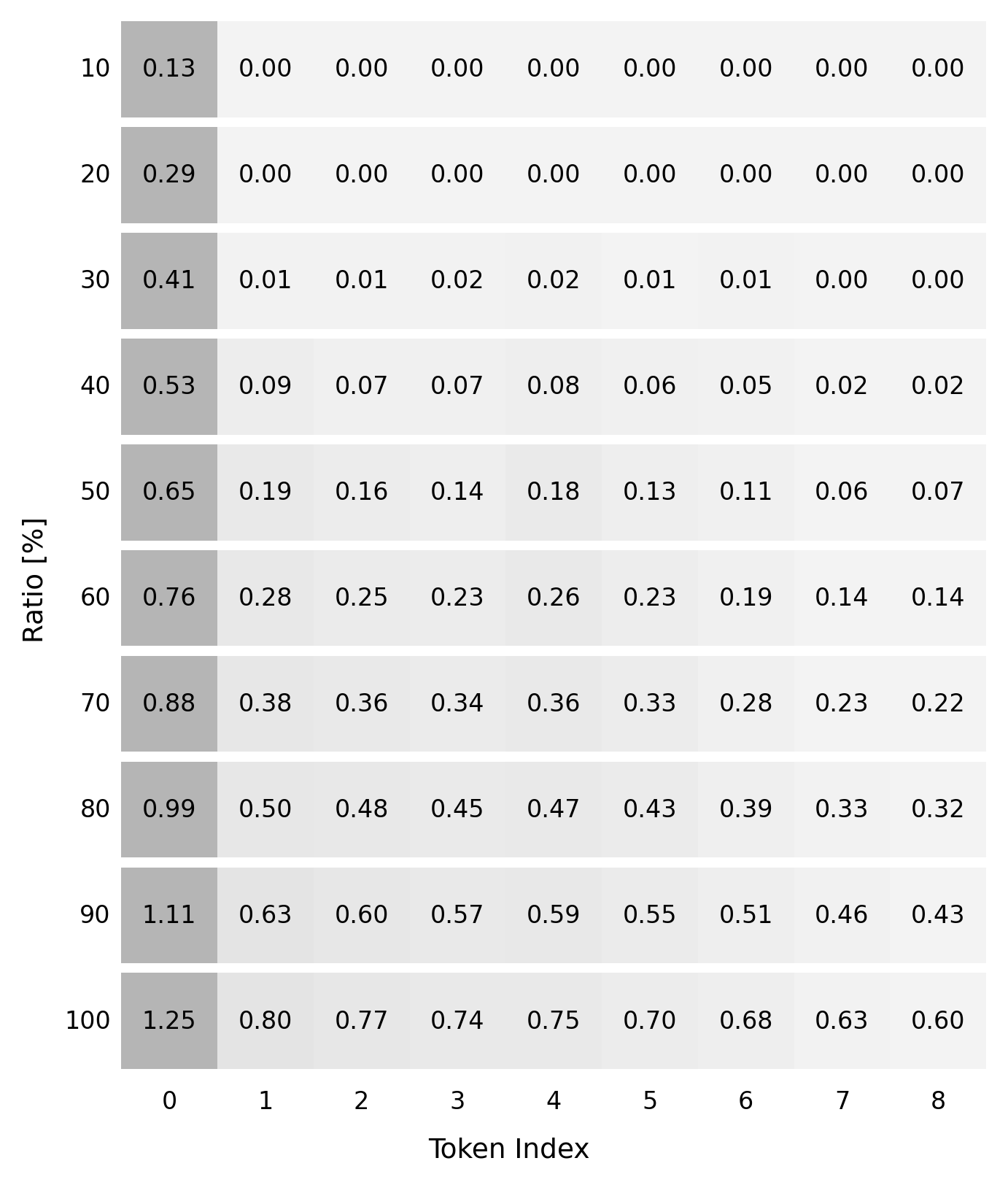}
        \caption{Mistral-7B-instruct}
        \label{mink:princ:e:third+:mistral-7b}
    \end{subfigure}
    \caption{\textbf{[third+]} Min-K Entropy scores across all percentiles over the first 9 tokens from third+ hallucination spans at global level.}
    \label{mink:princ:e:third+}
\end{figure}

\begin{figure}[htbp]
    \centering
    \begin{subfigure}{0.49\textwidth}
        \centering
        \includegraphics[width=\linewidth]{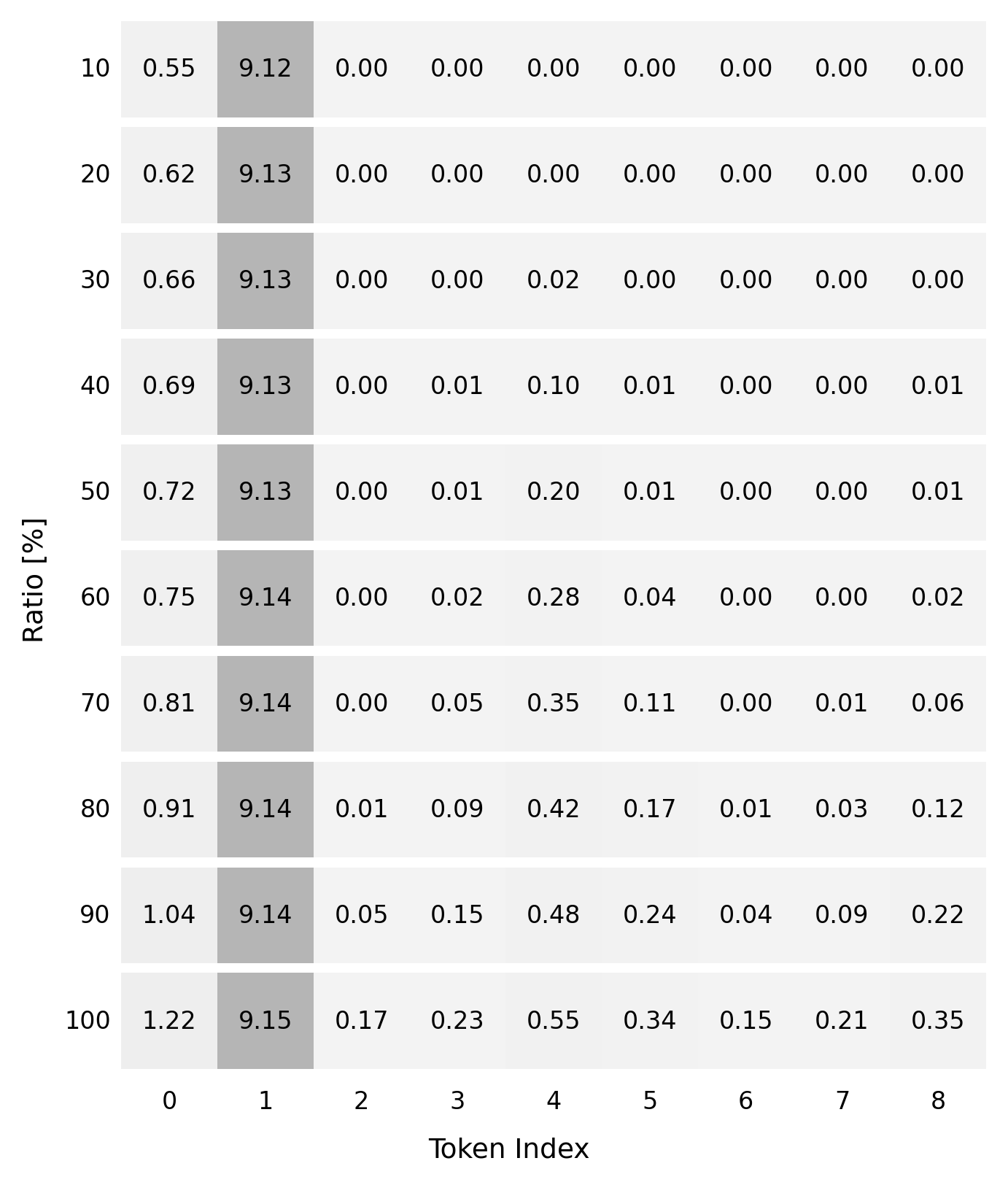}
        \caption{LLaMA-2-7B-chat}
        \label{mink:princ:e:pre:llama-7b}
    \end{subfigure}
    \hfill
    \begin{subfigure}{0.49\textwidth}
        \centering
        \includegraphics[width=\linewidth]{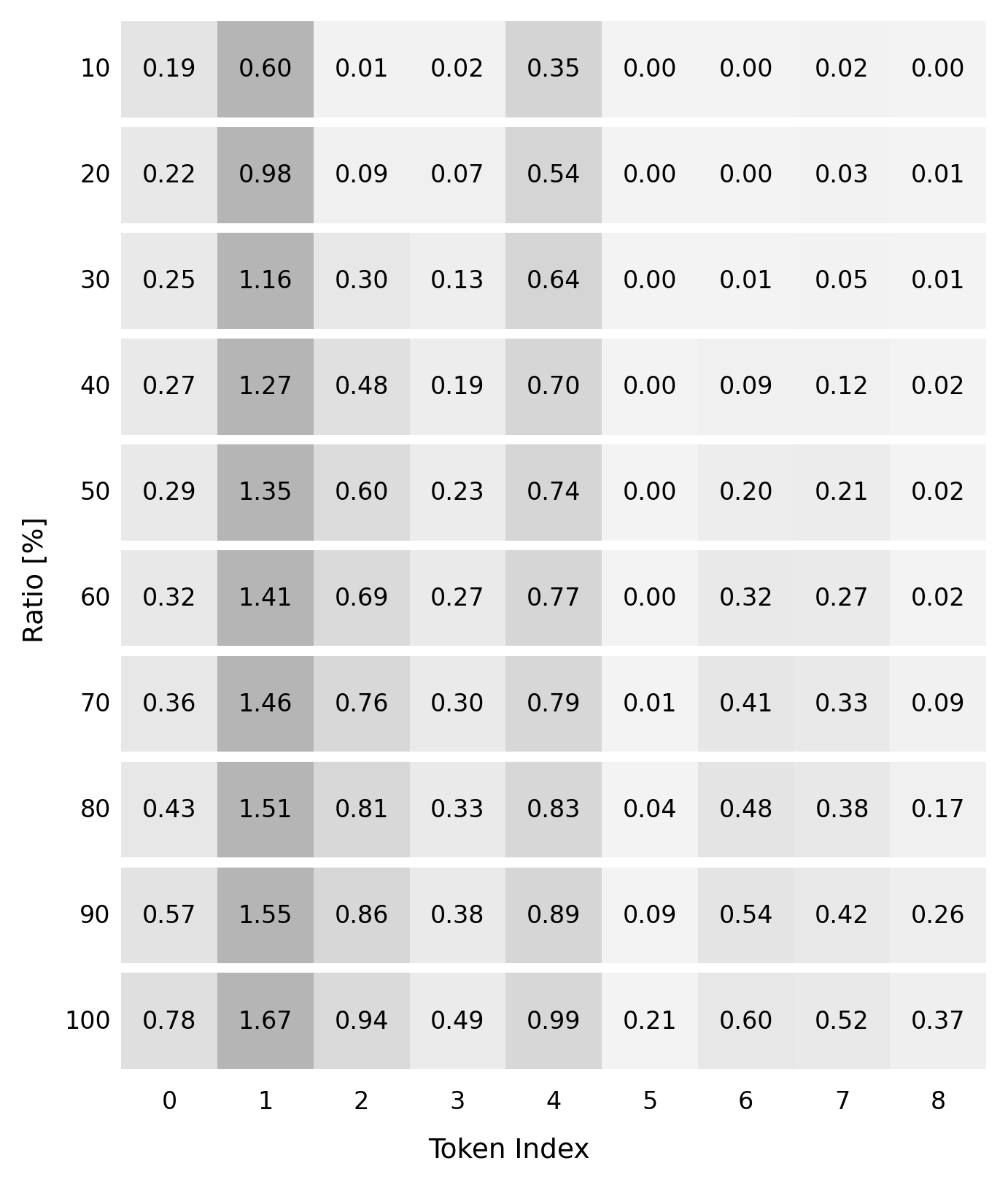}
        \caption{LLaMA-2-13B-chat}
        \label{mink:princ:e:pre:llama-13b}
    \end{subfigure}
    \medskip
    \begin{subfigure}{0.49\textwidth}
        \centering
        \includegraphics[width=\linewidth]{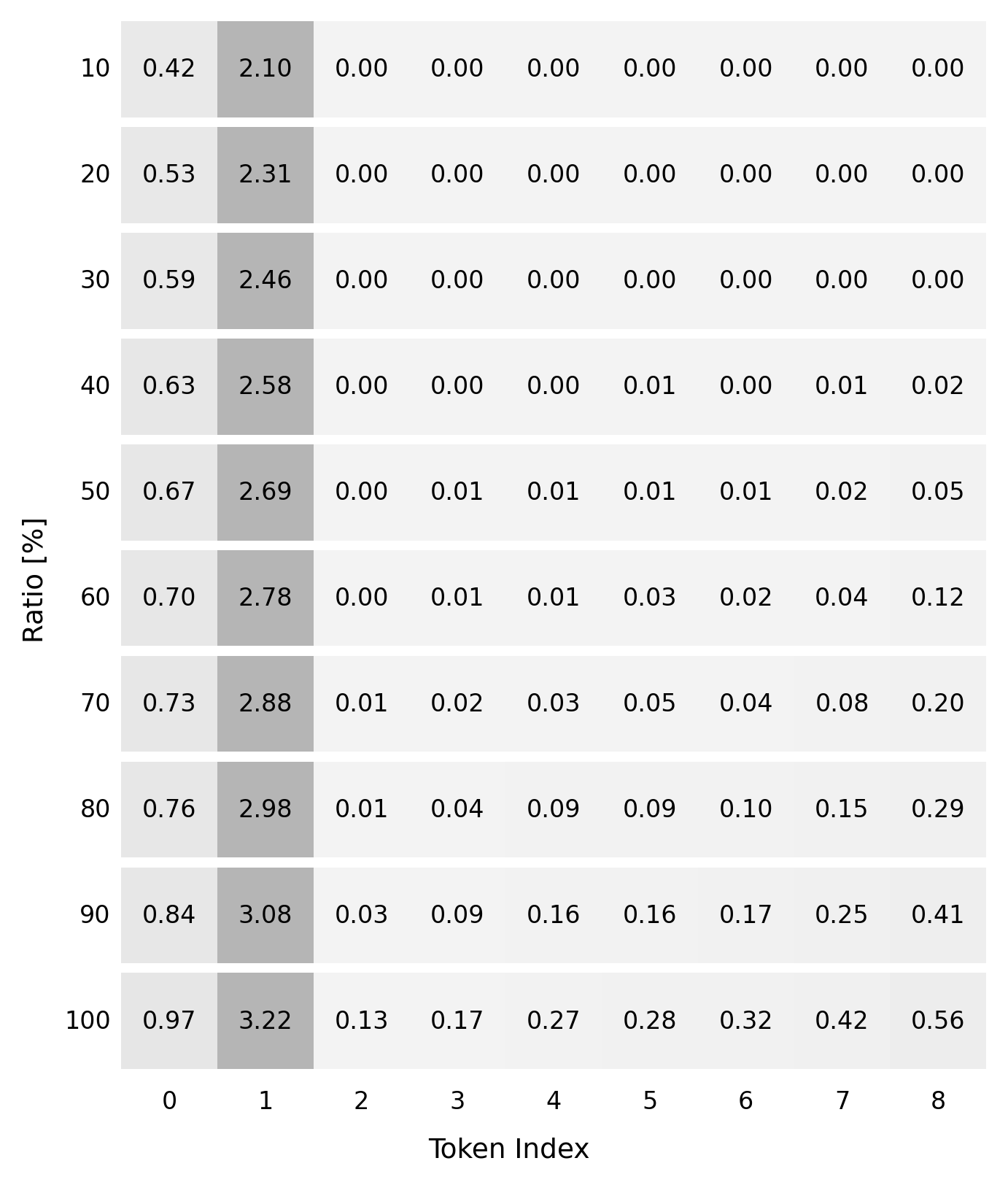}
        \caption{LLaMA-2-70B-chat}
        \label{mink:princ:e:pre:llama-70b}
    \end{subfigure}
    \hfill
    \begin{subfigure}{0.49\textwidth}
        \centering
        \includegraphics[width=\linewidth]{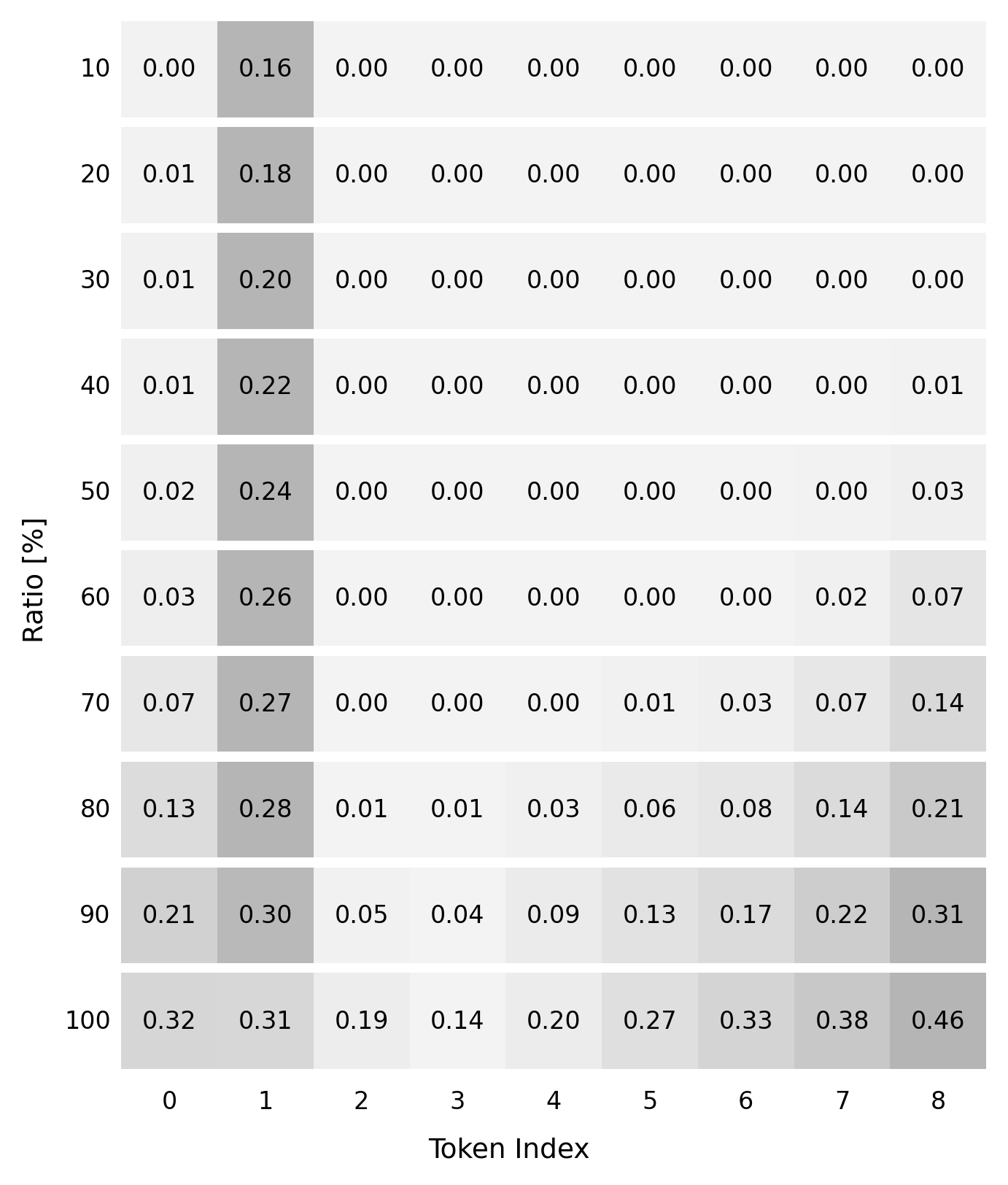}
        \caption{Mistral-7B-instruct}
        \label{mink:princ:e:pre:mistral-7b}
    \end{subfigure}
    \caption{\textbf{[pre]} Min-K Entropy scores across all percentiles over the first 9 pre-hallucination tokens at global level.}
    \label{mink:princ:e:pre}
\end{figure}

\begin{figure}[htbp]
    \centering
    \begin{subfigure}{0.49\textwidth}
        \centering
        \includegraphics[width=\linewidth]{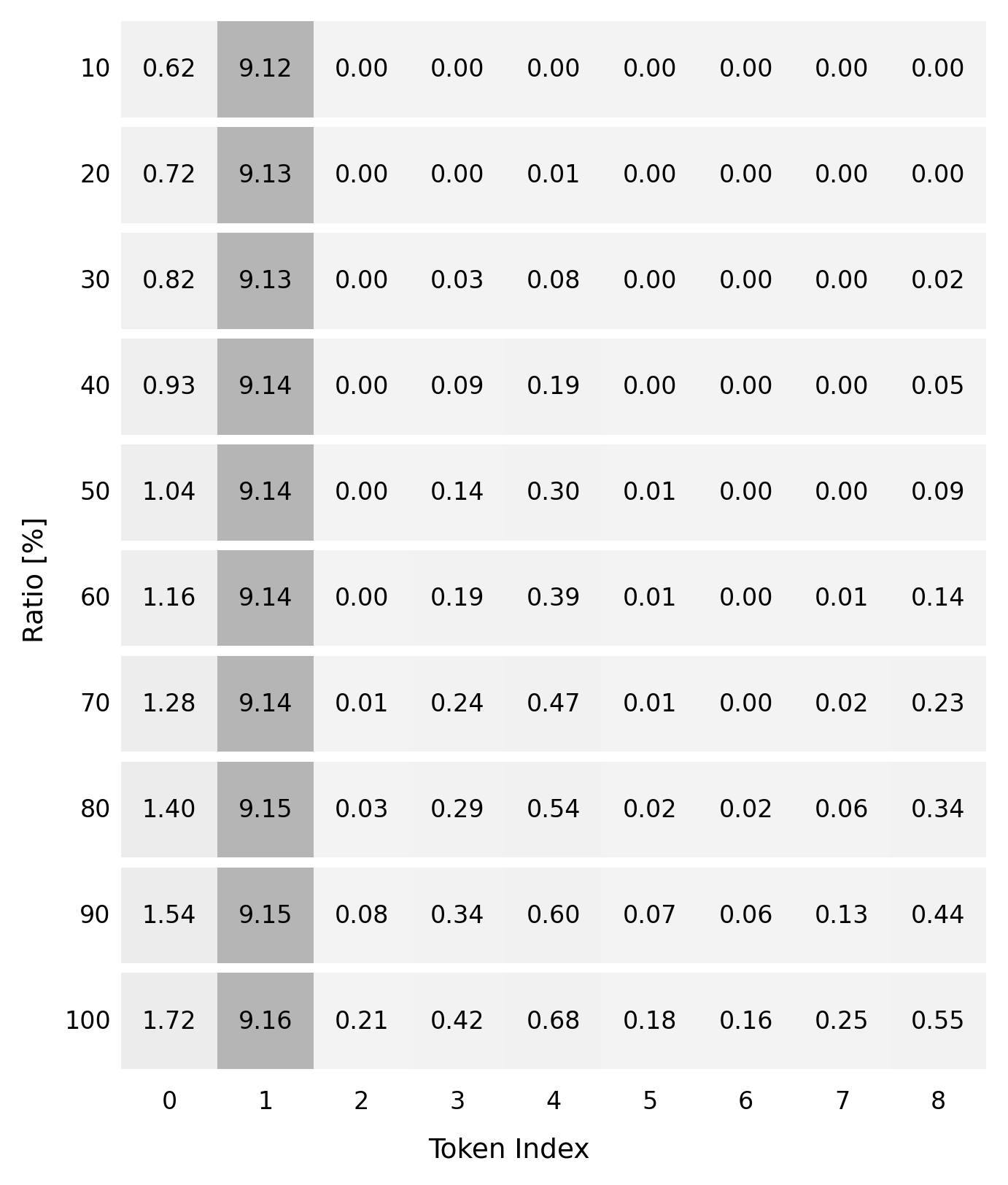}
        \caption{LLaMA-2-7B-chat}
        \label{mink:princ:e:no:llama-7b}
    \end{subfigure}
    \hfill
    \begin{subfigure}{0.49\textwidth}
        \centering
        \includegraphics[width=\linewidth]{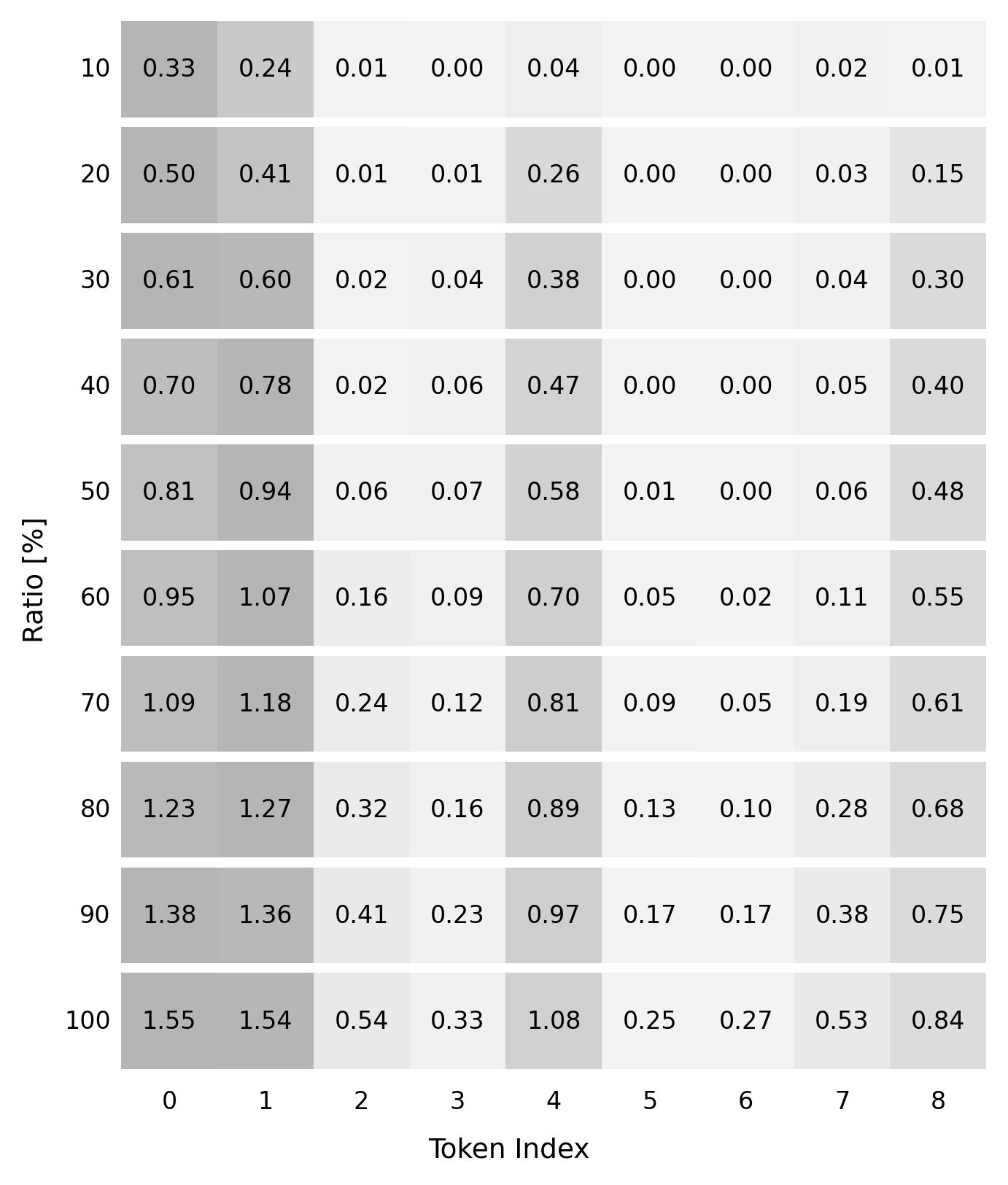}
        \caption{LLaMA-2-13B-chat}
        \label{mink:princ:e:no:llama-13b}
    \end{subfigure}
    \medskip
    \begin{subfigure}{0.49\textwidth}
        \centering
        \includegraphics[width=\linewidth]{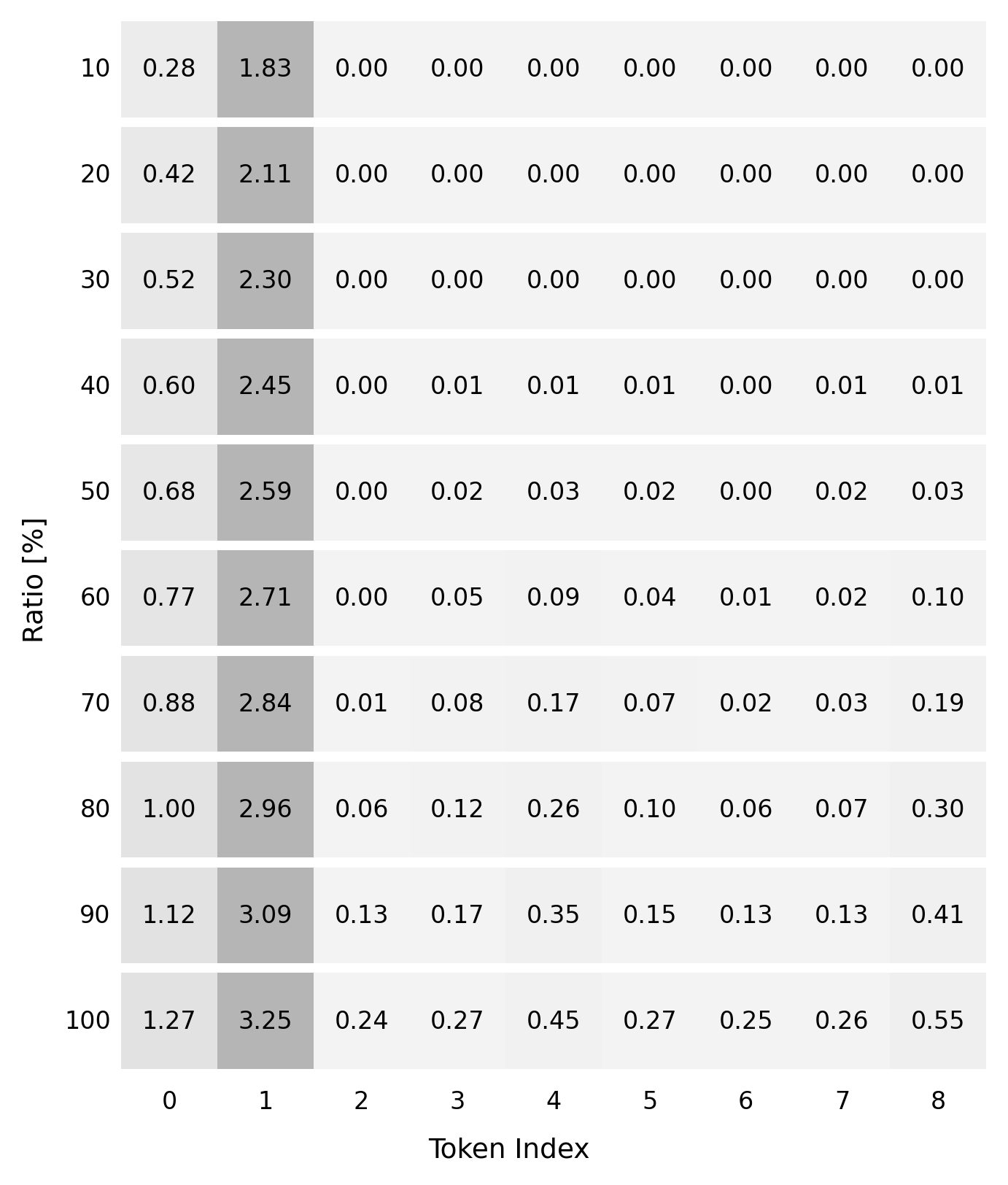}
        \caption{LLaMA-2-70B-chat}
        \label{mink:princ:e:no:llama-70b}
    \end{subfigure}
    \hfill
    \begin{subfigure}{0.49\textwidth}
        \centering
        \includegraphics[width=\linewidth]{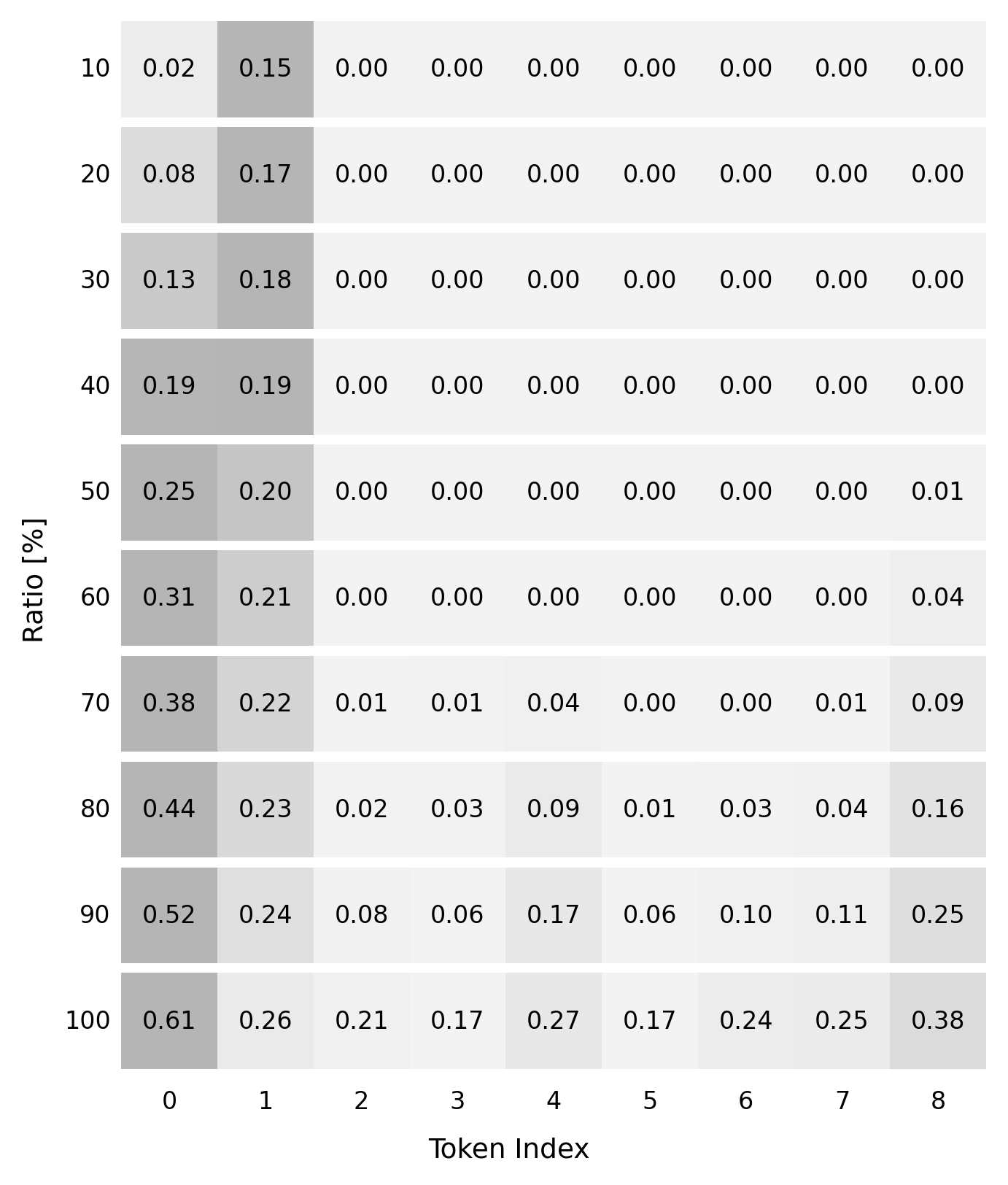}
        \caption{Mistral-7B-instruct}
        \label{mink:princ:e:no:mistral-7b}
    \end{subfigure}
    \caption{\textbf{[no]} Min-K Entropy scores across all percentiles over the first 9 tokens from responses without hallucination at global level.}
    \label{mink:princ:e:no}
\end{figure}

\FloatBarrier
\endgroup
\end{document}